%% file: thesis.tex
\documentclass[a4paper,11pt,authoryear,index]{Classes/PhDThesisPSnPDF}

\input{Preamble/preamble}

\input{thesis-info}

\ifdefineAbstract
\fi

\ifdefineChapter
\fi

\begin{document}
\titleformat{\chapter}[display]{\raggedleft\normalfont\fontsize{50}{60}\selectfont\bfseries}{\textcolor{Blue!60}{\thechapter}}{-10pt}{\huge}
  
\frontmatter

\maketitle

\iftrue

\include{Dedication/dedication}

\include{Acknowledgement/acknowledgement}\include{Abstract/abstract}

\include{Nomenclature/nomenclature}
\fi

\begin{spacing}{1.25}
\tableofcontents

\listoffigures

\listoftables
\printnomenclature[8em]
\end{spacing}

\mainmatter

\include{Chapter1/chapter1}

\include{Chapter2/chapter2}
\include{Chapter6/chapter6}
\include{Chapter7/chapter7}
\include{Chapter8/chapter8}

\include{Chapter9/chapter9}

\begin{spacing}{0.5}
\setlength{\bibsep}{6pt plus 0.5ex}

\bibliographystyle{abbrvnat} %
\cleardoublepage
 \bibliography{References/references} %

\end{spacing}

\begin{appendices} %

\include{Appendix1/appendix1}

\end{appendices}

\printthesisindex %

\end{document}

%% file: thesis-info.tex
\title{Identifiability Results for\\ Causal Representation Learning}
\title{Identifiable Causal Representation Learning}
\subtitle{\vspace{1em}
Unsupervised, Multi-View, and Multi-Environment}
\author{Julius von K\"ugelgen}

\dept{Department of Engineering}

\university{University of Cambridge}
\crest{\includegraphics[width=0.2\textwidth]{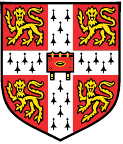}}
\supervisorlinewidth{0.45\textwidth}

\advisorlinewidth{0.38\textwidth}

\degreetitle{Doctor of Philosophy}

\college{Churchill College}

\degreedate{November 2023} 

\subject{Causal Inference and Machine Learning} \keywords{{Causal Inference} {Machine Learning} {PhD Thesis} {Engineering} {University of
Cambridge}}

%% file: Dedication/dedication.tex
\begin{dedication} 
To my loving parents, who set me on the joyful path of inquiry. 
\end{dedication}

%% file: Acknowledgement/acknowledgement.tex
\begin{acknowledgements}      
\iftrue
\vspace{-2.5em}
\looseness-1 Pursuing a PhD has been a formative period, and I have many people to thank for making it such a rewarding experience.
First and foremost, I would like to express my sincere gratitude to my supervisors for allowing me
to conduct research at two such wonderful institutions as the University of Cambridge and the Max Planck Institute for Intelligent Systems, T\"ubingen.
Bernhard Sch\"olkopf is responsible for sparking my interest in and excitement about causality.
Besides cultivating a lab where collaboration and fundamental research thrive, his insights have crucially shaped my personal research direction and understanding of the field.
Adrian Weller welcomed me to Cambridge and gave me the freedom to read and explore topics I am passionate about. 
While we only worked together on a couple of projects,
I am truly grateful for his continued advice, unconditional support, and reassuring words, which have always been a source of encouragement.
Both of them have supported my development as a researcher in numerous ways.
I would also like to wholeheartedly thank my advisor Richard Turner 
for sharing his experience,
and my examiners Pradeep Ravikumar and Jos\'e Miguel Hern\'andez-Lobato for taking the time to scrutinize and provide insightful comments on this thesis.
Further, I thank Google for supporting me through a PhD Fellowship in Machine Learning.

Research can be a lonely endeavour, and I have been incredibly fortunate to meet many inspiring colleagues and mentors along the way.
My time in T\"ubingen would not have been the same
without 
Luigi Gresele, who has over the years become both a close friend and co-leading author on numerous projects, and who first introduced me to ICA and identifiability.
A special thanks also goes to Michel Besserve, who has been very generous in sharing his insights and mathematical prowess.
Both 
were heavily involved in most of the work presented in this thesis, and I have learnt a lot from our many discussions and deep dives.
I would also like to highlight Francesco Locatello, both for his scientific input and advice on other aspects of academia, and Cian Eastwood, who has kindly involved me in several interesting projects---I have greatly enjoyed working with you both and benefited from our many conversations in and outside the office. 
Further, I am very grateful to
Isabel Valera for her kind mentorship during my early days in T\"ubingen, 
and to Amir-Hossein Karimi for our joint projects on algorithmic recourse. 

During my doctoral studies, I also had the privilege to visit some other world-class institutions, and I want to thank: 
David M.\ Blei and Elias Bareinboim for kindly welcoming me into their labs and hosting me for a research visit at Columbia University, Peter Gehler for being a great manager during my internships at Amazon, and Frederik Eberhardt for organising the Simons Institute Causality program at UC Berkeley, part of which I had the pleasure to attend. 
I am also grateful to all the other senior researchers and professors who took the time to meet with me and share their scientific insights, including 
Yoshua Bengio, 
Matthias Bethge, 
Dominik Janzing,
Chris Russell, 
Uri Shalit,
Ricardo Silva,
Jin Tian,
Richard Zemel, and
Kun Zhang.

\looseness-1 Collaborating with many smart people has been one of my favourite aspects of the PhD, 
and I am indebted to all my other co-authors for their contributions to our projects, 
including but not limited to:
Sander Beckers, 
Umang Bhatt, 
Jack Brady,
Wieland Brendel, 
Simon Buchholz,
Shubhangi Ghosh,
Zhijing Jin, 
Jonas K\"ubler, 
Armin Keki\'c, 
Klaus-Rudolf Kladny, 
S\'ebastien Lachapelle,
Felix Laumann,
Marco Loog, 
Lars Lorch,
Sara Magliacane,
Osama Makansi, 
Georg Martius,
Alexander Mey,
Abdirisak Mohamed,
Michael Muehlebach, 
Andrei Nicolicioiu, 
Junhyung Park,
Robert Peharz,
Ronan Perry, 
Patrik Reizinger,
Paul Rubenstein,
Lukas Schott,
Yash Sharma, 
Shashank Singh,
Vincent Stimper, 
Matthias Tangemann,
Perouz Taslakian, 
Christian Toth, 
Frederik Tr\"auble,
Ivan Ustyushaninov,
Liang Wendong,
Danru Xu,
and
Dingling Yao.
In addition, I would like to thank some other colleagues and fellow researchers met during conferences, research visits, or workshops, whom I did not get to work with directly, but who have positively shaped my PhD experience through friendly encounters and insightful discussions: 
Kartik Ahuja, 
Bryon Aragam, 
Philippe Brouillard, 
Patrick Burauel,
Nicolas Chapados, 
Taco Cohen, 
Alexander D'Amour,
Pim de Haan, 
Alexandre Drouin,
Sergio Garrido, 
Jason Hartford, 
Niki Kilbertus, 
Alexandre Lacoste, 
Daniel Malinsky, 
Gemma Moran, 
Krikamol Muandet, 
Alexander Neitz, 
Elizabeth Ogburn, 
Giambattista Parascandolo, 
Mateo Rojas-Carulla, 
Aaron Schein, 
Dhanya Sridhar, 
Chandler Squires,
Victor Veitch, 
and Sebastian Weichwald.

Further, I am grateful to Jack Brady, Cian Eastwood, Luigi Gresele, and Michele Tonutti for providing useful feedback on earlier versions of this thesis.

I also want to extend a warm thank you to our secretaries and administrative staff---Ann-Sophie,
Camelia,
Lidia,
and Sabrina 
in T\"ubingen, and Catherine, Kim,  
and Rachel 
in Cambridge---for help with bureaucratic issues and travel arrangements; to Sebastian and Vincent 
for support with IT and equipment; and to
Alex, Armin, Diego, Fred, Heiner, Jonas K., Jonas W., Luigi, Max, Nasim, Sergio, Timmy, and Yassine for EI office banter and good lunch conversations.

This journey would not have been nearly as enjoyable without many wonderful friends, both new and old.
Adri\'an, Ahmad, 
Diego, Gigi, Heiner, Marie,
Michael,
Sabrina, Sofia, and Yves---living in a small student town like T\"ubingen never got boring with you. 
Dan and Miguel---your friendship has kept me sane while in Cambridge.
Mic and Tom---the dinner discussions back in our undergrad days first got me excited about AI.
Adrian, Bille, Danny, Fynn, Gage, Gio, Jan, Juli, Leo, Marlene, Melis, Nani, Olivier, and Robin---thanks for being there all these years and for all the good times, I hope there will be many more.

\looseness-1 Finally, but most importantly of all, I want to express my deepest gratitude to all my family and loved ones, especially:
To my godfather Rainer B., for being a role model in science and inspiring me to want to become an academic.
To my parents,
for their unwavering encouragement to find my passion, and for all their advice, and emotional and financial support
along the way. 
And to Ina, for her love and kindness, for being an amazing person and partner, and for sticking with me also through all the stressful and difficult times. 
\fi
\end{acknowledgements}

%% file: Abstract/abstract.tex
\begin{abstract}
\vspace{-1em}

This thesis brings together ideas from \text{causality} and \text{representation learning}. 
Causal models provide rich descriptions of complex systems as sets of
\text{mechanisms} by which each variable is influenced by its direct causes.
They support reasoning about manipulating parts of the system, capture a whole range of \text{interventional} distributions, 
and thus hold promise for addressing some of the open challenges of artificial intelligence (AI), such as \text{planning}, \text{transferring knowledge} in changing environments, or \text{robustness to distribution shifts}.
However, a key obstacle to a more widespread use of causal models in AI is the requirement that the relevant variables need to be specified a priori, which is typically not the case for the high-dimensional, unstructured data processed by modern AI systems.
At the same time, machine learning (ML) has proven quite successful at automatically extracting useful and compact \text{representations} of such complex data. 
\textit{Causal representation learning} (CRL) aims to combine the core strengths of ML and causality by learning representations in the form of latent variables endowed with causal model semantics. 

In this thesis, we study and present new results for different CRL settings.
A central theme is the question of \textit{identifiability}: Given infinite data, when are representations satisfying the same learning objective guaranteed to be equivalent? 
This is arguably an important prerequisite for CRL, as it formally characterises if and when a learning task is, at least in principle, feasible.
Since learning causal models---even without a representation learning component---is notoriously difficult, we require additional assumptions on the model class or rich data beyond the classical i.i.d.\ setting.
For \textit{unsupervised} representation learning from i.i.d.\ data, we develop independent mechanism analysis, a  constraint on the mixing function mapping latent to observed variables, which is shown to promote the identifiability of independent latents.
For a \textit{multi-view} setting of learning from pairs of non-independent observations, we prove that the invariant block of latents that are always shared across views can be identified. 
Finally, for a \textit{multi-environment} setting of learning from non-identically distributed datasets arising from perfect single-node interventions, we show that the latents and their causal graph are identifiable. 

By studying and partially characterising identifiability for different settings, this thesis investigates what is possible and impossible for CRL without direct supervision, and thus contributes to its theoretical foundations. 
Ideally, the developed insights can help inform data collection practices or inspire the design of new practical estimation methods and algorithms.
\end{abstract}

%% file: Nomenclature/nomenclature.tex
\nomenclature[z-CRL]{CRL}{Causal Representation Learning}
\nomenclature[z-ML]{ML}{Machine Learning} 
\nomenclature[z-AI]{AI}{Artificial Intelligence}   
\nomenclature[z-SCM]{SCM}{Structural Causal Model}
 \nomenclature[z-IMA]{IMA}{Independent Mechanism Analysis}
 \nomenclature[z-SSL]{SSL}{Self-Supervised Learning}
 \nomenclature[z-VAE]{VAE}{Variational Autoencoder}
\nomenclature[z-iid]{i.i.d.}{independent and identically distributed}
\nomenclature[z-st]{s.t.}{such that (also written as ``:'' within equations)}
\nomenclature[z-ie]{i.e.}{that is (Latin: ``id est'')}
\nomenclature[z-iff]{iff.}{if and only if}
\nomenclature[z-eg]{e.g.}{for example (Latin: ``exempli gratia'')}
\nomenclature[z-wrt]{w.r.t.}{with respect to}
\nomenclature[z-ERM]{ERM}{Empirical Risk Minimisation}
\nomenclature[z-ICM]{ICM}{Independent Causal Mechanism(s)}
\nomenclature[z-ICA]{ICA}{Independent Component  Analysis}
\nomenclature[z-DAG]{DAG}{Directed Acyclic Graph}
\nomenclature[z-LHS]{LHS}{Left-Hand Side (of an equation)}
\nomenclature[z-LVM]{LVM}{Latent Variable Model}
\nomenclature[z-RHS]{RHS}{Right-Hand Side (of an equation)}
\nomenclature[z-wlog]{w.l.o.g.}{without loss of generality}
\nomenclature[z-ELBO]{ELBO}{Evidence Lower Bound}
\nomenclature[z-CL]{CL}{Contrastive Learning}
\nomenclature[z-QRM]{QRM}{Quantile Risk Minimization}
\nomenclature[z-SFB]{SFB}{Spurious Feature Boosting}
\nomenclature[z-DG]{DG}{Domain Generalisation}
\nomenclature[z-CauCA]{CauCA}{Causal Component Analysis}
\nomenclature[z-IGCI]{IGCI}{Information Geometric Causal Inference}

\nomenclature[A, 11]{$V$}{random variable (r.v.), can be observed or unobserved/latent}
\nomenclature[A, 13]{$U_V$}{exogenous (or noise) variable affecting $V$, always unobserved}
\nomenclature[A, 14]{$Y$}{usually a label or target r.v.}
\nomenclature[A, 15]{$\mathbf{X}$}{random vector $\mathbf{X}=(X_1, ..., X_d)^\intercal$ of observed variables}
\nomenclature[A, 16]{$\Zb$}{random vector $\Zb=(Z_1,...,Z_n)^\intercal$ of latent variables \\(sometimes corresponding to ground truth generative factors)}
\nomenclature[C, 10]{$v$}{scalar, e.g., a realization of $V$}
\nomenclature[C, 11]{$\xb, \zb$}{vectors, e.g., realizations of $\Xb,\Zb$}
\nomenclature[C, 12]{$\Ab, \mathbf{B}$}{matrices, e.g., weighted adjacency matrix}
\nomenclature[C, 13]{$\ab^\intercal,\Ab^\intercal$}{transpose of vector $\ab$ or matrix $\Ab$}
\nomenclature[C, 14]{$\det \Ab$}{determinant of a square matrix $\Ab$}
\nomenclature[C, 15]{$\tr\Ab$}{trace of a square matrix $\Ab$}

\nomenclature[D, 01]{$P_\Zb$ or $P(\Zb)$}{probability distribution of $\Zb$}
\nomenclature[D, 02]{$p_\Zb$ or $p(\Zb)$}{probability (mass or density) function of $\Zb$}
\nomenclature[D, 03]{$p(\zb)$ or $p_\Zb(\zb)$}{probability (mass or density) function of $\Zb$ evaluated at $\zb$}
\nomenclature[D, 04]{$P_{Y\mid\Xb}$ or $P(Y\mid\Xb)$}{collection of $P(Y\mid\Xb=\xb)$ for all $\xb$; conditional of $Y$ given $\Xb$}
\nomenclature[D, 05]{$p(y\mid\xb)$}{conditional probability function of $Y$ given $\Xb=\xb$ evaluated at $y$}
\nomenclature[D, 06]{$\fb_*(P_\Zb)$ or $\fb_*(p_\Zb)$}{pushforward distribution or density of the r.v.\ $\fb(\Zb)$}
\nomenclature[D, 07]{$\Ncal(\bm{\mu},\bm\Sigma)$}{multivariate Gaussian distribution with mean $\bm\mu$ and covariance~$\bm\Sigma$}
\nomenclature[D, 08]{$\Ebb[X]$}{expectation of $X$}
\nomenclature[D, 09]{$\Var[X]$}{variance of $X$}
\nomenclature[D, 10]{$\Cov[X,Y]$}{covariance of $X$ and $Y$}
\nomenclature[D, 11]{$X \independent Y$}{independence between random variables $X$ and $Y$}
\nomenclature[D, 12]{$X \independent Y\,\mid \,W$}{conditional independence between $X$ and $Y$ given $W$}

\nomenclature[G, 1]{$G$}{causal graph or causal diagram (usually a DAG)}
\nomenclature[G, 2]{$\Mcal$}{structural causal model}
\nomenclature[G, 3]{$\PA^\Gcal_{i}$, $\DE^\Gcal_{i}$, $\AN^\Gcal_{i}$}{parents, descendants, and ancestors of node $i$ in graph $\Gcal$}

\nomenclature[E, 1]{$f,g,h$}{scalar functions}
\nomenclature[E, 2]{$\fb,\gb,\hb$}{vector-valued functions}
\nomenclature[E, 3]{$f_i$}{$i$\textsuperscript{th} component of $\fb$}
\nomenclature[F, 4]{$f'$}{derivative of $f$}
\nomenclature[F, 5]{$\frac{\partial \fb}{\partial z_j}$}{partial derivative of $\fb$ w.r.t.\ $z_j$}
\nomenclature[F, 6]{$\Jb_\fb$}{Jacobian matrix $(\Jb_\fb)_{ij}=\frac{\partial f_i}{\partial z_j}$}

\nomenclature[S, Xcal]{$\Xcal$}{observation space, usually a subset of $\mathbb{R}^d$}
\nomenclature[S, Zcal]{$\Zcal$}{latent or representation space, usually $\mathbb{R}^n$}

%% file: Chapter1/chapter1.tex
\ifpdf
    \graphicspath{{Chapter1/Figs/Raster/}{Chapter1/Figs/PDF/}{Chapter1/Figs/}}
\else
    \graphicspath{{Chapter1/Figs/Vector/}{Chapter1/Figs/}}
\fi

\chapter{Introduction}  %
\label{chap:introduction}

This thesis presents contributions in \textit{causal representation learning} (CRL), a subfield of artificial intelligence (AI) research that brings together ideas from machine learning (ML) and causality. 
To set the scene, we start with a brief historical account of different approaches to AI~(\cref{sec:history_of_AI}) and describe some limitations of current ML-based systems~(\cref{sec:weaknesses_of_current_ML}).
We then sketch
how \textit{causal}, as opposed to purely statistical, modelling may, in principle, help address some of these challenges~(\cref{sec:from_correlation_to_causation}).
However, to handle high-dimensional unstructured data, causal models need to be combined with representation learning techniques, giving rise to CRL~(\cref{sec:from_causal_models_to_causal_representations}).
 Finally, we motivate the study of \textit{identifiability} as a principled mathematical tool for understanding learnt representations~(\cref{sec:intro_identifiability}).
We summarise
 the organisation and main contributions of this thesis in~\cref{sec:overview_and_contributions}, and conclude by listing the underlying  publications in \cref{sec:publications}, as well as other publications not covered in detail in this thesis in~\cref{sec:other_publications}.

\section{A Brief History of Symbolic vs.\ Learning-Based AI}
\label{sec:history_of_AI}
Since the mid 20\textsuperscript{th} century, the quest for \textit{artificial intelligence} (AI) 
has been to build machines that exhibit intelligent behaviour~\citep{turing1950computing,mccarthy-proposal-for-dartmouth-1955}.
This includes core capabilities of human intelligence such as learning, planning, and reasoning.
Some of the first approaches known as 
 \textit{symbolic AI} involved manipulating a prespecified set of symbols based on typically hand-crafted logical rules. This led to early successes in tasks such as playing checkers~\citep{samuel1959checkers}, automated theorem proving~\citep{newell1956logic}, or problem solving~\citep{newell1959gps}, and later to the development of expert systems~\citep{russell2020artificial}. 
Limited by the need
 to program the symbols and rules by hand, however, these approaches were not easily scalable to the exponentially large number of concepts, relations, and exceptions involved in more complex tasks.

In parallel, the field of \textit{machine learning} (ML) developed as an alternative, data-driven approach to AI. 
Instead of symbolic representations and instructions specified a priori by a human expert, ML systems learn from experience by finding regularities, or patterns, in the data---similar to many natural forms of intelligence. 
Some of the first learning machines were shallow
 artificial neural networks (NNs) with adjustable synaptic weights~\citep{rosenblatt1958perceptron}, based on earlier models of Boolean NNs~\citep{mcculloch1943logical,hebb1949behavior}. 
With advances in training algorithms~\citep{rumelhart1986learning} and architectures~\citep{lecun1989backpropagation,hochreiter1997long}, later NNs got progressively deeper and more expressive.

For the first decades, symbolic approaches prevailed in many areas of AI, with purely learning-based systems only able to solve relatively simple predictive tasks. 
In hindsight, this can to a large extent be attributed to a lack of sufficient labelled data and compute power (combined with multiple periods of decreased  funding for AI research, the so-called ``AI winters'').
Then, the invention and wide-spread adaptation of the World Wide Web in the 1980s and 90s, together with the exponential increase in processing power of modern computer chips driven by \textit{Moore's law}~\citep{moore1965experts} heralded in the digital age. 
In the era of \textit{big data}, the vast majority of information is no longer found in analogue sources such as books, but instead stored digitally. With massive amounts of text and images available online, as well as immense computing resources (and several algorithmic advances), \textit{deep learning}~\citep{schmidhuber2015deep,lecun2015deep,goodfellow2016deep}, an ML technique based on deep NNs, has emerged in the 2010s and 2020s as the dominant and most successful approach in almost all areas of AI.

\section{Weaknesses of Current ML Systems: The I.I.D. Assumption}
\label{sec:weaknesses_of_current_ML}
Similar to statistical models, most ML approaches rely on the assumption that the examples from which they learn (the training data) are \textit{independent and identically distributed} (i.i.d.). 
ML models then aim to approximate the joint distribution of the data, or some property thereof,  such as a conditional expectation (``regression'') or a categorical conditional distribution (``classification'').
When sufficient data is available to train an expressive ML model, this  yields highly accurate predictions on new, unseen examples (the test data) from the \textit{same} distribution, as backed up by a rigorous mathematical theory of in-distribution generalization~\citep{vapnik1999nature}.
However, the assumption that the training and test data are i.i.d.\ is almost always violated in practice. 
Just to give a few examples, it might be that:
\begin{itemize}[]
	 \item the underlying system generating the data is not stable or stationary; 
	 \item the population on which the model is deployed differs from the training population; 
	 \item measurement devices, data collection practices, or environmental conditions change;
	 \item data points are not independent but 
	  linked, e.g., through a social or biological network.
\end{itemize}
For this reason, the i.i.d.\ assumption has also been referred to as ``the big lie of machine learning''~\citep{gharamani2017biglie}.

Due to their reliance on the i.i.d.\ assumption, ML models often lack \textit{robustness} and do not generalize well \textit{out-of-distribution} (OOD).  
This is exhibited by their reliance on \textit{spurious features}, typically simple patterns which, in the training data, are associated with a task and thus help make better predictions on i.i.d.\ data, but whose relationship with the variable of interest may change at test time. 
This strategy of relying on easy-to-learn but unstable features is also referred to as \textit{shortcut learning}~\citep{geirhos2020shortcuts}.
In image classification and object recognition, for example, background~\citep{beery2018recognition}, texture~\citep{geirhos2019imagenettrained}, pose~\citep{alcorn2019strike}, and context~\citep{rosenfeld2018elephant} are commonly used as spurious features.
Another case of lacking robustness is vulnerability to \textit{adversarial examples}, imperceptible input perturbations which induce erroneous predictions%
~\citep{szegedy2014intriguing,goodfellow14,ilyas2019adversarial}.
Models that rely on spurious associations instead of learning stable mechanisms can fail catastrophically when deployed in new conditions. 
This is particularly problematic when it comes to high-stakes contexts such as  medical applications.
\begin{figure}[t]
\centering
\includegraphics[width=0.725\textwidth]{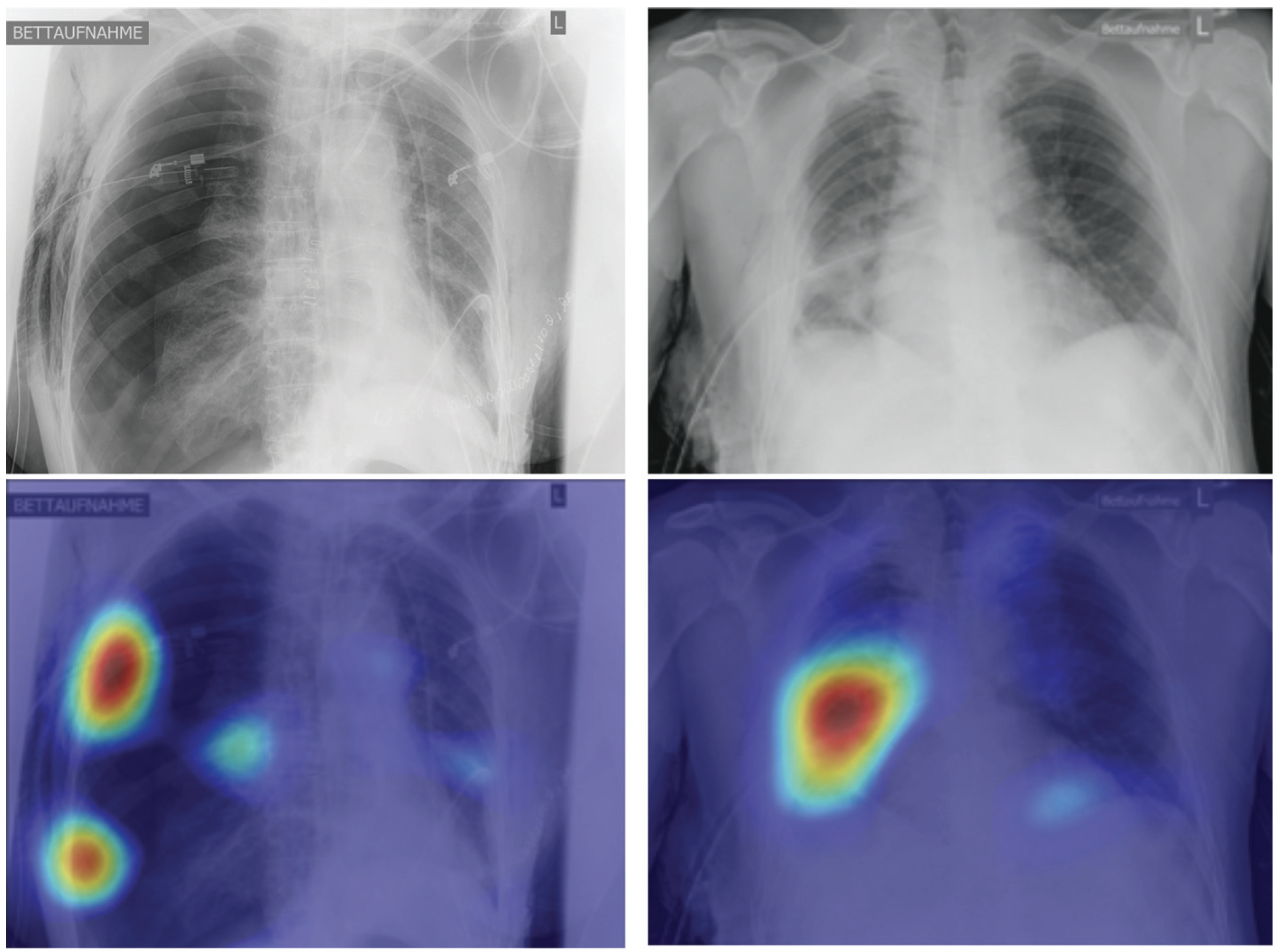}	
\caption[Thoracic Tubes as Spurious Features in Pneumothorax Detection]{\textbf{Thoracic Tubes as Spurious Features in Pneumothorax Detection.} This figure from~\cite{rueckel2020impact} shows two examples of chest X-ray scans (top) and the corresponding heatmaps (bottom) highlighting image regions that were used by their algorithm 
to make the respective predictions. 
The left image shows a case of pneumothorax; the right one does not. 
Yet, in both cases, the strongest feature activations correspond to the inserted thoracic tubes.
As a result of relying on thoracic tube detection as a shortcut,  performance
degrades drastically when evaluated on different patient populations, reaching chance level for distinguishing positive scans without tubes from negative scans with tubes.}
\label{fig:chest_xray}
\end{figure}

\begin{example}[Spurious Associations in Medical AI]
\label{ex:medical_AI}
Chest radiography is routinely used to screen for pulmonary disorders. With more than 2 billion yearly examinations world-wide, it is the most commonly performed medical imaging procedure~\citep{rueckel2020impact}.
Yet, its low sensitivity can complicate a correct interpretation by non-expert clinicians~\citep{raoof2012interpretation}, highlighting the potential for AI-based decision support systems~\citep{kallianos2019far}.
The publicly available ChestX-ray14 dataset~\citep{wang2017chestx} contains over $100$k frontal-view chest X-ray images (similar to~\cref{fig:chest_xray}, top) with annotations\footnote{automatically extracted from the text radiological reports via natural language processing techniques} for $14$ common chest diseases.
In a breakthrough announcement, a deep learning model trained on ChestX-ray14 was reported to detect many of these diseases as well as or better than 
expert radiologists~\citep{rajpurkar2017chexnet,rajpurkar2018deep}, as measured by average performance on held out examples from the \emph{same} dataset. 

A subsequent study by~\citet{rueckel2020impact} set out to analyse the behaviour and robustness of AI models for pneumothorax (i.e., collapsed lung) detection in more detail. 
They created a new benchmark, 
in which they additionally recorded age, sex, pneumothorax size, and the presence of \emph{thoracic tubes}, plastic tubes for draining fluid or air from the chest that are commonly inserted into patients with  pneumothorax. 
Using interpretability tools to inspect which image features most strongly influence the predictions, they found that models had actually learnt to detect and rely on the presence of the tubes as a shortcut, see~\cref{fig:chest_xray}. 
As expected, when evaluating their own model and the celebrated ChexNet model of~\cite{rajpurkar2017chexnet} on the new benchmark, performance varied heavily across different patient subgroups. 
In the most extreme case of classifying scans of pneumothorax-positive cases \emph{without} tubes against scans \emph{with} visible tubes but without any remaining signs of pneumothorax, performance degraded to that of random guessing~\citep[][Fig.~2A]{rueckel2020impact}. 

Clearly, such a system cannot be trusted to accurately and reliably detect the condition in yet-to-be-diagnosed patients, that is, before a thoracic tube is inserted. 
While a more careful curation of training data may help mitigate some of these issues, several other studies have reported similar cases of AI systems that achieve seemingly strong performance by relying on spurious features. 
In the context of predicting medical conditions from X-ray scans, other examples include the use of image borders and laterality markers\footnote{metal tokens placed on the X-rayed area to distinguish right from left or front from back; while not directly related to the presence of disease, these spurious features can help predict, e.g., the hospital in which the scans were taken, which might, in turn, contain information about the prevalence of the disease} in detecting pneumonia~\citep{zech2018variable} or COVID-19~\citep{degrave2021ai}. 
\end{example}

\Cref{ex:medical_AI} highlights that strongly predictive \textit{statistical} patterns can be unstable (and can become completely useless or even harmful) when parts of the data-generating process change.
The reliance of ML systems on the i.i.d.\ assumption and the resulting lack of robustness, therefore, poses serious challenges to building intelligent systems that---like biological forms of intelligence---can generalize to new domains, transfer knowledge across tasks,  learn continually over time, or plan and reason about the effect of changes and interventions.

\section{From Association to Causation}
\label{sec:from_correlation_to_causation}
\looseness-1 Many desirable properties of AI systems
such as robustness, modularity, and invariance are closely linked to \textit{causality}~\citep[][see~\cref{sec:ICM} for a more detailed account]{Pearl2009,simon1954spurious,aldrich1989autonomy}:
while statistical relations may change, the underlying causal mechanisms are thought to mostly remain stable when parts of the system are perturbed.
It has therefore been argued that, to reliably support tasks such as knowledge transfer and
OOD generalization, ML systems ought to take  
 causal structure into account~\citep{scholkopf2012causal}.
How does the underlying causal structure manifest itself in the data though, given that it is commonplace that ``correlation\footnote{or, more generally, association or statistical dependence} does not imply causation''?
The connection between observable associations and possible underlying causal structures can be summarised as follows.

\begin{principle}[Common Cause Principle~\citep{reichenbach1956direction}]
\label{princ:reichenbach}
If two random variables $X$ and~$Y$ are \emph{statistically} dependent ($X\not\independent Y$), then there exists a random variable $Z$ which \emph{causally} influences both of them and which explains all their dependence in the sense of rendering them conditionally independent ($X\independent Y\mid Z$).
As a special case, $Z$ may coincide with $X$ or $Y$.\footnote{In principle, there is another option: $X$ and $Y$ both causally influence $Z$, and we have conditioned on $Z$, which consitutes a form of selection bias. For example, when conditioning on the presence of a disease $Z$, two very rare causes $X$ and $Y$ that were originally independent become negatively correlated, since it is likely that either one or the other is present but, due to their rarity, not both.} 
\end{principle}

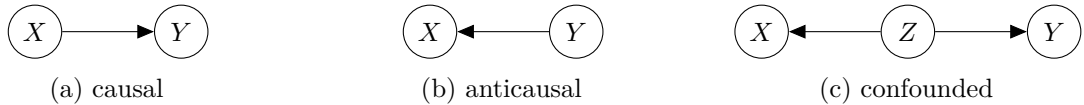
\begin{figure}[t]
	\newcommand{\xshift}{4em}
	\newcommand{\yshift}{3em}
	\newcommand{\nodescale}{1}
	\centering
	\begin{subfigure}{0.33\textwidth}
		\centering
		\begin{tikzpicture}
			\centering
			\node (X) [latent, scale=\nodescale] {$X$};
			\node (Y) [latent , xshift=1.25*\xshift, scale=\nodescale] {$Y$};
			\edge{X}{Y};
		\end{tikzpicture}
		\caption{causal}
		\label{fig:causal_learning}
	\end{subfigure}%
	\begin{subfigure}{0.33\textwidth}
		\centering
		\begin{tikzpicture}
			\centering
			\node (Y) [latent, scale=\nodescale] {$Y$};
			\node (X) [latent, xshift=-1.25*\xshift, scale=\nodescale] {$X$};
			\edge{Y}{X};
		\end{tikzpicture}
		\caption{anticausal}
		\label{fig:anticausal}
	\end{subfigure}
	\begin{subfigure}{0.33\textwidth}
		\centering
		\begin{tikzpicture}
			\centering
			\node (X) [latent, scale=\nodescale] {$X$};
			\node (Z) [latent, xshift=1.25*\xshift, scale=\nodescale] {$Z$};
			\node (Y) [latent, xshift=2.5*\xshift, scale=\nodescale] {$Y$};
			\edge{Z}{X,Y};
		\end{tikzpicture}
		\caption{confounded}
	\end{subfigure}
	\caption[Reichenbach's Common Cause Principle and Its Implications for ML]{\textbf{Reichenbach's Common Cause Principle and Its Implications for ML.} \Cref{princ:reichenbach} postulates three elementary\footnote{In principle, combinations of (a) or (b) with (c) are also possible.} possible causal explanations for statistical dependence between two observables $X$ and $Y$, shown as causal graphs in (a)--(c). In (a) the common cause $Z$ coincides with $X$, and in (b) it coincides with $Y$. 
		Thus, different causal structures can give rise to the same statistical properties. However, they behave differently under interventions, i.e., when parts of the system are manipulated.
		When the task is to predict $Y$ from $X$, only the causal learning task in (a) is stable under changes to the feature distribution~$P_X$, but the anticausal and confounded settings in (b) and (c) are not~\citep{scholkopf2012causal}.
	}
	\label{fig:reichenbach}
\end{figure}

\Cref{princ:reichenbach} characterizes statistical dependence as an epiphenomenon that always results from causal relationships through which variables, including potentially unobserved ones, influence each other.
Importantly, multiple causal structures can give rise to the same dependence (see~\cref{fig:reichenbach}), highlighting why statistical information alone is insufficient to infer causation. 

But what exactly do we mean by
``causal influences''?
While different causal graphs can induce the same dependence structures, 
they have different implications for what happens under \textit{interventions}, i.e., when some of the variables are \textit{externally manipulated}. 
Causal relationships, unlike statistical ones, are fundamentally directed and asymmetric. 
Changing a cause variable will typically also result in changes to its effects (since the mechanisms producing effects from causes remain invariant), but, conversely, changing an effect will leave its causes unaffected.
\begin{example}[Temperature $\to$ Thermometer]
	Consider two variables: room temperature and the reading of a thermometer placed inside the same room. Statistically speaking, the two variables are simply very closely correlated. At a causal level of description, there is additional structure, though. Changing the room temperature (e.g., by turning on the radiator)
	would induce a changed thermometer reading. However, changing the thermometer reading (e.g., by hacking
	  the device)
	would not meaningfully affect the room temperature. 
	We therefore say that temperature causally influences the thermometer reading and not vice versa.
\end{example}
These notions of intervention---controllability and manipulability---are central to most \textit{counterfactual} theories of causation~\citep{hume1748philosophical,woodward2005making,lewis1973causation}, which broadly state that ``a variable~$X$ is a cause of another variable $Y$ if 
there are  interventions on $X$ that would change (the distribution of) $Y$''. 
We will provide a more formal mathematical account in~\cref{sec:causal_modelling}.

Compared to statistical models, causal models are additionally endowed with explicit causal structure, typically in the form of a directed acyclic causal graph like those shown in~\cref{fig:reichenbach}.
The causal graph encodes information about how the system can be decomposed into independent modules, which in turn entails certain stability properties~\citep{peters2017elements}.
Specifically, the \textit{causal mechanisms that generate each variable from its parents} in the causal graph remain invariant, unless directly intervened upon.
In~\cref{ex:medical_AI}, the presence of thoracic tubes in the image ($X$) can be seen as an effect of the target variable pneumothorax ($Y$).\footnote{since inserting or removing a tube will not (immediately) influence the pneumothorax status, whereas the latter will often trigger the insertion of a thoracic tube}
In such an ``anti-causal'' learning task (see~\cref{fig:anticausal}), the 
direction of prediction does not align with the causal direction ($Y\to X$)%
~\citep{scholkopf2012causal}. 
As a result, an ML model trained to predict $Y$ from $X$ does not learn to approximate a causal mechanism (cf.\ the causal learning task in~\cref{fig:causal_learning}) and can easily become unstable when the input distribution changes.

\begin{figure}
\centering
\newcommand{\xshift}{2.5em}
\newcommand{\yshift}{1em}    
\begin{subfigure}[b]{0.275\textwidth}
    \centering
    \begin{tikzpicture}
        \centering
        \node (X_1) [obs] {$X_1$};
        \node (X_2) [det, below=of X_1, xshift=-\xshift, yshift=\yshift] {$x_2$};
        \node (Q) [const, above=of X_2, yshift=-2.5*\yshift] {$\EE[X_3|do(x_2)]$\textbf{?}};
        \node (X_3) [obs, below=of X_1, xshift=\xshift, yshift=\yshift] {$X_3$};
        \edge{X_1, X_2}{X_3};
    \end{tikzpicture}
    \caption{Causal Reasoning}
    \label{subfig:causal_reasoning}
\end{subfigure}%
\begin{subfigure}[b]{0.275\textwidth}
    \centering
    \begin{tikzpicture}
        \centering
        \node (X_1) [obs] {$X_1$};
        \node (X_2) [obs, below=of X_1, xshift=-\xshift, yshift=\yshift] {$X_2$};
        \node (X_3) [obs, below=of X_1, xshift=\xshift, yshift=\yshift] {$X_3$};
        \path[<->] (X_1) edge[dashed] node[xshift=-.5em, yshift=.5em] {\textbf{?}} (X_2);
        \path[<->] (X_1) edge[dashed] node[xshift=.5em, yshift=.5em] {\textbf{?}} (X_3);
        \path[<->] (X_2) edge[dashed] node[yshift=.5em] {\textbf{?}} (X_3);
    \end{tikzpicture} 
    \caption{Causal Discovery}
    \label{subfig:causal_discovery}
\end{subfigure}%
 \begin{subfigure}[b]{0.45\textwidth}
    \centering
    \begin{tikzpicture}
        \centering
        \node (X_1) [latent, dashed] {$Z_1$};
        \node (X_2) [latent, dashed, below=of X_1, xshift=-\xshift, yshift=\yshift] {$Z_2$};
        \node (X_3) [latent, dashed, below=of X_1, xshift=\xshift, yshift=\yshift] {$Z_3$};
        \edge[dashed]{X_1, X_2}{X_3};
        \edge[dashed]{X_1}{X_2};
        \plate[inner sep=0.1em,
    yshift=0.2em, dashed] {plate}{(X_1) (X_2) (X_3)}{};
    \node (x) [right=of plate,xshift=-0.25*\xshift]{\hspace{-.75em}\includegraphics[width=0.44\textwidth]{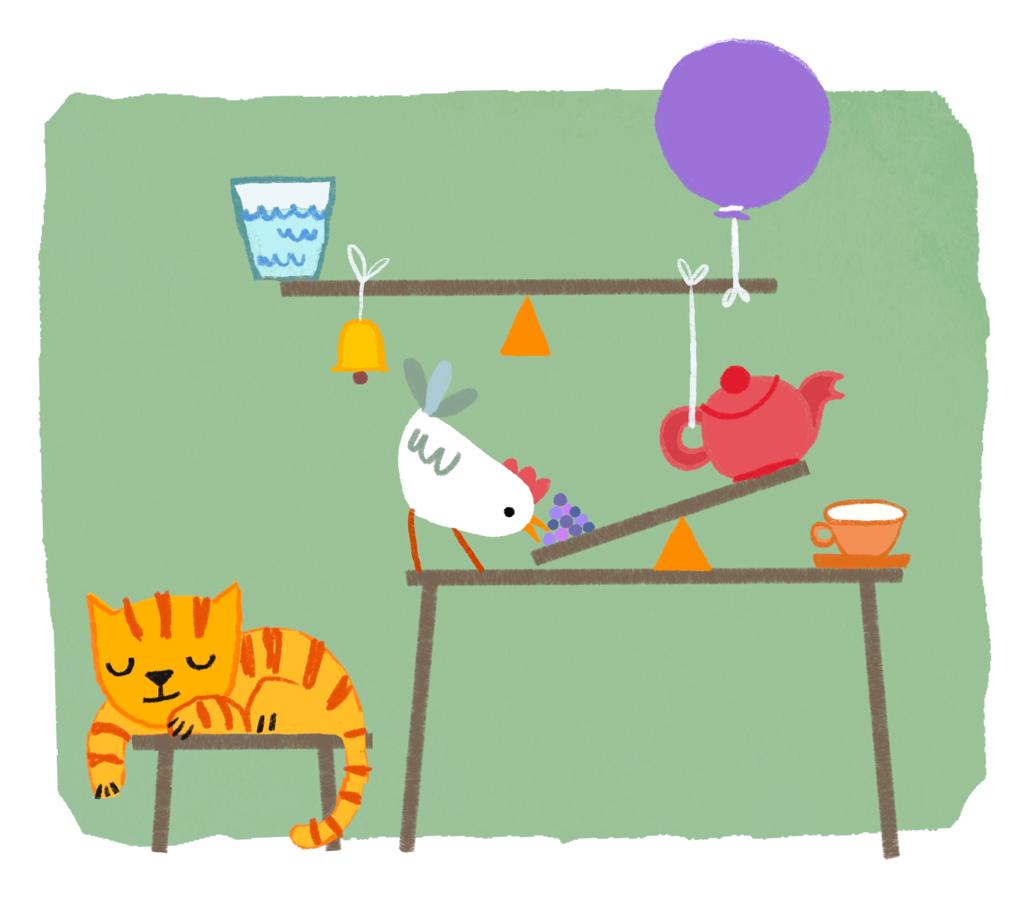}};
    \edge[dashed]{plate}{x};
    \node (f) [const, right=of plate, xshift=-0.75*\xshift, yshift=-\yshift] {$\fb$};
    \end{tikzpicture}
    \vspace{-.75em}
    \caption{Causal Representation Learning}
    \label{subfig:causal_rep_learning}
\end{subfigure}%
\caption[Overview of Different Causal Learning Tasks]{\looseness-1 \textbf{Overview of Different Causal Learning Tasks.} (a)~\textit{Causal reasoning} aims to answer interventional or counterfactual queries based on a (partial) causal model over observed variables~$X_i$. (b)~\textit{Causal discovery} aims to learn the causal graph connecting a set of \textit{observed} variables. (c)~\textit{Causal representation learning} (CRL) aims to infer a causal model consisting of a small number of \textit{unobserved} high-level, abstract causal variables~$Z_i$ and their relations from potentially very high-dimensional, low-level observations~$\Xb=\fb(\Zb)$. Illustration by Ana Mart\'in Larra\~naga.
}
\label{fig:causal_learning_tasks}
\end{figure}
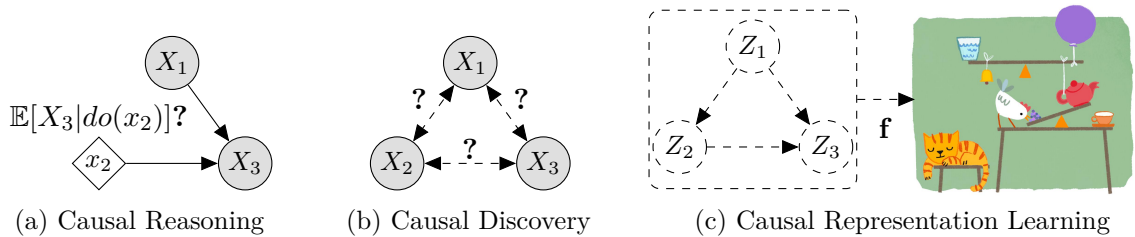

In addition to determining stability and invariance, a sound and complete set of graphical rules known as the ``do-calculus'' describes
how the causal mechanisms can be recombined and manipulated to compute the effect of interventions~\citep{pearl1995causal,huang2006pearl,shpitser2006identification,Pearl2009}, an example of causal reasoning (\cref{subfig:causal_reasoning}).
A causal model thus captures a whole set of distributions corresponding to all the different interventions on subcomponents of the system, and can also be viewed as a (potentially infinite) structured family of statistical models.
As such, causal models naturally provide a principled way of talking about distribution shifts~\citep{bareinboim2016causal,pearl2014external}.

\looseness-1 Since causal models rely on graphical representations and inference rules,
they do not fit the purely learning-based paradigm of ML and statistical modelling
driven by large amounts of i.i.d.\ data, 
but also share certain characteristics of
 \textit{symbolic} approaches.
As such, they inherit some of the weaknesses of symbolic AI discussed in~\cref{sec:history_of_AI}: they require human input to specify parts of the model and do not scale easily.
While many methods  for automatically inferring the causal graph from data have been proposed, so-called \textit{causal discovery} (\cref{subfig:causal_discovery}) is very challenging and only partially possible in general~\citep{spirtes2001causation}. 
The remaining ambiguities then need to be resolved by human domain experts or through carefully controlled experiments, which are often impractical or unethical. 
As the number of possible causal graphs grows super-exponentially in the number of variables~\citep{robinson1973counting}, search procedures or the construction of expert graphs can quickly become infeasible, even for a moderate number of variables.
Yet, for several settings in econometrics, social, and biomedical science, the number of relevant variables is limited, and the causal relations can reasonably confidently be specified by domain-experts. In many such cases, causal methodology has been successfully applied~\citep{angrist2009mostly,morgan2014counterfactuals,hernan2020causal,imbens2015causal}.

\section{From Causal Models to Causal Representations}
\label{sec:from_causal_models_to_causal_representations}
Aside from the difficulty of inferring causal structure, another key obstacle preventing a more widespread use of causal models in \textit{AI} is the requirement that the relevant causal variables---the \textit{symbols}---can be specified a priori, and that most (or at least some) of them are directly observed.
However, the data processed by modern AI, such as images, videos, speech, or text, is usually unstructured, very high-dimensional, and only provides a noisy view of more high-level underlying concepts, such as objects in a scene~\citep{locatello2020object}, or topics in a document~\citep{blei2003latent}. 
In the context of~\cref{ex:medical_AI}, AI systems have to operate on low-level input in the form of large matrices of numbers between zero and one (the grey-scale pixels). 
Relevant patterns such as thoracic tubes or laterality markers are not labelled or directly apparent, but first need to be discovered as groups of co-occurring pixels in different locations.
The assumption of access to a handful of observables on which interventions are meaningfully defined therefore poses serious challenges to the application of causal methodology to complex data such as raw pixels~\citep{LopNisChiSchBot17}.

Machine learning, on the other hand, has proven very successful at automatically extracting useful \textit{representations} of high-dimensional, low-level observations%
~\citep{bengio2013representation}. %
These representations are usually lower-dimensional, related to the observed data via some nonlinear (deterministic or probabilistic) map, and \text{distributed}, meaning that they take the form of vectors of continuous numbers~\citep{mikolov2013distributed,hinton1986distributed}. 
Unsupervised representation learning objectives tend to amount to different forms of \textit{compression}, for example, by feeding observations through a bottleneck and then reconstructing them as in \textit{autoencoders}~\citep{hinton2006reducing,hinton2006fast,kingma2013auto,vincent2008extracting}. 
Sometimes this is combined with an additional constraint that the inferred representation be \textit{disentangled}~\citep{bengio2013representation}, traditionally interpreted as consisting of statistically independent explanatory factors~\citep{higgins2017beta}, as in independent component analysis~\citep[]{hyvarinen2001independent}.
More recently, \textit{self-supervised} objectives such as predicting part of an observation~\citep{GPT3,baevski2020wav2vec} or jointly embedding similar views~\citep{chen2020simple,zbontar2021barlow} have become the dominant approach to representation learning. 
However, representations learned with such objectives from a single large dataset that is treated as i.i.d.\ only extract statistical information and do not make any of the underlying causal structure explicit.
ML models based on such \textit{statistical representations} therefore exhibit the same limitations discussed in~\cref{sec:weaknesses_of_current_ML} and do not support the causal machinery discussed in~\cref{sec:from_correlation_to_causation}.

\text{Causal representation learning} (CRL; \cref{subfig:causal_rep_learning}) aims to combine the core strengths of modern AI and causal inference: the representation learning capabilities of ML to make sense of high-dimensional data and the principled view on interventions and distribution shifts offered by causal models~\citep{scholkopf2021toward}.
Its core goal is to learn \textit{representations endowed with causal semantics} in that they can support causal notions such as \textit{planning} and \textit{reasoning under interventions}.
One approach is to view different components of a learnt representation as  variables in a causal model, for example, by augmenting them with symbolic structure such as a causal graph, that can be used to compute causal queries of interest.
Returning to~\cref{ex:medical_AI}, one inferred variable may correspond to pneumothorax and another to thoracic tubes. A causal representation should then capture that the latter is an effect of the former and not vice versa. 

A core difference to classical work on causal inference with unobserved variables, or hidden confounders, is that in CRL, \textit{none of the causal variables of interest are directly observed}.
In fact, the high-dimensional observations  are sometimes even not considered 
part of a causal model  at all, since intervening on them directly often does not seem plausible. 
 For example, it is unclear what the causal interpretation of intervening on individual pixels in an image should be. 

\section{Identifiability of Representations}
\label{sec:intro_identifiability}
One of the goals of unsupervised representation learning
  is to uncover properties of the true data generating process.
But how do we know whether a learnt representation (causal or not) is meaningfully related to the underlying ground truth?
One way to formally answer this question---and the central approach explored in this thesis---is through the lens of \textit{identifiability}~\citep{lehmann2006theory}.
Intuitively, identifiability guarantees that a learning task is well-specified, in that any two models that explain the data equally well are equivalent in a certain sense. 
As such, it can be viewed as a prerequisite for subsequent estimation procedures, data analysis, and interpretation. 
For a non-identifiable setting, two separately inferred representations can be equally good according to the learning objective but still be completely different. 
For example, identifiability is an important aspect of blind source separation in digital signal processing to ensure that a given split into separate signals
indeed recovers the original sources,
rather than mixtures thereof~\citep{comon1994independent,jutten1991blind}. 

 To formally study the identifiability of representations,
 a common approach---and the one pursued throughout this thesis---is to first specify how the observed data was generated from
 latent (unobserved) variables.
Subject to assumptions on the considered model class, we can then analyze whether representations that are solutions to a given learning objective can be shown to provably recover (parts of) the ground truth (up to pre-specified ambiguities). 
 If the answer is positive, the model class is said to be (weakly) identifiable.
Unlike statistical learning theory, such identifiability studies typically ignore finite sample issues and instead operate at the distribution level. 
They thus characterize under which circumstances aspects of the true data generating process can, at least in principle, be recovered given infinite data.

Already in classical causal inference~\citep{Pearl2009},  identifiability---for example, of the causal graph or of a certain treatment effect---is considered crucial to provide a mathematical justification for causal claims made on the basis of modelling assumptions and non-experimental data.
For similar reasons, it also plays a central role in the context of CRL since, arguably, it is  identifiability that substantiates any claims to a given representation actually being causal. 

For CRL, the data generating process takes the form of a latent variable model with additional causal structure (a ``latent causal model''). 
However, similar to causal discovery in the classical, fully-observed case~(see~\cref{princ:reichenbach} and~\cref{fig:causal_graphs}), learning representations from  i.i.d.\ data alone is very challenging. 
In fact, recovering the underlying latent variables is impossible for general nonlinear models, even in the absence of causal relations, that is, even when all latents are assumed to be independent~\citep{hyvarinen1999nonlinear,locatello2019challenging}.
To overcome this fundamental negative result and make progress towards identifying (causal) representations, we therefore need (i) \textit{additional constraints on the model class} or (ii) \textit{non-i.i.d.\ data}, which in CRL, is often linked to interventions in the latent causal model. 
This thesis explores both of these paths towards identifiability. 
\section{Overview and Main Contributions}
\label{sec:overview_and_contributions}

The organisation and main contributions of this thesis can be summarised as follows.
\Cref{chap:background} fist provides a self-contained review of relevant background material in the two fields which this thesis brings together: causal modelling and identifiable representation learning. 

\Cref{chap:IMA,chap:SSL_content_style,chap:CRL} then present the main original contributions in the form of identifiability studies for different representation learning settings (unsupervised, multi-view, and multi-environment).
Our results highlight different paths towards identifying causal representations, each requiring different assumptions on the underlying data generating process, see~\cref{tab:structure} for an 
 overview.

In \cref{chap:IMA}, we study \textit{unsupervised} representation learning from \textit{i.i.d.}\ data by proposing a new \textit{constraint on the class of mixing functions} mapping latent to observed variables.
Specifically, we focus on the task of nonlinear independent component analysis (ICA), which can be viewed as the simplest special case of CRL with an empty causal graph and jointly independent latents. 
Motivated by the principle of independent causal mechanisms, we posit that each latent source should independently influence the mixing process. 
We formalize this assumption as an orthogonality constraint on the partial derivatives of the mixing function (i.e., the columns of its Jacobian matrix), and refer to it as  \textit{independent mechanism analysis} (IMA).
The restricted IMA function class consists of {orthogonal coordinate transformations}, which contains  isometries and conformal maps are special cases, but is strictly larger. We show that it is invariant to permutation and element-wise reparametrisation of the sources, the fundamentally irresolvable ambiguities of nonlinear ICA.
We then provide theoretical and empirical evidence that IMA circumvents a number
of nonidentifiability issues by ruling out common spurious solutions.
In follow-up works, we connect IMA to the training objective of variational autoencoders
and leverage a similar functional constraint to identifiy object-centric representations.

In~\cref{chap:SSL_content_style}, we study \textit{multi-view} CRL from pairs of simultaneously observed \textit{non-independent} measurements. 
The considered setting is inspired by common practices in data augmentation via hand-crafted transformations intended to leave semantic aspects of the data intact.
We formalize a multi-view generative process as a latent causal model with ``\textit{content}'' variables that are invariant to augmentation and thus always shared, and ``\textit{style}'' variables that may change  across views.
When the style variables are non-ancestors of content variables, the views can then be interpreted as \textit{counterfactual} data pairs resulting from imperfect style interventions. 
We introduce the notion of \textit{block-identifiability} as the recovery of a group of variables up to an invertible mapping, and prove that generative or contrastive self-supervised learning algorithms block-identify the invariant content variables.
In this sense, our results lend some theoretical support for the empirically observed success of such methods. 
Further, we introduce Causal3DIdent, a semi-synthetic image dataset with dependent latents, and use it to study the effect of data augmentations performed in practice.
In follow-up works, we extend our analysis to different modalities, more than two views, and identifiability of individual style variables. 

\begin{table}[t]
    \centering
    \caption[Thesis Structure.]{\textbf{Thesis Structure.} Overview of the different settings studied in the main chapters of this thesis. PCH stands for the Pearl Causal Hierarchy~\citep{bareinboim2022pch}.}
    \label{tab:structure}
    \def\arraystretch{1.25}
    \resizebox{\textwidth}{!}{
    \begin{tabular}{l l m{4.5cm} m{2.7cm} m{4.5cm}}
    \toprule 
    \textbf{Chapter} & \textbf{Setting} & \textbf{Data} & \textbf{Layer in PCH} & \textbf{Main Reference}
    \\[.25em]
    \midrule 
    \Cref{chap:IMA}
    & Unsupervised & independent \& identically distributed (i.i.d.) & Observational (Layer 1) & \citet{gresele2021independent}
    \\[.75em]
    \hline
    \Cref{chap:SSL_content_style} 
    & Multi-View & {non-independent\newline  (paired data)} & Counterfactual (Layer 3) & \citet{von2021self}
    \\[.75em]
    \hline
    \Cref{chap:CRL} 
    & Multi-Environment & non-identically distributed (multiple datasets) & Interventional (Layer 2) & \citet{von2023nonparametric}
    \\[.75em]
    \bottomrule 
    \end{tabular}
    }
\end{table}

In \cref{chap:CRL}, we study \textit{multi-environment} CRL from multiple \textit{non-identically distributed} datasets, arising from unknown \textit{interventions} in an underlying latent causal model. 
Specifically, we consider a fully nonparametric setting in which both the mixing function and the latent causal model are completely unconstrained, thus complementing existing studies relying on parametric assumptions such as linearity.
Our goal is to identify both the ground truth latents and the unknown causal graph up to a set of ambiguities which we show to be irresolvable from interventional data alone.
For the fundamental setting of two causal variables, we prove that the observational distribution and one perfect intervention per node suffice for identifiability, subject to a genericity condition which rules out spurious solutions that involve fine-tuning of the intervened and observational distributions, mirroring similar conditions for nonlinear cause-effect inference.
For an arbitrary number of variables, we show that two distinct paired perfect interventions per node guarantee identifiability.
Further, we demonstrate that the strengths of causal influences among the latent variables are preserved by all equivalent solutions, rendering the inferred representation appropriate for drawing causal conclusions from new data. 
Finally, we discuss other multi-environment works in which we focus on  tasks such as causal component analysis, structure learning, or domain generalization.

\Cref{chap:conclusion} discusses our findings in a broader context. It highlights some works that have built on the contributions presented in this thesis, and lays out directions and open problems that future work on CRL might seek to address. 

\clearpage
\section{Publications Covered in This Thesis}
\label{sec:publications}
Below, an asterisk ($^*$) is used to indicate shared first authorship (equal contribution) and a dagger ($^\dagger$) to denote shared last authorship (joint supervision). 

\Cref{chap:background} is partly based on background material from the main publications underlying~\cref{chap:IMA,chap:SSL_content_style,chap:CRL} listed below, as well as on~\cite{scholkopf2022statistical}:
\begin{selfcitebox}
\href{https://arxiv.org/abs/2204.00607}{\ul{From statistical to causal learning}}
\\
Bernhard Sch\"olkopf$^*$ and \textbf{Julius von K\"ugelgen}$^*$ %
\\
\textit{Proceedings of the International Congress of Mathematicians}, 2022
\end{selfcitebox}

\Cref{chap:IMA} is \text{mainly} based on~\citet{gresele2021independent}:
\begin{selfcitebox}
\href{https://arxiv.org/abs/2106.05200}{\ul{Independent mechanism analysis, a new concept?}}
\\
Luigi Gresele$^*$, \textbf{Julius von K\"ugelgen}$^*$, Vincent Stimper, Bernhard Schölkopf, \\
{Michel Besserve} %
\\
\textit{Advances in Neural Information Processing Systems (NeurIPS)}, 2021
\end{selfcitebox}

It also discusses some material from~\citet{reizinger2022embrace} and \citet{Brady2023Provably}:
\begin{selfcitebox}
\href{https://arxiv.org/abs/2206.02416}{\ul{Embrace the gap: VAEs perform independent mechanism analysis}}
\\
Patrik Reizinger$^*$, Luigi Gresele$^*$, Jack Brady$^*$, \textbf{Julius von K\"ugelgen}, Dominik Zietlow, Bernhard Sch\"olkopf, Georg Martius, Wieland Brendel,  Michel Besserve
\\
\textit{Advances in Neural Information Processing Systems  (NeurIPS)}, 2022,
\end{selfcitebox}

\begin{selfcitebox}
\href{https://arxiv.org/abs/2305.14229}{\ul{Provably learning object-centric representations}}
\\
Jack Brady$^*$, Roland S.\ Zimmermann$^*$, Yash Sharma, Bernhard Sch\"olkopf, \\
\textbf{Julius von K\"ugelgen}$^\dagger$, Wieland Brendel$^\dagger$ %
\\
\textit{International Conference on Machine Learning (ICML)}, 2023
\end{selfcitebox}

\Cref{chap:SSL_content_style} is mainly based on \citet{von2021self}:
\begin{selfcitebox}
\href{https://arxiv.org/abs/2106.04619}{\ul{Self-supervised learning with data augmentations provably isolates content from style}}
\\
\textbf{Julius von K\"ugelgen}$^*$, Yash Sharma$^*$, Luigi Gresele$^*$, Wieland Brendel, \\
Bernhard Sch\"olkopf$^\dagger$, Michel Besserve$^\dagger$, Francesco Locatello$^\dagger$ %
\\
\textit{Advances in Neural Information Processing Systems (NeurIPS)}, 2021
\end{selfcitebox}

\newpage
It also discusses some material from \citet{eastwood2023self} and \citet{yao2023multiview}:
\begin{selfcitebox}
\href{https://arxiv.org/abs/2311.08815}{\ul{Self-supervised disentanglement	by leveraging structure in data augmentations}}
\\
Cian Eastwood, \textbf{Julius von K\"ugelgen}, Linus Ericsson, Diane Bouchacourt, Pascal Vincent, Mark Ibrahim, Bernhard Sch\"olkopf
\\
\textit{NeurIPS Workshop ``Causal Representation Learning'',} 2023
\end{selfcitebox}

\begin{selfcitebox}
\href{https://arxiv.org/abs/2311.04056}{\ul{Multi-view causal representation learning with partial observability}}
\\
Dingling Yao, Danru Xu, S\'ebastien Lachapelle, Sara Magliacane, Perouz Taslakian,\\ Georg Martius, \textbf{Julius von K\"ugelgen}, Francesco Locatello
\\
\textit{International Conference on Learning Representations (ICLR),} 2023
\end{selfcitebox}

\Cref{chap:CRL} is \text{mainly} based on~\citet{von2023nonparametric}:
\begin{selfcitebox}
\href{https://arxiv.org/abs/2306.00542}{\ul{Nonparametric identifiability of causal representations from unknown interventions}}
\\
\textbf{Julius von K\"ugelgen}, Michel Besserve, Wendong Liang, Luigi Gresele, Armin Keki\'c,\\ Elias Bareinboim, David M Blei, Bernhard Sch\"olkopf
\\
\textit{Advances in Neural Information Processing Systems (NeurIPS)}, 2023
\end{selfcitebox}

It also discusses some material from \citet{perry2022causal},  \citet{Liang2023cca}, and \citet{eastwood2022probable,eastwood2023spuriosity}:
\begin{selfcitebox}
\href{https://arxiv.org/abs/2206.02013}{\ul{Causal discovery in heterogeneous environments under the sparse mechanism shift hypothesis}}
\\
Ronan Perry, \textbf{Julius von K\"ugelgen}$^{\dagger}$, Bernhard Sch\"olkopf$^{\dagger}$ %
\\
\textit{Advances in Neural Information Processing Systems (NeurIPS)}, 2022
\end{selfcitebox}

\begin{selfcitebox}
\href{https://arxiv.org/abs/2305.17225}{\ul{Causal component analysis}}
\\
Liang Wendong, Armin Keki\'c, \textbf{Julius von K\"ugelgen}, Simon Buchholz, Michel Besserve, Luigi Gresele$^\dagger$, Bernhard Sch\"olkopf$^\dagger$
\\
\textit{Advances in Neural Information Processing Systems (NeurIPS)}, 2023
\end{selfcitebox}

\begin{selfcitebox}
\href{https://arxiv.org/abs/2207.09944}{\ul{Probable domain generalization via quantile risk minimization}}
\\
Cian Eastwood$^*$, Alexander Robey$^*$, Shashank Singh, \textbf{Julius von K\"ugelgen}, \\
Hamed Hassani, George J.\ Pappas, Bernhard Sch\"olkopf
\\
\textit{Advances in Neural Information Processing Systems (NeurIPS)}, 2022
\end{selfcitebox}

\begin{selfcitebox}
\href{https://arxiv.org/abs/2307.09933}{\ul{Spuriosity didn't kill the classifier: Using invariant predictions to harness spurious features}}
\\
Cian Eastwood$^*$, Shashank Singh$^*$, Andrei Liviu Nicolicioiu, Marin Vlastelica, \\ \textbf{Julius von K\"ugelgen}, Bernhard Sch\"olkopf
\\
\textit{Advances in Neural Information Processing Systems (NeurIPS)}, 2023
\end{selfcitebox}

\looseness-1 As (shared) first author of the underlying publications,
I contributed to the conceptualisation, theory, design of experiments, analysis of results, and writing of the main material presented in~\cref{chap:IMA,chap:SSL_content_style,chap:CRL}. The experiments presented in~\cref{sec:IMA_experiments} were implemented by Luigi Gresele and Vincent Stimper, those in~\cref{sec:ssl_experiments} by Yash Sharma, and those in~\cref{sec:experiments} by Armin Keki\'c---with input from the respective co-authors. 
I contributed to varying degrees to the other works discussed in~\cref{sec:extensions_ima,sec:multiview_extensions,sec:extensions_multi_env}, with the main contributions made by the respective (co-)first authors.

\section{Other Publications and Preprints}
\label{sec:other_publications}
During my PhD, I have contributed to other peer-reviewed publications, workshop papers, and preprints (listed chronologically below) which are not covered in this thesis. Most of them revolve around different topics at the intersection of causality and machine learning, including:
\begin{itemize}[itemsep=-.35em,topsep=.1em]
\item 
implications of the principle of independent causal mechanisms for domain adaptation~\citep{kugelgen2019semi}, semi-supervised learning~\citep{kugelgen2020semi}, and natural language processing~\citep{jin2021causal}; 
\item 
causal reasoning for algorithmic recourse~\citep{karimi2020algorithmic,karimi2022towards,von2021algorithmic}, fairness~\citep{von2022fairness}, and explainability~\citep{von2023backtracking,kladny2023deep}; 
\item identifiable (causal) representation learning on manifolds~\citep{ghosh2023independent}, or with masked~\citep{xu2023a} or support-constrained~\citep{ghosh2022pitfalls} latents;
\item 
Bayesian optimal experimental design for causal discovery and inference
~\citep{von2019optimal,toth2022active}; 
\item 
causal generative scene modelling and object-centric learning~\citep{von2020towards,tangemann2023unsupervised}; 
\item 
causal data analysis of Covid-19 case fatality rates~\citep{von2021simpson} and vaccination scenarios~\citep{kekic2023evaluating};
\item merging marginal causal models over overlapping variable 
sets~\citep{gresele2022causal}; and
\item statistically efficient treatment effect estimation from mixed data~\citep{kladny2023causal}.
\end{itemize}
Others are on topics are not directly related to causality, such as:
\begin{itemize}[itemsep=-.35em,topsep=.1em]
\item backward-compatible updating of predictions~\citep{trauble2021backward}; 
\item kernel (conditional) independence tests for functional data~\citep{laumann2021kernel,laumann2023kernel};
\item network analysis of the sustainable development goals~\citep{laumann2022complex}; 
\item evaluation of learnt representations~\citep{eastwood2022dcies,schott2022visual}; and
\item explainable trajectory prediction~\citep{makansi2022you}.  
\end{itemize}
\clearpage
\begin{selfcitebox}
\href{https://arxiv.org/abs/1807.07879}{\ul{Semi-generative modelling: covariate-shift adaptation with cause and effect features}}
\\
\textbf{Julius von K\"ugelgen}, Alexander Mey, Marco Loog
\\
\textit{International Conference on Artificial Intelligence and Statistics (AISTATS)}, 2019
\end{selfcitebox}

\begin{selfcitebox}
\href{https://arxiv.org/abs/1910.03962}{\ul{Optimal experimental design via Bayesian optimization: active causal structure learning for Gaussian process networks}}
\\
\textbf{Julius von K\"ugelgen}, Paul K Rubenstein, Bernhard Sch\"olkopf, Adrian Weller
\\
\textit{NeurIPS Workshop ``Do the Right Thing''},~2019
\end{selfcitebox}

\begin{selfcitebox}
\href{https://arxiv.org/abs/1905.12081}{\ul{Semi-supervised learning, causality, and the conditional cluster assumption}}
\\
\textbf{Julius von K\"ugelgen}, Alexander Mey, Marco Loog, Bernhard Sch\"olkopf
\\
\textit{Conference on  Uncertainty in Artificial Intelligence (UAI)}, 2020
\end{selfcitebox}

\begin{selfcitebox}
\href{https://arxiv.org/abs/2004.12906}{\ul{Towards Causal generative scene models via competition of experts}}
\\
\textbf{Julius von K\"ugelgen}$^*$, Ivan Ustyuzhaninov$^*$, Peter Gehler$^\dagger$, Matthias Bethge$^\dagger$, \\ Bernhard Sch\"olkopf$^\dagger$ %
\\
\textit{ICLR  Workshop ``Causal learning for decision making''}, 2020
\end{selfcitebox}

\begin{selfcitebox}
\href{https://arxiv.org/abs/2006.06831}{\ul{Algorithmic recourse under imperfect causal knowledge: a probabilistic approach}}
\\
Amir-Hossein Karimi$^*$, \textbf{Julius von K\"ugelgen}$^*$, Bernhard Sch\"olkopf, Isabel Valera %
\\
\textit{Advances in Neural Information Processing Systems (NeurIPS)}, 2020
\end{selfcitebox}

\begin{selfcitebox}
    \href{https://www.mdpi.com/2673-4591/5/1/31}{\ul{Kernel two-sample and independence tests for nonstationary random processes}}\\
    Felix Laumann, \textbf{Julius von K\"ugelgen}, Mauricio Barahona\\
    \textit{Engineering Proceedings}, 2021
\end{selfcitebox}

\begin{selfcitebox}
\href{https://arxiv.org/abs/2110.03618}{\ul{Causal direction of data collection matters: implications of causal and anticausal learning for~NLP}}\\
Zhijing Jin$^*$, \textbf{Julius von Kügelgen}$^*$, Jingwei Ni, Tejas Vaidhya, Ayush Kaushal, \\Mrinmaya Sachan, Bernhard Sch\"olkopf %
\\
\textit{Conference on Empirical Methods in Natural Language Processing (EMNLP),} 2021
\end{selfcitebox}

\begin{selfcitebox}
\href{https://arxiv.org/abs/2005.07180}{\ul{Simpson's paradox in Covid-19 case fatality rates: a mediation analysis of age-related causal effects}}
\\
\textbf{Julius von K\"ugelgen}$^*$, Luigi Gresele$^*$, Bernhard Sch\"olkopf %
\\
\textit{IEEE Transactions on Artificial Intelligence}, 2021
\end{selfcitebox}

\begin{selfcitebox}
\href{https://arxiv.org/abs/2106.11849}{\ul{Algorithmic recourse in partially and fully confounded settings through bounding\\ counterfactual effects}}
\\
\textbf{Julius von K\"ugelgen}, Nikita Agarwal, Jakob Zeitler, Afsaneh Mastouri, Bernhard Sch\"olkopf
\\
\textit{ICML Workshop ``Algorithmic Recourse''}, 2021
\end{selfcitebox}

\begin{selfcitebox}
\href{https://arxiv.org/abs/2107.01057}{\ul{Backward-compatible prediction updates: a probabilistic approach}}
\\
Frederik Träuble, \textbf{Julius von K\"ugelgen}, Matthäus Kleindessner, Francesco Locatello, Bernhard Sch\"olkopf, Peter Gehler
\\
\textit{Advances in Neural Information Processing Systems (NeurIPS)}, 2021
\end{selfcitebox}

\begin{selfcitebox}
\href{https://arxiv.org/abs/2107.08221}{\ul{Visual representation learning does not generalize strongly within the same domain}}
\\
Lukas Schott, \textbf{Julius von K\"ugelgen}, Frederik Träuble, Peter Gehler, Chris Russell, \\ Matthias Bethge, Bernhard Sch\"olkopf, Francesco Locatello$^\dagger$, Wieland Brendel$^\dagger$
\\
\textit{International Conference on Learning Representations (ICLR)}, 2022
\end{selfcitebox}

\begin{selfcitebox}
\href{https://arxiv.org/abs/2110.05304}{\ul{You mostly walk alone: analyzing feature attribution in trajectory prediction}}
\\
Osama Makansi, \textbf{Julius von K\"ugelgen}, Francesco Locatello, Peter Gehler, \\Dominik Janzing, Thomas Brox$^{\dagger}$, Bernhard Sch\"olkopf$^{\dagger}$
\\
\textit{International Conference on Learning Representations (ICRL)}, 2022
\end{selfcitebox}

\begin{selfcitebox}
\href{https://arxiv.org/abs/2202.06844}{\ul{On pitfalls of identifiability in unsupervised learning. \\
A note on: ``Desiderata for representation learning: A causal perspective''}}
\\
Shubhangi Ghosh, Luigi Gresele, \textbf{Julius von K\"ugelgen}, Michel Besserve, Bernhard Sch\"olkopf
\\
\textit{arXiv}, 2022
\end{selfcitebox}

\begin{selfcitebox}
\href{https://www.thelancet.com/journals/lanplh/article/PIIS2542-5196(22)00070-5/fulltext}{\ul{Complex interlinkages, key objectives, and nexuses among the Sustainable Development Goals and climate change: a network analysis}}
\\
Felix Laumann, \textbf{Julius von K\"ugelgen}, Thiago Uehara, Mauricio Barahona
\\
\textit{The Lancet Planetary Health}, 2022
\end{selfcitebox}

\begin{selfcitebox}
\href{https://arxiv.org/abs/2010.06529}{\ul{On the fairness of causal algorithmic recourse}}
\\
\textbf{Julius von K\"ugelgen}, Amir-Hossein Karimi, Umang Bhatt, Isabel Valera, Adrian Weller, Bernhard Sch\"olkopf
\\
\textit{Proceedings of the AAAI Conference on Artificial Intelligence}, 2022
\end{selfcitebox}

\begin{selfcitebox}
\href{https://arxiv.org/abs/2202.01300}{\ul{Causal inference through the structural causal marginal problem}}
\\
Luigi Gresele$^*$, \textbf{Julius von K\"ugelgen}$^*$, Jonas K\"ubler$^*$, Elke Kirschbaum, \\Bernhard Sch\"olkopf, Dominik Janzing %
\\
\textit{International Conference on Machine Learning (ICML)}, 2022
\end{selfcitebox}

\begin{selfcitebox}
\href{https://link.springer.com/chapter/10.1007/978-3-031-04083-2_8}{\ul{Towards causal algorithmic recourse}}
\\
Amir-Hossein Karimi$^*$, \textbf{Julius von K\"ugelgen}$^*$, Bernhard Sch\"olkopf, Isabel Valera %
\\
\textit{xxAI---Beyond Explainable AI (Springer Lecture Notes in Computer Science vol.\ 13200)}, 2022
\end{selfcitebox}

\begin{selfcitebox}
\href{https://arxiv.org/abs/2206.02063}{\ul{Active Bayesian causal inference}}
\\
Christian Toth, Lars Lorch, Christian Knoll, Andreas Krause, Franz Pernkopf, \\Robert Peharz$^{\dagger}$, \textbf{Julius von K\"ugelgen}$^{\dagger}$ %
\\
\textit{Advances in Neural Information Processing Systems (NeurIPS)}, 2022
\end{selfcitebox}

\begin{selfcitebox}
\href{https://arxiv.org/abs/2212.08498}{\ul{Evaluating vaccine allocation strategies using simulation-assisted causal modeling}}
\\
Armin Keki\'c, Jonas Dehning, Luigi Gresele, \textbf{Julius von K\"ugelgen}, Viola Priesemann$^{\dagger}$, Bernhard Sch\"olkopf$^{\dagger}$
\\
\textit{Patterns}, 2023
\end{selfcitebox}

\begin{selfcitebox}
\href{https://arxiv.org/abs/2211.00472}{\ul{Backtracking counterfactuals}}
\\
\textbf{Julius von K\"ugelgen}, Abdirisak Mohamed, Sander Beckers
\\
\textit{Conference on Causal Learning and Reasoning (CLeaR)}, 2023
\end{selfcitebox}

\begin{selfcitebox}
\href{https://arxiv.org/abs/2110.06562}{\ul{Unsupervised object learning via common fate}}
\\
Matthias Tangemann, Steffen Schneider, \textbf{Julius von K\"ugelgen}, Francesco Locatello, \\Peter Gehler, Thomas Brox, Matthias K\"ummerer$^{\dagger}$, Matthias Bethge$^{\dagger}$, Bernhard Sch\"olkopf$^{\dagger}$
\\
\textit{Conference on Causal Learning and Reasoning (CLeaR)}, 2023
\end{selfcitebox}

\begin{selfcitebox}
\href{https://openreview.net/forum?id=xyz123}{\ul{DCI-ES: an extended disentanglement framework with connections to identifiability}}
\\
Cian Eastwood$^*$, Andrei Liviu Nicolicioiu$^*$, \textbf{Julius von K\"ugelgen}$^*$, Armin Keki\'c, \\ Frederik Tr\"auble, Andrea Dittadi, Bernhard Sch\"olkopf %
\\
\textit{International Conference on Learning Representations (ICLR)}, 2023
\end{selfcitebox}

\begin{selfcitebox}
\href{https://arxiv.org/abs/2306.06002}{\ul{Causal effect estimation from observational and interventional data through matrix weighted linear estimators}}
\\
Klaus-Rudolf Kladny, \textbf{Julius von K\"ugelgen}, Bernhard Sch\"olkopf, Michael Muehlebach
\\
\textit{Conference on Uncertainty in Artificial Intelligence (UAI)}, 2023
\end{selfcitebox}

\begin{selfcitebox}
\href{https://arxiv.org/abs/2310.07665}{\ul{Deep backtracking counterfactuals for causally compliant explanations}}
\\
Klaus-Rudolf Kladny, \textbf{Julius von K\"ugelgen}, Bernhard Sch\"olkopf, Michael Muehlebach
\\
\textit{Preprint}, 2023
\end{selfcitebox}

\begin{selfcitebox}
\href{https://www.preprints.org/manuscript/202310.1816/v1}{\ul{Kernel-based independence tests for causal structure learning on functional data}}\\
Felix Laumann, \textbf{Julius von K\"ugelgen}, Junhyung Park, Bernhard Sch\"olkopf, \\Mauricio Barahona
\\
\textit{Entropy}, 2023
\end{selfcitebox}

\begin{selfcitebox}
\href{https://openreview.net/forum?id=Whr6uobelR}{\ul{A sparsity principle for partially observable causal representation learning}}\\
Danru Xu, Dingling Yao, S\'ebastien Lachapelle, Perouz Taslakian, \textbf{Julius von K\"ugelgen}, Francesco Locatello, Sara Magliacane
\\
\textit{NeurIPS Workshop ``Causal Representation Learning''}, 2023
\end{selfcitebox}

\begin{selfcitebox}
\href{https://arxiv.org/abs/2312.13438}{\ul{Independent mechanism analysis and the manifold hypothesis}}\\
Shubhangi Ghosh, Luigi Gresele, \textbf{Julius von K\"ugelgen}, Michel Besserve, Bernhard Sch\"olkopf
\\
\textit{NeurIPS Workshop ``Causal Representation Learning''}, 2023
\end{selfcitebox}

%% file: Chapter2/chapter2.tex
\graphicspath{{Chapter2/Figs/}}

\chapter{Background}
\label{chap:background}
This chapter formally introduces %
relevant background material that the rest of the thesis will build upon. This mainly comprises two fields: causal modelling~(\cref{sec:causal_modelling}) and identifiable representation learning~(\cref{sec:identifiable_rep_learning}).

\section[Causal Modelling]{Causal Modelling}
\label{sec:causal_modelling}

The starting point for our formal treatment of causality is the \textit{structural causal model} (SCM)\footnote{SCMs are also referred to as functional causal models or \textit{non-parametric} structural equation models (SEMs) to emphasise the difference to  the often linear SEMs used in econometrics.} framework of~\citet{Pearl2009}, which has roots in path analysis~\citep{wright1920relative,wright1921correlation,wright1934method} and econometrics~\citep{haavelmo1944probability}.

\subsection{Structural Causal Models}
An SCM specifies a generative process for so-called \textit{endogenous} random variables $\Vb=(V_1, ..., V_n)^\intercal$.
This process captures the causal relationships among the $V_i$s through functional mechanisms that describe how each $V_i$ is determined by its direct causes and so-called \textit{exogenous} random variables $\Ub=(U_1, ..., U_n)^\intercal$ which are determined by factors outside the model.

\begin{definition}[SCM]
	\label{def:SCM}
	An SCM $\Mcal=(\Scal,P_\Ub)$ consists of (i) a set of assignments, or \emph{structural equations}, 
	\begin{equation}
		\label{eq:structural_equations}
		\Scal=\{V_i := f_i (\PA_i, U_i)\}_{i=1}^n
	\end{equation}
	\looseness-1 where $\PA_i\subseteq\Vb\setminus\{V_i\}$ are the direct causes, or \emph{causal parents}, of $V_i$, and $f_i$ are deterministic functions;
	and (ii) a %
	joint distribution $P_\Ub$ over the exogenous variables.
\end{definition}

The paradigm of SCMs views the processes $f_i$ by which each $V_i$ is generated from others as (abstractions or coarse-grainings of) physical mechanisms.
All randomness comes from the exogenous variables $U_i$ which capture possible stochasticity of the process, as well as uncertainty due to unmeasured parts of the system.

The assignment symbol ``$:=$'' is used instead of an equality sign to indicate that  the LHS quantity is defined to take on the RHS value, reflecting the asymmetry of causal relationships.
In particular, one cannot simply rewrite a structural equation $V_2:=V_1+U_2$ as $V_1=V_2-U_2$, as would be the case for a standard equation.

In principle, each of the $V_i$s and $U_i$s may be multi-dimensional, but they are often assumed to be real-valued scalar random variables.

Associated with each SCM is a \textbf{causal graph} $G$, which provides a compact representation of the qualitative causal relationships encoded by the model.

\begin{definition}[Causal graph]
	\label{def:causal_graph}
	The causal graph $G$ induced by $\Mcal$ is a directed graph with vertices~$\Vb$ and edges $V_j\to V_i$ if and only if $V_j\in\PA_i$.
\end{definition}

\Cref{def:SCM} allows for a rich class of causal models, including ones with cyclic causal relations.%
\footnote{See, e.g., \citet{bongers2021foundations} for a treatment of cyclic causal models.}
Throughout this thesis, unless explicitly stated otherwise, we will make the following common simplifying assumption.

\begin{assumption}[Acyclicity]
	\label{ass:acyclicity}
	The causal graph $G$ does not contain cycles, i.e., it is a directed acyclic graph (DAG).
\end{assumption}

\Cref{ass:acyclicity} guarantees\footnote{Acyclicity is a sufficient, but not a necessary condition.} that the SCM induces a unique distribution $P(\Vb)$ over the endogenous variables $\Vb$ through the exogenous distribution $P(\Ub)$ and the structural equations $\Scal$. We can draw from this distribution via \textit{ancestral sampling}:
first, we draw a realization of the exogenous variables from $P(\Ub)$, and then we iteratively compute the values of the $V_i$s according to~\eqref{eq:structural_equations} in (partial) topological order of the causal graph (i.e., starting at a root node of the graph), substituting previously computed $V_i$s into the structural equations where necessary.

Recursive substitution of the $V_i$s on the RHS of~\eqref{eq:structural_equations} in the same order also yields an expression for $\Vb$ directly in terms of the exogenous variables $\Ub$ as
\begin{equation}
	\label{eq:reduced_form}
	\Vb=\fb_\textsc{rf}(\Ub)\,.
\end{equation}
This mapping $\fb_\textsc{rf}$ is called the \textbf{reduced form} (or solution function) of the SCM.
Through~\eqref{eq:reduced_form}, the induced distribution $P(\Vb)$ can be formally defined as follows.

\begin{definition}[Distribution induced by an acyclic SCM]
	The distribution $P(\Vb)$ over $\Vb$ induced by an (acyclic) SCM is given by the push-forward of the exogenous distribution through the structural equations,
	\begin{equation}
		P_\Vb=\fb_{\textsc{rf}*}(P_\Ub)\,.
	\end{equation}
\end{definition}

\begin{figure}[t]
	\newcommand{\xshift}{4em}
	\newcommand{\yshift}{3em}
	\newcommand{\nodescale}{0.85}
	\centering
	\begin{subfigure}{0.5\textwidth}
		\centering
		\begin{tikzpicture}
			\centering
			\node (X_1) [latent] {$V_1$};
			\node (X_2) [latent, yshift=-\yshift, xshift=-0.5*\xshift] {$V_2$};
			\node (X_3) [latent, yshift=-\yshift, xshift=0.5*\xshift] {$V_3$};
			\edge {X_1} {X_2,X_3};
			\edge {X_2} {X_3};
		\end{tikzpicture} 
		\caption{}
		\label{fig:ex_causal_graph}
	\end{subfigure}%
	\begin{subfigure}{0.5\textwidth}
		\centering
		\begin{tikzpicture}
			\centering
			\node (X_1) [latent] {$V_1$};
			\node (X_2) [det, yshift=-\yshift, xshift=-0.5*\xshift] {$V_2$};
			\node (X_3) [latent, yshift=-\yshift, xshift=0.5*\xshift] {$V_3$};
			\edge {X_1} {X_3};
			\edge {X_2} {X_3};
		\end{tikzpicture} 
		\caption{}
		\label{fig:ex_post_intervention_graph}
	\end{subfigure}
	\caption[Causal Graphs]{\textbf{Causal Graphs.} (a)~Causal graph $G$ for the SCM from~\cref{ex:simple_SCM}. (b)~Post-intervention graph after a perfect or hard intervention $do(V_2=v_2)$.}
	\label{fig:causal_graphs}
\end{figure}
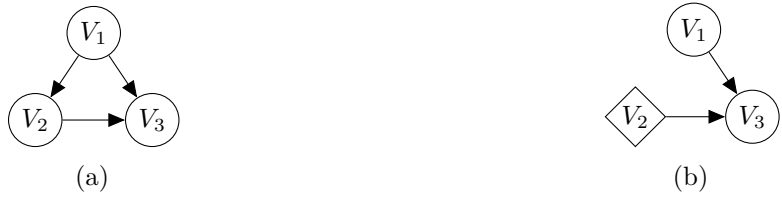

\begin{example}
	\label{ex:simple_SCM}
	Consider the following linear SCM $\Mcal$ over endogenous variables $\Vb=(V_1,V_2, V_3)$ and exogenous variables $\Ub=(U_1,U_2,U_3)$ with structural equations
	\begin{equation}
		\label{eq:ex_full_scm}
		\begin{aligned}
			V_1&:=U_1,\\
			V_2&:=V_1+U_2,\\
			V_3&:=V_1+V_2+U_3,
		\end{aligned}
	\end{equation}
	and exogenous distribution $P(\Ub)=\Ncal(\mathbf{0},\Ib_3)$.
	The causal graph $G$ induced by $\Mcal$ is shown in~\cref{fig:ex_causal_graph}.
	The reduced form $\Vb=\fb_\textsc{rf}(\Ub)$ 
	for~\eqref{eq:ex_full_scm} is given by
	\begin{equation}
		\label{eq:ex_rf_scm_matrix}
		\Vb=
		\begin{pmatrix}
			V_1\\V_2\\V_3
		\end{pmatrix}
		=
		\begin{pmatrix}
			1 & 0 & 0 \\1 & 1 & 0\\2 & 1 & 1
		\end{pmatrix}
		\begin{pmatrix}
			U_1\\U_2\\U_3
		\end{pmatrix}
		=:\Ab \Ub.
	\end{equation}
	The induced distribution is therefore given by $P_\Vb=\Ab_*(\Ncal(\mathbf{0},\Ib))=\Ncal(\mathbf{0},\Ab\Ab^\intercal)$.
\end{example}

\paragraph{Terminology.} 
The endogenous variables $\Vb$ which appear on the LHS of the structural equations in~\eqref{eq:structural_equations} are often assumed to be fully observed, while all unobserved variables are treated as exogenous.
Since the focus of this thesis is on causal representation learning, which involves talking about causal structure among latent (unobserved) variables, we will instead consider endogenous variables that can be either \textit{observed} or \textit{latent}. We will typically refer to observed endogenous variables as $\Xb$ and to latent endogenous variables as $\Zb$. To avoid potential confusion with the traditional view of endogenous variables as observed, we will also refer to $\Vb=(\Zb,\Xb)$ as \text{causal variables}. 
Moreover, since we allow for latent variables $\Zb$ to be part of the model, we will assume that the exogenous variables $\Ub$ are mutually independent (as in~\cref{ex:simple_SCM}) and will also refer to them as \text{noise variables} (sometimes denoted $N$). 

\subsection{Causal Sufficiency and Markovianity}
Throughout this thesis, unless explicitly stated otherwise, we will make the following assumption.  

\begin{assumption}[Independent noises]
	\label{ass:independent_noise}
	The $U_i$s are mutually independent, i.e., their joint distribution is fully factorised with density given by: 
	\begin{equation}
		\label{eq:factorised_noise_distribution}
		p(\Ub)=\prod_{i=1}^n p_i(U_i)
	\end{equation}
	for some suitable marginal densities $p_i$.
\end{assumption}

\Cref{ass:independent_noise} implies that any variable that directly influences two or more of the causal variables, called a \textbf{confounder}, is included in the model, for otherwise this would constitute a dependence between some $U_i$s (which account for any exogenous influences). 
In this case, the system is also called \textbf{causally sufficient}.

\begin{definition}[Causal sufficiency]
	\label{def:causal_sufficiency}
	A set of variables $\Wb$ is \emph{causally sufficient} if any direct common cause $C$ of two  (or more) variables in $\Wb$ is also in $\Wb$, that is, if the following implication holds for any $C$ and any $W_1 \neq W_2\in\Wb$:
	\begin{equation}
		(C\to W_1) \land (C\to W_2) \quad  \implies \quad  C\in \Wb.
	\end{equation}
\end{definition}
The distribution $P_\Vb$ induced by an SCM then satisfies the following \textbf{causal Markov condition} w.r.t.\ the induced causal graph~$G$~\citep[][Thm.~1.4.1]{Pearl2009}.

\begin{definition}[Causal Markov condition; \citet{kiiveri_speed_carlin_1984}]
	\label{def:causal_markov_condition}
	A distribution $P_\Vb$ over variables $\Vb$ satisfies the \emph{causal Markov condition} w.r.t.\ a causal graph $G$ if every $V_i\in \Vb$ is conditionally independent of its non-descendants in $G$, given its parents in $G$:
	\begin{equation}
		V_i \independent \Nb\DE^G_i ~|~ \PA^G_i\,.
	\end{equation}
\end{definition}

\Cref{def:causal_markov_condition} is also known as the \textit{local} Markov property.
It is equivalent~\citep[][Thm.~3.27]{lauritzen1996graphical} to saying that the induced distribution is Markovian w.r.t.\ $G$, meaning that its density obeys the following \textbf{causal Markov factorization}:
\begin{equation}
	\label{eq:causal_Markov_factorisation}
	p(\Vb)=\prod_{i=1}^n p(V_i~|~\PA^G_i)
\end{equation}

\begin{remark}
	Even if $\Vb$ is causally sufficient and $P_\Vb$ obeys the causal Markov factorization, this need not hold for subsets $\Wb\subset \Vb$. In particular, if $\Vb=(\Zb,\Xb)$, the observed distribution $P_\Xb$ will typically \emph{not} possess a simple factorisation like~\eqref{eq:causal_Markov_factorisation} due to latent confounders in~$\Zb$.
\end{remark}

\subsection{Independence of Causal Mechanisms, Modularity, and Invariance}
\label{sec:ICM}
By the chain rule, we have that
\begin{equation}
	\label{eq:non_causal_factorization}
	p(\Vb)=\prod_{i=1}^n p\left(V_{\pi(i)}~|~V_{\pi(1)}, ..., V_{\pi(i-1)}\right)\,.    
\end{equation}
for any permutation $\pi$ of $\{1,...,n\}$.
Hence, according to~\cref{def:causal_markov_condition}, any distribution $P_\Vb$ satisfies the causal Markov condition w.r.t.\ any complete graph~$G$.
What is it then that sets the causal Markov factorization in~\eqref{eq:causal_Markov_factorisation} apart from other \textbf{non-causal factorizations} like those in~\eqref{eq:non_causal_factorization}?
A key conceptual difference is that, as probabilistic analogues of the assignments $V_i:=f_i(\PA_i,N_i)$, the causal Markov kernels $p(V_i~|~\PA_i)$ typically capture \textit{independent physical processes}, rather than being mere mathematical objects like the conditionals in~\eqref{eq:non_causal_factorization}.
This ``independence of causal mechanisms'' (ICM) can be summarized as follows.

\begin{principle}[ICM principle; \cite{peters2017elements}]
	\label{principle:ICM}
	The causal generative process of a system's variables is composed of autonomous modules that do not inform or influence each other. 
	In the probabilistic case, this means that the conditional distribution of each
	variable given its causes (i.e., its mechanism) 
	does not inform or influence the
	other conditional distributions. 
\end{principle}%

Crucially, independence in the sense of ICM does not refer to \textit{statistical} independence of random variables, but rather to independence of the underlying mechanisms, distributions, or structural equations considered as algorithmic objects.

The first implication of ICM concerns the \textbf{lack of shared information}, which is best understood through the lens of \textit{compression}. Two mechanisms are considered independent if encoding them jointly does not admit a shorter description than encoding them separately. One formalization of this idea relies on Kolmogorov complexity as a measure of algorithmic information~\citep{janzing2010causal}.
Since Kolmogorov complexity is not computable, several alternative criteria have been proposed ~\citep{besserve2018group,janzing2021causal,guo2022causal,marx2017telling}, often relying on additional assumptions such as linear or deterministic relations~\citep{janzing2010telling,zscheischler2011testing,daniuvsis2010inferring,janzing2012information,janzing2015justifying,shajarisales2015telling}.
One aspect of no information sharing is the independence of the noise terms $U_i$ (\cref{ass:independent_noise}), which forms the basis for several constraint- and regression-based causal discovery methods~\citep{peters2014causal,peters2014identifiability,spirtes2001causation,shimizu2006linear,hoyer2008nonlinear,mooij2016distinguishing}.

The second implication of ICM concerns the \textbf{lack of shared influences}. This property is also known as \textbf{modularity}~\citep{Pearl2009} and is best understood through the lens of \textit{manipulation} or \textit{intervention}. Since the causal mechanisms are autonomous modules, 
changing one of the $p(V_i~|~\PA_i)$ will not affect the other causal conditionals $p(V_j~|~\PA_j)$ on the RHS of~\eqref{eq:causal_Markov_factorisation} but will leave the rest of the system \textbf{invariant}~\citep{aldrich1989autonomy,frisch1948autonomy,marschak1950statistical,simon1953causal,hurwicz1966structural,hoover2008causality}.
In analogy, swapping one machine in a production line will leave the other machines unaffected. 
The same will generally not be true for changing a non-causal conditional in the factorization in~\eqref{eq:non_causal_factorization}, which typically does not correspond to a meaningful intervention on a subpart of a causal system.

\subsection{Interventions}
Interventions refer to external manipulations or changes that modify parts of a causal system.
To emphasise the difference between such external manipulation and more conventional conditioning, interventions are typically denoted using the $do(\cdot)$ operator~\citep{Pearl2009}. 
For example, an intervention that fixes the value of one of the $V_i$ to some constant $v_i$ is denoted $do(V_i=v_i)$.
Due to the assignment character of structural equations, interventions are naturally expressed within the SCM framework.
To model an intervention, one simply replaces the corresponding structural equations, which gives rise to a new interventional SCM.

\begin{definition}[Interventional SCM]
	\label{def:intervention}
	Let $\Mcal=(\Scal,P_\Ub)$ be an SCM (\cref{def:SCM}). The interventional SCM arising from an intervention $do(\Wb=\wb)$ on a subset of variables $\Wb\subseteq \Vb$ is given by $\Mcal^{do(\Wb=\wb)}=(\Scal^{do(\Wb=\wb)},P_\Ub)$, where $\Scal^{do(\Wb=\wb)}$ is obtained by replacing the corresponding structural equations in $\Scal$:
	\begin{equation}
		\label{eq:intervened_structural_equations}
		\Scal^{do(\Wb=\wb)}=\{\Wb:=\wb\} \,\cup\, \{V_i:=f_i(\PA_i, U_i)\}_{V_i\in \Vb\setminus \Wb}\,. 
	\end{equation}
	The interventional distribution~$P(\Vb~|~do(\Wb=\wb))$ and the post-intervention graph~$G'$ are those induced by $\mathcal{M}^{do(\Wb=\wb)}$.
\end{definition}
For such \textit{hard} interventions, which force a subset of the variables to constants, the interventional distribution (in the Markovian case) is given in terms of the original causal mechanisms by the \textit{g-computation formula}~\citep{robins1986new}, also known as the \textit{manipulation theorem}~\citep{spirtes2001causation} or \textit{truncated factorization}~\citep{Pearl2009}:
\begin{equation}
	\label{eq:truncated_factorization}
	p(\Vb~|~do(\Wb=\wb))=\delta(\Wb=\wb)\prod_{V_i\in\Vb\setminus \Wb} p(V_i~|~\PA_i)\,,
\end{equation}
where $\delta(\cdot)$ denotes the Dirac delta distribution.

As discussed in the context of modularity and the ICM principle, all non-intervened mechanisms remain unchanged in~\eqref{eq:intervened_structural_equations} and~\eqref{eq:truncated_factorization}.

Since hard interventions remove all influences from the original causal parents, the post-intervention graph is obtained from the original graph by simply removing all arrows pointing into intervened-upon variables, a procedure known as \textit{graph surgery}~\citep{spirtes2001causation}, see~\cref{fig:ex_post_intervention_graph} for an example.

\paragraph{Soft Interventions.} Rather than setting variables to constants through hard interventions, one can also consider more general \textit{soft} interventions~\citep{dawid2002influence,eberhardt2007interventions}, which enforce new assignments $do(V_j=\tilde f_j(\widetilde \PA_j, \tilde U_j))$, giving rise to new mechanisms $\tilde p(V_j~|~\widetilde\PA_j)$. Typically, this is done under the constraint that the new parent set $\widetilde \PA_j\subseteq \Vb\setminus \{V_j\}$ does not introduce cycles, and that the new noise variable $\tilde U_j$ is mutually independent of all other noises.
In this case, the interventional distribution $\tilde p(\Vb)$ is given by
\begin{equation}
	\tilde p(\Vb)= \tilde p(V_j~|~\widetilde\PA_j) \, \prod_{i\neq j} p(V_i~|~\PA_i)\,.
\end{equation}
If $\widetilde\PA_j=\varnothing$, the intervention is referred to as \textit{perfect}, otherwise as \textit{imperfect}. Further, if $\Var_{\tilde p}[V_j]>0$, the intervention is called \textit{stochastic}.

\subsection{Counterfactuals}
Counterfactuals are statements about hypothetical interventions which are at odds with what was actually observed (i.e., counter-to-fact). 
They involve ``imagining'' alternative scenarios that could have played out had the world been different in some regards but equal in all others. 
Counterfactuals occupy the third and highest rung or level in the Ladder of Causation~\citep{pearl2018book} or Pearl Causal Hierarchy~\citep{bareinboim2022pch}---after association (``seeing''; rung 1) and intervention (``doing''; rung 2).

As an example, consider the following counterfactual query: 
\begin{quote}
	Given that a patient received treatment A and their health got worse, what would have happened  if they had been given treatment B instead, \textit{all else being equal}?
\end{quote}
The ``all else being equal'' part highlights the difference between interventions and counterfactuals:
observing the factual outcome (i.e., what actually happened) provides information about the background state of the system (as captured by the noise terms)
which is used to reason about alternative, counterfactual outcomes.
This differs from an intervention where such background information is not available, but the system is manipulated without prior partial observation thereof.
In the above example, observing that treatment A did not work may suggest that the patient has a rare condition and that treatment B would have therefore worked.
However, given that treatment A has been prescribed, the patient's condition may have changed, and B may no longer work in a future intervention.

Counterfactuals are computed in SCMs through the following three-step ``abduction-action-prediction'' procedure:
\begin{enumerate}
	\item Update the noise distribution to  its posterior given 
	the observed evidence (``abduction'').
	\item Manipulate the structural equations to capture the hypothetical intervention (``action'').
	\item Use the modified SCM to infer the quantity of interest (``prediction'').
\end{enumerate}
\begin{definition}[Counterfactual SCM]
	Let $\Mcal=(\Scal,P_\Ub)$ be an SCM (\cref{def:SCM}).
	Given observed evidence $\Eb=\eb$ for some subset of variables $\Eb\subseteq\Vb$, the counterfactual SCM  $\Mcal^{\Eb=\eb}$ is obtained by updating $P_\Ub$ with its posterior: $\Mcal^{\Eb=\eb}=(\Scal,P_{\Ub|\eb})$.
	Counterfactuals are then computed by performing interventions in the counterfactual SCM $\Mcal^{\Eb=\eb}$ according to~\cref{def:intervention}.
\end{definition}
For example, the counterfactual distribution of what would have happened under $do(\Wb=\wb)$ given $\Eb=\eb$ is given by $\Mcal^{\Eb=\eb; \, do(\Wb=\wb)}=(\Scal^{do(\Wb=\wb)}, P_{\Ub|\eb})$.
Importantly, $\Wb$ and $\Eb$ need not be disjoint, as the former is used to modify the structural equations and the latter to update the noise distribution.

\begin{example}%
	\label{ex:counterfactual_SCM}
	Let $\Mcal$ be the linear SCM from~\cref{ex:simple_SCM}.
	Suppose that we observe $(V_1=1, V_2=2, V_3=2)$ and wish to reason about the counterfactual ``what would have been, had $V_2$ been $3$''.
	\begin{enumerate}[]
		\item \textbf{Abduction:} from the original structural equations in~\cref{eq:ex_full_scm}, 
		we find the posterior~$P(\Ub~|~V_1=1,V_2=2,V_3=2)$ to be a point mass on $(U_1=1,U_2=1, U_3=-1)$.
		\item \textbf{Action:} 
		we modify the structural equations $\Scal$ to obtain $\Scal^{do(V_2=3)}$ as
		\begin{equation}
			\begin{aligned}
				\label{eq:ex_IC}
				V_1&:=U_1,\qquad 
				V_2&:=3, \qquad 
				V_3&:=V_1+V_2+U_3;
			\end{aligned}
		\end{equation}
		\item \textbf{Prediction:}
		we compute the push-forward of $P(\Ub~|~~V_1=1,V_2=2,V_3=2)$ via~\eqref{eq:ex_IC} which yields a point mass on $(V_1=1, V_2=3, V_3=3)$. That is, $V_1$ would have remained unaffected, but $V_3$ would have increased by~one.
	\end{enumerate}
\end{example}

\clearpage 
\section{Identifiable Representation Learning}
\label{sec:identifiable_rep_learning}
Rather than employing the SCM framework, 
another approach to modelling data 
is to directly specify a statistical model $P_{\thetab}(\Xb)$. Here, $\thetab\in\Thetab$ is a parameter vector which specifies the considered family of distributions, e.g., the mean and covariance of a multi-variate Gaussian distribution.
When $\Xb$ is very high-dimensional or is expected to follow a very complex distribution, as is commonly the case in ML, a popular approach is to instead specify its generative process indirectly 
in the form of a latent variable model (LVM). 

\subsection{Latent Variable Models and Identifiability}
An LVM postulates the existence of (often lower dimensional) latent random variables $\Zb$ that give rise to the observed variables $\Xb$.
An LVM typically assumes the following generative process
\begin{align}
	\Zb&\sim P_{\thetab_\Zb}(\Zb) \label{eq:latent_prior} \\
	\Xb~|~\Zb &\sim P_{\thetab_\Xb}(\Xb~|~\Zb)
	\label{eq:conditional_X_given_Z}
\end{align}
where $P_{\thetab_\Zb}(\Zb)$ is often a simple distribution, whereas the conditional $P_{\thetab_\Xb}(\Xb~|~\Zb)$ can be more complex.

\begin{remark}
	In the spirit of SCMs, another way of expressing~\eqref{eq:conditional_X_given_Z} is as a structural equation
	\begin{equation}
		\label{eq:noisy_mixing}
		\Xb:=\fb(\Zb,\Ub_\Xb)
	\end{equation}
	with noise $\Ub_\Xb$.
	In this case, $\thetab_\Xb$ would specify the generator or mixing function $\fb$ and the distribution of~$\Ub_\Xb$. (If $\fb$ is nonparametric, $\thetab_\Xb$ will be infinite-dimensional.)
\end{remark}

A common way of learning an LVM is to fit $\thetab$ by maximising the (marginal) likelihood given data $\xb_1, ..., \xb_n$ sampled \iid from $P_\Xb$.
According to~\eqref{eq:latent_prior} and~\eqref{eq:conditional_X_given_Z}, the likelihood is given by
\begin{equation}
	\label{eq:marginal_likelihood}
	p_{\thetab}(\Xb)=\int p_{\thetab_\Xb}(\Xb|\Zb)p_{\thetab_\Zb}(\Zb) \diff \Zb \,.
\end{equation}
For discrete $\Zb$, the integral in \eqref{eq:marginal_likelihood} becomes a sum and $\thetab$ can be estimated through expectation maximisation~\citep{dempster1977maximum}.
For continuous $\Zb$, the integral in~\eqref{eq:marginal_likelihood} is intractable in general. In this case, it is common to employ approximate inference techniques, such as variational inference~\citep{blei2017variational}.

Often, an unknown true model parametrized by $\thetab^*$ is postulated to have generated the observations and the goal is to infer $\thetab^*$ (or properties thereof) from data.
However, in principle, different $\thetab\neq\thetab'$ can give rise to the same distribution $P_{\thetab}(\Xb)$ over observations.
A crucial question is thus whether it is at least in principle possible to recover $\thetab^*$, subject to assumptions on the model class. 
The answer to this question is related to the identifiability of the model class.

\begin{definition}[Identifiability, adapted from~\cite{lehmann2006theory}, Defn.~5.2]
	\label{def:point_identifiability}
	Let $\Xb\sim P_{\thetab}(\Xb)$. Then $\thetab$ is \emph{point} (or \emph{uniquely}) \emph{identifiable} (based on $\Xb$) if 
	\begin{equation}
		P_{\thetab}(\Xb)=P_{\thetab'}(\Xb) \implies \thetab=\thetab'\, ,
	\end{equation}
	that is, if the mapping $\thetab\mapsto P_{\thetab}$ is injective.
\end{definition}

Point identifiability ensures that, given  infinite data sampled from $P_{\thetab^*}(\Xb)$, globally maximizing the likelihood will recover $\thetab^*$ exactly.
However, this is typically not achievable in LVMs (without strong additional assumptions) as there are often certain symmetries, such as permutation or rescaling of the latent variables~$\Zb$, that cannot be uniquely resolved.
A more practical notion is therefore that of %
identifiability up to some level of ambiguity.

\begin{definition}[Identifiability up to $\sim$; from~\cite{khemakhem2020variational}, Defn.~1]
	\label{def:identifiability_up_to_equiv}
	Let $\Xb\sim P_{\thetab}(\Xb)$, and let $\sim$ be an equivalence relation\footnote{A binary relation that is reflexive ($a\sim a$), symmetric ($a\sim b \iff b\sim a$), and transitive ($a\sim b$ and $b\sim c \implies a\sim c$).} defined on the parameter space~$\Thetab$. Then  $\thetab$ is \emph{identifiable up to $\sim$} (based on $\Xb$) if
	\begin{equation}
		P_{\thetab}(\Xb)=P_{\thetab'}(\Xb) \quad  \implies \quad \thetab\sim\thetab'\,.
	\end{equation}
	The elements of the quotient space $\Thetab / \sim$ are called the identifiability classes.
\end{definition}

Identifiability studies aim to characterize these identifiability classes, subject to suitable assumptions on the model class (and the available data).
A type of LVM for which identifiability has been studied extensively, and which is of particular interest (since it can be viewed as a special case of CRL) is \textit{independent component analysis}.

\subsection{Independent Component Analysis (ICA)}
\label{sec:background_ICA}
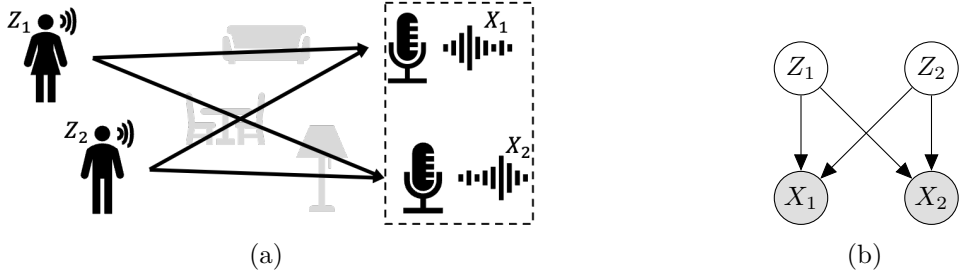
\begin{figure}
	\centering
	\begin{subfigure}{0.5\textwidth}
		\centering
		\includegraphics[width=0.9\textwidth]{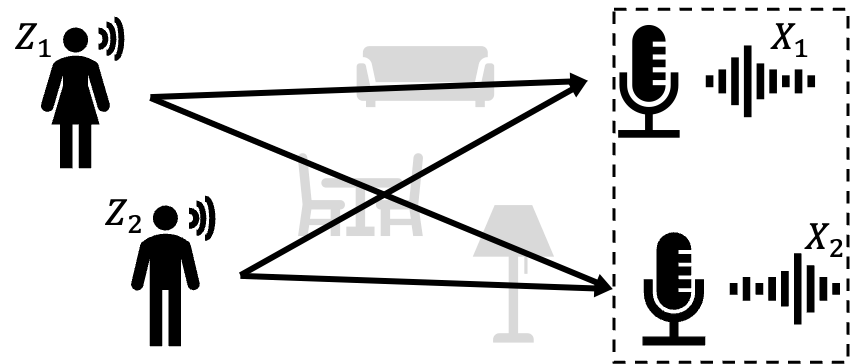}
		\caption{}
		\label{fig:cocktailparty}
	\end{subfigure}%
	\begin{subfigure}{0.5\textwidth}
		\centering
		\begin{tikzpicture}
			\centering
			\node (s1) [latent] {$Z_1$};
			\node (s2) [latent, right=of s1] {$Z_2$};
			\node (x1) [obs, below=of s1] {$X_1$};
			\node (x2) [obs, below=of s2] {$X_2$};
			\edge {s1,s2}{x1,x2};
		\end{tikzpicture}
		\caption{}
		\label{fig:ICA_graph}
	\end{subfigure}
	\caption[Independent Component Analysis (ICA) Problem Setting]{\textbf{Independent Component Analysis (ICA) Problem Setting.} (a) Illustration of the cocktail party problem, a motivating example for ICA: here, the task is to recover the individual voices, based only on recorded mixtures thereof.
		(b) Graphical model representation of ICA with $n=2$ sources $Z_i$ and $d=2$ mixtures $X_i$. The sources are mutually independent and unobserved (white nodes), while the mixtures are observed (grey nodes).}
\end{figure}

Independent component analysis (ICA) considers a setting in which the latent variables $Z_i$ are assumed to be mutually independent,
\begin{equation}
	\label{eq:independent_sources}
	p(\Zb)=\prod_{i=1}^n p(Z_i)\,.
\end{equation}
Moreover,
  ICA typically assumes an invertible \textit{deterministic} mapping between latents and observations,
\begin{equation}
	\label{eq:mixing function}
	\Xb:=\fb(\Zb)
\end{equation}
as well as an equal number of observed and latent variables ($d=n$).
In ICA, typically all variables are considered continuous. Moreover, the latent variables $\Zb$ are generally referred to as \textit{sources}, the observed variables~$\Xb$ as \textit{mixtures}, and $\fb$ as a \textit{mixing function}. 
The canonical example of ICA is the \textit{cocktail party problem}, illustrated in~\cref{fig:cocktailparty}.
Here, a number of conversations are happening in parallel, and the task is to recover the individual voices $Z_i$ from the mixtures $X_i$ recorded by microphones placed at different locations throughout the room. 
Since only the $X_i$s are observed, this task is also referred to as \textit{blind source separation} (BSS).

Independent component estimation aims to learn an encoder or unmixing function $\gb$ such that the components of the estimated sources
\begin{equation}
	\label{eq:independent_component_estimation}
	\hat \Zb = \gb(\Xb)
\end{equation}
are mutually independent.
Whether the estimated independent components recover the true sources (i.e., solve the blind source separation task), or up to which ambiguities, is the question of identifiability, which has been studied for ICA under different assumptions on the model class. 

In general, an ICA model is specified by a pair $(\fb,P_\Zb)$, consisting of a mixing function and a (factorized) source distribution.
The distribution of observations (based on which identifiability is defined, see~\cref{def:point_identifiability,def:identifiability_up_to_equiv}) is then given by the push-forward 
\begin{equation}
	P_\Xb=\fb_*(P_\Zb)
\end{equation}
and its density by the change of variable formula
\begin{equation}
	\label{eq:change_of_variable_formula}
	p_\Xb(\xb)=p_\Zb\left(\fbinv(\xb)\right)\abs{\det \Jb_\fb\left(\fbinv(\xb)\right)}^{-1}
\end{equation}
where $\Jb_\fb$ denotes the Jacobian matrix of $\fb$.

In linear ICA, (i.e., if the mixing function is constrained to be an invertible matrix), the model is identifiable up to permutation and (linear) rescaling of the sources, provided that at most one source is Gaussian~\citep{darmois1953analyse,skitovich1954linear,comon1994independent}.

In nonlinear ICA, the mixing function in~\eqref{eq:mixing function} is not constrained to be linear but generally taken to be any (smooth) invertible function.\footnote{To facilitate the analysis, it is typically assumed that $\fb$ is smooth or continuously differentiable.}
Similar to the linear case, the ordering of the sources is fundamentally irresolvable. 
Moreover, the scale ambiguity becomes \textit{nonlinear} in the sense of blindness to \textit{element-wise reparametrisation}, for if $Z_i$ and $Z_j$ are independent, then so are $h_i(Z_i)$ and $h_j(Z_j)$ for any functions $h_i$ and $h_j$.
Thus, for any invertible element-wise function $\hb(\Zb)=(h_1(Z_1), ..., h_n(Z_n))$ and permutation $\Pb$, we have
\begin{equation}
\Xb=\fb(\Zb)=\left(\fb\circ\Pb^{-1}\circ\hb^{-1}\right)\left(\hb(\Pb\Zb)\right)=\tilde \fb (\tilde\Zb)\,,
\end{equation}
where  $\tilde\Zb=\hb(\Pb\Zb)$ also has independent components.
Permutation and element-wise (nonlinear) reparametrization are therefore fundamentally irresolvable ambiguities of nonlinear ICA, and the following notion of equivalence is of interest.

\begin{definition}[Equivalence up to permutation and element-wise reparametrization]
\label{def:bss_identifiability}
Let $\Fcal_n$ be the set of all invertible functions $\fb:\RR^n\rightarrow \RR^n$, and $\Pcal_n$ the space of all (non-degenerate) factorized densities on $\RR^n$.
The equivalence relation $\sim_\BSS$ on $\Fcal_n\times\Pcal_n$ is given by:
\begin{equation}
	(\fb, p)\sim_\BSS(\tilde \fb,\tilde p) \qquad \iff \qquad 
	(\fb, p)=(\tilde \fb\circ \hb^{-1}\circ\Pb^{-1}, (\Pb\circ\hb)_*\tilde p)
\end{equation}%
for some permutation matrix $\Pb\in\RR^{n\times n}$ and some element-wise, invertible function $\hb(\Zb)=(h_1(Z_1), ..., h_n(Z_n))\in\Fcal_n$.
\end{definition}%

\subsection{Non-Identifiability of Nonlinear ICA from I.I.D. Data}
\label{sec:background_nonlinear_ICA_nonidentifiable}
A major obstacle to \textit{nonlinear} blind source separation (in the sense of  identifiability up to $\sim_\textsc{bss}$) is that nonlinear mixtures of $Z_i$ and $Z_j$ can still be independent, even if all sources are non-Gaussian. 
As a result, independent component estimation~\eqref{eq:independent_component_estimation} does not necessarily  recover the original sources.

\begin{theorem}[Non-identifiability of nonlinear ICA; \citet{hyvarinen1999nonlinear}]
\label{thm:nonidentifiability_nonlinear_ICA}
Nonlinear ICA~\eqref{eq:mixing function} with $(\fb,p_\Zb)\in \Fcal_n\times \Pcal_n$ is \emph{not} identifiable up to~$\sim_\textsc{bss}$.
\end{theorem}

\begin{figure}
\centering
\includegraphics[width=0.5\textwidth]{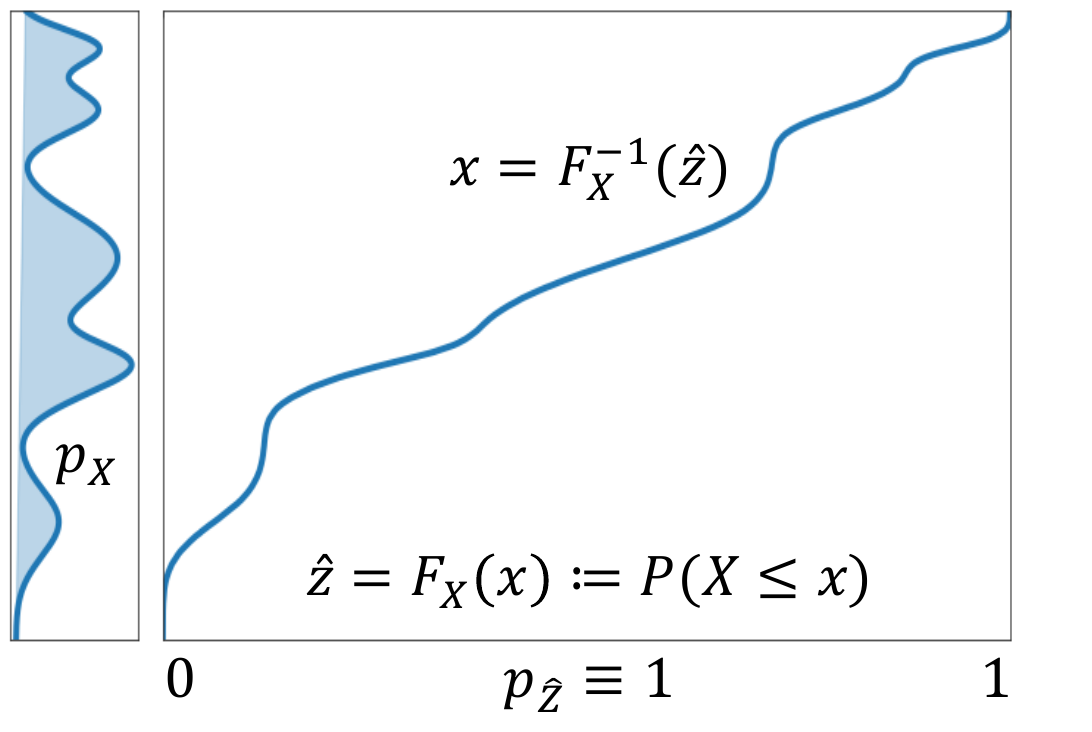}
\caption[Illustration of the Cumulative Distribution Function (CDF) Transform]{\textbf{Illustration of the Cumulative Distribution Function (CDF) Transform.} Applying the CDF transform~$F_X(x)=P(X\leq x)$ to any random variable $X$ with non-degenerate density $p_X$ (left) yields a random variable $\hat Z=F_X(X)$ that is uniform on $(0,1)$
	  (bottom).
}
\label{fig:IGCI_Darmois}
\end{figure}

The proof of~\cref{thm:nonidentifiability_nonlinear_ICA} builds upon a construction due to~\cite{darmois1951construction}, which employs the cumulative distribution function (CDF) to transform a given random variable into a uniformly distributed one, as illustrated in~\cref{fig:IGCI_Darmois}. 

\begin{definition}[Darmois construction]
\label{def:darmois_solution}
For $i=1,...,n$, the \emph{Darmois construction} $\gb^\mathrm{D}:\RR^n\rightarrow(0,1)^n$ is defined via:
\begin{equation}
	\label{eq:Darmois_construction}
	g^\mathrm{D}_i(\xb_{1:i})
	:=P(X_i\leq x_i~|~\xb_{1:i-1})
	=\int_{-\infty}^{x_i}p(x_i'~|~\xb_{1:i-1}) \diff x_i'
	\, .
\end{equation}%
\end{definition}%
The Darmois construction repeatedly applies the conditional CDF transform in a procedure that can be viewed as a probabilistic analogue of Gram-Schmidt orthogonalization~\citep{hyvarinen1999nonlinear}. It yields a set of estimated components $\hat \Zb=\gb^\mathrm{D}(\Xb)$ which are jointly uniform on the unit hypercube $(0,1)^n$, and thus mutually independent.
To see this, first note that $\gb^\mathrm{D}$ has lower-triangular Jacobian (i.e., $\nicefrac{\partial g_i^\mathrm{D}}{\partial x_j}=0\,$ for~$i<j$) since $g^\mathrm{D}_i$ in~\eqref{eq:Darmois_construction} only depends on~$\xb_{1:i}$. 
Hence, its determinant is given by the product of the diagonal terms,
\begin{equation}
\det \Jb_{\gb^\mathrm{D}}(\xb)=\prod_{i=1}^n \frac{\partial g_i^\mathrm{D}}{\partial x_i}(\xb)=\prod_{i=1}^n p(x_i~|~\xb_{1:i-1})=p(\xb)
\end{equation}
where the second equality follows from $\frac{\partial}{\partial x_i}P(X_i\leq x_i~|~\xb_{1:i-1})=p(x_i~|~\xb_{1:i-1})$, and the third from the chain rule.\footnote{This exposition is partly based on~\citet[][\S2.2]{papamakarios2021normalizing}.}
Thus, the density of $\hat \Zb$ evaluated at $\hat\zb=\gb^\mathrm{D}(\xb)$ is given by the change of variable formula~\eqref{eq:change_of_variable_formula} as
\begin{equation}
p_{\hat\Zb}(\hat\zb)=p_\Xb\left(\xb\right)\abs{\det \Jb_{\gb^\mathrm{D}}\left(\xb\right)}^{-1}=p_\Xb\left(\xb\right)\abs{p_\Xb\left(\xb\right)}^{-1}=1\,,
\end{equation}
which is uniform.
Denoting the corresponding mixing function by $\fb^\text{D}=(\gb^\text{D})^{-1}$,
and the uniform density on $(0,1)^n$ by $p_\Ub$, 
the pair $(\fb^\text{D}, p_\Ub)$ thus forms a valid nonlinear ICA solution.
However, the independent components $\hat\Zb$ thus estimated will generally not be meaningfully related to the \textit{true} sources $\Zb$.
Since the order of the observed $X_i$ is arbitrary,  applying~\eqref{eq:Darmois_construction} to a permuted version of $\Xb$ yields different Darmois solutions.
Hence, any $X_i$ can be taken to be one of the original sources, which violates identifiability up to $\sim_\textsc{bss}$ unless the true mixing $\fb$ has diagonal Jacobian (i.e., does not actually mix the sources).

Additional spurious solutions can be constructed through \textit{measure preserving automorphisms} (MPAs), functions $\ab$ which map the source space to itself without affecting the source distribution, i.e.,  $\ab_* (P_\Zb)=P_\Zb$~\citep{hyvarinen1999nonlinear}.
An instructive class of MPAs is the following~\citep{locatello2019challenging,khemakhem2020variational}.

\begin{definition}[``Rotated-Gaussian'' MPA]
\label{def:measure_preserving_automorphism_Gaussian}
Let $\Rb\in O(n)$ be an orthogonal matrix, and denote by $\Fb_\Zb(\zb)=(F_{Z_1}(z_1), ..., F_{Z_n}(z_n))$ and $\bm\Phi(\zb)=(\Phi(z_1), ..., \Phi(z_n))$ the element-wise CDFs
of a smooth, factorized density $p_\Zb$ and of a standard isotropic  Gaussian, respectively.
Then the ``rotated-Gaussian'' MPA $\ab^{\Rb}(p_\Zb)$ is given by
\begin{equation}
	\label{eq:measure_preserving_automorphism_Gaussian}
	\ab^{\Rb}(p_\Zb) =\Fb_\Zb^{-1} \circ \bm\Phi \circ \Rb \circ \bm\Phi^{-1} \circ \Fb_\Zb\,.
\end{equation}
\end{definition}
$\ab^{\Rb}(p_\Zb)$ first maps to the rotationally invariant isotropic Gaussian distribution via $\bm\Phi^{-1} \circ \Fb_\Zb$, then applies a rotation, and finally maps back, without affecting the source distribution.
Hence, if $(\tilde\fb, \tilde p_{\Zb})$ is a valid solution, then so is~$(\tilde\fb\circ\ab^\Rb(\tilde p_{\Zb}),\tilde p_{\Zb})$ for any $\Rb\in O(n)$.
Unless $\Rb$ is a permutation, $\ab^\Rb(p_\Zb)$ is not an element-wise function,
contradicting
 $\sim_\BSS$-identifiability.
\begin{remark}
The Darmois construction~\eqref{eq:Darmois_construction} yields $n$ independent components, even if the original sources 
 were not independent to begin with. Combined with MPAs such as~\eqref{eq:measure_preserving_automorphism_Gaussian}, this implies that not only factorized but any (unconditional, see below) prior $P_\Zb$ is insufficient for identifiability under arbitrary nonlinear mixing~\citep[][\S~D]{khemakhem2020variational}.
\end{remark}

\subsection{Identifiability Results Under Additional Assumptions}
\label{sec:background_nonlinear_ICA_auxiliary}
While nonlinear ICA is not identifiable from i.i.d.\ data,
this impossibility can be overcome by \textit{deviations from the i.i.d.\ assumption}.
For example, identifiability has been shown for (i)~\textbf{not identically distributed} observations in the form of (segment-wise) non-stationary time-series~\citep{hyvarinen2016unsupervised}, or (ii)~temporally \textbf{dependent} observations in the form of autocorrelated (non-Gaussian, stationary) time-series~\citep{hyvarinen2017nonlinear}. Both results are proven constructively, giving rise to the time-contrastive learning (TCL) and permutation-contrastive learning (PCL) algorithms, respectively. PCL and TCL are instances of \textit{contrastive self-supervised learning} in which a classifier is trained on top of a feature extractor to discriminate between real and fake data~\citep{gutmann2010noise,oord2018representation}. For TCL and PCL, fake data is created by permuting the time segment or index, respectively, and the true sources are provably recovered in the process.

These approaches can be unified by introducing an auxiliary variable, which renders the sources \textit{conditionally} independent~\citep{hyvarinen2019nonlinear}. 
The assumption of joint independence in~\eqref{eq:independent_sources} is thus replaced with
\begin{equation}
\label{eq:conditionally_independent_sources}
p(\Zb~|~\ub)=\prod_{i=1}^n p(Z_i~|~\ub)\,,
\end{equation}
where the auxiliary variable $\ub$ can represent relevant side-information such as an environment index, a partial history of past observations, temporal or spatial structure, or a class label.
Typically, $\ub$ is required to be observed, but there are also some exceptions~\citep{halva2020hidden,halva2021disentangling}.

\begin{theorem}[Identifiability of nonlinear ICA with auxiliary variables;~informal]
\label{thm:nonlinear_ICA_auxiliary_ID}
Nonlinear ICA with a conditional prior as in~\eqref{eq:conditionally_independent_sources} is identifiable up to $\sim_\textsc{bss}$; see~Thm.~1 of~\citet{hyvarinen2019nonlinear} for details on the technical assumptions. 
\end{theorem}

A key requirement for~\cref{thm:nonlinear_ICA_auxiliary_ID} is a so-called \textit{assumption of variability}, which roughly states that the auxiliary variable needs to have ``a sufficiently strong and diverse
effect on the distributions of the independent components''~\citep{hyvarinen2019nonlinear}. More formally, it states that there exist at least $2n+1$ distinct values of $\ub$ such that a collection of vectors of first and second-order partial derivatives of $\log p(z_i~|~\ub)$ w.r.t.\ $z_j$ are linearly independent.

These developments on positive results for the general nonlinear case have sparked a renewed interest in identifiability in the context of deep representation learning~\citep{khemakhem2020variational,khemakhem2020ice,klindt2020slowvae,sorrenson2020disentanglement,roeder2021linear,kivva2022identifiability}.

Instead of temporal structure or other auxiliary information, several other methods are based on the related idea of using weak supervision in the form of multiple simultaneously observed views~\citep{gresele2019incomplete,locatello2020weakly,shu2019weakly,zimmermann2021contrastive}. %

\paragraph{Restricting the Mixing Function.}
\looseness-1 Instead of constraining the nonlinear ICA model by placing additional assumptions such as~\eqref{eq:conditionally_independent_sources} on the source distribution, another option is to instead restrict the class of considered mixing functions $\fb$. 
The following chapter explores this idea.

%% file: Chapter6/chapter6.tex
\graphicspath{{Chapter6/Figs/}}
\chapter{Unsupervised Representation Learning}
\label{chap:IMA}
In this chapter, we begin our investigation of identifiable CRL by considering the fully unsupervised setting of learning from a single i.i.d.\ dataset of unlabelled observations. 
As discussed in~\cref{chap:background}, identifiability is not achievable in this case if arbitrary nonlinear mixings are considered.
We therefore explore \emph{assumptions that constrain the class of mixing functions} by drawing on ideas from the field of causality.
Our main efforts are spent on the setting of identifying independent latent factors (ICA), 
a special case of CRL with trivial causal graph,
for which we propose a constraint that is inspired by the idea of independent mechanisms or influences rooted in the ICM~\cref{principle:ICM}.
Importantly, the resulting function class still allows for non-trivial nonlinear maps. 
At the same time, it promotes identifiability by ruling out common spurious solutions.
We establish connections of our assumption to the learning objectives of variational autoencoders, and discuss a related constraint on the class of mixing functions for identifiable object-centric representation learning.

\text{The main content of this chapter has been published in the following paper:}
\begin{selfcitebox}
\href{https://arxiv.org/abs/2106.05200}{\ul{Independent mechanism analysis, a new concept?}}
\\
Luigi Gresele$^*$, \textbf{Julius von K\"ugelgen}$^*$, Vincent Stimper, Bernhard Schölkopf, \\
{Michel Besserve} ($^*$equal contribution)
\\
\textit{Advances in Neural Information Processing Systems (NeurIPS)}, 2021
\end{selfcitebox}

\section{Introduction}
\label{sec:IMA_introduction}

Independent component analysis (ICA) provides a principled framework for unsupervised representation learning,  with solid theory on the identifiability of the latent code that generated the data, given only observations of mixtures thereof~(\cref{sec:identifiable_rep_learning}).
Here, we consider the nonlinear ICA setting, that is, 
\begin{equation}
	\label{eq:gen}
	\xb = \fb(\sbb)\,, 
	\quad\quad\quad\quad\quad
	p_\sb(\sb) = \prod_{i=1}^n p_{s_i}(s_i)\,.
\end{equation}

\begin{remark}[Notation]
    In line with the ICA and deep generative modelling literature, in this chapter and the next, we sometimes slightly abuse the notation by also using lower-case symbols when referring to random variables rather than their realizations; we hope that it is clear from the context which is meant. Further, $\sb$ is used here instead of $\zb$ to denote the sources.
\end{remark}

The model in~\eqref{eq:gen} is provably non-identifiable in general, since statistical independence of the latent variables alone does not sufficiently constrain the problem~(\cref{sec:background_nonlinear_ICA_nonidentifiable}).
However, recent work has shown that identifiability can be recovered in settings where additional, typically observed auxiliary variables are included in the generative process~(\cref{sec:background_nonlinear_ICA_auxiliary}). 
Here, we investigate a different route to identifiability by drawing inspiration from the field of \text{causality}~(\cref{sec:causal_modelling}).
To this end, we \textit{interpret the ICA mixing as a causal process} and apply the principle of independent causal mechanisms (ICM) which postulates that the generative process consists of independent modules which do not share information.
As discussed in~\cref{sec:ICM}, here ``independent'' does not refer to \text{statistical} independence 
of random variables, but rather to %
the notion that 
the distributions and functions composing the generative process are chosen independently by Nature~\citep{JanChaSch16}.
While there is a formalisation of ICM %
in terms of algorithmic (Kolmogorov) complexity~\citep{janzing2010causal},
 it is not computable, and hence applying ICM in practice requires 
assessing such non-statistical independence with suitable domain-specific criteria~\citep{SteJanSch10}.
Our goal is thus to \textit{constrain the nonlinear ICA problem, in particular the mixing function, via suitable ICM measures}, thereby ruling out common counterexamples to identifiability which intuitively violate the ICM principle.

Traditionally, ICM criteria have been developed for 
causal discovery, where \textit{both  cause and effect are observed}~\citep{daniuvsis2010inferring,janzing2012information,janzing2010telling,zscheischler2011testing}.
They enforce independence between (i) the cause (source) distribution and (ii) the conditional or mechanism (mixing function) generating the effect (observations), and thus rely on the fact that the \textit{observed} cause distribution is informative. 
As we will show, this renders them insufficient for nonlinear ICA, since the constraints they impose 
are satisfied by common counterexamples to identifiability discussed in~\cref{sec:background_nonlinear_ICA_nonidentifiable}.
With this in mind, we introduce a new way to characterise or \textit{refine} the ICM principle for
unsupervised representation learning tasks such as nonlinear ICA. 

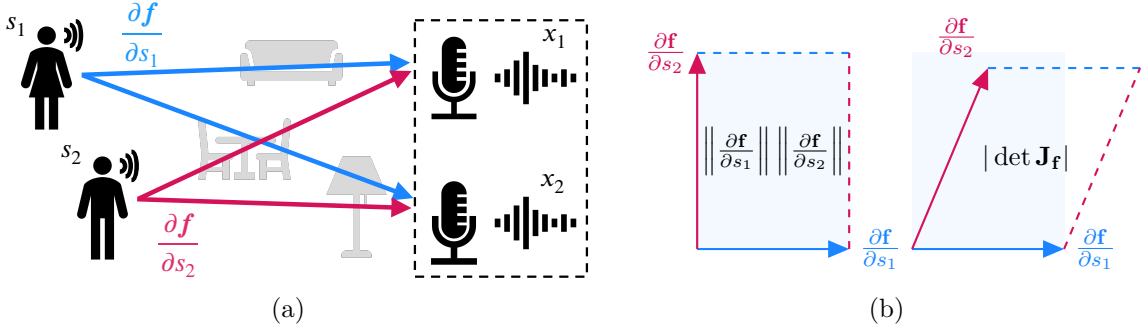
\begin{figure}
\begin{subfigure}[b]{0.5\textwidth}
	\centering
	\includegraphics[width=\textwidth]{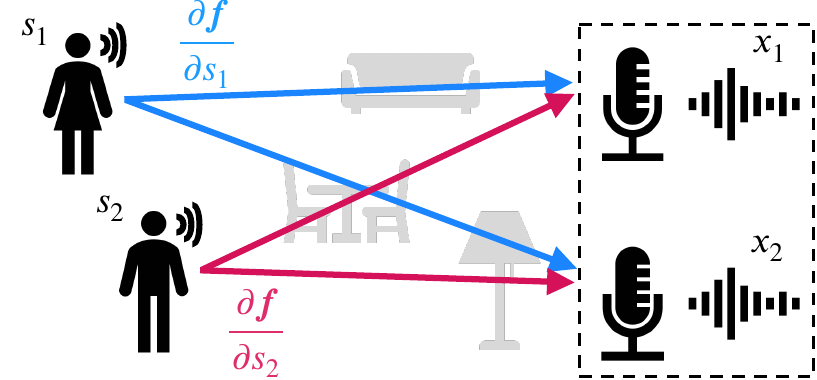}
        \caption{}
        \label{fig:intuition_pic}
\end{subfigure}%
\begin{subfigure}[b]{0.5\textwidth}
	\centering
	\begin{tikzpicture}
		\fill[blue_cblind!5!white] (0,0) rectangle (2,2.6);
		\draw[->,thick,  color=blue_cblind] (0,0) -- (2,0) node[anchor= west] {$\frac{\partial \fb}{\partial s_1}$};
		\draw[->,thick,  color=red_cblind] (0,0) -- (0,2.6) node[anchor= east] {$\frac{\partial \fb}{\partial s_2}$} ;
		\draw[-,thick,  dashed, color=blue_cblind] (0,2.6) -- (2,2.6);
		\draw[-,thick,  dashed, color=red_cblind] (2,0) -- (2,2.6);
		\draw[] (1,1.3) node[] {\mbox{\small$\norm{\frac{\partial \fb}{\partial s_1}}\norm{\frac{\partial \fb}{\partial s_2}}$}};
	\end{tikzpicture}%
	\begin{tikzpicture}
		\fill[blue_cblind!5!white] (0,0) rectangle (2,2.6);
		\draw[->, thick, color=blue_cblind] (0,0) -- (2,0) 
		node[anchor= west] {$\frac{\partial \fb}{\partial s_1}$}
		;
		\draw[->, thick, color=red_cblind] (0,0) -- (1,2.4) 
		node[anchor= south east] {$\frac{\partial \fb}{\partial s_2}$}
		;
		\draw[-, thick, dashed, color=blue_cblind] (1,2.4) -- (3,2.4);
		\draw[-, thick, dashed, color=red_cblind] (2,0) -- (3,2.4);
		\draw[] (1.5,1.2) node[] {\mbox{\small $|\det \Jb_\fb|$}};
	\end{tikzpicture}
        \caption{}
        \label{fig:intuition_orthogonal}
\end{subfigure}%
\caption[Overview of Independent Mechanism Analysis (IMA)]{\textbf{Overview of Independent Mechanism Analysis (IMA).} (a)
For the cocktail party problem, the ICM principle \textit{as traditionally understood} would say that the content of speech $p_\sb$ is independent of the mixing or recording process $\fb$ (microphone placement, room acoustics).
IMA refines, or extends, this idea \textit{at the level of the mixing function} by postulating that the contributions~$\nicefrac{\partial \fb}{\partial s_i}$ of each source to $\fb$, as captured
by the speakers' positions relative to the recording process, should not be fine-tuned to each other.
(b) We formalise this independence between the~$\nicefrac{\partial \fb}{\partial s_i}$, which are the columns of the Jacobian~$\Jb_\fb$, as an \textit{orthogonality condition}: the absolute value of the determinant~$|\det \Jb_\fb|$, i.e., the volume of the parallelepiped spanned by $\nicefrac{\partial \fb}{\partial s_i}$, should decompose as the product of the norms of the~$\nicefrac{\partial \fb}{\partial s_i}$.}
\label{fig:intuition}
\end{figure}

\paragraph{Motivating Example.}
To build intuition, we return to 
the cocktail party problem, see \cref{fig:intuition_pic}, 
where  a number of conversations are happening in parallel, and the task is to recover the individual voices $s_i$ from the recorded mixtures $x_i$.
The mixing or recording process~$\fb$ is primarily determined by the
room acoustics and the locations at which microphones are placed.
Moreover, each speaker influences the recording through their positioning in the room, and we may think of this influence as $\nicefrac{\partial \fb}{\partial s_i}$.
Our independence postulate then amounts to stating that the speakers' positions are not fine-tuned to the
room acoustics and microphone placement, or to each other, i.e., \textit{the contributions $\nicefrac{\partial \fb}{\partial s_i}$ should be independent (in a non-statistical sense).%
}%

\paragraph{Our Approach.}
We formalise this notion of
independence between the contributions~$\nicefrac{\partial \fb}{\partial s_i}$ of each source 
to the mixing process
(i.e., the columns of the Jacobian matrix $\Jb_\fb$ of partial derivatives) as an orthogonality condition, see~\cref{fig:intuition_orthogonal}.  
Specifically, the absolute value of the determinant $|\det \Jb_\fb|$, which describes the local change in infinitesimal volume induced by mixing the sources, should factorise or decompose as the product of the norms of its columns. 
This can be seen as a decoupling of the local influence of each partial derivative in the pushforward operation (mixing function) mapping the source distribution to the observed one, and gives rise to a novel framework which we term independent mechanism analysis (IMA).
IMA can be understood as a refinement of the ICM principle that applies the idea of independence of mechanisms 
at the level of the mixing function.

\paragraph{Contributions.} Our main contributions can be summarised as follows:
\begin{itemize}%
\item We
  show that existing ICM criteria such as IGCI~\citep{janzing2012information} do not sufficiently constrain unsupervised representation learning tasks such as nonlinear ICA~(\cref{sec:unsuitability_of_existing_ICM_measures}).
\item We propose a more suitable ICM criterion 
for unsupervised representation learning, which gives rise to a new framework that we term independent mechanism analysis (IMA;~\cref{sec:IMA}). 
We provide geometric and information-theoretic interpretations of IMA~(\cref{sec:ima_intuition}),
introduce an IMA contrast function, which is invariant to the inherent ambiguities of nonlinear ICA%
~(\cref{sec:ima_definition_properties}), %
and 
show that it rules out a large class of counterexamples and is consistent with existing identifiability results~(\cref{sec:ima_theory}).
\item We experimentally validate our theoretical claims 
and propose a regularised maximum-likelihood learning approach based on the IMA contrast which compares favourably to an unregularised baseline~(\cref{sec:IMA_experiments}).
\end{itemize}%

\section{Existing ICM Measures Are Insufficient for Nonlinear ICA}
\label{sec:unsuitability_of_existing_ICM_measures}

Allowing for deterministic relations between cause (sources) and effect (observations),  the criterion which is most closely related to the ICA setting in~\eqref{eq:gen} is
		\textit{information-geometric causal inference} (IGCI)~\citep{daniuvsis2010inferring,janzing2012information}.
		IGCI assumes a nonlinear relation $\eb=\fb(\cb)$ and formulates 
		a notion of independence between the cause distribution $p_\cb$ and the deterministic mechanism $\fb$ (which we think of as a degenerate conditional $p_{\eb|\cb}$) via the following condition (in practice, assumed to hold approximately),%
		\begin{equation}
			\label{eq:IGCI_condition}
			C_\IGCI(\fb,p_\cb):=
			\int \log  \left|\det\Jb_{\fb}(\cb)\right|p_\cb(\cb)d\cb 
			-
			\int \log \left|\det\Jb_{\fb}(\cb)\right|d\cb
			= 0 \, ,
		\end{equation}%
		where $(\Jb_\fb(\cb))_{ij}=\nicefrac{\partial f_i}{\partial c_j}(\cb)$ is the Jacobian matrix.
		$C_\IGCI$
		can
		be understood as the covariance between $p_\cb$ and $\log \left|\det\Jb_{\fb}\right|$ (viewed as r.v.s on 
		the unit cube w.r.t.\ the Lebesgue measure), so that 
		$C_\IGCI=0$ rules out a form of fine-tuning between $p_\cb$ and $|\det \Jb_\fb|$. 
		As its name suggests,  IGCI can, from an information-geometric perspective, also be seen as an orthogonality condition between cause and mechanism in the space of probability distributions~\citep{janzing2012information}.
Our aim is to use the ICM~\cref{principle:ICM} to further constrain the space of models $\Mcal\subseteq\Fcal\times\Pcal$  (where $\Fcal$ is the set of all smooth, invertible functions $\fb:\RR^n\rightarrow \RR^n$,
and $\Pcal$ the set of all smooth, factorised densities $p_\sb$ with connected support on $\RR^n$), and rule out common counterexamples to identifiability such as those presented in~\cref{sec:background_ICA}.
Intuitively, both the Darmois construction~\eqref{eq:Darmois_construction} and the rotated Gaussian MPA~\eqref{eq:measure_preserving_automorphism_Gaussian} give rise to ``\textit{non-generic}'' solutions which should violate ICM: the former, $(\fb^\text{D},p_\ub)$, due the triangular Jacobian of $\fb^\text{D}$, meaning that each observation
$x_i=f^\text{D}_i(\yb_{1:i})$ only depends on a subset of the inferred independent components~$\yb_{1:i}$,
and the latter, $(\fb\circ\ab^\Rb(p_\sb),p_\sb)$, due to the dependence of $\fb\circ\ab^\Rb(p_\sb)$ on $p_\sb$ in~\eqref{eq:measure_preserving_automorphism_Gaussian}.

However, previous ICM criteria~(\cref{sec:ICM}) were developed for 
the task of cause-effect inference where \textit{both} variables are observed.
In contrast, here we consider
an unsupervised representation learning task
where \textit{only the effects} (the mixtures $\xb$) \text{are observed}, but the causes (the sources $\sb$) are not. 
It turns out that this
renders existing ICM criteria insufficient for BSS in that
they can easily 
be satisfied by spurious solutions which are not equivalent to the true one.
We can show this for IGCI.
Denote the class of nonlinear ICA models satisfying IGCI~\eqref{eq:IGCI_condition} by \[ \Mcal_{\IGCI}=\{(\fb, p_\sb)\in\Fcal\times\Pcal: C_\IGCI(\fb, p_\sb)=0\}\subset \Fcal\times\Pcal\,.\] Then the following negative result holds.%
\begin{proposition}[IGCI is insufficient for $\sim_\BSS$-identifiability]
	\label{prop:IGCI_insufficient_for_BSS}
	The model in \eqref{eq:gen} is not identifiable up to $\sim_\BSS$ (\cref{def:bss_identifiability}) on $\Mcal_{\IGCI}$.
	\begin{proof}
		IGCI~\eqref{eq:IGCI_condition} is
		satisfied when $p_\sb$ is uniform.
		However, the Darmois construction~\eqref{eq:Darmois_construction}
		yields uniform sources, see~\cref{fig:IGCI_Darmois}.
		This means that $(\fb^\text{D}\circ\ab^\Rb(p_\ub), p_\ub)\in\Mcal_\IGCI$%
		, so
		IGCI can be satisfied by solutions which do not 
		separate the sources in the sense of~\cref{def:bss_identifiability}.
	\end{proof}%
\end{proposition}%

\begin{figure}[t]
	\centering
\begin{subfigure}[b]{0.3\textwidth}
	\centering
	\begin{tikzpicture}
		\centering
		\node (s1) [obs, draw=blue_cblind,line width=0.5mm] {$\cb$};
		\node (x1) [obs, below=of s1] {$\eb$};
		\edge[color=red_cblind] {s1}{x1};
	\end{tikzpicture}
	\caption{ICM}
	\label{fig:ICM_graph}
\end{subfigure}%
\begin{subfigure}[b]{0.3\textwidth}
	\centering
	\begin{tikzpicture}
		\centering
		\node (s1) [latent] {$s_1$};
		\node (s2) [latent, xshift=4em] {$s_2$};
		\node (x) [obs, below=of s1, xshift=2em] {$\xb$};
		\edge [color=blue_cblind] {s1}{x};
		\edge [color=red_cblind] {s2}{x};
	\end{tikzpicture}
	\caption{IMA}
	\label{fig:IMA_graph}
\end{subfigure}%
\caption[Relation of IMA to Classical ICM Criteria]{\textbf{Relation of IMA to Classical ICM Criteria.} (a) Existing ICM criteria typically enforce independence between an observed input or cause distribution $p_\cb$ and a mechanism $p_{\eb|\cb}$ (independent objects are highlighted in blue and red). (b)  IMA enforces independence between the contributions of different sources $s_i$ to the mixing function $\fb$ as captured by $\nicefrac{\partial \fb}{\partial s_i}$.}
\end{figure}
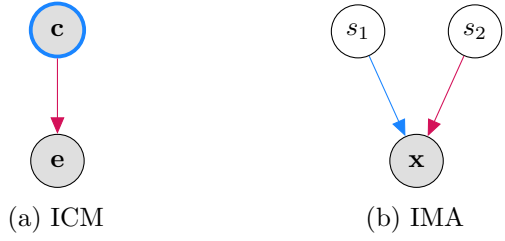

\looseness-1 As illustrated in~\cref{fig:ICM_graph}, condition%
~\eqref{eq:IGCI_condition} and other similar criteria\footnote{Many ICM criteria can be phrased as special cases of a unifying group-invariance framework~\citep{besserve2018group}.} enforce a notion of ``genericity'' or ``decoupling'' of the mechanism w.r.t.\ the \emph{observed} input distribution. 
They thus rely on the fact that the cause (source) distribution is informative, and are generally not invariant to reparametrisation of the cause variables.
In the (nonlinear) ICA setting, on the other hand, the \emph{learnt} source distribution may be fairly uninformative.
This poses a challenge for existing ICM criteria since any mechanism is generic w.r.t.\ an uninformative (uniform) input distribution. 

\section{Independent Mechanism Analysis (IMA)}
\label{sec:IMA}
\looseness-1 As argued in~\cref{sec:unsuitability_of_existing_ICM_measures}, enforcing independence between the input distribution and the mechanism (\cref{fig:ICM_graph}), as existing ICM criteria do, is insufficient for ruling out spurious solutions to nonlinear ICA.
We therefore propose a new ICM-inspired framework which
is more suitable for BSS and which
we term \textit{independent mechanism analysis} (IMA).
All proofs are provided in~\Cref{app:proofs_IMA}.

\subsection{Intuition Behind IMA
}
\label{sec:ima_intuition}
As motivated using the cocktail party example in~\cref{sec:IMA_introduction} and~\cref{fig:intuition_pic},
our main idea is to enforce
a notion of 
\textit{independence between the contributions or influences of the different sources $s_i$ on the observations} $\xb=\fb(\sb)$
as illustrated in~\cref{fig:IMA_graph}---as opposed to between the source distribution and mixing function, cf.~\cref{fig:ICM_graph}.
These contributions or influences are captured by the vectors  of partial derivatives $\nicefrac{\partial \fb}{\partial s_i}$.
IMA can thus be understood as a \textit{refinement of ICM at the level of the mixing $\fb$}:
in addition to
\textit{statistically independent components}~$s_i$, we look for a mixing with \textit{contributions $\nicefrac{\partial \fb}{\partial s_i}$ which are independent}, in a non-statistical sense which we formalise %
as follows.
\begin{principle}[IMA]
\label{principle:IMA}
The mechanisms by which each source $s_i$ influences the observed distribution, as captured by the 
partial derivatives $\nicefrac{\partial\fb}{\partial s_i}$, are independent of each other in the sense that for all~$\sb$:%
\begin{equation}
\label{eq:IMA_principle}
  \log  |\det \Jb_\fb(\sb)|
    = 
    \sum_{i=1}^n
    \log \norm{\frac{\partial \fb}{\partial s_i}(\sb)} 
\end{equation}
\end{principle}%

\paragraph{Geometric Interpretation.}
Geometrically, the IMA principle can be understood as an \textit{orthogonality condition}, as illustrated for $n=2$ in~\cref{fig:intuition_orthogonal}.
First, the vectors of partial derivatives~$\nicefrac{\partial \fb}{\partial s_i}$, for which the IMA principle postulates independence, are the \textit{columns} of%
~$\Jb_\fb$.
$|\det \Jb_\fb|$ 
thus measures the volume of the $n-$dimensional parallelepiped spanned by these columns, as shown on the right of~\cref{fig:intuition_orthogonal}.
The product of their norms, 
on the other hand, corresponds to the volume of an $n$-dimensional box, or rectangular parallelepiped with side lengths $\smallnorm{\nicefrac{\partial \fb}{\partial s_i}}$, as shown on the left of~\cref{fig:intuition_orthogonal}.
The two volumes are equal if and only if all columns $\nicefrac{\partial \fb}{\partial s_i}$ of $\Jb_\fb$ are orthogonal.
\begin{remark}[IMA in one dimension]
Note that%
~\eqref{eq:IMA_principle} is trivially satisfied for $n=1$, that is, if there is no mixing. This further highlights its difference from ICM for causal discovery.
\end{remark}

\paragraph{Independent Influences and Orthogonality.}
\looseness-1 In a high dimensional setting (large $n$), this orthogonality can be intuitively interpreted from the ICM perspective as \emph{Nature choosing the direction of the influence of each source component in the observation space independently and from an isotropic prior}. Indeed, it can be shown that the scalar product of two independent isotropic random vectors in $\RR^n$ vanishes as the dimensionality $n$ increases (equivalently: two high-dimensional isotropic vectors are typically orthogonal). This property was previously exploited in other linear ICM-based criteria, see~\citet[][Lemma 5]{janzing2018detecting} and%
~\citet[][Lemma 1 \& Thm.\ 1]{janzing2010telling}%
, and has also been used as a \text{``leading intuition''} to interpret IGCI~\citep{janzing2012information}. %
The principle in~\eqref{eq:IMA_principle} can be seen as a constraint on the function space, enforcing such orthogonality between the columns of the Jacobian of $\fb$ at all points in the source domain, %
thus approximating the high-dimensional behaviour described above.

\paragraph{Information-Geometric Interpretation and Comparison to IGCI.}
The additive contribution of the sources' influences $\nicefrac{\partial \fb}{\partial s_i}$ in~\eqref{eq:IMA_principle}
suggests their local \textit{decoupling at the level of the mechanism}~$\fb$.
\looseness-1 Note that IGCI~\eqref{eq:IGCI_condition}, on the other hand, postulates a different type of decoupling: one between $\log |\det \Jb_\fb|$ and~$p_\sb$. There, dependence between  cause and mechanism can be conceived as a fine tuning between the derivative of the mechanism and the input density. %
The IMA principle leads to a complementary, non-statistical measure of independence between the influences~$\nicefrac{\partial \fb}{\partial s_i}$ of the individual sources on the vector of observations.
Both the IGCI and IMA postulates have an information-geometric interpretation related to the influence of (``non-statistically'') independent modules on the observations: both lead to an \textit{additive decomposition of a KL-divergence between the effect distribution and a reference distribution.} 
For IGCI, independent modules correspond to the cause distribution and the mechanism mapping the cause to the effect.
 For IMA, on the other hand, these are the influences of each source component on the observations in an interventional setting (under soft interventions on individual sources), as measured by the KL-divergences between the original and intervened distributions. See Appendix B.3 of~\citet{gresele2021independent} for a more detailed account.  %

\begin{remark}[``Mechanism'' terminology]  
While recent work based on the ICM principle has mostly used the term ``mechanism'' to refer to causal Markov kernels $p(X_i~|~\PA_i)$
or structural equations~\citep{peters2017elements}, we here employ it %
in line with the broader use of this concept in the philosophical literature, see~\citet[][Table 1]{mahoney2001beyond} for a long list of definitions. %
To highlight just two examples: {``Causal processes, causal interactions, and causal laws provide the mechanisms by which the world works; to understand why certain things happen, we need to see how they are produced by these mechanisms''}~\citep{salmon2020scientific}; and ``Mechanisms are events that alter relations among some specified set of elements''~\citep{tilly2001historical}. Following this perspective, we argue that a causal mechanism can more generally denote \textit{any process that describes the way in which causes influence their effects}: the partial derivative $\nicefrac{\partial\fb}{\partial s_i}$
thus reflects a causal mechanism in the sense that it describes the infinitesimal changes in the observations $\xb$, when an infinitesimal perturbation is applied to $s_i$.
\end{remark}

\subsection{Definition and Useful Properties of the IMA Contrast}
\label{sec:ima_definition_properties}
We now introduce a contrast function based on the IMA principle~\eqref{eq:IMA_principle} 
and show that it possesses several desirable properties in the context of nonlinear ICA.
First, we define a local contrast as the difference between the two sides of~\eqref{eq:IMA_principle} for a particular value of the sources $\sb$.
\begin{definition}[Local IMA contrast]
\label{def:local_IMA_contrast}
The local IMA contrast $c_\IMA(\fb,\sb)$ of
$\fb$ at a point $\sb$ is
given by
\begin{equation}
\label{eq:adm_single_point} 
c_\IMA(\fb,\sb) = \sum_{i = 1}^n \log \norm{\frac{\partial \fb}{\partial s_i}(\sb)} - \log \left |\det \Jb_\fb(\sb)\right|\,.
\end{equation}%
\end{definition}%
\begin{remark}
\label{remark:left_KL}
This corresponds to the left KL measure of diagonality~\citep{alyani2017diagonality} for
$\sqrt{\Jb_\fb(\sb)^\top\Jb_\fb(\sb)}$.
\end{remark}%
The local IMA contrast $c_\IMA(\fb,\sb)$ 
quantifies the extent to which the IMA principle is violated at a given point $\sb$.
We summarise some of its properties in the following proposition.
\begin{restatable}[Properties of $c_\IMA(\fb,\sb)$]
{proposition}{impropties}
    \label{prop:local_IMA_contrast_properties}
    The local IMA contrast $c_\IMA(\fb,\sb)$ defined in~\eqref{eq:adm_single_point} satisfies:%
    \begin{enumerate}[(i), topsep=0pt,itemsep=0pt]
        \item $c_\IMA(\fb,\sb) \geq 0$, with equality if and only if all columns $\nicefrac{\partial \fb}{\partial s_i}(\sb)$ of $\Jb_\fb(\sb)$ are orthogonal. %
        \item $c_\IMA(\fb,\sb)$ is invariant to left multiplication of $\Jb_\fb(\sb)$ by an orthogonal matrix and to right multiplication by permutation and diagonal matrices.%
    \end{enumerate}%
\end{restatable}
Property \textit{(i)} formalises the geometric interpretation of IMA as an orthogonality condition on the columns of the Jacobian from~\cref{sec:ima_intuition}, and property \textit{(ii)} intuitively states that changes of orthonormal basis and permutations or rescalings of the columns of $\Jb_\fb$ do not affect their orthogonality. 

\looseness-1 Next, we define a global IMA contrast w.r.t.\ a source distribution $p_\sb$ as the expected local IMA contrast.
\begin{definition}[Global IMA contrast]
\label{def:global_IMA_contrast}
The 
global IMA contrast $C_\IMA(\fb,p_\sb)$ of
$\fb$ 
w.r.t.\
$p_\sbb$ is given by 
\begin{equation}
    C_\IMA(\fb, p_\sbb) = \EE_{\sb\sim p_\sb}[c_\IMA(\fb,\sb)] = \medint\int c_\IMA(\fb,\sb) p_\sb(\sb)  d\sb\,.
    \label{eq:adm_metric}
\end{equation}%
\end{definition}%
The global IMA contrast $C_\IMA(\fb,p_\sb)$ thus quantifies the extent to which the IMA principle is violated for a particular solution $(\fb, p_\sb)$ to the nonlinear ICA problem.
We summarise its properties as follows.%
\begin{restatable}[Properties of $C_\IMA(\fb,p_\sb)$]
{proposition}{admproperties}
\label{prop:global_IMA_contrast_properties}
The global IMA contrast $C_\IMA(\fb, p_\sbb)$ from~\eqref{eq:adm_metric}
satisfies:%
    \begin{enumerate}[(i), topsep=0pt,itemsep=0pt]
        \item $C_\IMA(\fb, p_\sbb) \geq 0$, with equality
        iff.\ $\Jb_{\fb}(\sb) = \Ob(\sb) \Db(\sb)$ almost surely w.r.t.\ $p_\sb$, where $\Ob(\sb), \Db(\sb)\in\RR^{n\times n}$ are orthogonal and  diagonal matrices, respectively;
        \item $C_\IMA(\fb, p_\sbb) = C_\IMA(\fbt, p_{\sbt})$ for any $\fbt=\fb\circ\hb^{-1}\circ \Pb^{-1}$ and $\sbt = \Pb\hb(\sb)$,  where $\Pb\in\RR^{n\times n}$ is a permutation and $\hb(\sb)=(h_1(s_1), ..., h_n(s_n))$ an invertible element-wise function.
    \end{enumerate}%
\end{restatable}%
\begin{figure}[t]
\centering
\begin{subfigure}{0.3\textwidth}
\includegraphics[height=\textwidth]{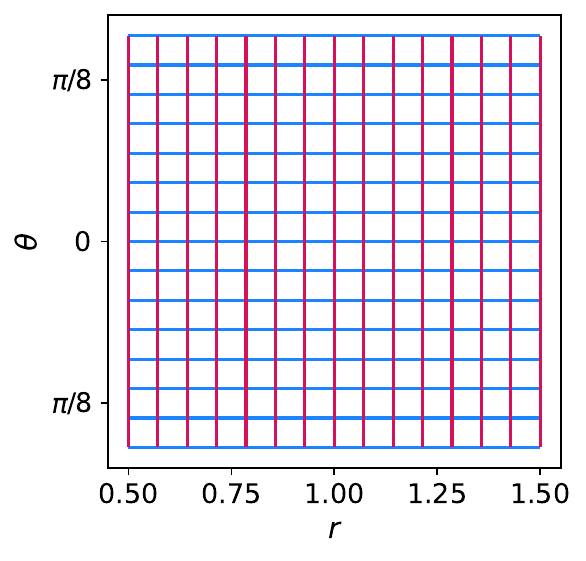}
\end{subfigure}%
\hspace{3em}
\begin{subfigure}{0.3\textwidth}
\includegraphics[height=\textwidth]{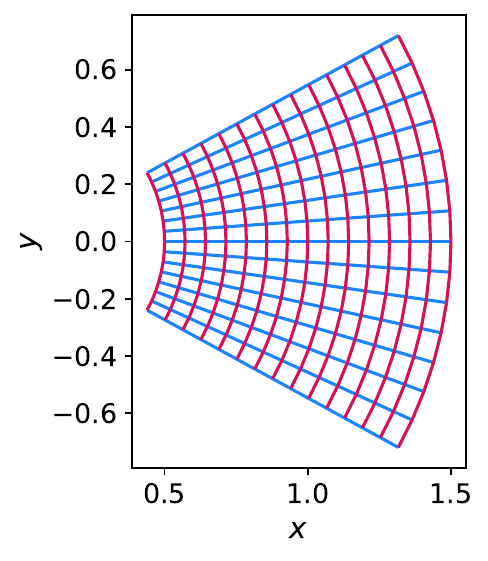}
\end{subfigure}%
\vspace{-0.5em}
\caption[Example of a (Non-Conformal) Orthogonal Coordinate Transformation]{\textbf{Orthogonal Coordinate Transformations.} An example of an orthogonal coordinate transformation from polar (left) to Cartesian (right) coordinates, which is not conformal.}
\label{fig:orthogonal_coordinate_transform}
\end{figure}
Property \textit{(i)} is the distribution-level analogue to \textit{(i)} of~\Cref{prop:local_IMA_contrast_properties} and only allows for orthogonality violations on sets of measure zero w.r.t.\ $p_\sb$.
This means that $C_\IMA$ can only be zero if $\fb$ is an \textit{orthogonal coordinate transformation} almost everywhere~\citep{lame1859leccons, darboux1910leccons,moon1971field}, see~\cref{fig:orthogonal_coordinate_transform} for an example.
We 
particularly
stress property \textit{(ii)}, as it precisely matches the inherent
indeterminacy
of nonlinear ICA: %
\textit{$C_\IMA$ is blind to reparametrisation of the sources by permutation and element wise transformation}.

\subsection{Theoretical Analysis and Justification of the IMA Contrast}
\label{sec:ima_theory}
We now show that, under suitable assumptions on the generative model~\eqref{eq:gen}, a large class of
spurious solutions---such as those based on the Darmois construction~\eqref{eq:Darmois_construction} or 
measure-preserving automorphisms such as $\ab^\Rb$ from~\eqref{eq:measure_preserving_automorphism_Gaussian} as described in~\cref{sec:background_ICA}---%
exhibit nonzero IMA contrast.
Denote the class of nonlinear ICA models satisfying~\eqref{eq:IMA_principle} (IMA) by 
\[\Mcal_{\IMA}=\{(\fb, p_\sb)\in \Fcal\times\Pcal: C_\IMA(\fb, p_\sb)=0\}\subset \Fcal\times\Pcal.\]
Our first main theoretical result is that, under mild assumptions on the observations, Darmois solutions will have strictly positive $C_\IMA$, making them distinguishable from those in~$\Mcal_\IMA$.%
\begin{restatable}{theorem}{admdarmois}
\label{thm:adm_darmois}
Assume the data generating process in~\eqref{eq:gen}
and assume that
$x_i \nindep x_j$ for some $i \neq j$. Then any  Darmois solution $(\fb^\text{D}, p_\ub)$ based on $\gb^\text{D}$ as defined in~\eqref{eq:Darmois_construction} satisfies $C_\IMA(\fb^\text{D}, p_\ub)>0$.
Thus a solution satisfying $C_\IMA(\fb, p_\sb)=0$ can be distinguished from $(\fb^\text{D}, p_\ub)$ based on the contrast $C_\IMA$.
\end{restatable}%
The proof is based on the fact that the Jacobian of $\gb^\text{D}$ is triangular 
and on the form of~\eqref{eq:Darmois_construction}.

A specific example %
of a mixing process satisfying the IMA assumption
is the case where $\fb$ is a conformal (angle-preserving) map. 

\begin{definition}[Conformal map]
\label{def:conformal_map}
A smooth map $\fb:\RR^n\rightarrow\RR^n$ is conformal if $\Jb_\fb(\sb)=\Ob(\sb)\lambda(\sb)$ $\forall\sb$, where $\lambda:\RR^n\rightarrow\RR$ is a scalar field,
and $\Ob\in O(n)$ is an orthogonal matrix.%
\end{definition}%

\begin{restatable}
{corollary}{confmapsadm}
\label{cor:IMA_identifiability_of_conformal_maps}
Under the assumptions of~\Cref{thm:adm_darmois}, if additionally $\fb$ is a conformal map, then $(\fb,p_\sb)\in\Mcal_\IMA$ for any $p_\sb\in\Pcal$ due to~\cref{prop:global_IMA_contrast_properties} \textit{(i)}, see~\cref{def:conformal_map}. 
Based on~\cref{thm:adm_darmois}, $(\fb,p_\sb)$ is thus distinguishable from  Darmois solutions~$(\fb^\text{D}, p_\ub)$.%
\end{restatable}%

This 
is consistent with
a result
that proves identifiability of conformal maps
for $n=2$ and conjectures it in general~\citep{hyvarinen1999nonlinear}. 
(Note that~\Cref{cor:IMA_identifiability_of_conformal_maps} holds for any dimensionality $n$.)
However, conformal maps are only a (small) subset of all maps for which $C_\IMA=0$. 
This is apparent from the more flexible condition of~\Cref{prop:global_IMA_contrast_properties}~\textit{(i)}, compared to the stricter~\Cref{def:conformal_map}:
unlike for conformal maps, the diagonal entries (Jacobian column norms) can be distinct for IMA functions.

\begin{example}[Polar to Cartesian coordinate transform]
\label{ex:polar}
Consider the \textit{non-conformal}
transformation from polar to Cartesian coordinates~(see~\cref{fig:orthogonal_coordinate_transform}), defined as 
\[(x,y)=\fb(r,\theta):=(r\cos(\theta),r\sin(\theta))\]
 with independent sources $\sb=(r,\theta)$, with $r\sim U(0,R)$ and $\theta\sim U(0, 2\pi)$.
Then, $C_\IMA(\fb,p_\sb)=0$ and $C_\IMA(\fb^\text{D}, p_\ub)>0$ for any
Darmois solution $(\fb^\text{D}, p_\ub)$~\citep[][Appendix D]{gresele2021independent}.
\end{example}

\begin{remark}
For different $p_\sb$, $(x,y)$ in~\cref{ex:polar} can be made to have independent Gaussian components~\citep[][II.B]{taleb1999source}, and
$C_\IMA$-identifiability is lost. This shows that the assumption of~\Cref{thm:adm_darmois} that $x_i \nindep x_j$ for some $i \neq j$ is crucial.
\end{remark}

Finally, for the case in which the true mixing is linear, we obtain the following result.%
\begin{restatable}{corollary}{admidentlinear}
\label{cor:IMA_identifiability_of_linear_ICA}
Consider a linear ICA model, $\xb=\Ab\sb$, with $\EE[\sb^\top\sb]=\Ib$, and $\Ab\in O(n)$ an orthogonal, non-trivial mixing matrix, i.e., not the product of a diagonal and a permutation matrix~$\Db \Pb$.
If at most one of the $s_i$ is Gaussian, then $C_\IMA(\Ab, p_\sb)=0$ and $C_\IMA(\fb^\text{D}, p_\ub)>0$.%
\end{restatable}

In a ``blind'' setting, we may not know a priori whether the true mixing is linear or not, and thus choose to learn a nonlinear unmixing.
\Cref{cor:IMA_identifiability_of_linear_ICA} shows that, in this case, Darmois solutions are still distinguishable from the true mixing via $C_\IMA$.
Note that unlike in~\cref{cor:IMA_identifiability_of_conformal_maps}, the assumption that $x_i \nindep x_j$ for some $i \neq j$ is not required for~\cref{cor:IMA_identifiability_of_linear_ICA}. In fact, due to Theorem 11 of~\cite{comon1994independent}, it follows from the assumed linear ICA model with non-Gaussian sources, and the fact that the mixing matrix  is not the product of a diagonal and a permutation matrix.

Having shown that the IMA principle allows distinguishing a class of models (including, but not limited to conformal maps) from Darmois solutions, we next turn to a second well-known counterexample to identifiability: the ``rotated-Gaussian'' MPA $\ab^\Rb(p_\sb)$~\eqref{eq:measure_preserving_automorphism_Gaussian} from~\Cref{def:measure_preserving_automorphism_Gaussian}.
Our second main theoretical result is that, under suitable assumptions, this class of MPAs can also be ruled out for ``non-trivial'' $\Rb$.%
\begin{restatable}{theorem}{thmMPA}
\label{thm:IMA_identifiability_measure_preserving_automorphism}
Let $(\fb,p_\sb)\in\Mcal_\IMA$ and assume that $\fb$ is a conformal map.
Given $\Rb\in O(n)$, assume
additionally
that 
    $\exists$ at least one non-Gaussian 
    $s_i$ whose associated canonical 
    basis vector $\eb_i$ is not transformed by $\Rb^{-1}=\Rb^\top$ into another canonical basis vector $\eb_j$. 
Then $C_\IMA(\fb\circ \ab^\Rb(p_\sb),p_\sb)>0$.%
\end{restatable}

\Cref{thm:IMA_identifiability_measure_preserving_automorphism} states that for conformal maps, applying the $\ab^\Rb(p_\sb)$ transformation at the level of the sources leads to an increase in $C_\IMA$, except for very specific rotations $\Rb$ that are ``fine-tuned'' to $p_\sb$ in the sense that they permute all non-Gaussian sources $s_i$ with another $s_j$.
Interestingly, as for the linear case, non-Gaussianity also plays an important role in the proof of~\Cref{thm:IMA_identifiability_measure_preserving_automorphism}.

\section{Experiments}
\label{sec:IMA_experiments}

Our theoretical results from~\cref{sec:IMA} suggest that $C_\IMA$ is a promising contrast function for nonlinear blind source separation. 
We test this empirically by evaluating the IMA contrast $C_\IMA$ of spurious nonlinear ICA solutions~(\cref{sec:experiment1_evaluation}), and using it as a learning objective to recover the true solution~(\cref{sec:experiment2_learning}).

We sample the ground truth sources from a uniform distribution in $[0,1]^n$; the reconstructed sources
are also mapped to the uniform hypercube as a reference measure via the CDF transform.
Unless otherwise specified, the ground truth mixing~$\fb$ is a M\"obius transformation~\citep{phillips1969liouville} (i.e., a conformal map) with randomly sampled parameters, thereby satisfying~\cref{principle:IMA}. In all of our experiments, we use JAX~\citep{jax2018github} and Distrax~\citep{distrax2021github}. For additional technical details, equations and plots see~Appendix E of~\citet{gresele2021independent}. The code to reproduce our experiments is available at \href{https://github.com/lgresele/independent-mechanism-analysis}{https://github.com/lgresele/independent-mechanism-analysis}.

\subsection{Numerical Evaluation of the IMA Contrast for Spurious Solutions
}
\label{sec:experiment1_evaluation}

\paragraph{Learning the Darmois Construction.} To
learn the Darmois construction from data, we use
 normalising flows~\citep{huang2018neural, papamakarios2021normalizing}. %
Since Darmois solutions have triangular Jacobian,
 we use an
architecture based on 
residual flows~\citep{chen2019residualflows} which we constrain such that the Jacobian of the full model is 
triangular. This yields an expressive model which we can 
train effectively via maximum likelihood.

\newcommand\width{2.25}
\newcommand\height{1.95}
\newcommand\gap{.005}
\newcommand\minipagewidth{.13}
\newcommand\folder{plots_hsv}
\newcommand\widthbottom{2.4}
\newcommand\heightbottom{2.85}
\newcommand\leftplace{-0.9}
\newcommand\lowplace{3.0}
\begin{figure}[t]
    \centering
        \begin{minipage}{\minipagewidth \textwidth}
            \begin{subfigure}{1.0\textwidth}
            \centering
            \includegraphics[height=\height cm, keepaspectratio]{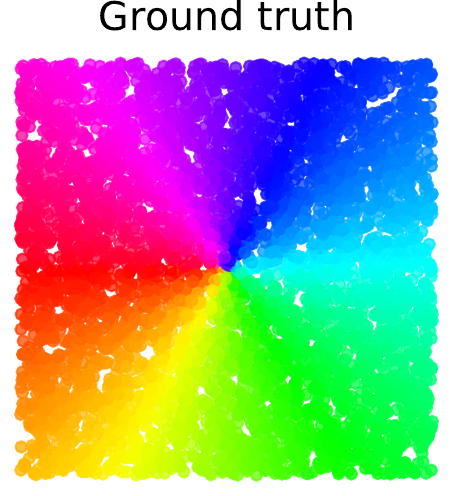}
            \end{subfigure}
        \end{minipage}%
        \hspace{\gap em}
        \begin{minipage}{\minipagewidth \textwidth}
            \begin{subfigure}{1.0\textwidth}
            \centering
            \includegraphics[height=\height cm, keepaspectratio]{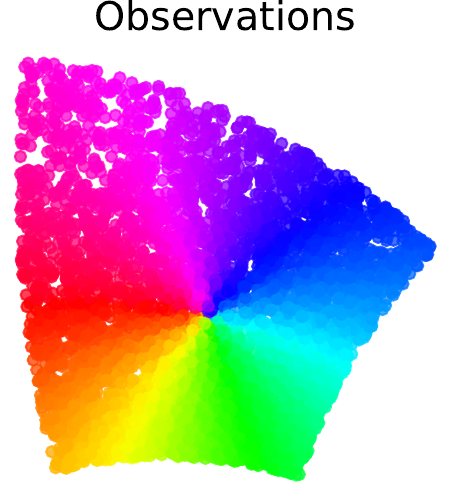}
            \end{subfigure}
        \end{minipage}%
        \hspace{\gap em}
        \begin{minipage}{\minipagewidth \textwidth}
            \begin{subfigure}{1.0\textwidth}
            \centering
            \includegraphics[height=\height cm, keepaspectratio]{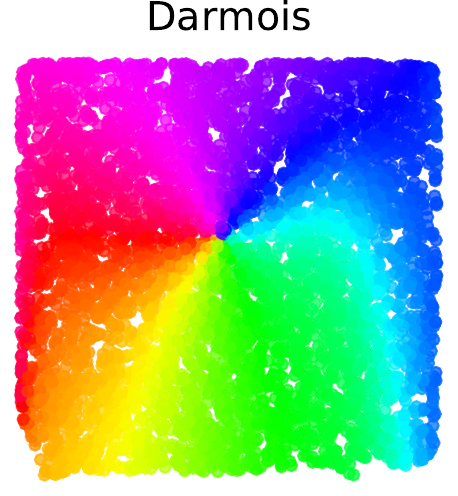}
            \end{subfigure}
        \end{minipage}%
        \hspace{\gap em}
        \begin{minipage}{\minipagewidth \textwidth}
            \begin{subfigure}{1.0\textwidth}
            \centering
            \includegraphics[height=\height cm, keepaspectratio]{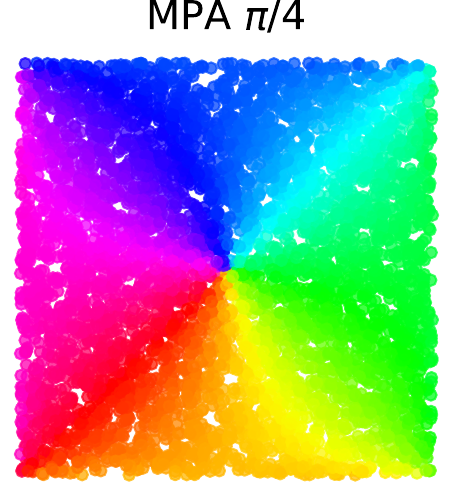}
            \end{subfigure}
        \end{minipage}%
        \hspace{\gap em}
        \begin{minipage}{\minipagewidth \textwidth}
            \begin{subfigure}{1.0\textwidth}
            \centering
            \includegraphics[height=\height cm, keepaspectratio]{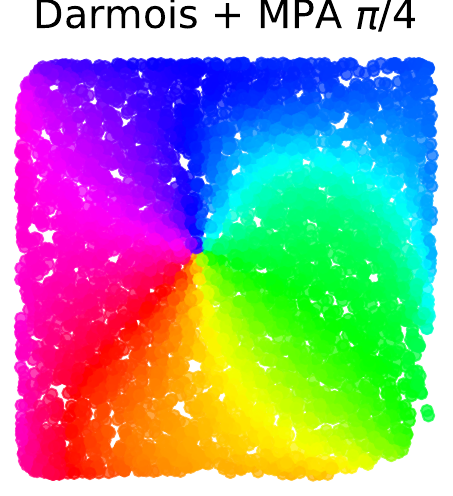}
            \end{subfigure}
        \end{minipage}%
        \hspace{\gap em}
        \begin{minipage}{\minipagewidth \textwidth}
            \begin{subfigure}{1.0\textwidth}
            \centering
            \includegraphics[height=\height cm, keepaspectratio]{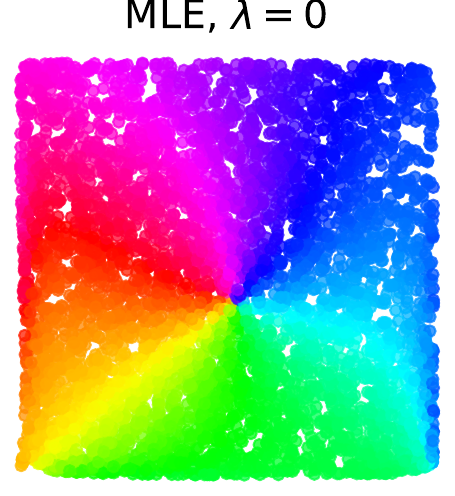}
            \end{subfigure}
        \end{minipage}
        \hspace{\gap em}
        \begin{minipage}{\minipagewidth \textwidth}
            \begin{subfigure}{1.0\textwidth}
            \centering
            \includegraphics[height=\height cm, keepaspectratio]{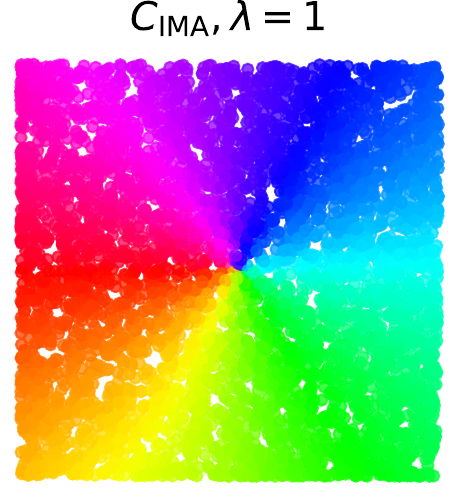}
            \end{subfigure}
        \end{minipage}
    \vspace{0.2em}
    \begin{subfigure}[b]{0.24\textwidth}
        \centering

        \begin{overpic}[height=\heightbottom cm]{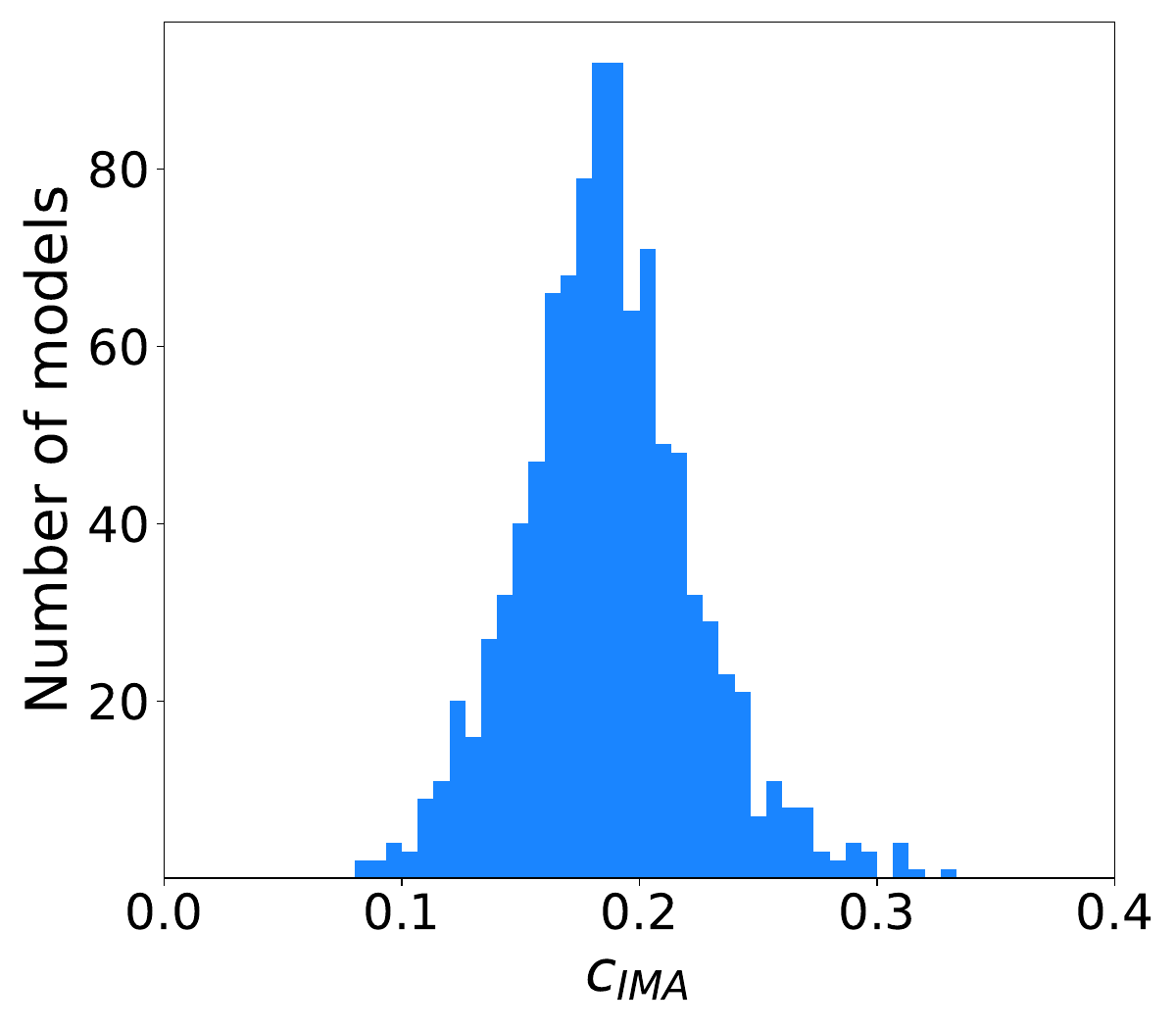}
 \put (\leftplace, \lowplace) {\textbf{\small(a)}}
\end{overpic}
    \end{subfigure}%
    \begin{subfigure}[b]{0.24\textwidth}
        \centering
        \begin{overpic}[height=\heightbottom cm]{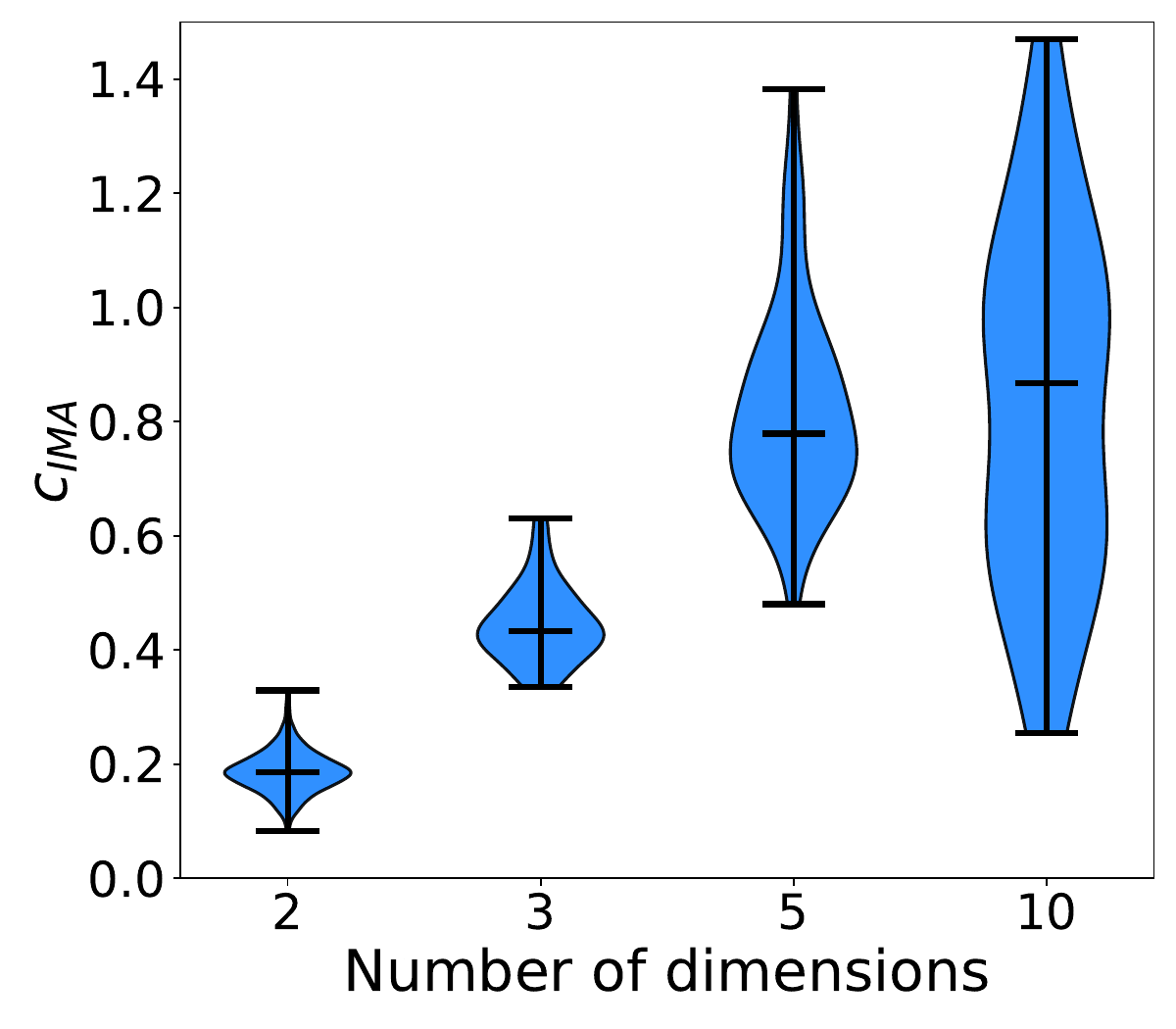}
 \put (\leftplace, \lowplace) {\textbf{\small(b)}}
\end{overpic}
    \end{subfigure}%
    \begin{subfigure}[b]{0.24\textwidth}
        \centering
                \begin{overpic}[height=\heightbottom cm]{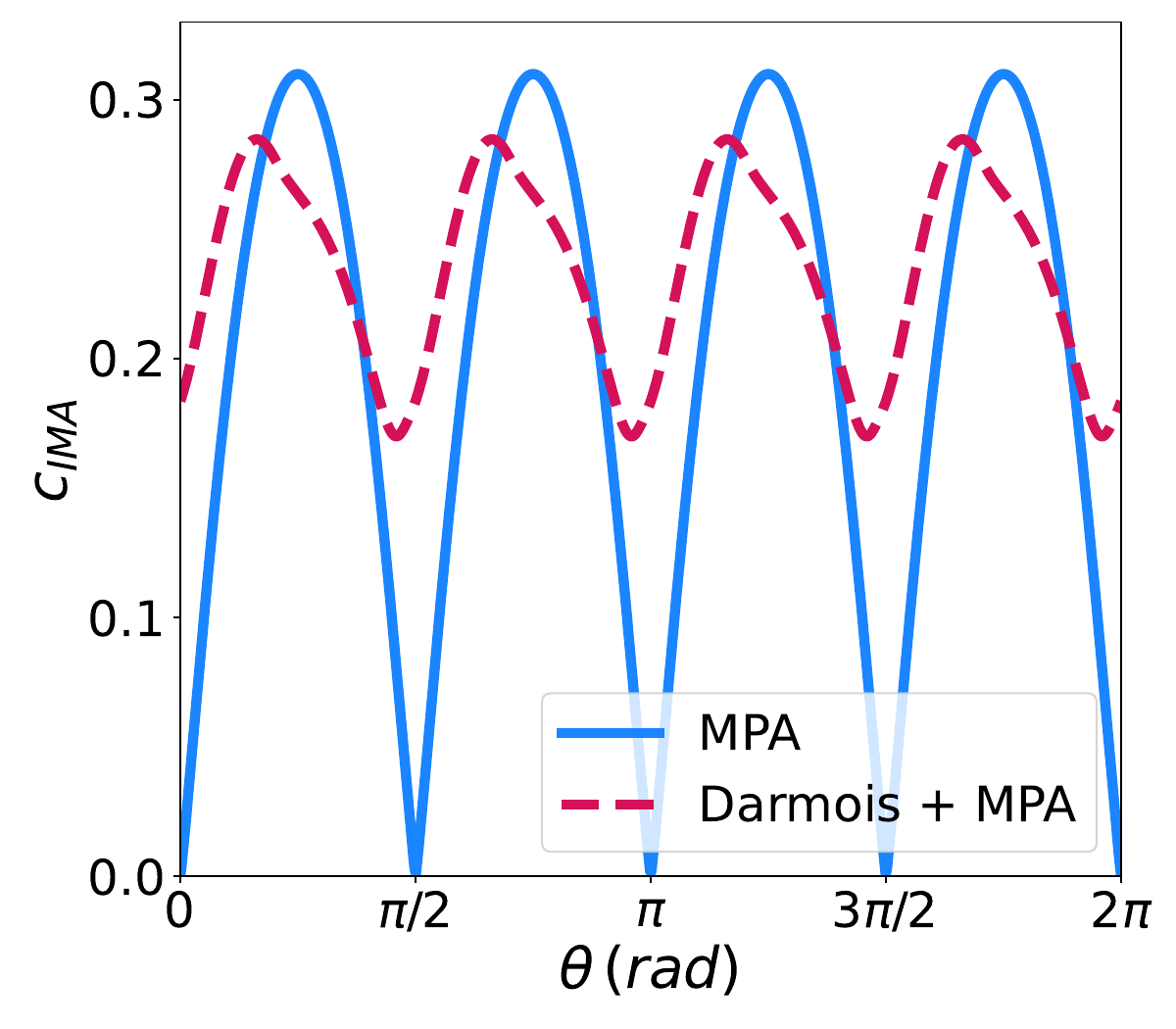}
 \put (\leftplace, \lowplace) {\textbf{\small(c)}}
\end{overpic}
    \end{subfigure}
    \begin{subfigure}[b]{0.24\textwidth}
        \centering
        \begin{overpic}[height=\heightbottom cm]{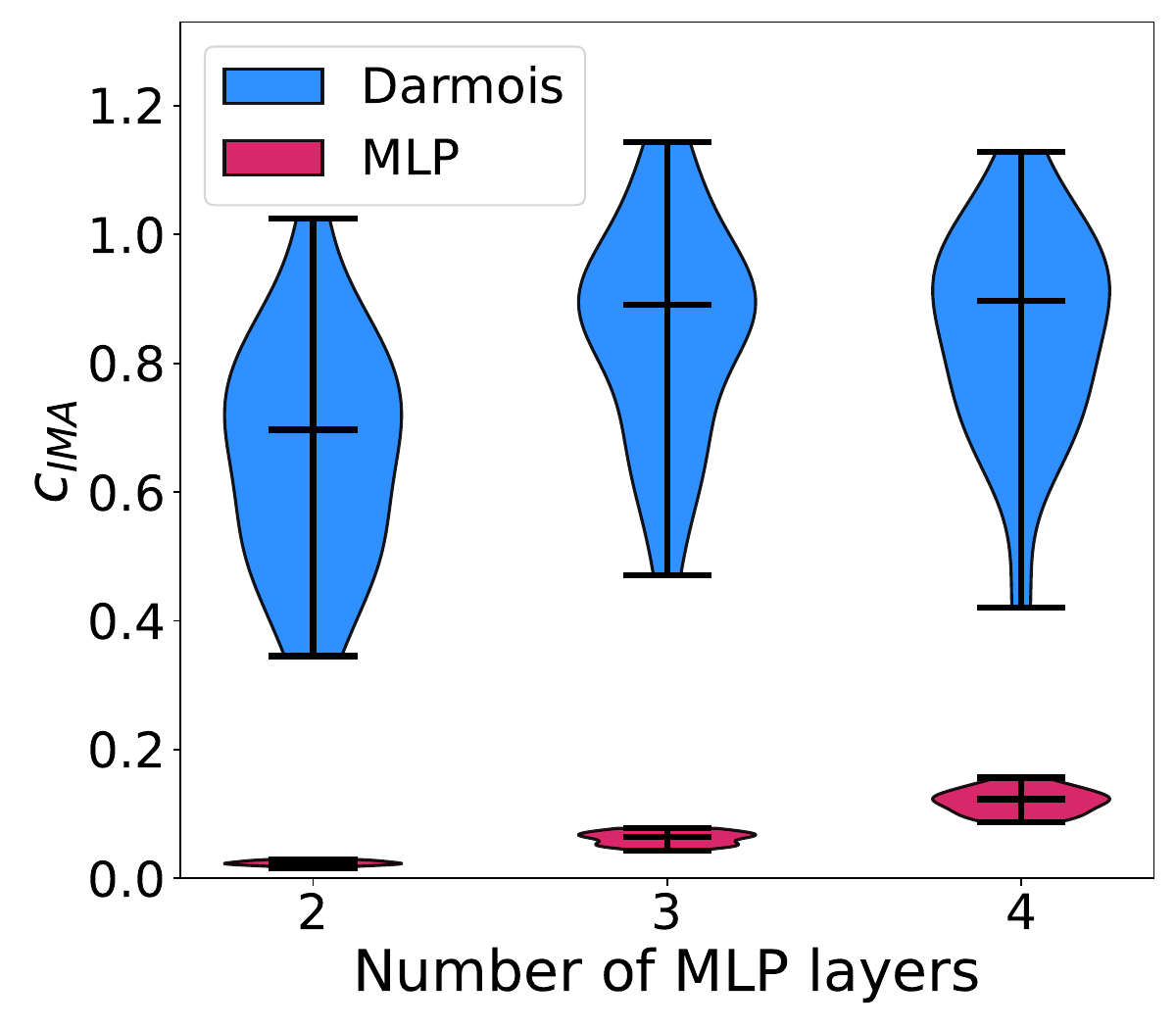}
 \put (\leftplace, \lowplace) {\textbf{\small(d)}}
\end{overpic}
    \end{subfigure}%
    \caption[Comparison of Different Nonlinear ICA solutions]{\textbf{Comparison of Different Nonlinear ICA solutions.} \textit{(Top)} Visual comparison of different nonlinear ICA solutions for $n=2$: \text{(from left to right)} true sources; observed mixtures;  Darmois solution; true unmixing, composed with the measure-preserving automorphism (MPA) from~\eqref{eq:measure_preserving_automorphism_Gaussian} with rotation by~$\nicefrac{\pi}{4}$; Darmois solution composed with the same MPA; maximum likelihood~($\lambda=0$); %
    and $C_\IMA$-regularised approach~($\lambda=1$).
    \textit{(Bottom)} Quantitative comparison of the IMA contrast $C_\IMA$ for different spurious solutions:
    learnt Darmois solutions for \textbf{(a)} $n=2$, and \textbf{(b)} $n\in\{2, 3, 5, 10\}$ dimensions;
    \textbf{(c)} composition of the MPA~\eqref{eq:measure_preserving_automorphism_Gaussian} in $n=2$ dimensions with the true solution (blue) and a  Darmois solution (red) for different angles. \textbf{(d)}~$C_\IMA$ distribution for true MLP mixings outside the IMA function class (red)  vs.\ the corresponding Darmois solution (blue) for $n=5$ dimensions and $L\in\{2,3,4\}$ layers.
    }
    \label{fig:results1}
\end{figure}

\paragraph{IMA Contrast for Darmois Solutions.}
To check whether Darmois solutions (learnt from finite data) can be distinguished from the true one, as predicted by~\cref{thm:adm_darmois}, we generate $1000$ random mixing functions for $n=2$, compute the
$C_{\IMA}$ values of learnt solutions, 
and find that all values are indeed significantly larger than zero, see~\cref{fig:results1} \textbf{(a)}.
The same holds for higher dimensions,
see~\cref{fig:results1} \textbf{(b)} for results with $50$ random mixings for $n\in \{2, 3, 5 ,10\}$: with higher dimensionality, both the mean and variance of the $C_\IMA$ distribution for the learnt Darmois solutions generally attain higher values, the latter possibly due to the increased difficulty of the learning task for larger $n$. We also confirmed these findings for mappings which are not conformal, while still satisfying~\eqref{eq:IMA_principle}~\citep[][Appendix E.5]{gresele2021independent}.

\paragraph{IMA Contrast for MPAs.}
We also investigate the effect on $C_\IMA$ of applying an MPA $\ab^{\Rb}(\cdot)$ from~\eqref{eq:measure_preserving_automorphism_Gaussian} to the true solution or a learnt Darmois solution.
Results for $n=2$ dimensions for different rotation matrices $\Rb$ (parametrised by the angle $\theta$) are shown in~\cref{fig:results1} \textbf{(c)}.
As expected, the behaviour is periodic in $\theta$, and vanishes for the true solution (blue) at multiples of $\nicefrac{\pi}{2}$, i.e., when $\Rb$ is a permutation matrix, as predicted by~\cref{thm:IMA_identifiability_measure_preserving_automorphism}. For the learnt Darmois solution (red, dashed) $C_\IMA$ remains larger than zero.

\paragraph{IMA Contrast for Random MLP mixings.} 
Lastly, we study
the behaviour 
of 
spurious solutions based on the Darmois construction
under deviations from our assumption of $C_\IMA=0$ for the true mixing function.
To this end, we use invertible MLPs with
orthogonal weight initialisation and \texttt{leaky\_tanh} activations~\citep{gresele2020relative} as mixing functions; 
the more layers $L$ are added to the mixing MLP, the larger a deviation from our assumptions is expected. We compare the true mixing and learnt Darmois solutions over $20$ realisations for each $L \in \{2, 3, 4\}$, $n=5$. 
Results are shown in figure~\cref{fig:results1} \textbf{(d)}: the IMA contrast $C_\IMA$ of the mixing MLPs grows with $L$, but its value for Darmois solutions is typically higher.

\paragraph{Summary.} We verify that common spurious solutions can be distinguished from the true one based on the value of the IMA contrast~$C_\IMA$.

\subsection{Learning with IMA-Regularised Maximum Likelihood
}
\label{sec:experiment2_learning}

\paragraph{Experimental Setup.} To use $C_\IMA$ as a learning signal, we consider a regularised maximum-likelihood approach to learn an umixing function $\gb$ with the following objective:
\[\Lcal(\gb) = \EE_{\xb}[\log p_\gb(\xb)] - \lambda \, C_\IMA(\gb^{-1}, p_\yb)\,,\] where $\yb = \gb(\xb)$ denotes the reconstructed sources, and $\lambda\geq0$ a Lagrange multiplier. %
For $\lambda=0$, this corresponds to standard maximum likelihood estimation, whereas for $\lambda>0$, $\Lcal$ lower-bounds the likelihood, and recovers it exactly if and only if $(\gb^{-1},p_\yb)\in\Mcal_\IMA$.
We train a residual flow~$\gb$ (with full Jacobian) to maximise $\Lcal$.
For evaluation, we compute (i) the KL divergence to the true data likelihood, as a measure of goodness of fit for the learnt flow model; and (ii)
the mean correlation coefficient (MCC) between ground truth and reconstructed sources~\citep{hyvarinen2016unsupervised, khemakhem2020variational}.
We also introduce
(iii) a nonlinear extension of the Amari distance~\citep{amari1996new}
between the true mixing and the learnt unmixing, which
is larger than or equal to zero, with equality iff.\ %
the learnt model belongs to the BSS equivalence class~(\cref{def:bss_identifiability}) of the true solution, see Appendix E.5 of~\cite{gresele2021independent} for details. %

\newcommand\heightrow{2.9}
\newcommand\heightrowtwo{2.9}
\newcommand\heightamari{2.853}
\newcommand\heightmcc{2.853}
\newcommand\gapfive{0.002}
\begin{figure}[t]
    \hspace{-0.8 em}
    \begin{subfigure}[b]{0.15\textwidth}
        \centering
        \includegraphics[height=\heightrow cm]{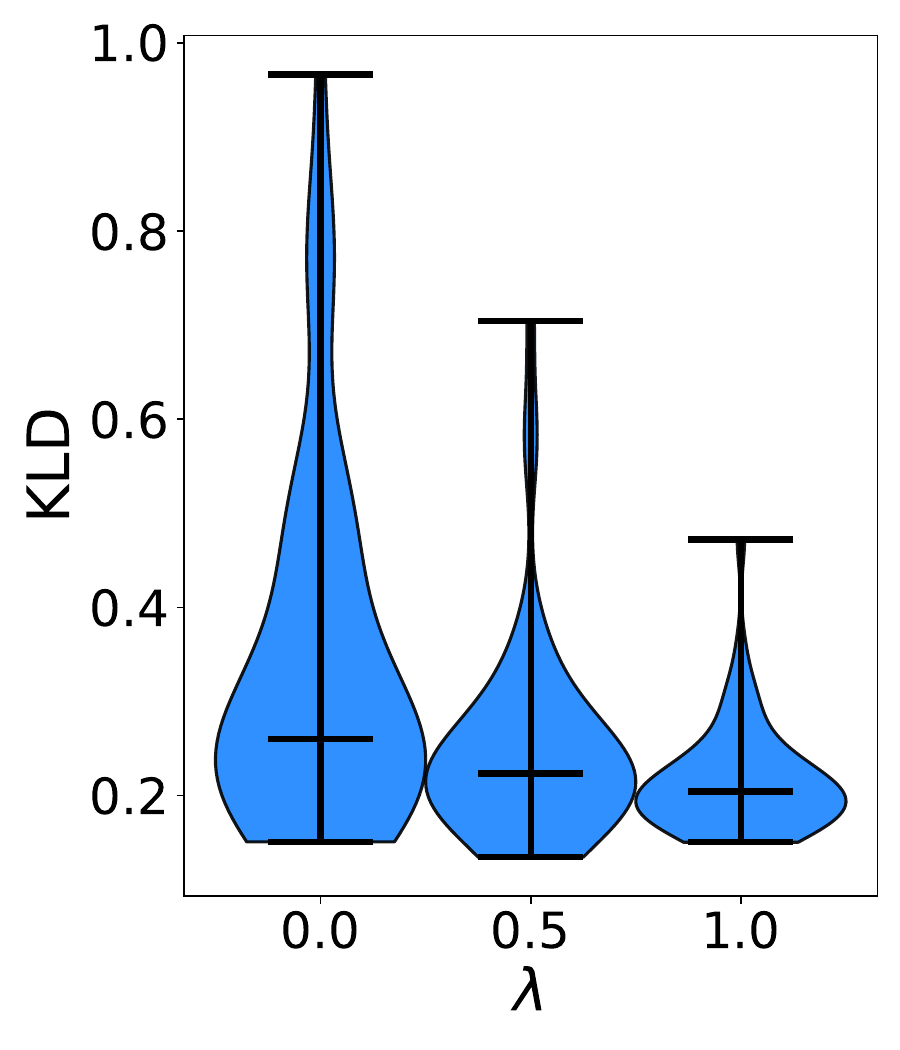}
    \end{subfigure}%
    \hspace{0.8 em}
    \begin{subfigure}[b]{0.15\textwidth}
        \centering
        \includegraphics[height=\heightrow cm]{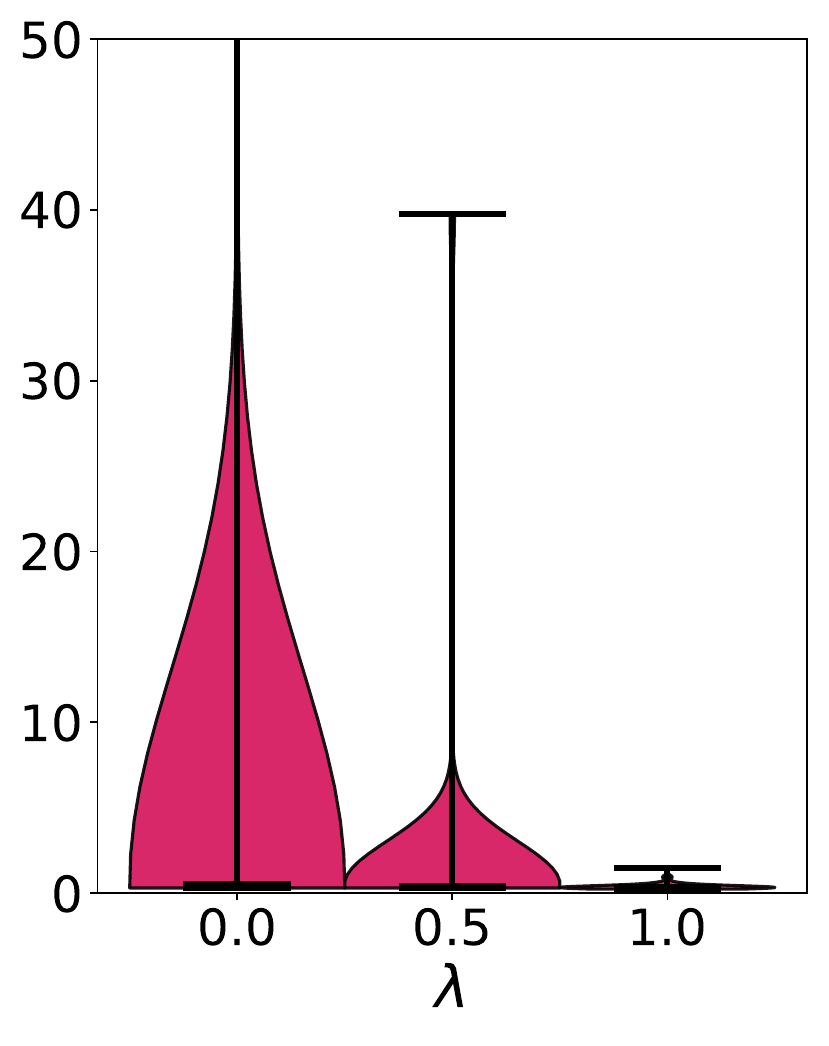}
    \end{subfigure}
    \hspace{0.005 em}
    \begin{subfigure}[b]{0.15\textwidth}
        \centering
        \includegraphics[height=\heightrow cm]{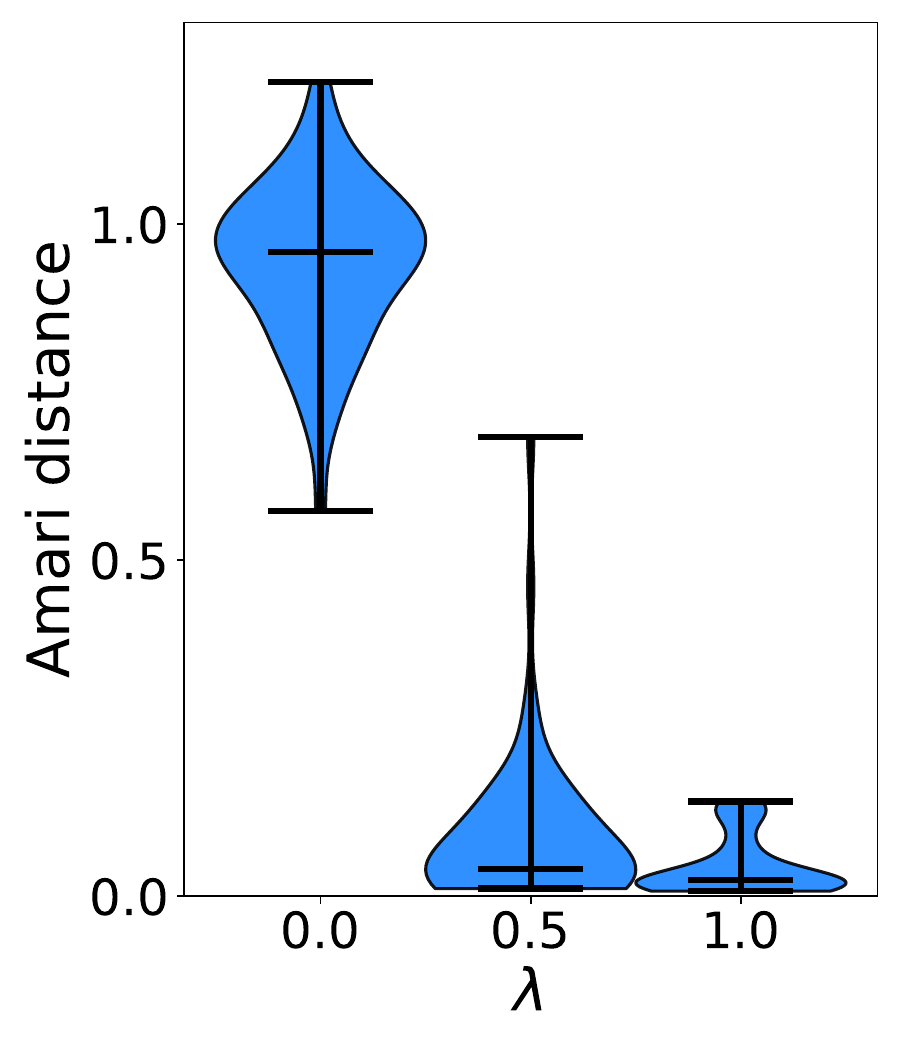}
    \end{subfigure}
    \hspace{0.5 em}
    \begin{subfigure}[b]{0.15\textwidth}
        \centering
        \includegraphics[height=\heightrow cm]{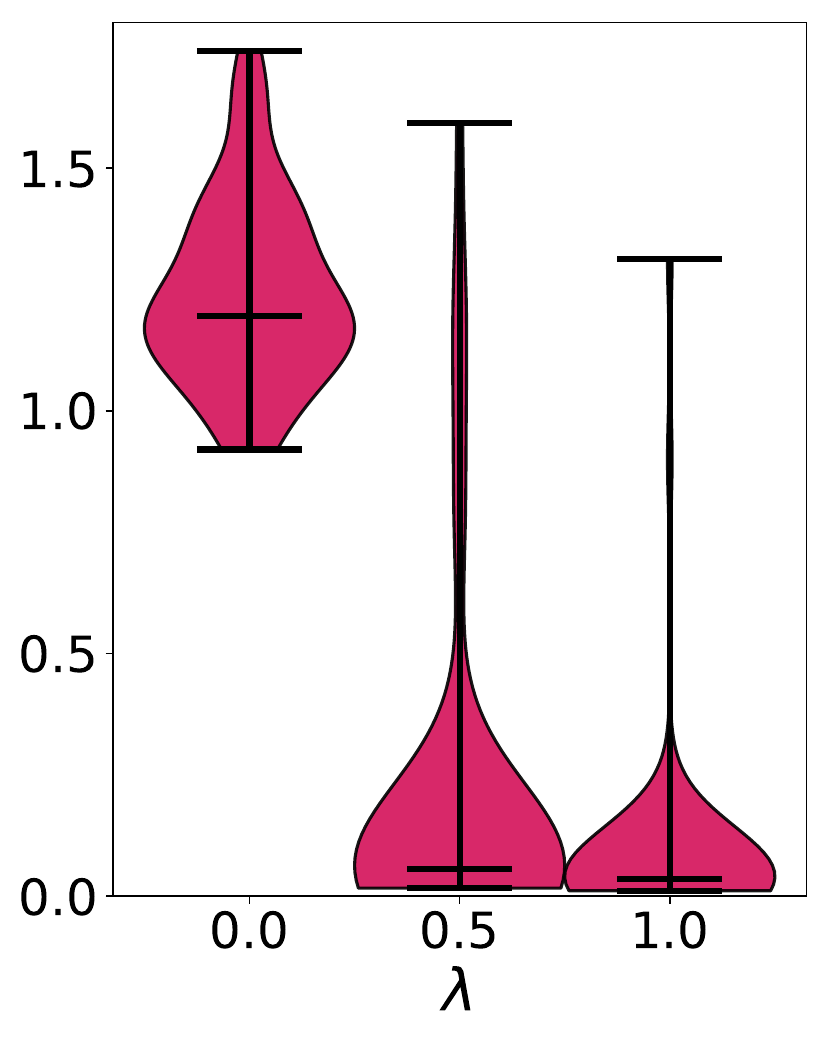}
    \end{subfigure}
    \hspace{0.005 em}
    \begin{subfigure}[b]{0.15\textwidth}
        \centering
        \includegraphics[height=\heightrow cm]{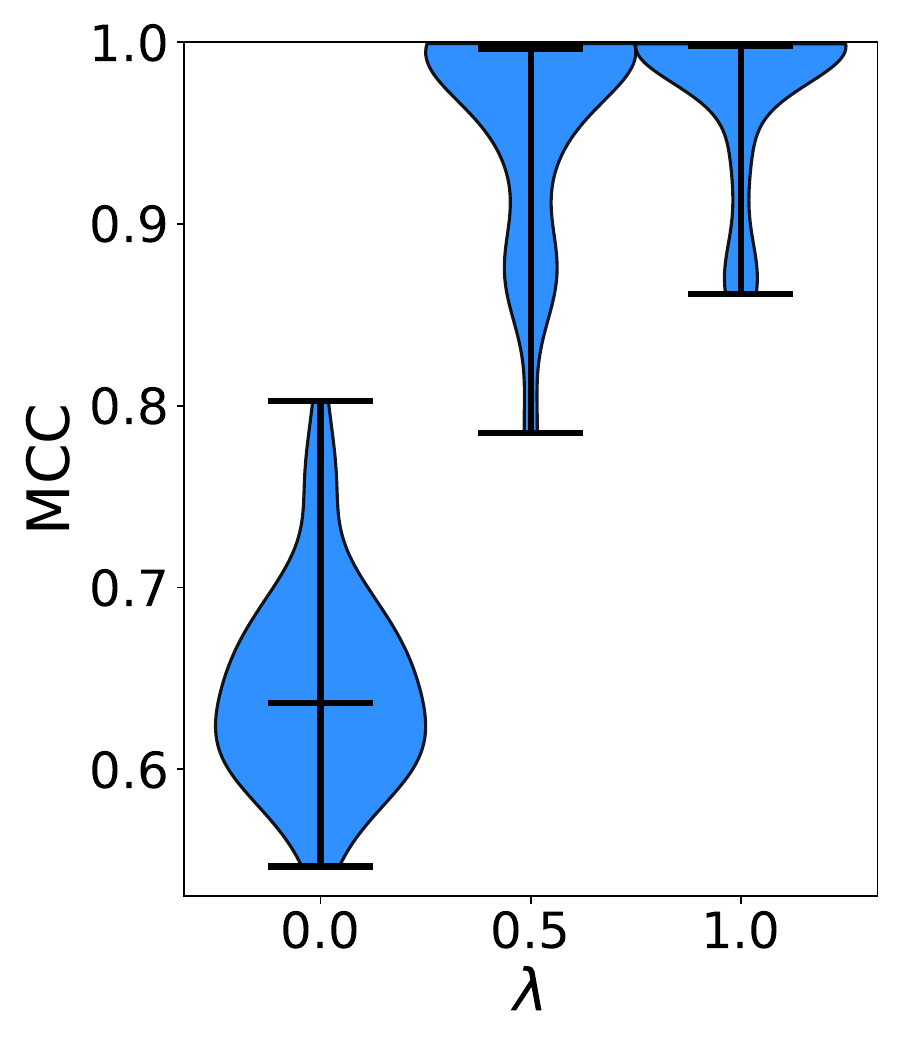}
    \end{subfigure}
    \hspace{0.55 em}
    \begin{subfigure}[b]{0.15\textwidth}
        \centering
        \includegraphics[height=\heightrow cm]{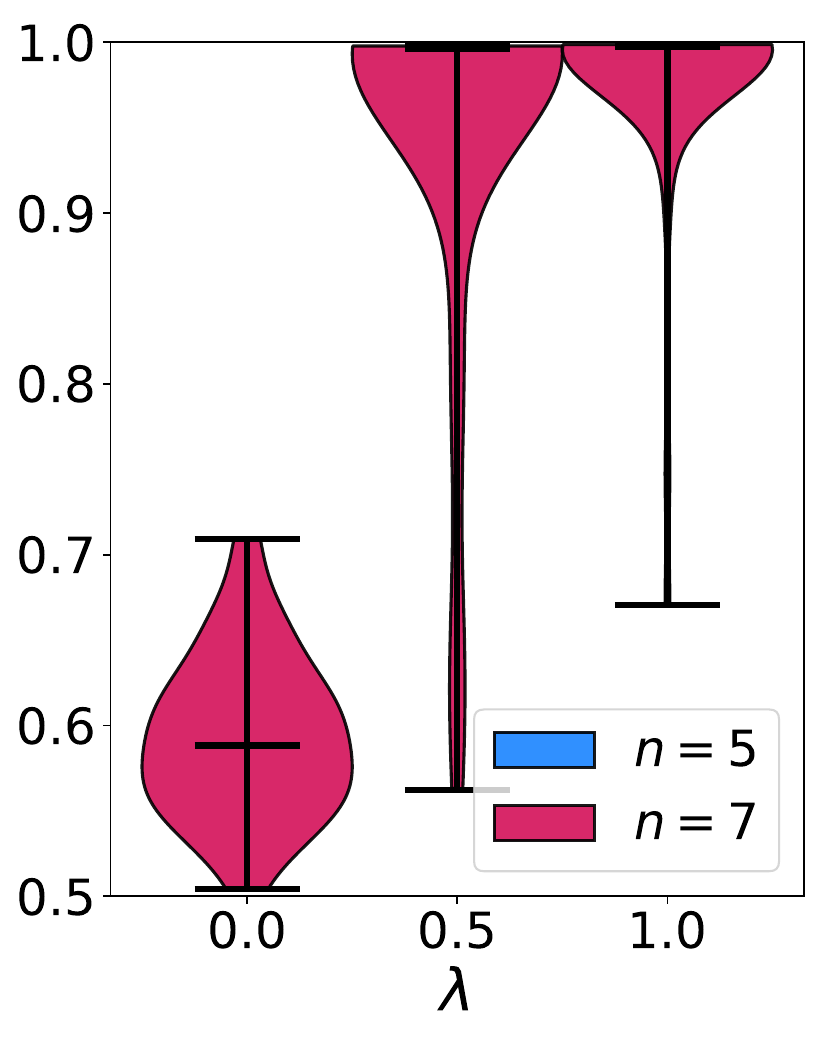}
    \end{subfigure}
    \caption[BSS via $C_\IMA$-Regularised MLE]{\textbf{BSS via $C_\IMA$-Regularised 
    MLE.}
    We show, side by side, $n=5$ (blue) and $n=7$ (red) dimensions with $\lambda\in \{0.0,0.5,1.0\}$.
    \textit{(Left)} KL-divergence between ground truth likelihood and learnt model;
    \textit{(centre)} nonlinear Amari distance given true mixing and learnt unmixing; \textit{(right)} MCC between true and reconstructed sources.
    }
    \label{fig:results2}
\end{figure}

\paragraph{Results.}
In~\cref{fig:results1} \textit{(Top)}, we show an 
example of the distortion induced by different \textit{spurious}
solutions for $n=2$, and contrast it with a solution learnt using our proposed objective \textit{(rightmost plot)}.
Visually, we find that the $C_\IMA$-regularised solution (with $\lambda=1$) recovers the true sources most faithfully.
Quantitative results for 50 learnt models for each $\lambda\in \{0.0, 0.5, 1.0 \}$ and $n\in\{5, 7\}$ are summarised in~\cref{fig:results2}. 
As indicated by the KL divergence values \textit{(left)}, most trained models achieve a good fit to the data across all values of~$\lambda$.
We observe that using $C_\IMA$ (i.e.,
$\lambda>0$) is beneficial for BSS, both in terms of our nonlinear Amari distance \textit{(centre, lower is better)} and  MCC \textit{(right, higher is better)}, though we do not observe a substantial difference between $\lambda=0.5$ and $\lambda=1$.

\paragraph{Summary.} $C_\IMA$ can be a useful learning signal to recover the true solution.

\section{Discussion}
\paragraph{Assumptions on the Mixing Function.} Instead of relying on weak supervision in the form of auxiliary variables~(\cref{sec:background_nonlinear_ICA_auxiliary}),
our IMA approach
places additional constraints on the functional form of the mixing process.
In a similar vein,
the \textit{minimal nonlinear distortion principle}~\citep{zhang2008minimal}
proposes to
favour solutions 
that are
as close to linear as possible. Another  %
example is the \textit{post-nonlinear model}~\citep{taleb1999source, zhang2009identifiability}, which assumes an element-wise nonlinearity applied after a linear mixing.
IMA is different in that it still allows for strongly nonlinear mixings~(see, e.g.,~\cref{fig:orthogonal_coordinate_transform}) provided that the columns of their Jacobians are (close to) orthogonal.
\looseness-1 In the related
field of disentanglement~\citep{bengio2013representation,locatello2019challenging}, a 
line of work that focuses on image generation with
adversarial networks~\citep{goodfellow2014generative} similarly proposes to constrain the ``generator'' function
via 
regularisation of its Jacobian~\citep{ramesh2018spectral} or Hessian~\citep{peebles2020hessian}, though mostly from an empirically-driven, rather than from an identifiability perspective.

\paragraph{Towards Identifiability with IMA.} The IMA principle rules out a large class of spurious solutions to nonlinear ICA, which does not amount to a full identifiability result. Yet, our experiments show that $C_\IMA$ can be used to recover the BSS equivalence class, suggesting that identifiability might indeed hold, possibly under additional assumptions---e.g., for conformal maps~\citep{hyvarinen1999nonlinear}, see~\cref{chap:conclusion} for further discussion of this point in the context of follow-up work by~\citet{buchholz2022function}.

\paragraph{IMA and Independence of Cause and Mechanism.} While inspired by measures of independence of cause and mechanism as traditionally used for cause-effect inference~\citep{daniuvsis2010inferring, janzing2012information, janzing2010telling, zscheischler2011testing}, we view the IMA principle as addressing a different question,
in the sense that they evaluate independence between different elements of the causal model. 
Any nonlinear ICA solution that satisfies the IMA~\cref{principle:IMA} can be turned into one with uniform reconstructed sources---thus satisfying IGCI as argued in~\cref{sec:unsuitability_of_existing_ICM_measures}---through composition with an element-wise transformation which, according to~\cref{prop:global_IMA_contrast_properties}~\textit{(ii)}, leaves the $C_\IMA$ value unchanged.
Both IGCI~\eqref{eq:IGCI_condition} and IMA~\eqref{eq:IMA_principle} can therefore be fulfilled simultaneosly, while the former on its own is inconsequential for BSS as shown in~\cref{prop:IGCI_insufficient_for_BSS}.

\paragraph{BSS Through Algorithmic Information.} 
Algorithmic information theory has previously been proposed as a unifying framework for identifiable approaches to \textit{linear} BSS~\citep{pajunen1998blind, pajunen1999blind}, in the sense that commonly used contrast functions could, under suitable assumptions, be interpreted as proxies for the total complexity of the mixing and the reconstructed sources.
However, to the best of our knowledge, the problem of specifying suitable proxies for the complexity of \textit{nonlinear} mixing functions has not yet been
addressed.
We conjecture that our framework could be linked to this view, based on the additional assumption of algorithmic independence of causal mechanisms~\citep{janzing2010causal}, thus potentially representing an approach to \textit{nonlinear} BSS~by minimisation of algorithmic complexity.

\paragraph{ICA for Causal Inference \& Causality for ICA.}
Past advances in ICA have inspired novel methods for causal discovery~\citep{shimizu2006linear,monti2020causal,khemakhem2021causal}.
Conversely, the present work constitutes, to the best of our knowledge, the first effort to use ideas from causality (specifically ICM) for BSS.

\paragraph{Possible Applications.}
Our IMA principle holds exactly for orthogonal coordinate transformations, and is thus of potential interest for learning spatial representations~\citep{hinton1981frames}, robot dynamics~\citep{mistry2010inverse}, or physics problems where orthogonal reference frames are common~\citep{moon1971field}.

\section{Extensions and Connections with Other Work}
\label{sec:extensions_ima}
We now briefly sketch connections between IMA and two of our other works on unsupervised representation learning. 
First, in~\cref{sec:ima_vae} we establish a connection between IMA and variational autoencoders~\citep[VAEs;][]{kingma2013auto}. Specifically, we show that, under certain conditions, the evidence lower bound (ELBO; the standard training objective of VAEs) corresponds to the IMA-regularised (rather than exact) likelihood.
In~\cref{sec:object-centric}, we then consider unsupervised representation learning for multi-object scenes. 
In a similar spirit to IMA, we show that by imposing certain constraints on the Jacobian of the mixing function, different slots of latents can be identified---even without requiring them to be statistically independent.

\subsection{ELBO Maximization in VAEs Enforces IMA}
\label{sec:ima_vae}
This subsection is based on the following publication, with all figures therein adopted without further modification. 
We briefly summarise the main points that are relevant to the context of this chapter, and refer to the full paper for further details.
\begin{selfcitebox}
\href{https://arxiv.org/abs/2206.02416}{\ul{Embrace the gap: VAEs perform independent mechanism analysis}}
\\
Patrik Reizinger$^*$, Luigi Gresele$^*$, Jack Brady$^*$, \textbf{Julius von K\"ugelgen}, Dominik Zietlow, Bernhard Sch\"olkopf, Georg Martius, Wieland Brendel,  Michel Besserve ($^*$equal contribution)
\\
\textit{Advances in Neural Information Processing Systems  (NeurIPS)}, 2022,
\end{selfcitebox}
Empirically, it has been shown that VAEs tend to learn disentangled representations, even though the underlying model class is not identifiable, see~\cref{sec:background_nonlinear_ICA_nonidentifiable} and~\citet{locatello2019challenging} for details.
However, VAEs are not actually trained by maximising the exact (log-)likelihood, but a lower bound to it (the ELBO), offering a possible resolution of this paradox.
Prior work has studied the gap between log-likelihood and ELBO for (near-)deterministic decoders, resulting in conflicting claims that the gap either disappears~\citep{nielsen_survae_2020} or implicitly encodes certain inductive biases~\citep{rolinek_variational_2019,kumar_implicit_2020}.

\begin{figure}[tb]
	\centering
	\includegraphics[width=\textwidth]{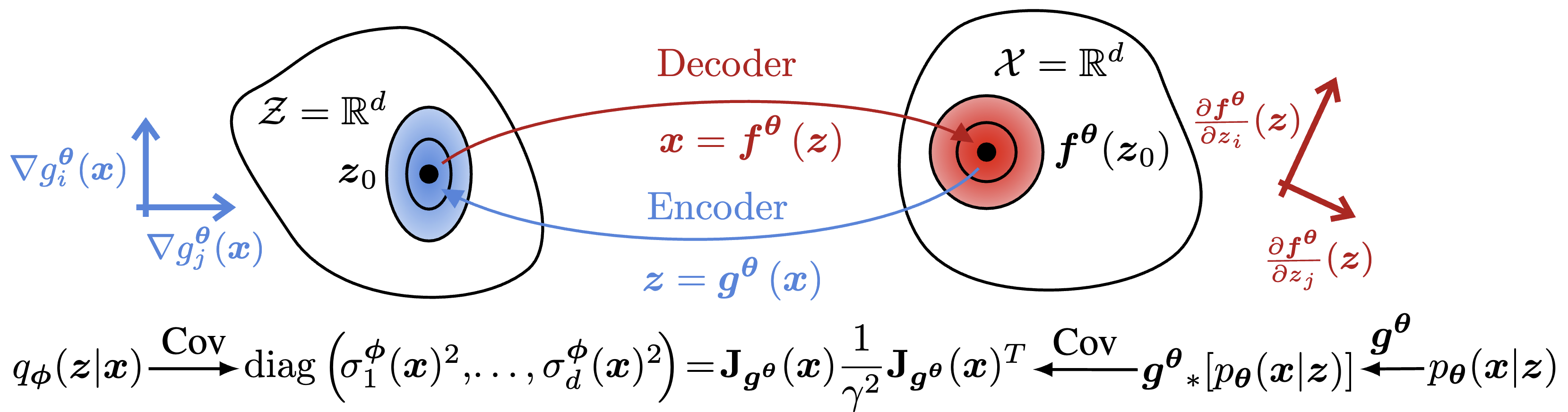}
	\caption[Modelling Choices in VAEs Promote IMA]{{\bf Modelling Choices in VAEs Promote IMA.}
		For a Gaussian VAE with isotropic decoder, in the near-deterministic regime (i.e., as the decoder precision $\gamma^2$ goes to infinity) the mean {encoder} approximately inverts the mean {decoder}, $\gb^{\thetab} \approx \fb^{\thetab -1}$ ({``self-consistency''}).
		To close the gap between the evidence lower bound (ELBO) and the exact likelihood, the  covariances of the variational posterior $q_\phi(\zb|\xb)$ and of the true posterior, approximated by the pushforward 
		$\gb^\theta_*[p_\theta(\xb|\zb)]$, need to match.
		As illustrated on the bottom, under self-consistency and diagonal encoder covariance, this enforces a row-orthogonal {encoder} Jacobian $\Jb_{\gb^\theta}(\xb)$---or equivalently, a \textit{column-orthogonal} {decoder} Jacobian $\Jb_{\fb^\theta}(\zb)$.
		The latter condition states that $\fb$ is in the IMA function class. This connection shows that the ELBO of (near-deterministic) VAEs actually corresponds to an IMA-regularised (rather than an exact) likelihood,
		and elucidates unintended potential benefits of using the ELBO for unsupervised representation learning.
	} 
	\label{figure:vae_ima}
\end{figure}

\looseness-1 We set out to rigorously study the effects of ELBO maximisation for unsupervised representation learning for the standard class of VAEs with (i) isotropic Gaussian decoder, (ii) log-concave (e.g., Gaussian) prior, and (iii) diagonal Gaussian encoder, see~\cref{figure:vae_ima} (top) for a visualisation. First, we show that in the near-deterministic regime as the decoder precision $\gamma^2 \to \infty$, the optimal encoder approximately inverts the decoder, in the sense that its mean and variances converge to the inverse of the decoder mean and zero at rates $O(1/\gamma)$ and $O(1/\gamma^2)$, respectively.
This property is also referred to as \textit{self-consistency} and had been conjectured but not formally shown in prior work. 
We refer to the objective resulting from the optimal encoder choice as self-consistent ELBO$^*$ and prove that, due to the modelling choices (i)-(iii) above, it converges to the IMA-regularised likelihood in the near-deterministic limit at rate $O(1/\gamma^2)$, see~\cref{figure:vae_ima} (bottom) for further intuition.
For small decoder variances (as commonly used in practice), the learning objective of VAEs  thus approximately implements an implicit bias for decoders from the IMA function class, which, as argued throughout this chapter, can benefit identifiability.

\begin{figure}[t]
\centering
\begin{subfigure}{0.36\textwidth}
\centering
    \includegraphics[width=\textwidth]{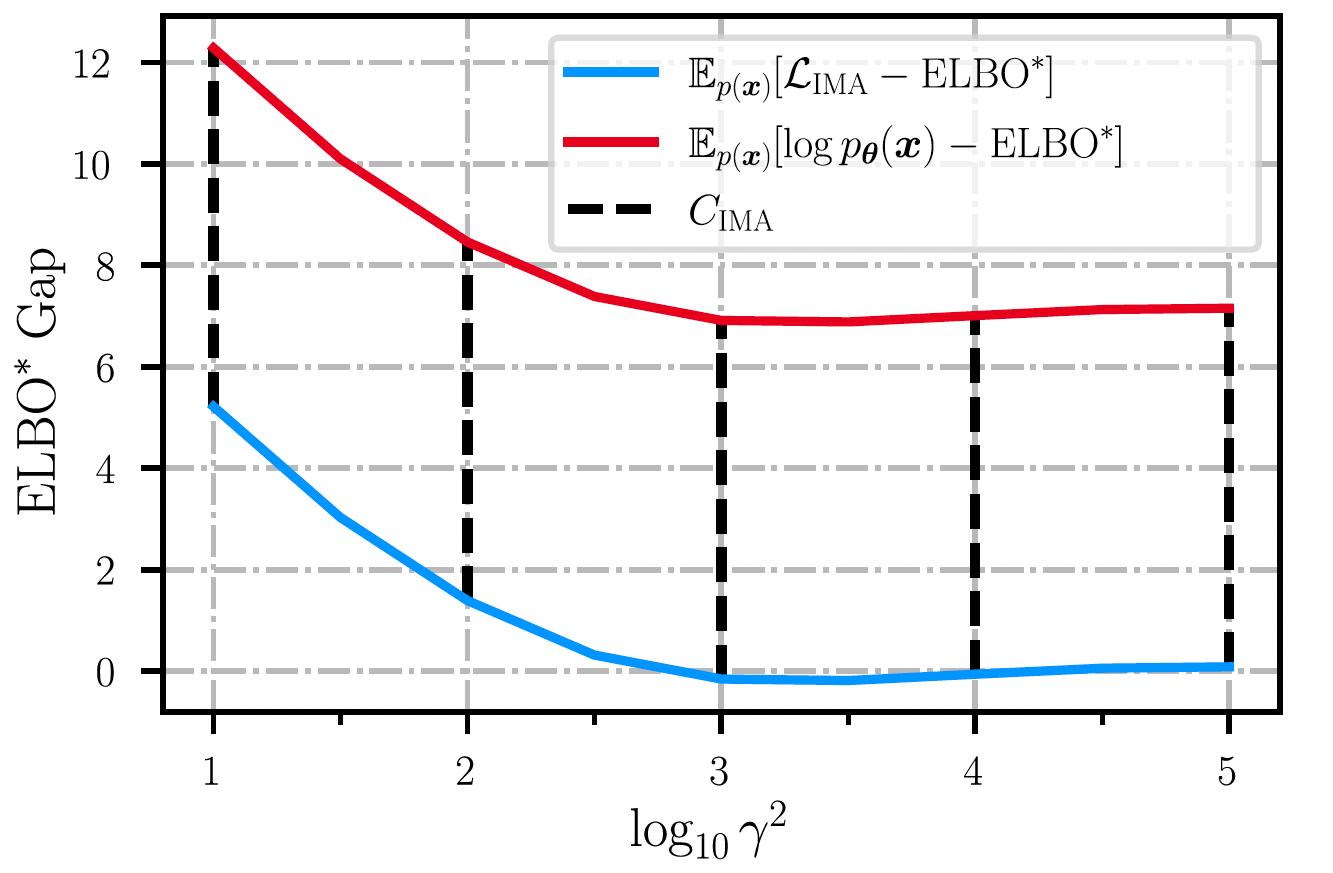}
\end{subfigure}%
\begin{subfigure}{0.64\textwidth}
\centering
    \includegraphics[width=\textwidth]{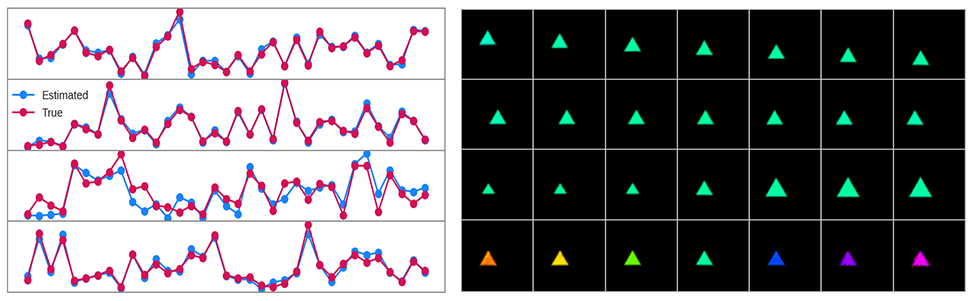}
    \vspace{.35em}
\end{subfigure}%
\caption[Empirical Results for the Relation Between ELBO and IMA]{\textbf{Empirical Results for the Relation Between ELBO and IMA.} \textit{(left)} Comparison of the gap between the self-consistent $\text{ELBO}^*$ and the IMA-regularised or unregularised log-likelihoods as a function of decoder precision $\gamma^2$.
\textit{(centre)} True and estimated
latent factors for a VAE trained on Sprites. \textit{(right)} The corresponding latent interpolations and MCC values (from top to bottom): $y$-position ($0.989$), $x$-position ($0.996$), scale ($0.933$), and colour ($0.989$).
}
\label{fig:ima_elbo_likelihood}
\end{figure} 

Empirically, we find that the ELBO$^*$ also converges to $\Lcal_\IMA$ and not to the exact log-likelihood for finite decoder precisions (when the true mixing is not in the IMA function class), see~\cref{fig:ima_elbo_likelihood} (left). Further, we find that VAEs successfully recover the true underlying factors for a simple image dataset, see~\cref{fig:ima_elbo_likelihood} (centre, right), suggesting that this data-generating process may approximately satisfy the IMA principle.

In summary, our findings show that widely adopted modelling choices and variational training objectives in deep generative models such as VAEs can implicitly enforce functional constraints like IMA, thus possibly explaining their empirical disentanglement properties.

\subsection{Identifying Object-Centric Representations}
\label{sec:object-centric}
This subsection is based on the following publication, with all figures therein adopted without further modification. 
We briefly summarise the main points that are relevant to the context of this chapter, and refer to the full paper for further details.
\begin{selfcitebox}
\href{https://arxiv.org/abs/2305.14229}{\ul{Provably learning object-centric representations}}
\\
Jack Brady$^*$, Roland S.\ Zimmermann$^*$, Yash Sharma, Bernhard Sch\"olkopf, \\
\textbf{Julius von K\"ugelgen}$^\dagger$, Wieland Brendel$^\dagger$ 
($^*$equal contribution, $^\dagger$shared last author)
\\
\textit{International Conference on Machine Learning (ICML)}, 2023
\end{selfcitebox}

The ability to represent visual scenes in terms of individual objects comes natural to humans and facilitates core cognitive abilities such as compositional generalization~\citep{fodor1988connectionism,lake2017building,battaglia2018relational,greff2020binding} and causal reasoning over discrete concepts~\citep{marcus2003algebraic,gopnik2004theory,gerstenberg2017intuitive,gerstenberg2021counterfactual}.
Learning such \textit{object-centric representations} promises to also  improve the generalization abilities of current ML models, and can be viewed as a prerequisite for CRL from image or video data.
While recent efforts to this end have shown promising empirical progress~\citep{locatello2020object}, a theoretical account of when unsupervised object-centric representation learning is possible is still lacking.

\begin{figure}[tbp]
    \centering
    \includegraphics[width=\textwidth]{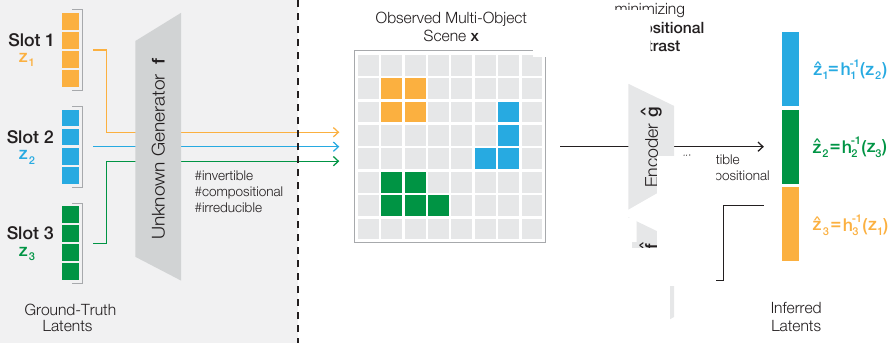}
    \caption[Identifiability of Object-Centric Representations]{\textbf{Identifiability of Object-Centric Representations.} Observed multi-object scenes~$\xb$ comprising $K$ objects are rendered by an unknown generator~$\fb$ from multiple ground-truth latent slots $\zb_1,..., \zb_K$ (here, $K=3$). If this generative model has two key properties, \textit{compositionality} (see~\cref{fig:compositional_function_visualisation}) and \textit{irreducibility} (see~\cref{fig:irreducibility}), we show that an \textit{invertible} inference model $\hat\gb$ with a compositional inverse $\hat \fb$ yields latent slots $\zbh_k$ which identify the ground-truth slots $\zb_k$ up to permutation and slot-wise invertible functions $\hb_i$ (\textit{slot identifiability}). 
    }
    \label{fig:multi_object_scenes}
\end{figure}

To study this task from an identifiability perspective, we formalize the generative process for multi-object scenes as a structured latent variable model with nonlinear mixing function $\fb$ and latents $\zb=(\zb_1, .., \zb_K)$ consisting of $K$ different multi-dimensional slots $\zb_k$, each capturing the properties of one of the objects, see~\cref{fig:multi_object_scenes} for a visualisation.
In contrast to the ICA setting studied in the context of IMA, the latent distribution $P_\zb$ is unconstrained, allowing for dependences both within and across slots in order to capture
relations and co-occurrence patterns among objects.
Similar to IMA, we impose additional constraints on (the Jacobian of) the mixing function to make progress towards identifiability.
Specifically, we introduce two assumptions on $\fb$, \emph{compositionality} and \emph{irreducibility}, which are inspired by the scene rendering process and together encode that each slot generates \textit{exactly} one object in the scene. 

\begin{figure}[tbp]
\centering
  \includegraphics[width=\textwidth]{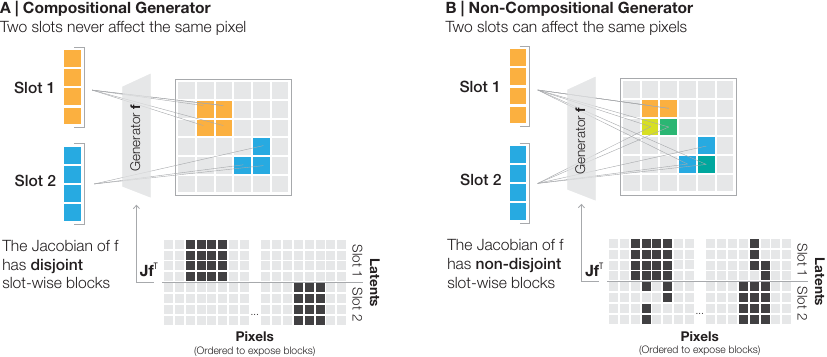}
  \caption[Compositionality]{\textbf{Compositionality.}
  \textbf{(A)} For a compositional generator $\fb$, every pixel is affected by at most one latent slot. As a result, there always exists an ordering of the pixels such that the Jacobian $\Jb_\fb$ consists of disjoint blocks, one for each latent slot \textit{(bottom)}. %
  Note that this ordering and the specific structure of the Jacobian is not fixed across scenes and might depend on the input~$\zb$. 
  \textbf{(B)} For a non-compositional generator, there exists no pixel ordering that exposes such a structure in the Jacobian, since the same pixel can be affected by more than one latent slot.}
  \label{fig:compositional_function_visualisation}
\end{figure}

Compositionality captures that, for any scene $\xb=\fb(\zb)$, each pixel $x_i$ belongs to \textit{at most} one object. Formally, it states that the partial derivative of $f_i$ is non-zero w.r.t.\ at most one slot $\zb_k$. This imposes a \textit{dynamic block-structure} on the Jacobian of $\fb$: for any $\zb$ there exists a (potentially distinct) permutation of the pixels which turns $\Jb_\fb(\zb)$ into a block-matrix, see~\cref{fig:compositional_function_visualisation}.

\begin{figure}[t]
\centering
  \includegraphics[width=\textwidth]{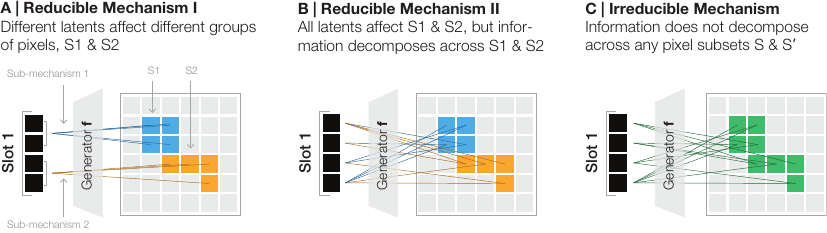}
 \caption[(Ir)reducible Mechanisms]{\textbf{(Ir)reducible Mechanisms.} \textbf{(A)} A simple example of a \textit{reducible mechanism} is one for which disjoint subsets of latents from the same slot render pixel groups $S_1$ and $S_2$ separately s.t.\ they form \textit{independent sub-mechanisms} according to~\cref{def:independent_dependent_submechanism},
 indicated by the difference in colours.  \textbf{(B)} Not all reducible mechanisms look as simple as in A: here, $S_1$ and $S_2$ depend on every latent component in the slot, but the information in $S_1\cup S_2$ may still decompose across $S_1$ and $S_2$.
 \textbf{(C)} For an irreducible mechanism, the information does not decompose across any pixel partition $S,  
 S^{'}$, and so it is impossible to separate it into independent sub-mechanisms.}
\label{fig:irreducibility}
\end{figure}

Irreducibility, captures that a slot should not generate multiple objects.  
Formally, it states that
the pixels affected by a given slot cannot be further partitioned in a way that results in independent sub-mechanisms in the following sense (see~\cref{fig:irreducibility} for a visualisation).
\begin{definition}[Independent Sub-Mechanisms]\label{def:independent_dependent_submechanism}
    Let $S_1, S_2 \subseteq [N]$ be disjoint. The Jacobian sub-matrices $\Jb_{\fb_{S_1}}(\zb)$ and $\Jb_{\fb_{S_2}}(\zb)$ are said to be \emph{independent sub-mechanisms} if:
    \begin{equation} \label{eq:independent_dependent_submechanism_independent}
        \rank\left(\Jb_{\fb_{S_1\cup S_2}}(\zb)\right)=\rank\left(\Jb_{\fb_{S_1}}(\zb)\right)+\rank\left(\Jb_{\fb_{S_2}}(\zb)\right).
    \end{equation}
\end{definition}
\Cref{def:independent_dependent_submechanism} can be viewed as yet another formalization (besides IMA)  of independence in the sense of the ICM~\cref{principle:ICM}: intuitively,
two sub-mechanisms $\Jb_{\fb_{S_1}}(\zb)$ and $\Jb_{\fb_{S_2}}(\zb)$ are considered independent if the information content of pixels $S_1\cup S_2$ decomposes across $S_1$ and~$S_2$ in the sense that the latent capacity required to \textit{jointly} generate $S_1\cup S_2$ (LHS of~\eqref{eq:independent_dependent_submechanism_independent}) is the same as that required to generate $S_1$ and~$S_2$ \textit{separately} (RHS of~\eqref{eq:independent_dependent_submechanism_independent}). 

\looseness-1 Under these assumptions, we prove that the ground-truth object representations are identified by an invertible encoder with compositional inverse---up to a permutation of the slots and invertible slot-wise nonlinear transformations, which we refer to as \textit{slot-identifiability}.\footnote{Slot-identifiability is inspired by the notion of \textit{block-identifiability}~(\cref{def:block-identifiability}) of a {group} or subset of latents introduced in the following chapter, and generalizes it to the simultaneous recovery of multiple blocks.}
To measure violations of compositionality, we introduce a contrast function which is zero if and only if a function is compositional (similar to the IMA contrast $C_\textsc{ima}$). To measure invertibility, we rely on the reconstruction loss in an auto-encoder framework~(see~\cref{fig:multi_object_scenes}).
Empirically, our theory holds predictive power for existing object-centric models in that we find a close correspondence between their compositionality and invertibility and the corresponding degree of identifiability.

Overall, these findings demonstrate that assumptions on the Jacobian of the mixing function are also useful in other scenarios where the latents are not assumed to be independent, such as for identifying object-centric representations.

\section{Summary}
\looseness-1 In this chapter, we have considered unsupervised representation learning from i.i.d.\ data and studied how \textit{constraining the model class} can be beneficial for identifiability.
Our main contribution, the causally inspired IMA constraint on the mixing function, posits that the \textit{influences} of different latents on the observed distribution should be approximately independent or orthogonal.
As we have shown, when combined with the assumption of statistically independent latents, this constraint helps promote identifiability in the sense that it is generally violated by common spurious solutions to nonlinear ICA.
As full identifiability is non-trivial to show even in this case, we did not study IMA for more difficult CRL tasks with causally related latents. 
However, our identifiability result for object-centric representations that leverages a different but related constraint (also involving the Jacobian of the mixing function) allows for arbitrary latent dependences.  We thus hope that IMA could also be a useful inductive bias for such more general settings.

Imposing some constraints on the model class is strictly necessary in an i.i.d.\ setting. 
As we have seen, under the right conditions, this can also be sufficient. 
Yet, placing restrictions on the model class can be undesirable, especially when dealing with complex phenomena. 
The following chapters will therefore explore a different learning signal: richer data arising from violations of the i.i.d.\ assumption.

%% file: Chapter7/chapter7.tex
 \graphicspath{{Chapter7/Figs/}}
\chapter{Multi-View Causal Representation Learning}
\label{chap:SSL_content_style}

In this chapter, we depart from the i.i.d.\ setting studied in~\cref{chap:IMA} by learning from \textit{multiple views} in the form of \textit{non-independent} observations.
This richer type of data allows us to consider both \textit{general nonlinear mixing functions} (in contrast to~\cref{chap:IMA}) and latent variables that are not necessarily independent but possibly \textit{causally related}. 
We mainly focus on the setting of learning from pairs of observations that share a fixed set of invariant content variables, inspired by data augmentation practices, and show that this latent block can be identified through self-supervised learning.
We then discuss extensions to learning from tuples to also recover varying style variables, or handling multiple modalities and partial observability.

The main content of this chapter has been published in the following paper:
\begin{selfcitebox}
\href{https://arxiv.org/abs/2106.04619}{\ul{Self-supervised learning with data augmentations provably isolates content from style}}
\\
\textbf{Julius von K\"ugelgen}$^*$, Yash Sharma$^*$, Luigi Gresele$^*$, Wieland Brendel, \\Bernhard Sch\"olkopf$^\dagger$, Michel Besserve$^\dagger$, Francesco Locatello$^\dagger$ \\($^*$equal contribution, $^\dagger$shared last author)
\\
\textit{Advances in Neural Information Processing Systems (NeurIPS)}, 2021
\end{selfcitebox}

\section{Introduction}
\label{sec:ssl_introduction}
Learning good representations of high-dimensional observations
from large amounts of unlabelled data
is widely recognised as an important step for
more capable and data-efficient learning systems~\citep{bengio2013representation,lake2017building}.
Over the last decade, \textit{self-supervised learning} (SSL)
has emerged as the dominant paradigm for such unsupervised representation learning~\citep{wu2018unsupervised,oord2018representation,henaff2020data,tian2019contrastive,he2019momentum,chen2020simple, grill2020byol, chen2020exploring,agrawal2015learning,doersch2015unsupervised,wang2015unsupervised,vincent2008extracting,noroozi2016unsupervised}.
The main idea behind SSL is to extract a supervisory signal from unlabelled observations by leveraging known structure of the data. This then allows for the application of supervised learning techniques. 
A common approach
is to directly predict some
part of the observation from 
another part (e.g., future from past, or original from corruption), thus forcing the model to learn a meaningful representation in the process.
While this technique has shown remarkable success in natural language processing~\citep{collobert2011natural,mikolov2013distributed,pennington2014glove,logeswaran2018efficient,devlin2019bert,radford2018improving,liu2019roberta,GPT3}
and speech recognition~\citep{baevski2020wav2vec,Baevski2020vqwav2vec,ravanelli2020multi,schneider2019wav2vec}, where
a finite dictionary allows one to output a distribution over the missing part, such \textit{predictive} SSL methods are not easily applied to continuous or high-dimensional domains such as vision.
Here, 
a common approach is to learn a \textit{joint embedding} of similar observations or \textit{views} such that their representation is close~\citep{becker1992self,hadsell2006dimensionality,chopra2005learning,bromley1993signature}.
Different views can come, for example, from different modalities (e.g., text and speech, or video and audio) or time points.
As still images lack such multi-modality or temporal structure, recent advances in representation learning have relied on generating similar views by means of \textit{data augmentation}. 
In order to be useful, data augmentation is thought to require the transformations applied to generate additional views to be generally chosen to \textit{preserve the semantic characteristics} of an observation, while changing other ``nuisance'' aspects.
While this intuitively makes sense and has shown remarkable empirical results, the success of data augmentation techniques in practice is still not very well understood from a theoretical perspective---despite some efforts~\citep{ChaSch02,dao2019kernel,chen2020group}.
In the present work, we seek to better understand the empirical success of SSL with data augmentation by formulating the generative process
as a latent variable model (LVM) and studying \textit{identifiability} of the representation, i.e., under which conditions the ground truth latent factors can provably be inferred from the data. 

\paragraph{Related Work and Its Relation to the Current.}
Prior work on unsupervised representation learning from an LVM perspective often postulates \textit{mutually independent latent factors}~(\cref{sec:background_ICA}).
Since it is 
impossible to
identify the true latent factors without any supervisory signal in the general nonlinear case~(\cref{sec:background_nonlinear_ICA_nonidentifiable}), recent work has turned to weakly- or self-supervised approaches~(\cref{sec:background_nonlinear_ICA_auxiliary}) which leverage additional information in the form of multiple views~\citep{gresele2019incomplete,locatello2020weakly,shu2019weakly,zimmermann2021contrastive}, auxiliary variables~\citep{hyvarinen2019nonlinear,khemakhem2020variational}, or temporal structure~\citep{halva2020hidden,hyvarinen2016unsupervised,hyvarinen2017nonlinear,klindt2020slowvae}.
To identify or disentangle the individual independent latent factors, it is typically assumed
that there is a chance
that \textit{each factor changes} across views,
environments, or time points.

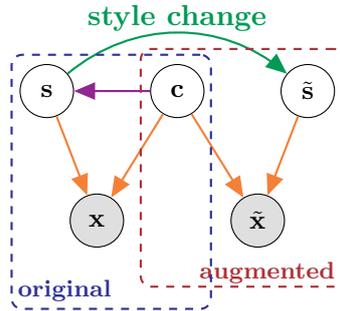
\begin{figure}[t]%
\newcommand{\dependencecolor}{Plum}
\newcommand{\decodingcolor}{Orange}
\newcommand{\stylechangecolor}{ForestGreen}
\newcommand{\originalcolor}{Blue}
\newcommand{\augmentationcolor}{Maroon}
\centering
    \begin{tikzpicture}
        \centering
        \node (c) [latent] {$\cb$};
        \node (s) [latent, left=of c] {$\sb$};
        \node (s') [latent, right=of c] {$\sbt$};
        \node (x) [obs, below=of c, xshift=-2.75em] {$\xb$};
        \node (x') [obs, below=of c, xshift=2.75em] {$\xbt$};
        \edge[color=\dependencecolor,thick]{c}{s};
        \path[->, color=\stylechangecolor,thick] (s) edge[bend right=-40] node[yshift=.5em] {\textbf{style change}} (s');
        \edge[color=\decodingcolor,thick]{c}{x,x'};
        \edge[color=\decodingcolor,thick]{s}{x};
        \edge[color=\decodingcolor,thick]{s'}{x'};
        \plate[inner sep=0.3em,yshift=0.2em,dashed,color=\augmentationcolor,thick] {augmented}{(c) (s') (x')}{\textcolor{\augmentationcolor}{\textbf{augmented}}}; 
        \tikzset{plate caption/.style={caption, node distance=0, inner sep=0pt, below left=5pt and 0pt of #1.south,text height=1.2em,text depth=0.3em}}
        \plate[inner sep=0.2em,yshift=0.1em,dashed,color=\originalcolor,thick] {original}{(c) (s) (x)}{\textcolor{\originalcolor}{\textbf{original}}};
    \end{tikzpicture}
    \caption[Overview of Our Problem Formulation]{\textbf{Overview of Our Problem Formulation.} We partition the latent variable $\zb$ into content~$\cb$ and style~$\sb$, and allow for \textcolor{\dependencecolor}{statistical and causal dependence of style on content}. We assume that \textcolor{\stylechangecolor}{only style changes between} \textcolor{\originalcolor}{the original view} $\xb$ and \textcolor{\augmentationcolor}{the augmented view} $\xbt$, i.e., they are obtained by \textcolor{\decodingcolor}{applying the same deterministic function} $\fb$ to $\zb=(\cb,\sb)$ and $\zbt=(\cb,\sbt)$.}
    \label{fig:our_assumption}
\end{figure}
Our work---being directly motivated by common practices in SSL with data augmentation---differs from these works in the following two key aspects (see~\cref{fig:our_assumption} for an overview).
First, we do not assume independence and instead \textit{allow for both nontrivial statistical and causal relations between latent variables}.
Second, instead of a scenario wherein all latent factors may change as a result of augmentation, we assume a \textit{partition of the latent space} into two
blocks: a \textit{content} block which is shared or \textit{invariant} across different augmented views, and a \textit{style} block that \textit{may change}. %
\looseness-1 This is aligned with the notion that  augmentations leave certain semantic aspects (i.e., content) intact and only affect style, and is thus a more appropriate assumption for studying SSL.

\paragraph{Structure and Contributions.}
Following a review of SSL with data augmentation
and identifiability theory~(\cref{sec:ssl_preliminaries}), we formalise the process of data generation and augmentation as an LVM with content and style variables~(\cref{sec:problem_formulation}).
We then establish identifiability results of the invariant content partition~(\cref{sec:ssl_theory}),
validate our theoretical insights experimentally~(\cref{sec:ssl_experiments}), and discuss our findings and their limitations in the broader context of SSL with data augmentation~(\cref{sec:ssl_discussion}).
We highlight the following contributions:%
\begin{itemize}
    \item We prove that SSL with data augmentations identifies the invariant content partition of the representation in generative~(\Cref{thm:main}) and discriminative learning with invertible~(\Cref{thm:CL}) and non-invertible encoders with entropy regularisation~(\Cref{thm:CL_MaxEnt}).
    In particular, \Cref{thm:CL_MaxEnt} provides a theoretical justification for the empirically observed effectiveness of contrastive SSL methods  that use data augmentation and InfoNCE~\citep{oord2018representation} as an objective, such as \texttt{SimCLR}~\citep{chen2020simple}.
    \item We show that our theory is consistent with results in simulating statistical dependencies within blocks of content and style variables, as well as with style causally dependent on content~(\cref{sec:experiment_1_numerical_simulation}).
    \item We introduce \textit{Causal3DIdent}, a 
    dataset of 3D objects 
    which allows for the study of identifiability in a causal representation learning setting, and use it to perform a systematic study of data augmentations used in practice, %
    yielding novel insights on what particular data augmentations are truly isolating as invariant content and discarding as varying style when applied~(\cref{sec:experiment_2_causal3dident}). 
\end{itemize}

\section{Preliminaries and Related Work}
\label{sec:ssl_preliminaries}

\paragraph{Self-Supervised Representation Learning with Data Augmentation.}
\label{sec:background_augmentation}
Given an unlabelled dataset of observations (e.g., images) $\xb$, data augmentation techniques proceed as follows.
First, a set of observation-level transformations $\tb\in\Tcal$ are specified together with a distribution $p_\tb$ over $\Tcal$. 
Both $\Tcal$ and $p_\tb$ are typically designed using human intelligence and domain knowledge with the intention of \textit{not changing the semantic characteristics} of the data (which arguably constitutes a form of weak supervision).\footnote{Recent work has investigated automatically discovering good augmentations~\citep{cubuk2019autoaugment,cubuk2020randaugment}.}
For images, for example, a common choice for $\Tcal$ are combinations of random crops~\citep{szegedy2015going}, horizontal or vertical flips, blurring, colour distortion~\citep{howard2013improvements,szegedy2015going}, or cutouts~\citep{devries2017improved}; and $p_\tb$ is a distribution over the parameterisation of these transformations, e.g., the centre and size of a crop~\citep{chen2020simple,devries2017improved}.
For each observation~$\xb$, a pair of transformations $\tb,\tb'
\sim
p_\tb$ is sampled and applied separately to $\xb$ to generate a pair of augmented views $(\xbt,\xbt')=(\tb(\xb),\tb'(\xb))$.

The joint-embedding approach to SSL then uses a pair of encoder functions $(\gb,\gb')$---typically, deep neural networks---to map the pair $(\xbt,\xbt')$ to a typically lower-dimensional representation $(\zbt,\zbt')=(\gb(\xbt),\gb'(\xbt'))$.
Often, the two encoders are either identical, $\gb=\gb'$, or directly related (e.g., via shared parameters or asynchronous updates). 
Then, the encoder(s) $(\gb,\gb')$ are trained  such that the representations $(\zbt,\zbt')$ are ``close'', i.e., such that $\text{sim}(\zbt,\zbt')$ is large for some  similarity metric $\text{sim}(\cdot)$, e.g., the cosine similarity~\citep{chen2020simple}, or negative L2 norm~\citep{zimmermann2021contrastive}.
The advantage of 
directly optimising for similarity in representation space over generative alternatives is that reconstruction can be very challenging for high-dimensional data.
The disadvantage
is the problem of \textit{collapsed representations}: if the only goal is to make representations of augmented views similar, a degenerate solution which simply maps any observation to the origin trivially achieves this goal.
To avoid 
collapsed representations and force the encoder(s) to learn a meaningful representation, two main
families of 
approaches have been used: (i) \textit{contrastive learning} (CL)~\citep{oord2018representation,he2019momentum,chen2020simple,wu2018unsupervised,henaff2020data,tian2019contrastive}; and (ii) \textit{regularisation-based} SSL~\citep{grill2020byol,chen2020exploring,zbontar2021barlow}.

The idea behind CL is to not only learn similar representations for augmented views $(\xbt_i, \xbt'_i)$ of the same $\xb_i$, or \textit{positive pairs}, but to also use other observations $\xb_j$ $(j\neq i)$ to contrast with, i.e., to enforce a dissimilar representation across \textit{negative pairs} $(\xbt_i, \xbt'_j)$.
In other words, CL pulls representations of positive pairs together, and pushes those of negative pairs apart.
Since both aims cannot be achieved simultaneously with a constant representation, collapse is avoided.
A popular CL objective function (used, e.g., in \texttt{SimCLR}~\citep{chen2020simple}%
)
is InfoNCE~\citep{oord2018representation},
based on noise-contrastive estimation
~\citep{gutmann2010noise,Gutmann12JMLR}:%
\begin{equation}
\label{eq:InfoNCE_objective}
    \Lcal_{\text{InfoNCE}}
    (\gb;\tau,K)
    =
    \EE_{\{\xb_i\}_{i=1}^K \sim p_\xb}
    \Big[
    -
    \sum_{i=1}^K
    \log 
    \frac{
    \exp\{\text{sim}(\zbt_i,\zbt'_i)/\tau\}
    }
    {
    \sum_{j=1}^K
    \exp\{\text{sim}(\zbt_i,\zbt'_j)/\tau\}
    }
    \Big]
\end{equation}
\looseness-1 where $\zbt=\EE_{\tb\sim p_\tb}[\gb(\tb(\xb))]$, $\tau$ is a temperature, and $K{-}1$ is the number of negative pairs.
InfoNCE~\eqref{eq:InfoNCE_objective} has an interpretation as multi-class logistic regression, and
lower bounds the mutual information across similar views $(\zbt,\zbt')$---a common representation learning objective~\citep{tschannen2019mutual,poole2019variational,hjelm2018learning,bachman2019learning,linsker1988self,linsker1989application,cardoso1997infomax,lee1999independent,bell1995information}.
Moreover,~\eqref{eq:InfoNCE_objective} can be interpreted as \textit{alignment} (numerator) and \textit{uniformity} (denominator) terms, the latter constituting a nonparametric entropy estimator of the representation as $K\rightarrow\infty$~\citep{wang2020understanding}.
{CL with InfoNCE can thus be seen as alignment of positive pairs with (approximate) entropy regularisation.}

Instead of using negative pairs, as in CL, a set of recent SSL methods only optimise for alignment and avoid collapsed representations through different forms of regularisation. 
For example, \texttt{BYOL}~\citep{grill2020byol} and \texttt{SimSiam}~\citep{chen2020exploring} rely on ``architectural regularisation'' in the form of moving-average updates for a separate ``target'' net $\gb'$ (\texttt{BYOL} only) or
a stop-gradient operation (both).
\texttt{BarlowTwins}~\citep{zbontar2021barlow}, on the other hand, optimises the cross correlation  between $(\zbt,\zbt')$ to be close to the identity matrix, thus enforcing redundancy reduction (zero off-diagonals) in addition to  alignment (ones on the diagonal).

\paragraph{Identifiability of Representations Learned from Multiple Views.}
Here, we seek to address the question of whether SSL with data augmentation 
can reveal or uncover properties of the underlying data generating process.
Most closely related to SSL with data augmentation are works which study identifiability when 
given 
a second view $\xbt$ of an observation $\xb$, resulting from a modified version $\zbt$ of the underlying latents or sources $\zb$~\citep{gresele2019incomplete,richard2020modeling,locatello2020weakly,shu2019weakly,zimmermann2021contrastive,klindt2020slowvae}.
Here, $\zbt$ is either an element-wise  corruption of $\zb$~\citep{gresele2019incomplete,richard2020modeling,zimmermann2021contrastive,klindt2020slowvae} or may share a random subset of its components~\citep{locatello2020weakly,shu2019weakly}.
Crucially, all previously mentioned works assume that \textit{any} of the independent latents (are allowed to) change, and aim to identify the individual factors.
However, in the context of SSL with data augmentation, where the semantic (content) part of the representation is intended to be shared between views, this assumption generally does not hold.

\section{Problem Formulation}
\label{sec:problem_formulation}

We specify our problem setting by formalising the processes of data generation and augmentation.
We take a latent-variable model perspective and
assume that observations $\xb$ (e.g., images) are generated by a \emph{mixing} function $\fb$ which takes a latent code $\zb$ as input.
Importantly, we describe the augmentation process through changes in this latent space as captured by a conditional distribution $p_{\zbt|\zb}$, as opposed to traditionally describing the transformations $\tb$ as acting directly at the observation level.

Formally, let $\zb$ be a continuous random variable taking values in an open, simply-connected $n$-dimensional \textit{representation space} $\Zcal\subseteq\RR^n$ with associated probability 
density $p_\zb$.
Moreover, let $\fb:\Zcal\rightarrow\Xcal$ be a \textit{smooth and invertible} mapping to an \textit{observation space} $\Xcal\subseteq \RR^d$
and let $\xb$ be the continuous random variable defined as $\xb=\fb(\zb)$.\footnote{While $\xb$ may be high-dimensional $n\ll d$, invertibility of $\fb$ implies that $\Xcal$ is an $n$-dim.\ sub-manifold of~$\RR^d$.}
The generative process for the dataset of original observations of $\xb$ is thus given by:%
\begin{align}
\textstyle
\label{eq:generative_process_original}
    \zb \sim p_\zb,
    \quad \quad \quad \quad 
    \xb =\fb(\zb).
\end{align}%
\looseness-1 Next, we formalise the data augmentation process.
As stated above, we take a representation-centric view, i.e., we assume that an augmentation $\xbt$ of the original $\xb$ is obtained by applying the same mixing or rendering function $\fb$ to a modified representation $\zbt$ which is (stochastically) related to the original representation $\zb$ of 
$\xb$.
Specifying the effect of data augmentation thus corresponds to specifying a conditional distribution $p_{\zbt|\zb}$ which captures the relation between $\zb$ and $\zbt$.

In terms of the transformation-centric view presented in~\cref{sec:background_augmentation}, we can view the modified representation $\zbt\in\Zcal$ as obtained by applying $\fbinv$ to a transformed observation $\xbt=\tb(\xb)\in\Xcal$ where $\tb\sim p_\tb$, i.e., $\zbt=\fbinv(\xbt)$.
The conditional distribution $p_{\zbt|\zb}$ in the representation space can thus be viewed as being induced by the distribution $p_\tb$ over transformations applied at the observation level.\footnote{We investigate this correspondence between changes in observation and latent space empirically in~\cref{sec:ssl_experiments}.}

We now encode the notion that the set of transformations $\Tcal$ used for augmentation is typically chosen such that any transformation $\tb\in\Tcal$ leaves certain aspects of the data invariant.
To this end, we assume that \textit{the representation $\zb$ can be uniquely partitioned into two disjoint parts}:%
\begin{enumerate}[label=(\roman*), topsep=-3pt,itemsep=0pt]
    \item an \textit{invariant} part $\cb$ which will \textit{always be shared} across $(\zb,\zbt)$, and which we refer to as \textit{content};
    \item a \textit{varying} part $\sb$ which \textit{may
    change} across $(\zb,\zbt)$, and which we refer to as \textit{style}.
\end{enumerate}%
We assume that $\cb$ and $\sb$ take values in content and style subspaces $\Ccal\subseteq \RR^{n_c}$ and 
$\Scal\subseteq\RR^{n_s}$, respectively, i.e., $n=n_c+n_s$ and $\Zcal=\Ccal\times\Scal$. 
W.l.o.g., we let $\cb$ corresponds to the first $n_c$ dimensions of~$\zb$:
\begin{equation*}
\textstyle
\label{eq:def_content_style}
    \zb = (\cb, \sb),
    \quad \quad \quad \quad 
    \cb := \zb_{1:n_c},
    \quad \quad \quad \quad 
    \sb := \zb_{(n_c+1):n},
\end{equation*}%
\looseness-1 We formalise the process of data augmentation with content-preserving transformations by defining the conditional $p_{\zbt|\zb}$ such that only a (random) subset of the style variables change at a time.

\begin{assumption}[Content-invariance]
\label{ass:content_invariance}
The conditional density $p_{\zbt|\zb}$ over $\Zcal\times\Zcal$ takes the form
\begin{equation*}
\textstyle
    p_{\zbt|\zb}(\zbt~|~\zb)=\delta(\cbt-\cb) \, p_{\sbt|\sb}(\sbt~|~\sb)
\end{equation*}
for some continuous density $p_{\sbt|\sb}$ on $\Scal\times\Scal$, where $\delta(\cdot)$ is the Dirac delta function, i.e., $\cbt=\cb$ a.e.
\end{assumption}

\begin{assumption}[Style changes]
\label{ass:style_changes}
Let $\Acal$ be the set of subsets of style variables $A\subseteq\{1, ..., n_s\}$ and let $p_A$ be a distribution on $\Acal$.
Then, the style conditional $p_{\sbt|\sb}$ is obtained via%
\begin{equation*}
\textstyle
    A\sim p_A,
    \quad \quad \quad \quad 
    p_{\sbt|\sb,A}(\sbt~|~\sb,A) = \delta(\sbt_{\Ac} - \sb_{\Ac}) \, p_{\sbt_\A|\sb_\A}(\sbt_\A~|~\sb_\A)\, ,
\end{equation*}%
where $p_{\sbt_\A|\sb_\A}$ is a continuous density on $\Scal_A\times\Scal_A$, $\Scal_A\subseteq\Scal$ denotes the subspace of changing style variables specified by $A$, and $\Ac
=\{1,...,n_s\}\setminus A
$ denotes the complement of $A$.
\end{assumption}%
Note that Assumption~\ref{ass:style_changes} is less restrictive than assuming that all style variables need to change, since it also allows for only a (possibly different) subset of style variables to change for any given observation.
This is in line with the intuition that not all transformations affect all changeable (i.e., style) properties of the data: e.g., a colour distortion should not affect positional information, and, in the same vein, a (horizontal or vertical) flip should not affect the colour spectrum.

The generative process of an augmentation or transformed observation $\xbt$ is thus given by
\begin{equation}
\label{eq:generative_process_augmentation}
    \textstyle
    A\sim p_A,
    \quad \quad \quad 
    \zbt~|~\zb, A \sim p_{\zbt|\zb,A},
    \quad \quad \quad
    \xbt = \fb(\zbt).
\end{equation}
\looseness-1 Our setting for modelling data augmentation differs from that commonly assumed in (multi-view) disentanglement and ICA in that \textit{we do not assume that the latent factors $\zb=(\cb,\sb)$ are mutually (or conditionally) independent}, i.e., we allow for \textit{arbitrary} (non-factorised) marginals $p_\zb$
in~\eqref{eq:generative_process_original}.\footnote{The recently proposed Independently \textit{Modulated} Component Analysis (IMCA)~\citep{khemakhem2020ice} extension of ICA is a notable exception, but only allows for very simple dependencies across $\zb$ in the form of a shared base measure.
}

\section{Counterfactual Interpretation of Data Augmentation}
\label{sec:counterfactual_interpretation_data_aug}
We now provide a causal account of the above data generating process by describing the (allowed) causal dependencies among latent variables using a 
structural causal model (SCM)~\citep{Pearl2009}. 
As we will see, this leads to an interpretation of data augmentations as counterfactuals in the underlying latent SCM.
The assumption that $\cb$ stays invariant as $\sb$ changes is consistent with the view that content may causally influence style, $\cb\rightarrow\sb$, but not vice versa, see~\cref{fig:our_assumption}. \looseness-1 We therefore formalise their relation as:%
\begin{align*}
\textstyle
    \cb := \fb_\cb(\ub_\cb), \quad \quad \quad \quad
    \sb := \fb_\sb(\cb, \ub_\sb),
    \quad \quad\quad \quad
    (\ub_\cb, \ub_\sb)\sim p_{\ub_\cb}\times p_{\ub_\sb}
\end{align*}%
where $\ub_\cb, \ub_\sb$ are independent exogenous variables, and $\fb_\cb,\fb_\sb$ are deterministic functions.
The latent causal variables $(\cb, \sb)$ are subsequently decoded into observations $\xb=\fb(\cb, \sb)$.
Given a factual observation $\xbf=
\fb(\cbf, \sbf)$ which resulted from $(\ub_\cb^\texttt{F}, \ub_\sb^\texttt{F})$, 
we may ask the counterfactual question: 
    ``\textit{what would have happened if the style variables had been
    (randomly)
    perturbed, all else being equal%
    ?}''.
\looseness-1 Consider, e.g.,
a \textit{soft intervention}~\citep{eberhardt2007interventions} on~$\sb$, i.e., an intervention that changes the mechanism~$\fb_\sb$ to%
\begin{equation*}
\textstyle
   do(\sb:=\Tilde{\fb}_\sb(\cb, \ub_\sb, \ub_\A)), 
\end{equation*}%
where $\ub_\A$ is an additional source of stochasticity accounting for the randomness of the augmentation process ($p_A\times  p_{\sbt|\sb,A}$).
The resulting distribution over counterfactual observations $\xbcf
=\fb(\cbf, \sbcf)$ can be computed from the modified SCM by fixing the exogenous variables to their factual values and performing the soft intervention:
\begin{align*}
\textstyle
    \cb^\texttt{CF} := \cbf, \quad \quad \quad \quad
    \sb^\texttt{CF} := \Tilde{\fb}_\sb(\cbf, \ub_\sb^\texttt{F}, \ub_\A),
    \quad    \quad     \quad    \quad 
    \ub_\A\sim p_{\ub_\A}.
\end{align*}%
This aligns with our intuition and assumed problem setting of data augmentations as style corruptions.
We note that the notion of augmentation as (hard) style interventions is also at the heart of \texttt{ReLIC}~\citep{mitrovic2020representation}, a recently proposed, causally-inspired SSL regularisation term for instance-discrimination~\citep{hadsell2006dimensionality,wu2018unsupervised}.
However, \texttt{ReLIC} assumes independence between content and style and does not address identifiability.
{For another causal perspective on data augmentation in the context of domain generalisation, c.f.~\citet{ilse2021selecting}.}

\section{Block-Identifiability of the Invariant Content Partition}
\label{sec:ssl_theory}
Our goal is to prove that we can identify the invariant content partition $\cb$ under a distinct, weaker set of assumptions, compared to existing results in
disentanglement and nonlinear ICA~\citep{zimmermann2021contrastive,gresele2019incomplete,locatello2020weakly,shu2019weakly,klindt2020slowvae}.
We stress again that our primary interest is not to identify or disentangle individual (and independent) latent factors $z_j$, but instead to separate content from style,
such that the content variables can be subsequently used for 
downstream tasks.
We first define this distinct notion of \textit{block-identifiability}.%
\begin{definition}[Block-identifiability]
\label{def:block-identifiability}
We say that the true content partition $\cb=\fbinv(\xb)_{1:n_c}$ is \emph{block-identified} by a function $\gb:\Xcal\rightarrow\Zcal$ if the inferred content partition $\cbh=\gb(\xb)_{1:n_c}$ contains \emph{all} and \emph{only} information about $\cb$, i.e., if there exists an \textit{invertible} function $\hb:\RR^{n_c}\rightarrow \RR^{n_c}$ s.t.\ $\cbh=\hb(\cb)$.%
\end{definition}%
\Cref{def:block-identifiability} is related to independent subspace analysis~\citep{hyvarinen2000emergence, le2011learning,theis2006towards,casey2000separation}, which also aims to identify blocks of random variables
as opposed to individual factors, though under an \textit{independence assumption across blocks}, and typically not within a multi-view setting as studied in the present work.

\subsection{Generative Self-Supervised Representation Learning}
First, we consider \textit{generative} SSL, i.e., fitting a generative model to pairs $(\xb,\xbt)$ of original and augmented views.\footnote{For notational simplicity, we present our theory for pairs $(\xb,
\xbt)$ rather than for two augmented views $(\xbt,\xbt')$, as typically used in practice but it also holds for the latter, see~\cref{sec:discussion} for further discussion.}
We show that under our specified data generation and augmentation process~(\cref{sec:problem_formulation}), as well as suitable additional assumptions (stated and discussed in more detail below), it is possible to isolate (i.e., block-identify) the invariant content partition.
Full proofs are included in~\Cref{app:proofs_SSL}.

\begin{restatable}[Identifying content with a generative model]
{theorem}{generative}
\label{thm:main}
Consider the data generating process described in~\cref{sec:problem_formulation}, i.e., the pairs $(\xb,\xbt)$ of original and augmented views are generated according to~\eqref{eq:generative_process_original} and~\eqref{eq:generative_process_augmentation} with $p_{\zbt|\zb}$ as defined in Assumptions~\ref{ass:content_invariance} and~\ref{ass:style_changes}.
Assume further that%
\begin{enumerate}[label=(\roman*), itemsep=0pt, topsep=0pt]
    \item
    $\fb:\Zcal\rightarrow \Xcal$ is smooth and invertible with smooth inverse (i.e., a diffeomorphism);
    \item $p_\zb$ is a smooth, continuous density on $\Zcal$ with $p_\zb(\zb)>0$ almost everywhere;
    \item for any $l\in\{1, ..., n_s\}$, there exists an $\A\subseteq\{1, ..., n_s\}$ such that: $l\in A$,  $p_A(A)>0$, $p_{\sbt_\A|\sb_\A}$ is smooth w.r.t.\ both $\sb_A$ and $\sbt_A$, and for any $\sb_A$,  $p_{\sbt_\A|\sb_\A}(\cdot~|~\sb_A)>0$ in some open, non-empty subset containing $\sb_A$.
\end{enumerate}%
If, for a given $n_s$ ($1\leq n_s<n$), a generative model $(\ph_\zb, \ph_A, \ph_{\sbt|\sb, A}, \fbh )$ assumes the same generative process~(\cref{sec:problem_formulation}), satisfies the above assumptions \textit{(i)-(iii)}, and matches the data likelihood,
\begin{equation*}
    \label{eq:match_likelihood}
    p_{\xb,\xbt}(\xb,\xbt)=\ph_{\xb,\xbt}(\xb,\xbt)
    \quad \quad \quad
    \forall (\xb,\xbt)\in \Xcal\times\Xcal,
\end{equation*}
then it block-identifies the true content variables via $\gb=\fbh^{-1}$ in the sense of~\cref{def:block-identifiability}.
\end{restatable}%

\paragraph{Proof Sketch.}
First, show (using \textit{(i)} and the matching likelihoods) that the representation $\zbh=\gb(\xb)$ extracted by $\gb$ is related to the true $\zb$ by a smooth invertible mapping $\hb=\gb\circ\fb$ such that $\cbh=\hb(\zb)_{1:n_c}$ is invariant across $(\zb,\zbt)$ almost surely w.r.t.~$p_{\zb,\zbt}$.\footnote{This step is partially inspired by~\citep{locatello2020weakly}; the technique used to prove the second \textit{main} step is entirely novel.}
Second, show by contradiction (using \textit{(ii)}, \textit{(iii)}) that $\hb(\cdot)_{1:n_c}$ can, in fact, only depend on the true content~$\cb$ and not on style~$\sb$, \looseness-1 for otherwise the invariance from step 1 would be violated in a region of the style (sub)space of measure greater than zero.

\paragraph{Intuition. 
}
\cref{thm:main} assumes that the number of content ($n_c$) and style ($n_s$) variables is known,
and that there is a positive probability that each \textit{style} variable may change, though not necessarily on its own, according to \textit{(iii)}.
In this case, training a generative model of the form specified in~\cref{sec:problem_formulation} 
(i.e., with an invariant content partition and subsets of changing style variables)
by maximum likelihood on pairs $(\xb,\xbt)$ will asymptotically (in the limit of infinite data) recover the true invariant content partition up to an invertible function, i.e., it isolates, or unmixes, content from style.

\paragraph{Discussion.} The identifiability result of~\cref{thm:main} for generative SSL is of potential relevance for existing variational autoencoder (VAE)~\citep{kingma2013auto} variants such as the \texttt{GroupVAE}~\citep{hosoya2019group},\footnote{which also uses a fixed content-style partition for multi-view data, but assumes that all latent factors are mutually independent, and that all style variables change between views, independent of the original style;} or its adaptive version \texttt{AdaGVAE}~\citep{locatello2020weakly}.
Since, contrary to existing results, \cref{thm:main} does not assume independent latents, it may also provide a principled basis for generative causal representation learning algorithms~\citep{leeb2020structural,yang2021causalvae,shen2022weakly}.
However, an important limitation to its practical applicability is that generative modelling does not tend to scale very well to complex high-dimensional observations,
such as images. 

\subsection{Discriminative Self-Supervised Representation Learning}
\label{sec:theory_discriminative}
We therefore next turn to a discriminative approach, i.e., directly learning an encoder function~$\gb$ which leads to a similar embedding across $(\xb,\xbt)$. 
As discussed in~\cref{sec:background_augmentation}, this is much more common for SSL with data augmentations.
First, we show that if an \textit{invertible} encoder $\gb$ is used, then learning a representation which is aligned in the first $n_c$ dimensions is sufficient to block-identify content.

\begin{restatable}[Identifying content with
an invertible encoder]{theorem}{discriminative}
\label{thm:CL}
Assume the same data generating process~(\cref{sec:problem_formulation}) and conditions \textit{(i)}-\textit{(iv)} as in~\Cref{thm:main}. 
Let $\gb:\Xcal \rightarrow \Zcal$ be any smooth and \emph{invertible} function 
which minimises the following functional:%
\begin{equation}
\label{eq:CL_MSE_objective}
\textstyle
\Lcal_\mathrm{Align}(\gb) := \EE_{(\xb,\xbt)\sim p_{\xb, \xbt}}
\left[
\bignorm{
\gb(\xb)_{1:n_c}-\gb(\xbt)_{1:n_c}
}^2_2
\right]
\end{equation}%
Then $\gb$ block-identifies the true content variables in the sense of Definition~\ref{def:block-identifiability}.%
\end{restatable}%

\paragraph{Proof Sketch.}
First, we show that the global minimum of~\eqref{eq:CL_MSE_objective} is reached by the smooth invertible function~$\fbinv$. Thus, any other minimiser $\gb$ must satisfy the same invariance across $(\xb,\xbt)$ used in step 1 of the proof of~\cref{thm:main}. The second step uses the same argument by contradiction as in~\cref{thm:main}.

\paragraph{Intuition.}
\looseness-1 \Cref{thm:CL} states that if---under the same assumptions on the generative process as in~\cref{thm:main}---we directly learn a representation with an \textit{invertible} encoder, then enforcing alignment between the first $n_c$ latents is sufficient to isolate the invariant content partition. Intuitively, invertibility guarantees that all information is preserved, thus avoiding a collapsed representation.

\paragraph{Discussion.}
\looseness-1 According to~\cref{thm:CL}, content can be isolated if, e.g.,
a flow-based architecture%
~\citep{dinh2016density,papamakarios2017masked,dinh2014nice,kingma2018glow, papamakarios2021normalizing} is used, or  invertibility is enforced otherwise during training~\citep{jacobsen2018revnet,behrmann2019invertible}.
However, the applicability of this approach is limited as it \textit{places strong constraints on the encoder architecture which makes it hard to scale these methods up to high-dimensional settings}.
As discussed in~\cref{sec:background_augmentation}, state-of-the-art SSL methods such as \texttt{SimCLR}~\citep{chen2020simple},
\texttt{BYOL}~\citep{grill2020byol},
\texttt{SimSiam}~\citep{chen2020exploring}, or
\texttt{BarlowTwins}~\citep{zbontar2021barlow} do not use invertible encoders, but instead avoid collapsed representations---which would result from naively optimising~\eqref{eq:CL_MSE_objective} for arbitrary, non-invertible $\gb$---using different forms of regularisation.

To close this gap between theory and practice, finally, we investigate how to block-identify content without assuming an invertible encoder.
We show that, if we add a regularisation term to~\eqref{eq:CL_MSE_objective} that encourages maximum entropy of the learnt representation, the invertibility assumption can be dropped.%
\begin{restatable}[Identifying content with discriminative learning and a non-invertible encoder]{theorem}{discriminativeMaxEnt}
\label{thm:CL_MaxEnt}
Assume the same data generating process~(\cref{sec:problem_formulation}) and conditions \textit{(i)}-\textit{(iv)} as in~\Cref{thm:main}. 
Let $\gb:\Xcal \rightarrow (0,1)^{n_c}$ be any smooth function which minimises the following functional:%
\begin{equation}
\label{eq:CL_MSE_MaxEnt_objective}
\textstyle
\Lcal_\mathrm{AlignMaxEnt}(\gb)
:= 
\EE_{(\xb,\xbt)\sim p_{\xb, \xbt}}
\left[
\bignorm{
\gb(\xb)-\gb(\xbt)
}^2_2
\right] - H\left(\gb(\xb)\right)
\end{equation}%
where $H(\cdot)$ denotes the differential entropy of the random variable $\gb(\xb)$ taking values in $(0,1)^{n_c}$.
Then $\gb$ block-identifies the true content variables in the sense of~\cref{def:block-identifiability}.%
\end{restatable}%
\paragraph{Proof Sketch.}
First, use the Darmois construction~\citep{darmois1951construction,hyvarinen1999nonlinear} to build a function $\db:\Ccal\rightarrow (0,1)^{n_c}$ mapping $\cb=\fbinv(\xb)_{1:n_c}$ to a uniform random variable. Then $\gb^\star=\db\circ \fb^{-1}_{1:n_c}$ attains the global minimum of~\eqref{eq:CL_MSE_MaxEnt_objective} because $\cb$ is invariant across $(\xb,\xbt)$ and  the uniform distribution is the maximum entropy distribution on $(0,1)^{n_c}$.
Thus, any other minimiser $\gb$ of~\eqref{eq:CL_MSE_MaxEnt_objective} must satisfy invariance across $(\xb,\xbt)$ and map to a uniform r.v. 
Then, use the same step 2 as in~\cref{thm:main,thm:CL} to show that $\hb=\gb\circ \fb:\Zcal\rightarrow (0,1)^{n_c}$ cannot depend on style, i.e., it is a function from $\Ccal$ to $(0,1)^{n_c}$.
Finally, we show that $\hb$ must be invertible 
since it maps 
$p_\cb$ to a uniform distribution,
using a result from~\citet{zimmermann2021contrastive}.

\paragraph{Intuition.}
\cref{thm:CL_MaxEnt} states that if we do not explicitly enforce invertibility of~$\gb$ as in~\cref{thm:CL}, additionally maximising the entropy of the learnt representation (i.e., optimising alignment \textit{and} uniformity~\citep{wang2020understanding})  avoids a collapsed representation and recovers the invariant content block. Intuitively, this is because any function that only depends on $\cb$ will be invariant across $(\xb,\xbt)$, so it is beneficial to preserve all content information to maximise entropy. 

\paragraph{Discussion.
}
Of our theoretical results, \cref{thm:CL_MaxEnt} requires the weakest set of assumptions, and is most closely aligned with common SSL practice.
As discussed in~\cref{sec:background_augmentation}, contrastive SSL with negative samples using InfoNCE~\eqref{eq:InfoNCE_objective} as an objective 
can asymptotically be understood as alignment with entropy regularisation~\citep{wang2020understanding}, i.e., objective~\eqref{eq:CL_MSE_MaxEnt_objective}.
\textit{\Cref{thm:CL_MaxEnt} thus provides a theoretical justification for the empirically observed effectiveness of CL with InfoNCE}:
subject to our assumptions, CL with InfoNCE asymptotically isolates content, i.e., the part of the representation that is always left invariant by augmentation.
For example, the 
strong image classification performance based on representations learned by \texttt{SimCLR}~\citep{chen2020simple},
which uses color distortion and random crops as augmentations, can be explained in that object class is a content variable in this case.
We extensively evaluate the effect of various augmentation techniques on different ground-truth latent factors in our experiments in~\cref{sec:ssl_experiments}.
There is also an interesting connection between~\cref{thm:CL_MaxEnt} and \texttt{BarlowTwins}~\citep{zbontar2021barlow}, which only uses positive pairs and combines alignment with a redundancy reduction regulariser that enforces decorrelation between the inferred latents.
Intuitively, redundancy reduction is related to increased entropy: $\gb^\star$ constructed in the proof of~\cref{thm:CL_MaxEnt}---and thus also any other minimiser of~\eqref{eq:CL_MSE_MaxEnt_objective}---attains the global optimum of the \texttt{BarlowTwins} objective, though the reverse implication may not hold.

\section{Experiments}
\label{sec:ssl_experiments}

We perform two main experiments.
First,
we numerically test our main result,~\cref{thm:CL_MaxEnt},
in a \textit{fully-controlled}, finite sample setting~(\cref{sec:experiment_1_numerical_simulation}), using CL to estimate the entropy term in~\eqref{eq:CL_MSE_MaxEnt_objective}.
Second, we seek to better understand the effect of data augmentations used \textit{in practice}~(\cref{sec:experiment_2_causal3dident}).
To this end, we introduce a new
dataset of 3D objects with 
dependencies between a number of known ground-truth latents, and use it to evaluate the effect of different augmentation techniques
on what is identified as content.
Additional results and analyses, as well as implementation details, are  described in Appendices~C and D of~\cite{von2021self}.

\subsection{Numerical Data}
\label{sec:experiment_1_numerical_simulation}

\paragraph{Experimental Setup.}
We generate synthetic data as described in~\cref{sec:problem_formulation}. We consider $n_c=n_s=5$, with content and style latents distributed as $\cb\sim\Ncal(0,\Sigma_c)$ and $\sb|\cb\sim\Ncal(\ab+B\cb, \Sigma_s)$, thus allowing for \emph{statistical dependence} within the two blocks (via $\Sigma_c$ and $\Sigma_s$) and \emph{causal dependence} between content and style (via $B$). For $\fb$, we use a
3-layer MLP with LeakyReLU activation functions.\footnote{chosen to lead to invertibility almost surely by following the settings used by previous work~\citep{hyvarinen2016unsupervised,hyvarinen2017nonlinear}}
The distribution $p_A$ over subsets of changing style variables is obtained by independently flipping the same biased coin for each $s_i$.
The conditional style distribution is taken as $p_{\sbt_A|\sb_A}=\Ncal(\sb_A,\Sigma_A)$.
We train an encoder $\gb$ on pairs $(\xb,\xbt)$ with InfoNCE using the negative L2 loss as the similarity measure, i.e., we approximate~\eqref{eq:CL_MSE_MaxEnt_objective} using empirical averages and negative samples. 
For evaluation, we 
use kernel ridge regression~\citep{murphybook} to predict the ground truth $\cb$ and $\sb$ from the learnt representation $\cbh=\gb(\xb)$ and report the $R^2$ coefficient of determination. 

\begin{table}[t]%
    \centering
    \caption[Numerical Simulation Results]{\textbf{Numerical simulation results.} $R^2$ (linear and nonlinear) for predicting the ground-truth content $\cb$ or style $\sb$ from the learnt representation $g(\xb)$; mean $\pm$ standard deviation over $3$ random seeds.  ``p(chg.)'' denotes the probability with which each style variable changes, ``Stat.'' stands for {statistical dependence} within blocks of content and style variables, and ``Cau.'' refers to style being {causally dependent} on content.}
    \label{tab:syn_sim}
    \begin{tabular}{ccccccc}
    \toprule
    \multicolumn{3}{c}{\textbf{Generative process}} & \multicolumn{2}{c}{$\bm{R^2}$ \textbf{(nonlinear)}} & \multicolumn{2}{c}{$\bm{R^2}$ \textbf{(linear)}}  \\
    \cmidrule(r){1-3}\cmidrule(r){4-5} \cmidrule(r){6-7}
    \textbf{p(chg.)} & \textbf{Stat.} & \textbf{Cau.} & \textbf{Content $\cb$} & \textbf{Style $\sb$} & \textbf{Content $\cb$} & \textbf{Style $\sb$} \\
    \midrule
    1.0 & \xmark & \xmark & $\textbf{1.00} \pm 0.00$ & $\textcolor{red}{0.07} \pm 0.00$ & $\textbf{1.00} \pm 0.00$ & $\textcolor{red}{0.00} \pm 0.00$\\
    0.75 & \xmark & \xmark & $\textbf{1.00} \pm 0.00$ & $\textcolor{red}{0.06} \pm 0.05$ & $\textbf{0.99} \pm 0.00$ & $\textcolor{red}{0.00} \pm 0.00$ \\
    0.75 & \cmark & \xmark & $\textbf{0.98} \pm 0.03$ & ${0.37} \pm 0.05$  & $\textbf{0.97} \pm 0.03$ & $0.37 \pm 0.05$ \\
    0.75 & \cmark & \cmark & $\textbf{0.99} \pm 0.01$ & $\textbf{0.80} \pm 0.08$ & $\textbf{0.98} \pm 0.01$ & $\textbf{0.78} \pm 0.07$ \\
    \bottomrule
    \end{tabular}
\end{table}

\begin{figure}[t]
\centering
\includegraphics[width=\textwidth]{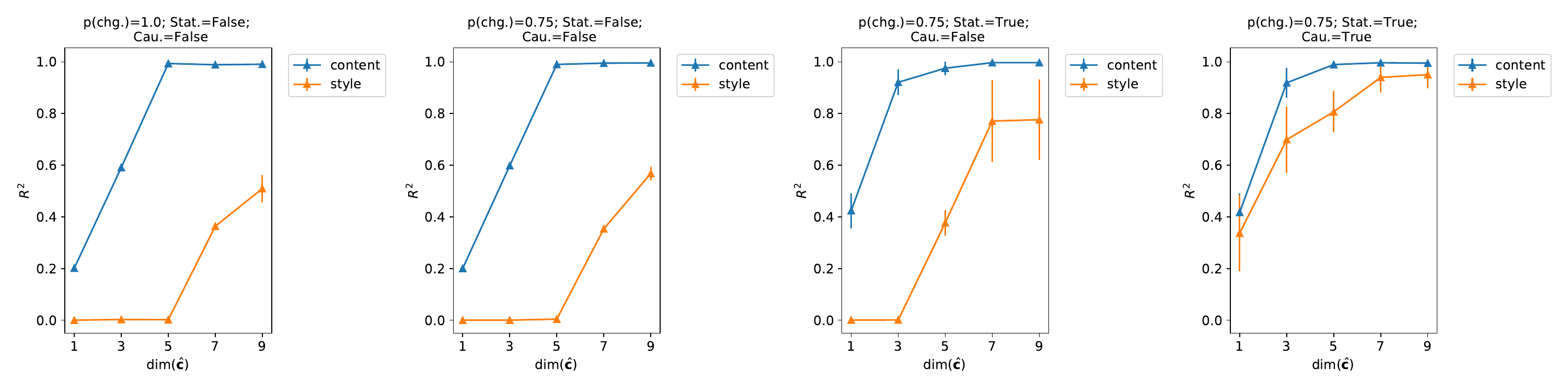}
\caption[Ablation on the Dimensionality of the Learnt Representation]{\textbf{Ablation on the Dimensionality of the Learnt Representation.} Identifiability scores (nonlinear $R^2$) for the content \& style blocks in the numerical experiment (\cref{sec:experiment_1_numerical_simulation}) as a function of the dimensionality $\text{dim}(\hat{\cb})$ of the learnt representation. The four plots (left to right) correspond to the different settings listed in~\cref{tab:syn_sim} (see plot titles). }
\label{fig:model_latent_ablation}
\end{figure}

\paragraph{Results.}
In~\cref{tab:syn_sim} we report mean $\pm$ std.\ dev.\ over $3$ random seeds across four generative processes of increasing complexity covered by~\cref{thm:CL_MaxEnt}: ``p(chg.)'', ``Stat.'', and ``Cau.'' denote respectively the change probability for each~$s_i$,
statistical dependence within blocks ($\Sigma_c\neq I\neq \Sigma_s$), and  causal dependence of style on content ($B\neq 0$).
An $R^2$ close to one indicates that almost all variation is explained by $\cbh$, i.e., that there is a 1-1 mapping, 
as required by~\cref{def:block-identifiability}.
As can be seen, \textit{across all settings, content is block-identified}. %
Note that a high $R^2$ score can be obtained even if we use linear regression to predict $\cb$ from $\cbh$.

We also perform an ablation and visualise in~\cref{fig:model_latent_ablation} how varying the dimensionality $\text{dim}(\cbh)$ of the learnt representation affects identifiability of the ground-truth content \& style partition. 
Generally,  we find that if $\text{dim}(\cbh)<n_c$,
there is insufficient capacity to encode all content, so a lower-dimensional mixture of content is learnt. 
Conversely, if $\text{dim}(\cbh)>n_c$, the excess capacity is used to encode some style information, as that increases entropy. 
As can be seen from~\cref{tab:syn_sim} and~\cref{fig:model_latent_ablation}, scores for identifying style increase substantially when dependences within and between blocks are included. 
If we compare the performance for small latent dimensionalities ($\text{dim}(\cbh)<n_c$) between the first two (without) and the third plot (with statistical dependence) of~\Cref{fig:model_latent_ablation},
we observe a significantly higher score in identifying content for the latter (e.g., $R^2$ of ca.\ 0.4 vs 0.2 at $\text{dim}(\cbh)=1$). 
This suggests that the introduction of statistical dependence among content variables  (as well as among style variables, and in how style variables change) in the third plot/row, reduces the effective dimensionality of the ground-truth latents and thus leads to higher content identifiability for the same~$\text{dim}(\cbh)<n_c$.
Since the $R^2$ for content is already close to 1 for $\text{dim}(\cbh)=3$ in the third plot of~\cref{fig:model_latent_ablation} (due to the smaller effective dimensionality induced by statistical dependence between~$\cb$), when $\text{dim}(\cbh)=n_c=5$ is used (as reported in~\cref{tab:syn_sim}), excess capacity is used to encode style, leading to a positive $R^2$.

Regarding causal dependence (i.e., the fourth plot in~\cref{fig:model_latent_ablation} and fourth row in~\cref{tab:syn_sim}), we note that the ground truth dependence between $\cb$ and $\sb$ is linear, i.e., $p(\sb|\cb)$ is centred at a linear transformation $\ab+B\cb$ of $\cb$.
Given that our evaluation consists of predicting the ground truth $\cb$ and $\sb$ from the learnt representation $\cbh=\gb(\xb)$, if we were to block-identify $\cb$ according to~\cref{def:block-identifiability}, we should be able to also predict some aspects of $\sb$ from $\cbh$.
This manifests in a relatively large $R^2$ for $\sb$ in the last row of~\cref{tab:syn_sim}.
To summarise, we highlight two main takeaways: (i) when latent dependence is present, this may reduce the effective dimensionality, so that some style is encoded in addition to content unless a smaller representation size is chosen; (ii) even though the learnt representation isolates content in the sense of~\cref{def:block-identifiability}, it may still be predictive of style when content and style are dependent.

\subsection{High-Dimensional Images: \textit{Causal3DIdent}}
\label{sec:experiment_2_causal3dident}

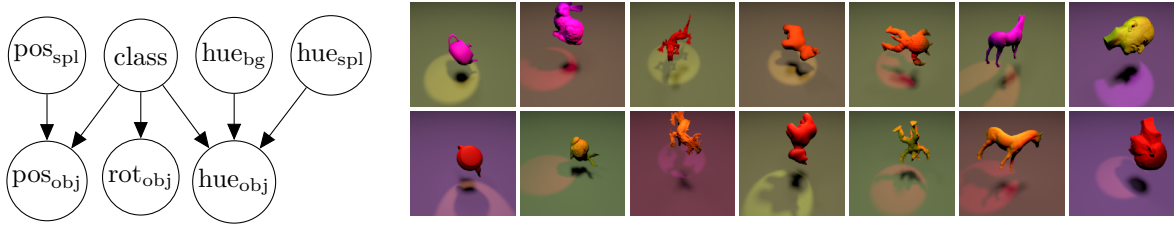
\begin{figure}[t]
    \newcommand{\xshift}{2em}
    \newcommand{\yshift}{1.em}
    \newcommand{\nodesize}{2.5em}
    \begin{subfigure}{0.35\textwidth}
    \centering
        \begin{tikzpicture}
            \centering
            \node (spotlight angle)
            [latent, minimum size=\nodesize]
            {\small $\text{pos}_\text{spl}$};
            \node (class)
            [latent, right=of spotlight angle, xshift=-\xshift, minimum size=\nodesize]
            {\small class};
            \node (object position)
            [latent, below=of spotlight angle, yshift=\yshift, minimum size=\nodesize]
            {\small $\text{pos}_\text{obj}$};
            \node (object rotation)
            [latent, below=of class, yshift=\yshift, minimum size=\nodesize]
            {\small $\text{rot}_\text{obj}$};
            \node (background hue) 
            [latent, right=of class, xshift=-\xshift, minimum size=\nodesize] 
            {\small $\text{hue}_\text{bg}$};
            \node (object hue)
            [latent, below=of background hue, yshift=\yshift, minimum size=\nodesize]
            {\small $\text{hue}_\text{obj}$};
            \node (spotlight hue)
            [latent, right=of background hue, xshift=-\xshift, minimum size=\nodesize]
            {\small $\text{hue}_\text{spl}$};
            \edge{class}{object hue, object position, object rotation};
            \edge[]{spotlight angle}{object position};
            \edge[]{spotlight hue, background hue}{object hue};
        \end{tikzpicture}
    \end{subfigure}%
    \begin{subfigure}{0.65\textwidth}
        \centering
        \includegraphics[width=\textwidth]{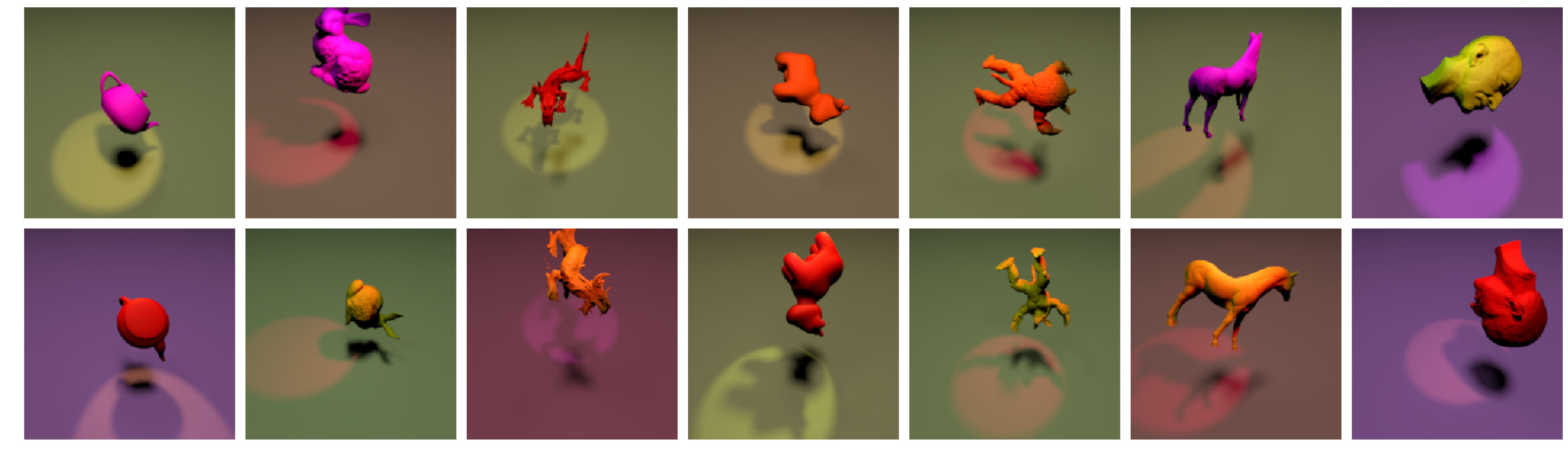}
    \end{subfigure}
    \caption[The Causal3DIdent Dataset]{\textbf{The Causal3DIdent Dataset.}
    \textit{(Left)} Causal graph for the \textit{Causal3DIdent} dataset. 
    \textit{(Right)} Two samples from each object class.
    }
    \label{fig:3DIdent_causal_graph}
\end{figure}

\paragraph{\textit{Causal3DIdent} Dataset.}
\textit{3DIdent}~\citep{zimmermann2021contrastive} is a benchmark for evaluating identifiability with rendered $224\times224$ images which contains hallmarks of natural environments (e.g., shadows, different lighting conditions, a 3D object). 
In \textit{3DIdent}, there is a single object class (Teapot), and all $10$ latents are sampled independently. For \textit{Causal3DIdent}, we introduce {six} additional classes (Hare, Dragon, Cow, Armadillo, Horse, and Head) and impose a causal graph over the latent variables, see~\cref{fig:3DIdent_causal_graph}. 
While object class and all environment variables (spotlight position \& hue, background hue) are sampled independently, all object latents are dependent.\footnote{E.g., our causal graph entails hares blend into the environment (object hue centred about background \& spotlight hue), a form of active camouflage observed for some hare species.}
Further details on Causal3DIdent can be found in Appendix B of~\citet{von2021self}. The dataset is publicly available at \href{https://zenodo.org/record/4784282}{https://zenodo.org/record/4784282}.  

\begin{figure}[t]
\centering
    \includegraphics[width=0.57\textwidth]{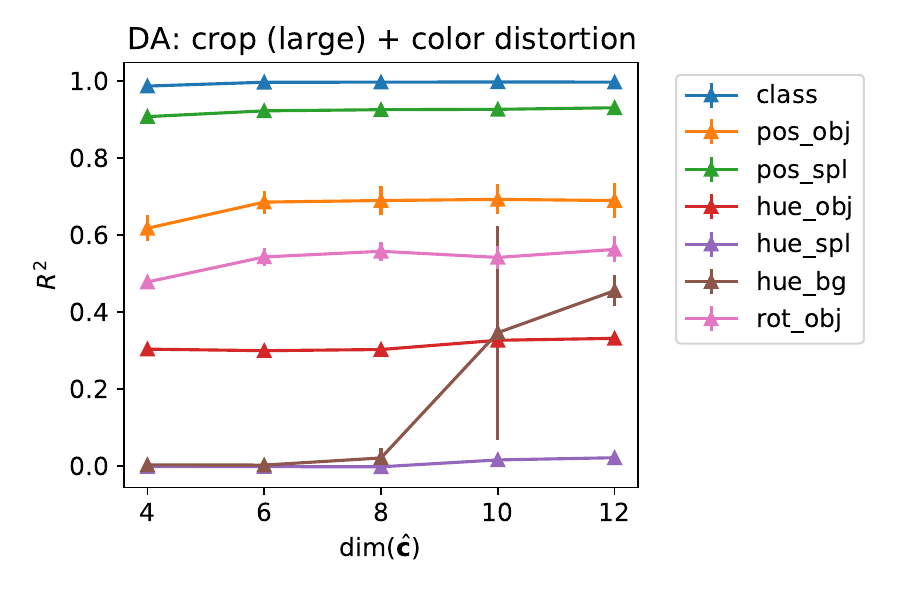}
    \vspace{-1.em}
    \caption[Ablation on the Content Dimension on Causal3DIdent]{\textbf{Ablation on the Content Dimension on Causal3DIdent.} Effect of the content dimension $\text{dim}(\cbh)$ on identifiability  scores for different generative factors on Causal3DIdent.}
    \label{fig:ablation}
\end{figure}

\paragraph{Experimental Setup.}
For $\gb$, we train a convolutional
encoder composed of a ResNet18%
~\citep{he2015deep} and an additional fully-connected layer, with LeakyReLU activation.
As in \texttt{SimCLR}~\citep{chen2020simple}, we use
InfoNCE with cosine 
similarity, 
and train on pairs of augmented examples~$(\xbt,\xbt')$.
As $n_c$ is unknown and variable depending on the augmentation, we fix $\text{dim}(\cbh)=8$ throughout.
Note that we
find the results to be, for the most part, robust to the choice of $\text{dim}(\cbh)$,
see~\cref{fig:ablation}.
We consider the following data augmentations (DA): crop, resize \& flip; colour distortion (jitter \& drop); and rotation {\small$\in\{90\degree,180\degree,270\degree\}$}. For comparison, we also consider directly imposing a content-style partition
by
performing a latent transformation (LT) to generate views. 
\looseness-1 For evaluation, we use linear logistic regression to predict object class, and 
kernel ridge 
to predict the other latents from $\cbh$.
(For results with linear regression, as well as evaluation using a higher-dimensional intermediate layer by considering a projection head~\citep{chen2020simple}, see~\citet[][Appendix C]{von2021self}.)

\newpage
\paragraph{Results.}
The results are presented in~\cref{tbl:final_nonlinear_abbrv}.
Overall, our main findings are that:
\begin{itemize}[topsep=0pt,itemsep=0pt,leftmargin=*]
\item it can be difficult to design image-level augmentations that leave \textit{specific} latents invariant;
\item augmentations \& latent transformations generally have a similar effect on groups of latents;
\item augmentations that yield good classification performance induce variation in all other latents.
\end{itemize}
We 
observe that, similar to directly varying the hue latents, colour distortion leads to
a discarding of hue information as style, and a preservation of (object) position
as content.
Crops, similar to varying the position latents, lead to a discarding of position
as style, and a preservation of background and object hue 
as content, the latter assuming crops are sufficiently large.
\looseness-1 In contrast, image-level rotation 
affects both the object rotation and position, and thus
deviates
from only varying the rotation latents.

Whereas class is always preserved as content when generating views with latent transformations,
when using data augmentations,
we can only reliably decode class when crops \& colour distortion are used in conjunction---a result which mirrors evaluation on ImageNet~\citep{chen2020simple}. As can be seen by our evaluation of crops \& colour distortion in isolation, while colour distortion leads to a discarding of hues as style, crops lead to a discarding of position \& rotation as style. Thus, when used in conjunction, class is isolated as the sole content variable. 

\begin{table}[tb]
\definecolor{LTcolor}{rgb}{.9,.9,.9}
\definecolor{DAcolor}{rgb}{1,1,1}
\centering
\caption[Causal3DIdent Results]{\textbf{{Causal3DIdent} Results.} $R^2$ mean $\pm$ std.\ dev.\  over $3$ random seeds. DA: data augmentation, LT: latent transformation, bold: $R^2\geq 0.5$, red: $R^2<0.25$.
Results for individual axes of object position \& rotation are aggregated.
}
\label{tbl:final_nonlinear_abbrv}
\resizebox{\textwidth}{!}{
\begin{tabular}{lc|cc|ccc|c}
\toprule
\multirow{2}{*}{\textbf{Views generated by}} & \multirow{2}{*}{\textbf{Class}} & \multicolumn{2}{c}{\textbf{Positions}} & \multicolumn{3}{c}{\textbf{Hues}} & \multirow{2}{*}{\textbf{Rotations}} \\
\cmidrule(r){3-4}\cmidrule(r){5-7}
& & $\text{object}$ & $\text{spotlight}$ & $\text{object}$ & $\text{spotlight}$ & $\text{background}$ & 
\\
\midrule
\rowcolor{DAcolor}
DA: colour distortion  & 
$0.42 \pm 0.01$ & $\textbf{0.61} \pm 0.10$ & $\textcolor{red}{0.17} \pm 0.00$ & $\textcolor{red}{0.10} \pm 0.01$ & $\textcolor{red}{0.01} \pm 0.00$ & $\textcolor{red}{0.01} \pm 0.00$ & $0.33 \pm 0.02$ \\
\rowcolor{LTcolor}
LT: change hues & 
$\textbf{1.00} \pm 0.00$ & $\textbf{0.59} \pm 0.33$ & $\textbf{0.91} \pm 0.00$ & $0.30 \pm 0.00$ & $\textcolor{red}{0.00} \pm 0.00$ & $\textcolor{red}{0.00} \pm 0.00$ & $0.30 \pm 0.01$ \\
\midrule
\rowcolor{DAcolor}
DA: crop (large) & 
$0.28 \pm 0.04$ & $\textcolor{red}{0.09} \pm 0.08$ & $\textcolor{red}{0.21} \pm 0.13$ & $\textbf{0.87} \pm 0.00$ & $\textcolor{red}{0.09} \pm 0.02$ & $\textbf{1.00} \pm 0.00$ & $\textcolor{red}{0.02} \pm 0.02$ \\
\rowcolor{DAcolor}
DA: crop (small) & 
$\textcolor{red}{0.14} \pm 0.00$ & $\textcolor{red}{0.00} \pm 0.01$ & $\textcolor{red}{0.00} \pm 0.01$ & $\textcolor{red}{0.00} \pm 0.00$ & $\textcolor{red}{0.00} \pm 0.00$ & $\textbf{1.00} \pm 0.00$ & $\textcolor{red}{0.00} \pm 0.00$ \\
\rowcolor{LTcolor}
LT: change positions & 
$\textbf{1.00} \pm 0.00$ & $\textcolor{red}{0.16} \pm 0.23$ & $\textcolor{red}{0.00} \pm 0.01$ & $0.46 \pm 0.02$ & $\textcolor{red}{0.00} \pm 0.00$ & $\textbf{0.97} \pm 0.00$ & $0.29 \pm 0.01$ \\
\midrule
\rowcolor{DAcolor}
DA: crop (large) + colour distortion & %
$\textbf{0.97} \pm 0.00$ & $\textbf{0.59} \pm 0.07$ & $\textbf{0.59} \pm 0.05$ & $0.28 \pm 0.00$ & $\textcolor{red}{0.01} \pm 0.01$ & $\textcolor{red}{0.01} \pm 0.00$ & $\textbf{0.74} \pm 0.03$ \\
\rowcolor{DAcolor}
DA: crop (small) + colour distortion & %
$\textbf{1.00} \pm 0.00$ & $\textbf{0.69} \pm 0.04$ & $\textbf{0.93} \pm 0.00$ & $0.30 \pm 0.01$ & $\textcolor{red}{0.00} \pm 0.00$ & $\textcolor{red}{0.02} \pm 0.03$ & $\textbf{0.56} \pm 0.03$ \\
\rowcolor{LTcolor}
LT: change positions + hues & 
$\textbf{1.00} \pm 0.00$ & $\textcolor{red}{0.22} \pm 0.22$ & $\textcolor{red}{0.07} \pm 0.08$ & $0.32 \pm 0.02$ & $\textcolor{red}{0.00} \pm 0.01$ & $\textcolor{red}{0.02} \pm 0.03$ & $0.34 \pm 0.06$ \\
\midrule
\rowcolor{DAcolor}
DA: rotation &
$0.33 \pm 0.06$ & $\textcolor{red}{0.17} \pm 0.09$ & $\textcolor{red}{0.23} \pm 0.12$ & $\textbf{0.83} \pm 0.01$ & $0.30 \pm 0.12$ & $\textbf{0.99} \pm 0.00$ & $\textcolor{red}{0.05} \pm 0.03$ \\
\rowcolor{LTcolor}
LT: change rotations & 
$\textbf{1.00} \pm 0.00$ & $\textbf{0.53} \pm 0.33$ & $\textbf{0.90} \pm 0.00$ & $0.41 \pm 0.00$ & $\textcolor{red}{0.00} \pm 0.00$ & $\textbf{0.97} \pm 0.00$ & $0.28 \pm 0.00$ \\
\midrule
\rowcolor{DAcolor}
DA: rotation + colour distortion & 
$\textbf{0.59} \pm 0.01$ & $\textbf{0.58} \pm 0.06$ & $\textcolor{red}{0.21} \pm 0.01$ & $\textcolor{red}{0.12} \pm 0.02$ & $\textcolor{red}{0.01} \pm 0.00$ & $\textcolor{red}{0.01} \pm 0.00$ & $0.33 \pm 0.04$ \\
\rowcolor{LTcolor}
LT: change rotations + hues &
$\textbf{1.00} \pm 0.00$ & $\textbf{0.57} \pm 0.34$ & $\textbf{0.91} \pm 0.00$ & $0.30 \pm 0.00$ & $\textcolor{red}{0.00} \pm 0.00$ & $\textcolor{red}{0.00} \pm 0.00$ & $0.28 \pm 0.00$ \\
\bottomrule
\end{tabular}}
\end{table}

\paragraph{Additional Experiments.}
\label{sec:additional_experiments}
\looseness-1 We repeat our analysis from~\cref{sec:experiment_2_causal3dident} using \texttt{BarlowTwins}~\citep{zbontar2021barlow} which, as discussed at the end of~\cref{sec:theory_discriminative}, is also loosely related to~\cref{thm:CL_MaxEnt}. The results mostly mirror those obtained for~\texttt{SimCLR} and  presented in~\cref{tbl:final_nonlinear_abbrv}.
Finally, we ran the same experimental setup as in~\cref{sec:experiment_2_causal3dident} also on the \textit{MPI3D-real} dataset~\citep{gondal2019transfer} containing $>1$ million \textit{real} images with ground-truth annotations of 3D objects being moved by a robotic arm.
Subject to some caveats, the results show a similar trend as those on \textit{Causal3DIdent}.

\section{Discussion}
\label{sec:ssl_discussion}

\paragraph{Theory vs Practice.}
We have made an effort to tailor our problem formulation~(\cref{sec:problem_formulation}) to the setting of data augmentation with content-preserving transformations.
However, some of our more technical assumptions, which are necessary to 
prove block-identifiability of the invariant content partition, may not hold exactly in practice.
This is apparent, e.g., from our second experiment~(\cref{sec:experiment_2_causal3dident}), where we observe that---while class should, in principle, always be invariant across views (i.e., content)---when using
\textit{only} crops, colour distortion, or rotation,  $\gb$~appears to encode \emph{shortcuts}~\citep{geirhos2020shortcuts,pezeshki2020gradient}.%
\footnote{class is distinguished by shape, a feature commonly 
unused in 
downstream tasks on natural images%
~\citep{geirhos2019imagenettrained}}
Data augmentation, unlike latent transformations, generates views~$\xbt$ which 
are not restricted to the 11-dim.\ image
manifold~$\Xcal$ 
corresponding to the generative process of \textit{Causal3DIdent}, but may introduce additional variation: e.g., colour distortion leads to a rich combination of colours, typically a 3-dimensional feature,
whereas \textit{Causal3DIdent} only contains one degree of freedom (hue).
With additional factors, any introduced invariances may be encoded as content in place of class.
Image-level augmentations also tend to change multiple latent factors in a correlated way, which may violate assumption \textit{(iii)} of our theorems, i.e., that $p_{\sbt_\A|\sb_\A}$ is fully-supported locally.
{We also assume that $\zb$ is continuous, even though \textit{Causal3DIdent} and most disentanglement datasets also contain discrete latents. This is a very common assumption in the related literature~\citep{locatello2020weakly,gresele2019incomplete,zimmermann2021contrastive,klindt2020slowvae,locatello2019challenging,khemakhem2020variational,hyvarinen1999nonlinear,hyvarinen2019nonlinear,hyvarinen2016unsupervised} that may be relaxed in future work.}
Moreover, our theory holds asymptotically and at the global optimum, whereas in practice we solve a non-convex optimisation problem with a finite sample and need to approximate the entropy term in~\eqref{eq:CL_MSE_MaxEnt_objective}, e.g., using a finite number of negative pairs. The resulting  challenges for optimisation may be further accentuated by the higher dimensionality of $\Xcal$ induced by image-level augmentations.
Finally, we remark that while, for simplicity, we have presented our theory for pairs $(\xb,\xbt)$ of original and augmented examples, in practice, using pairs $(\xbt,\xbt')$ of two augmented views typically yields better performance. All of our assumptions (content invariance, changing style, etc) and theoretical results still apply to the latter case. We believe that using two augmented views helps because it leads to \textit{increased variability} across the pair: for if $\xbt$ and $\xbt'$ differ from $\xb$ in style subsets $A$ and $A'$, respectively, then $(\xbt,\xbt')$ differ from each other (a.s.) in the union~$A\cup A'$.

\paragraph{Beyond Entropy Regularisation.}
We have established a link between an identifiable maximum entropy approach to SSL~(\cref{thm:CL_MaxEnt}) and \texttt{SimCLR}~\citep{chen2020simple} based on the analysis of~\citep{wang2020understanding}, and have discussed an intuitive connection to the notion of redundancy reduction used in \texttt{BarlowTwins}~\citep{zbontar2021barlow}. Whether other types of regularisation such as the architectural approach pursued in \texttt{BYOL}~\citep{grill2020byol} and \texttt{SimSiam}~\citep{chen2020exploring} can also be linked to entropy maximisation remains an open question, though recent work has started exploring this direction~\citep{wen2022mechanism,liu2022bridging,tian2021understanding}.
Deriving similar results to~\cref{thm:CL_MaxEnt} with other regularisers is a promising direction for future research.

\paragraph{The Choice of Augmentation Technique Implicitly Defines Content and Style.}
As we have defined content as the part of the representation which is always left invariant across views, the choice of augmentation implicitly determines the content-style partition.
This is particularly important
to keep in mind
when applying SSL with data augmentation to safety-critical domains, such as medical imaging.
We also advise caution when using data augmentation to identify specific latent properties, since, as observed in~\cref{sec:experiment_2_causal3dident}, image-level transformations may affect the underlying ground-truth factors in unanticipated ways.
Also note that, \textit{for a given downstream task}, we may not want to discard all style information since style variables may still be correlated with the task of interest and may thus help improve predictive performance. \textit{For arbitrary downstream tasks}, however, where style may change in an adversarial way, it can be shown that only using content is optimal~\citep{rojas2018invariant}.

\paragraph{\textit{What} vs \textit{How} Information is Encoded.}
We focus on \textit{what} information is learnt by SSL with data augmentations by specifying a generative process and studying identifiability of the latent representation.
Orthogonal to this, a different line of work instead studies \textit{how} information is encoded by analysing the sample complexity needed to solve a \textit{given downstream task} using a \textit{linear} predictor~\citep{arora2019theoretical,tosh2020contrastive,lee2020predicting,tosh2021contrastive,tsai2020self,tian2021understanding}.
Provided that downstream tasks only involve content, we can draw some comparisons. 
Whereas our results recover content only up to arbitrary invertible nonlinear functions~(see~\cref{def:block-identifiability}), our problem setting is more general: \citet{arora2019theoretical,lee2020predicting} assume (approximate) independence of views $(\xb,\xbt)$ given the task (content), while \citet{tosh2021contrastive,tsai2020self} assume (approximate) independence between one view and the task (content) given the other view, neither of which hold in our setting.

\section{Extensions and Connections with Other Publications}
\label{sec:multiview_extensions}
\looseness-1 We briefly present two extensions of the multi-view CRL setting studied throughout this chapter.
In~\cref{sec:structured_data_aug}, we propose a framework for SSL that makes a more \textit{structured use of different data augmentations}
to not only recover the block of content variables, but also to \textit{identify and separate individual style components}.
In~\cref{sec:multiview_CRL_partial_obs}, we depart from SSL with data augmentation and study a more general case with an arbitrary number of views, each potentially generated by a different mixing function (\textit{multi-modality}) and depending only on a subset of latents (\textit{partial observability}). 

\subsection{Disentangling Style Variables Through Structured Data Augmentation}
\label{sec:structured_data_aug}
This subsection is based on the following publication, with all figures therein adopted without further modification.
We briefly summarise the main points that are relevant to the context of this chapter, and refer to the full paper for further details.
\begin{selfcitebox}
\href{https://arxiv.org/abs/2311.08815}{\ul{Self-supervised disentanglement	by leveraging structure in data augmentations}}
\\
Cian Eastwood, \textbf{Julius von K\"ugelgen}, Linus Ericsson, Diane Bouchacourt, Pascal Vincent, Mark Ibrahim, Bernhard Sch\"olkopf
\\
\textit{NeurIPS Workshop ``Causal Representation Learning'',} 2023
\end{selfcitebox}

\definecolor{posGreen}{rgb}{0,0.6,0}
\definecolor{negRed}{rgb}{0.83,0.17,0.16}
\definecolor{mypurple}{rgb}{0.6,0.196,1}
\definecolor{myorange}{rgb}{1,0.5,0}
\definecolor{myblue}{rgb}{0,0.4,0.8}
\newcommand{\hlr}[1]{{\color{BrickRed} #1}}
\newcommand{\hlg}[1]{{\color{gray} #1}}
\newcommand{\hlp}[1]{{\color{mypurple} #1}}
\newcommand{\hlo}[1]{{\color{myorange} #1}}
\newcommand{\hlb}[1]{{\color{myblue} #1}}

As discussed throughout this chapter, SSL typically uses data augmentations to induce invariance to style attributes.
However, with downstream tasks unknown {a priori}, it is unclear what information can be safely discarded---\textit{one task's style may be another's content}.
\citet{ericsson2021well} illustrated this point,
finding ImageNet object-classification accuracy (the task optimized for in pre-training) to be poorly correlated with downstream object-detection and dense-prediction tasks, concluding that ``universal pre-training is still unsolved''.
To address this and learn more universal representations, we introduce a new SSL framework which uses data augmentations to \textit{disentangle style features rather than discard them}.

\begin{figure}[t]
    \centering
    \includegraphics[width=\textwidth]{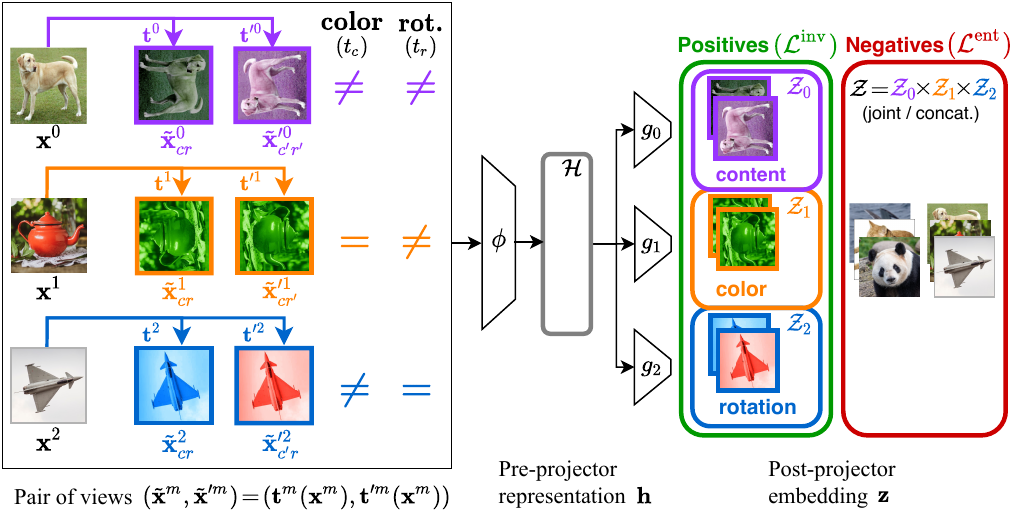}
    \caption[Structured Data Augmentation Framework for Style Disentanglement]{\textbf{Structured Data Augmentation Framework for Style Disentanglement.} 
    \textit{(Left)}
    Given $M$ atomic transformations such as \textcolor{myorange}{color distortion} or \textcolor{myblue}{rotation} (here, $M\! =\! 2$), we construct $M\! +\! 1$ transformation pairs
    $(\tb^m, \tb'^m)$
    {sharing different transformation parameters}
    and use these to create $M{+}1$ transformed image pairs
    $(\xbt^m, \xbt'^m)$ 
    {sharing different features}. \textit{(Center)} Each pair is routed to a different embedding space $\Zcal_m$. \textit{(Right)} We then (i)~enforce \textcolor{posGreen}{\textit{invariance within}} each embedding space; and (ii) maximize \textcolor{negRed}{\textit{entropy across}} the joint embedding spaces. Together, this allows us to learn a \textit{disentangled} embedding space that captures \textcolor{mypurple}{content} ($m\! =\! 0$) and $M$ style features ($m\! >\! 0$, one per atomic transformation).
    }
    \label{fig:ssl_style_overview}
\end{figure}

As illustrated in \Cref{fig:ssl_style_overview}, the key idea is to leverage different transformations in a structured way to learn multiple separate embedding spaces, each being invariant to \textit{all-but-one} augmentation.
Unlike most prior approaches to SSL with data augmentations~(\cref{sec:ssl_preliminaries}), we do \textit{not} create a single dataset of (``positive'') pairs $(\xbt,\xbt')$, but construct transformation pairs $(\tb^m,\tb'^m)$ in $M{+}1$ different ways, giving rise to $M{+}1$ datasets of pairs $(\xbt^m,\xbt'^m)$, each differing in the shared (style) properties.
To this end, we assume access to $M$ transformation distributions $\Tcal_1, \dots, \Tcal_M$ from which to sample $M$ atomic transformations $t_1, \dots, t_M$ with $t_m \sim \Tcal_m$. These are composed together to form  a final transformation 
$\tb = t_1 \circ \dots \circ t_M$. Critically, each atomic transformation $t_m$ is designed to perturb a different style attribute of the data.
For example, we could sample parameters for a \textcolor{myorange}{color distortion $t_c \sim \Tcal_c$} and \textcolor{myblue}{rotation $t_r \sim \Tcal_r$}, and then compose them as $\tb = \textcolor{myorange}{t_c} \circ \textcolor{myblue}{t_r}$.
We then construct $M + 1$ transformation pairs $\{(\tb^m,\tb'^m)\}_{m=0}^M$ which \textit{share different transformation parameters}. For $m\! = \!0$, we independently sample two transformations $\tb^0, \tb'^0 \sim \Tcal$, which will generally not share any transformation parameters (i.e., $t^0_k\neq t^{'0}_k$ $\forall k$).
For $1 \leq m \leq M$, we also independently sample two transformations $\tb^m, \tb'^m \sim \Tcal$, but then enforce that \textbf{the parameters of the $m^{\text{th}}$ transformation are shared} by setting $t'^m_m:=t^m_m$.
We apply each transformation pair to a different image $\xb^m$ to form a pair of views $(\xbt^m,\xbt'^m)=(\tb^m(\xb^m),\tb'^m(\xb^m))$.

\begin{example}[\textcolor{myorange}{Color} and \textcolor{myblue}{Rotation}]
\label{ex:color_rotation}
While some invariance to (or discarding of) an image's color and orientation features can be \textit{beneficial} for ImageNet object classification~\citep{chen2020simple}, it can also be \textit{detrimental} for other tasks like segmentation or fine-grained species classification~\citep{cole2022does}.
Suppose we can sample parameters for two transformations: \textcolor{myorange}{color distortion $t_c \sim \Tcal_c$} and \textcolor{myblue}{rotation $t_r \sim \Tcal_r$}.
As depicted in \cref{fig:ssl_style_overview}, we can then construct three transformation pairs \textit{sharing different parameters}: 
\begin{itemize}
    \item $\textcolor{mypurple}{(\tb^0,\tb'^0) = (t^0_c \circ t^0_r, t'^0_c \circ t'^0_r)}$ with \textcolor{mypurple}{no shared parameters}; 
    \item $\textcolor{myorange}{(\tb^1,\tb'^1)= (\hlg{t^1_c} \circ t^1_r,\hlg{t^1_c} \circ t'^1_r)}$ with \textcolor{myorange}{shared color parameters} \hlg{$t^1_c$}; and 
    \item $\textcolor{myblue}{(\tb^2,\tb'^2) = (t^2_c \circ \hlg{t^2_r},t'^2_c \circ \hlg{t^2_r})}$ with \textcolor{myblue}{shared rotation parameters}~$\hlg{t^2_r}$. 
\end{itemize}
Applying each transformation pair to a different image, we get three pairs of views: 
\begin{itemize}
    \item $\textcolor{mypurple}{(\Tilde{\xb}^0_{cr},\Tilde{\xb}^{0}_{c'r'})}$ for which only \textcolor{mypurple}{``content'' information is shared}  as both color and rotation differ;
    \item $\textcolor{myorange}{(\Tilde{\xb}^1_{cr},\Tilde{\xb}^{1}_{cr'})}$ for which ``content'' and \textcolor{myorange}{color information is shared}, but rotation differs; and
    \item $\textcolor{myblue}{(\Tilde{\xb}^2_{cr},\Tilde{\xb}^{2}_{c'r})}$ for which ``content'' and \textcolor{myblue}{rotation information is shared}, but color differs.
\end{itemize}
\end{example}

Each of the $M{+}1$ pairs $(\xbt^m,\xbt'^m)$ is encoded to a different embedding space $\Zcal_m$. We then propose an adapted SSL objective of maximising (i) \textcolor{posGreen}{invariance, or alignment, within each embedding space $\Zcal_m$} and (ii) \textcolor{negRed}{joint entropy across all spaces $\Zcal_0\times ... \times \Zcal_M$}.
To study what is learnt by this approach, we formalize the proposed augmentation procedure from a latent-variable-model perspective, building on the setting studied in this chapter. First, we partition the style block $\sb = (\sb_1, ..., \sb_M)$ into more fine-grained individual \textit{style components}~$\sb_m$.
Second, since the $m$\textsuperscript{th} atomic transformation is shared across $(\tb^m,\tb'^m)$ by construction, we design $M{+}1$ style conditionals $p^{(m)}_{\sbt,\sbt'|\sb}$ such that $(\xbt^m,\xbt'^m)=(\fb(
\cb,\sbt^m), \fb(\cb, \sbt'^m))$ share not only content, but also the same perturbed $m$\textsuperscript{th} style components $\sbt^m_m=\sbt'^m_m$---\textit{regardless of its original value $\sb_m$}.

\textbf{\Cref{ex:color_rotation} (continued).}
Denote the style component capturing \hlo{color} by \hlo{$\sb_c$} and that capturing \hlb{rotation} by \hlb{$\sb_r$}.
For $m=0,1,2$, let $\zb^m=(\cb^m,\sb_c^m,\sb_r^m)$ be the latents underlying separate images~$\xb^m$.
Then the augmentations shown in~\cref{fig:ssl_style_overview}~(left) are captured by the following changes to the latents:

\vspace{.5em}
\begin{minipage}{0.75\textwidth}
\small 
\centering
\renewcommand{\arraystretch}{1.5}%
\captionof{table}[Latents Style Changes for~\cref{ex:color_rotation}]{\textbf{Latents Style Changes for~\cref{ex:color_rotation}.}}
\begin{tabular}{ccccc}
\toprule
$m$ 
& $\zb^m$ 
& $\zbt^m$ & $\zbt'^m$ & \textbf{Shared Latents} \\
\midrule 
$0$ 
& $(\hlp{\cb^0},\sb_c^0,\sb_r^0)$ 
& $(\hlp{\cb^0},\tilde\sb^0_c,\tilde\sb^0_r)$ & $(\hlp{\cb^0},\tilde\sb'^0_c,\tilde\sb'^0_r)$ & only \hlp{content}
\\
$1$ 
& $(\hlp{\cb^1},\sb_c^1,\sb_r^1)$ 
& $(\hlp{\cb^1},\hlo{\tilde\sb^1_c},\tilde\sb^1_r)$ & $(\hlp{\cb^1},\hlo{\tilde\sb^1_c},\tilde\sb'^1_r)$ &
\hlp{content} \& \hlo{color}
\\
$2$
& $(\hlp{\cb^2},\sb_c^2,\sb_r^2)$ 
& $(\hlp{\cb^2},\tilde\sb^2_c,\hlb{\tilde\sb^2_r})$ & $(\hlp{\cb^2},\tilde\sb'^2_c,\hlb{\tilde\sb^2_r})$ &
\hlp{content} \& \hlb{rotation}
\\
\bottomrule 
\end{tabular}
\end{minipage}%
\begin{minipage}{0.25\textwidth}
\centering
\begin{tikzpicture}
\centering
\newcommand{\xshift}{3em}
\newcommand{\yshift}{3em}
\node (C) [latent] {$\hlp{\cb}$};
\node (S_C) [latent, yshift=-\yshift, xshift=-\xshift] {$\hlo{\sb_c}$};
\node (S_R) [latent, yshift=-\yshift, xshift=\xshift] {$\hlb{\sb_r}$};
\node (X) [obs, yshift=-2*\yshift, ] {$\xb$};
\edge{C}{S_C,S_R};
\edge{C,S_C,S_R}{X};
\end{tikzpicture}
\captionof{figure}[Causal Graph for~\cref{ex:color_rotation}]{\textbf{Causal Graph for~\cref{ex:color_rotation}.}}
\label{fig:ssl_style_causal_graph}
\end{minipage}
\vspace{1em}

Similar to~\cref{sec:counterfactual_interpretation_data_aug}, this generative process can be interpreted counterfactually if different style components do not causally influence each other, as in~\cref{fig:ssl_style_causal_graph}.
We then prove that indeed both content \textit{and} the different style components can be identified, provided that $\cb$ and $\{\sbt_m\}_{m\in [M]}$  are jointly independent, as is the case for perfect interventions~\citep{brehmer2022weakly} or if content and style are independent to begin with~\citep{lyu2021understanding}. 

In summary, we show that the structure underlying different data augmentations can be exploited to design SSL methods that do not only recover content, but also recover and disentangle style features.

\subsection{Multi-View, Multi-Modal CRL with Partial Observability}
\label{sec:multiview_CRL_partial_obs}
This subsection is based on the following publication, with all figures therein adopted without further modification.
We briefly summarise the main points that are relevant to the context of this chapter, and refer to the full paper for further details.
\begin{selfcitebox}
\href{https://openreview.net/forum?id=OGtnhKQJms}{\ul{Multi-view causal representation learning with partial observability}}
\\
Dingling Yao, Danru Xu, S\'ebastien Lachapelle, Sara Magliacane, Perouz Taslakian,\\ Georg Martius, \textbf{Julius von K\"ugelgen}, Francesco Locatello
\\
\textit{International Conference on Learning Representations (ICLR),} 2023
\end{selfcitebox}

\begin{figure}
    \centering
    \begin{subfigure}{0.375\textwidth}
        \centering
        \includegraphics[width=\textwidth]{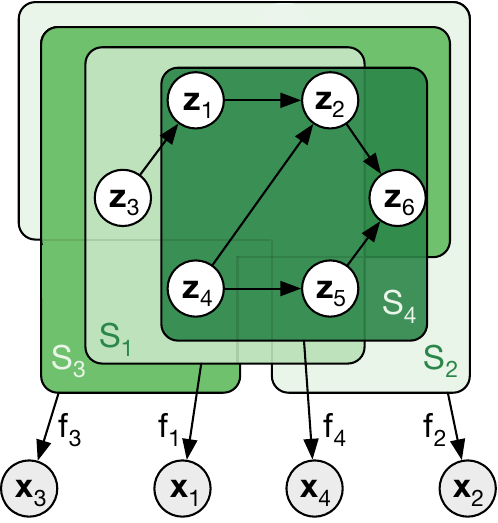}
        \caption{}
        \label{fig:multi_view_graph}
    \end{subfigure}%
    \begin{subfigure}{0.625\textwidth}
        \centering
        \includegraphics[width=\textwidth]{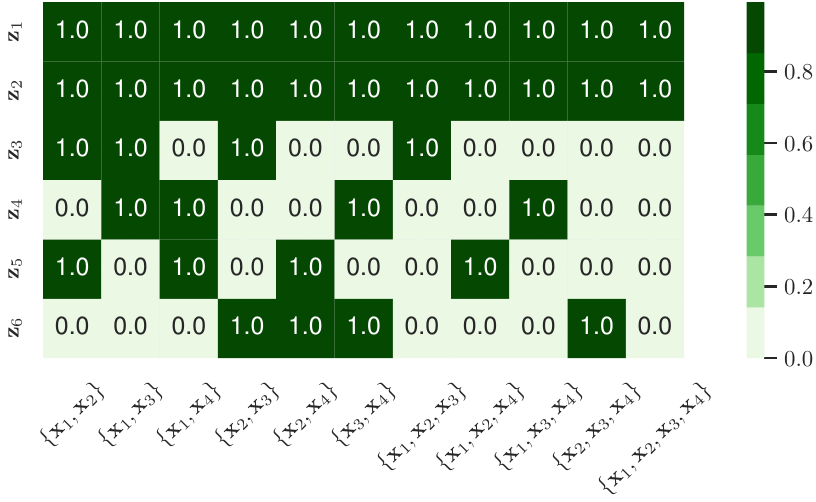}   
        \caption{}
        \label{fig:multi_view_results}
    \end{subfigure}
    \caption[Multi-View Setting with Partial Observability]{\textbf{Multi-View Setting with Partial Observability.} (a) Example setting with $K{=}4$ views and $N{=}6$ latents. Each view $\xb_k$ is generated by a subset $\zb_{S_k}$ of the latent variables through a {view-specific mixing function}~$f_k$.
    Directed arrows between latents indicate causal relations. (b) Average $R^2$ across multiple views generated from \emph{independent} latents.
    }
    \label{fig:partial_CRL}
\end{figure}

Whereas the generative processes considered thus far in this chapter were inspired by data augmentation practices in SSL, we now depart from this motivation and consider a multi-view CRL setting that relaxes  previous assumptions along different axes. First, we consider learning from an \textit{arbitrary number of} $K\geq 2$ \textit{views} $(\xb_1, \dots, \xb_K)$, rather than from pairs $(\xbt,\xbt')$. Second, each view $\xb_k$ may be generated by a different mixing function $\fb_k$, thus allowing us to learn from \textit{different data modalities}~\citep{daunhawer2023identifiability,lyu2021understanding}. Third, each view may only depend on a subset $S_k\subseteq [n]$ of the underlying latents (\textit{partial observability}), thus relaxing the assumption that $\fb$ is invertible in all latents. 
As an example, a person may undertake different medical exams, each shedding some light on their overall health status but none offering a comprehensive view:  an X-ray may show a broken bone, an MRI may show 
how the fracture affected nearby tissues, and a blood sample may inform about ongoing infections.
We model this through the following data generating process, illustrated in~\cref{fig:multi_view_graph}.
\begin{equation*}
    \zb \sim p_\zb, \qquad\qquad \xb_k:=\fb_k(\zb_{S_k}) \qquad\qquad \forall k \in [K],
\end{equation*}

For more than two views, there is no unique content block but multiple sets of content variables, defined as the latents shared among a given set of views---one for each subset $\xb_V=\{\xb_k\}_{k\in V}$ indexed by $V\subseteq[K]$ with $|V|\geq 2$.
We denote the content variables for $V_i$ by $\zb_{C_i}=\{\zb_j\}_{j\in C_i}$ with $C_i=\cap_{k\in V_i} S_k$, and the content among all views by $\zb_C$.
As before, our main focus is to provably and efficiently recover and disentangle as much latent information as possible.

\newpage 
\begin{example}
\label{example:intuitive}
Consider the following example with $K=4$ views, $N=6$ latents, and dependencies among the $\zb_k$ shown as a graphical model in~\cref{fig:multi_view_graph}. 
\begin{equation}
\label{eq:example}
\begin{aligned}
     \xb_1 &= f_1(\zb_1, \zb_2, \zb_3, \zb_4, \zb_5), \\
     \xb_2 &= f_2(\zb_1, \zb_2, \zb_3, \zb_5, \zb_6), \\
     \xb_3 &= f_3(\zb_1, \zb_2, \zb_3, \zb_4, \zb_6), \\
     \xb_4 &= f_4(\zb_1, \zb_2, \zb_4, \zb_5, \zb_6) .
\end{aligned}
\end{equation}
With this, the content of $\xb_{V_1} = (\xb_1,\xb_2)$ is $\zb_{C_1}
= (\zb_1,\zb_2,\zb_3,\zb_5)$;  that of $\xb_{V_2} = (\xb_1,\xb_2,\xb_3)$ is $\zb_{C_2}
= (\zb_1,\zb_2,\zb_3)$; and that of all four views $\xb_{V_3} = (\xb_1,\xb_2,\xb_3,\xb_4)$  is $\zb_{C_3} = (\zb_1,\zb_2)$.
\end{example}

For this setting, we first establish block-identifiability of the content $\zb_{C_i}$ shared among \emph{any set of views} $\xb_{V_i}$ through contrastive learning. 
The downside of this approach is that to learn different content blocks, we need to train an \textit{exponential} number of encoders $\gb_k^{V}$ for the same modality~$k$, one for each subset of views $V\subseteq [K]$. 
For~\cref{example:intuitive}, we need different encodings of $\xb_1$ depending on whether we seek to recover $\zb_{C_1}$ from $\xb_{V_1}$ or $\zb_{C_2}$ from $\xb_{V_2}$.
Hence, we extend this result and show that all content blocks can still be identified when using a \textit{single encoder $\rb_{k}$ per modality}, combined with \textit{binary content selectors} $\phi_k^{V}$ for different subsets of views $V$.

\textbf{\Cref{example:intuitive} (continued).}
For $\rb_1(\xb_1) = \rb_1 \circ \fb_1(\zb_1, \zb_2, \zb_3, \zb_4, \zb_5)$, if $\rb_1$ inverts $\fb_1$ (up to element-wise reparametrisation), 
the following $\phi_1^{V_i}$ are optimal selectors of content blocks $\zb_{C_i}$:
\begin{equation*}
\textstyle
    \phi_1^{V_1} = [1, 1, 1, 0, 1]\,, \qquad \qquad
    \phi_1^{V_2} = [1, 1, 1, 0, 0]\,, \qquad \qquad
    \phi_1^{V_3} = [1,1,0,0,0]\,.
\end{equation*}
We also develop an \textit{identifiability algebra} in the form of a set of simple rules that can be used to determine to what extent individual latents or groups thereof can be identified and disentangled.
For example, if $C$ and $C'$ are two identified content blocks, then their intersection $C\cap C'$ can also be block-identified; under additional independence constraints, the complements $C\setminus C'$ and $C'\setminus C$ can also be recovered. 
Evaluating these rules can thus be done based on graphical criteria and only requires a qualitative description of the data generative process, such as~\cref{fig:multi_view_graph}.

We validate our claims experimentally, see for example the identifiability scores in the first, seventh, and last column of~\cref{fig:multi_view_results}, which demonstrate that $\zb_{C_1}$, $\zb_{C_2}$, and $\zb_{C_3}$ from~\cref{example:intuitive} are indeed perfectly recovered.
As our framework provides a unified perspective on some previous work on multi-view nonlinear ICA~\citep{gresele2019incomplete}, disentanglement~\citep{locatello2020weakly,ahuja2022weakly},  CRL~\citep{daunhawer2023identifiability}, and multi-task learning~\citep{fumero2023leveraging,lachapelle2022synergies}, we also compare with some of these prior methods and find that the respective performances are recovered in different special cases of our setup. 

Overall, this study highlights that access to multiple views allows us to  identify a more fine-grained representation under generally milder assumptions such as partial observability.

\section{Summary}
In this chapter, we have demonstrated that in a \textit{non-i.i.d.}~scenario with multiple non-independent views of the same example we can learn useful representations, even in the presence of dependences among latents. 
Unlike in the previous~\cref{chap:IMA}, this did not require constraints on the mixing function or latent distribution.
Instead, the main theme has been to exploit \textit{invariance} as a learning signal by leveraging and enforcing the assumption that the values of certain latent variables are fixed and perfectly shared across views.
This contrasts with prior works which have usually leveraged change, rather than invariance for identifiability.
From a causal perspective, the considered multi-view setting can be understood as \textit{counterfactual} in nature, as only certain parts of the generative process are modified, \textit{all else being equal}. 
Our focus has mainly been on data augmentation as one of the few settings where such perfect sharing of some features across views plausibly arises, but the  multi-modal, partially observed setting constitutes an alternative. Depending on the timescale, it has also been argued that pre- and post-intervention views or temporally successive observations may \textit{approximately} satisfy the invariance assumption. 
Still, counterfactual data is, strictly speaking, impossible to obtain by definition. 
In the following chapter, we therefore consider a less restrictive data requirement by descending one rung in the ladder of causation and learning from \textit{interventional} data in the form of multiple experimental datasets or environments.

%% file: Chapter8/chapter8.tex
 \graphicspath{{Chapter8/Figs/}}
\chapter[Multi-Environment Causal Representation Learning]{Multi-Environment Causal Representation Learning}  %
\label{chap:CRL}
\looseness-1 
In this chapter, we consider a different violation of the i.i.d.\ assumption by learning from \textit{non-identically distributed} observations in the form of multiple datasets from different domains, or environments.
Unlike the multi-view data from~\cref{chap:SSL_content_style} consisting of non-independent counterfactual pairs of observations,  the multi-environment data we consider in this chapter is not paired and interventional in nature, with different environment-specific distributions arising through interventions in a shared underlying causal model.
Whereas previous chapters considered special cases of CRL by studying the recovery of independent latents~(\cref{chap:IMA}) or a subset of invariant root nodes~(\cref{chap:SSL_content_style}), here we focus on the full CRL task of identifying all causal latents and the causal graph in a general nonparametric setting.
We then discuss different special cases of this setting, as well as connections to supervised CRL and domain generalisation. 

The main content of this chapter has been published in the following paper:
\begin{selfcitebox}
\href{https://arxiv.org/abs/2306.00542}{\ul{Nonparametric identifiability of causal representations from unknown interventions}}
\\
\textbf{Julius von K\"ugelgen}, Michel Besserve, Wendong Liang, Luigi Gresele, Armin Keki\'c,\\ Elias Bareinboim, David M Blei, Bernhard Sch\"olkopf
\\
\textit{Advances in Neural Information Processing Systems (NeurIPS)}, 2023
\end{selfcitebox}

\section{Introduction}
\begin{figure}[t]
\begin{subfigure}{0.625\textwidth}
\newcommand{\decodercolor}{Plum}
\newcommand{\xshift}{4.25em}
\newcommand{\yshift}{2.65em}
    \centering
    \begin{tikzpicture}
        \centering
        \node (u_1) [latent, yshift=\yshift] {$U_1$};
        \node (u_i) [latent] {$\dots$};
        \node (u_n) [latent, yshift=-\yshift] {$U_n$};
        \node (v_1) [latent, yshift=\yshift, xshift=\xshift] {$V_1$};
        \node (s_i) [latent, xshift=\xshift] {$\dots$};
        \node (s_n) [latent, yshift=-\yshift, xshift=\xshift] {$V_n$};
        \draw [-stealth, thick] (u_1) -- (v_1);
        \draw [-stealth, thick] (u_i) -- (s_i);
        \draw [-stealth, thick] (u_n) -- (s_n);
        \draw [-stealth, thick] (v_1) -- (s_i);
        \draw [-stealth, thick] (s_i) -- (s_n);
        \draw [-stealth, thick] (v_1) to [out=220,in=140] (s_n);
        \coordinate (A) at (1.75*\xshift,1.2*\yshift);
        \coordinate (B) at (2.5*\xshift,1.7*\yshift);
        \coordinate (C) at (2.5*\xshift,-1.7*\yshift);
        \coordinate (D) at (1.75*\xshift,-1.2*\yshift);
        \draw[fill=\decodercolor!20] (A) -- (B) -- (C) -- (D) -- cycle;
        \coordinate (input_1) at (1.75*\xshift,\yshift);
        \coordinate (input_i) at (1.75*\xshift,0);
        \coordinate (input_n) at (1.75*\xshift,-1*\yshift);
        \draw [-stealth, thick, color=\decodercolor] (v_1) -- (input_1);
        \draw [-stealth, thick, color=\decodercolor] (s_i) -- (input_i);
        \draw [-stealth, thick, color=\decodercolor] (s_n) -- (input_n);
        \node[const,xshift=2.125*\xshift] {$\fb$};
        \node (pic) [xshift=3.8*\xshift]{\includegraphics[width=9em]
        {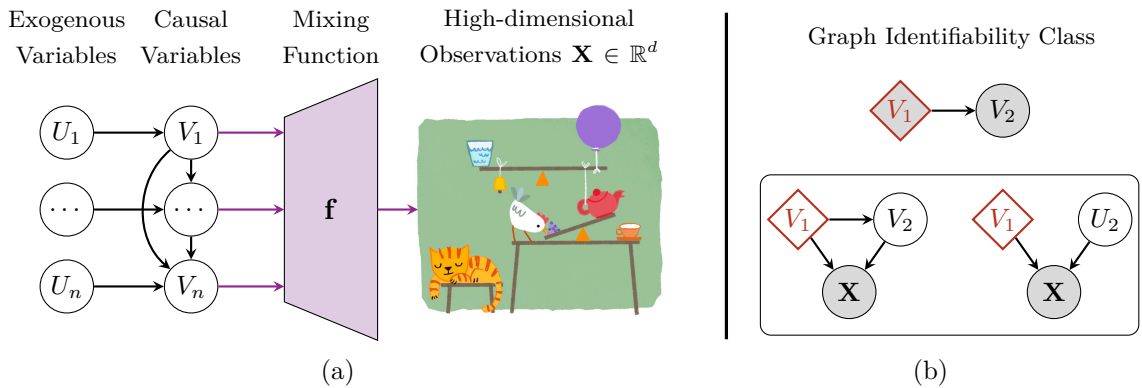}}; 
        \coordinate (output) at  (2.5*\xshift,0);
        \coordinate (data) at  (2.825*\xshift,0);
        \draw [-stealth, thick, color=\decodercolor] (output) -- (data);
        \node[const,xshift=3.8*\xshift, yshift=2.25*\yshift, text width=10em, align=center] {\footnotesize High-dimensional\\ {Observations~$\Xb\in\RR^d$}};
        \node[const,xshift=2.125*\xshift, yshift=2.25*\yshift, text width=5em, align=center] {\textcolor{black}{\footnotesize Mixing\\ Function}};
        \node[const, yshift=2.25*\yshift, text width=5em, align=center] {\textcolor{black}{\footnotesize Exogenous Variables}};
        \node[const,xshift=1*\xshift, yshift=2.25*\yshift, text width=5em,align=center] {\textcolor{black}{\footnotesize Causal\\ Variables}};
    \end{tikzpicture}
    \caption{}
    \label{fig:full_crl_setup_left}
\end{subfigure}%
\begin{subfigure}{0.365\textwidth}
\newcommand{\xshift}{3.5em}
\newcommand{\yshift}{2.5em}
\newcommand{\ivcolor}{BrickRed}
\centering
    \begin{tikzpicture}
    \centering
        \coordinate (bottom) at  (-1.7*\xshift,-3.15*\yshift);
        \coordinate (top) at  (-1.7*\xshift,1.35*\yshift);
        \draw[very thick] (bottom) -- (top);
        \node (graph) [const, xshift=0.5*\xshift, yshift=\yshift] {\footnotesize Graph Identifiability Class};
        \node (V_1) [det, draw=\ivcolor,thick, fill=gray!30] {\textcolor{\ivcolor}{$V_1$}};
        \node (V_2) [latent, fill=gray!30, xshift=\xshift] {$V_2$};
        \draw [-stealth, thick] (V_1) -- (V_2);

        \node (W_1) [det, yshift=-1.5*\yshift, xshift=-1*\xshift,draw=\ivcolor,thick] {\textcolor{\ivcolor}{$V_1$}};
        \node (W_2) [latent, yshift=-1.5*\yshift] {$V_2$};
        \draw [-stealth, thick] (W_1) -- (W_2);
        \node (X) [latent, fill=gray!30, xshift=-0.5*\xshift, yshift=-2.5*\yshift] {$\Xb$};
        \draw [-stealth, thick] (W_1) -- (X);
        \draw [-stealth, thick] (W_2) -- (X);

        \node (Y_1) [det, yshift=-1.5*\yshift,xshift=1*\xshift,draw=\ivcolor,thick] {\textcolor{\ivcolor}{$V_1$}};
        \node (Y_2) [latent, xshift=2*\xshift, yshift=-1.5*\yshift] {$U_2$};
        \node (Z) [latent, fill=gray!30, xshift=1.5*\xshift, yshift=-2.5*\yshift] {$\Xb$};
        \draw [-stealth, thick] (Y_1) -- (Z);
        \draw [-stealth, thick] (Y_2) -- (Z);
        \plate[inner sep=0.2em, yshift=0.25em] {plate2}{(W_1) (W_2) (X) (Y_1) (Y_2) (Z)}{};
    \end{tikzpicture}
    \caption{}
    \label{fig:full_crl_setup_right}
\end{subfigure}
\caption[CRL Data Generating Process and Comparison to Structure Learning]{\looseness-1 %
(a) \textbf{Data-Generating Process for Causal Representation Learning.} The observations~$\Xb$ are generated by applying a nonlinear mixing function $\fb$ to a set of causal latent variables $\Vb=\{V_1, ..., V_n\}$, which are in turn generated by a structural causal model (SCM) with independent exogenous
variables~$\Ub=\{U_1, ..., U_n\}$.
Illustration by Ana Mart\'in Larra\~naga.
(b) \textbf{Comparison to Structure Learning.} The causal direction between two unconfounded \textit{observed} variables (top) is uniquely identified from a single intervention~\citep{eberhardt2006n}; for causal representation learning (bottom), this is not the case as the nonlinear mixing introduces additional ambiguity due to spurious representations. 
Shaded nodes are observed, white ones unoberved, and interventions highlighted as red diamonds.
}
\label{fig:full_crl_setup}
\end{figure}

In CRL, we need not only identify the latents, but also the causal graph encoding their relations.
Even in the fully observed case, this task of causal discovery, or structure learning, is very challenging:  the graph can only be recovered up to Markov equivalence~\citep{spirtes2001causation} based only on (observational) i.i.d.\ data~\citep{squires2022causal}, meaning that the direction of some edges cannot be determined.
For CRL, the task gets strictly harder.
For instance, if the causal latents~$V_i$ in~\cref{fig:full_crl_setup_left} form a valid representation, then replacing them by the \textit{independent} exogeneous~$U_i$ might be considered an equally valid alternative.

\looseness-1 
\paragraph{What sets \textit{causal} representations apart?}
A crucial  feature of causal variables is that they are the ones on which \textit{interventions} are defined and whose relations we are interested in~\citep{Pearl2009}.
Causal discovery and CRL thus often rely on non-i.i.d.\ data linked to interventions on the underlying causal variables~\citep{kocaoglu2019characterization,jaber2020causal}.
Unless all variables are subject to intervention, however, some fundamental differences between the fully observed and representation learning settings in the level of ambiguity in the graph remain~\citep{squires2023linear}, as illustrated in~\cref{fig:full_crl_setup_right}.
In a sense, the non-identifiabilities of representation and structure learning combine, and both need to be addressed in conjunction.

\looseness-1 
\paragraph{Problem Setting.} %
We study the general \textit{nonparametric} CRL problem~(\cref{subsec:data_generating_process}) in which both the causal mechanisms and the mixing function are completely unconstrained. %
Our goal~(\cref{subsec:learning_target}) 
is to \text{identify} the latent causal variables up to element-wise nonlinear rescaling and their graph up to isomorphism~(\cref{def:identifiability}).
As motivated above, doing so without further supervision requires access to interventional data.
To this end, we consider learning from heterogeneous data from multiple related domains, or \textit{environments}, that arise from 
interventions in a shared underlying causal model~(\cref{subsec:multi_environment_data}).

\paragraph{Contributions.} 
\looseness-1 
Our main contributions are theoretical in nature~(\cref{sec:theory}). 
First, we establish the minimality of the targeted equivalence class~(\cref{prop:minimality_of_CRL}) in the sense that its ambiguities cannot be resolved from interventional data. 
We then present our main identifiability results. 
For the case of two latent causal variables, we show that an observational environment and one for each perfect intervention on either variable suffice~(\cref{thm:bivariate})---provided that the intervened and unintervened mechanisms are not ``fine-tuned'' to each other, which we formalize in the form of a \textit{genericity condition}~\eqref{eq:genericity_condition}.
For any number of latent causal variables, we prove that access to pairs of environments corresponding to two distinct perfect interventions on each node guarantees identifiability~(\cref{thm:general}).
We then question how to use or interpret causal representations~(\cref{sec:interpreting}), and show that certain quantities, such as the strengths of causal influences among variables, are preserved by all equivalent solutions~(\cref{prop:influence}). 
We sketch possible learning objectives~(\cref{sec:objectives}), and empirically investigate training different generative models~(\cref{sec:experiments}), finding that only those based on the correct causal structure attain the best fit and identify the ground truth.
We then discuss limitations and extensions in~\cref{sec:discussion}.

\section{Related Work}
\label{sec:related_work} 
Prior work on causal representation learning with general nonlinear relationships (both among latents and between latents and observations) and without an explicit task or label typically relies on some form of {\em weak supervision}. 
One example of weak supervision is ``multi-view'' data consisting of tuples of related observations, as discussed in~\cref{chap:SSL_content_style}.
For example, \Citet{brehmer2022weakly} use
counterfactual pre- and post-intervention views and show that the latent SCM can be identified
given all single-node perfect stochastic interventions. 
Another type of weak-supervision is 
temporal structure%
~\citep{ahuja2021properties}, possibly combined with nonstationarity~\citep{yao2021learning,yao2022temporally}, interventions on known targets~\citep{lippe2022citris,lippe2022icitris}, or observed actions inducing sparse mechanism shifts~\citep{scholkopf2021toward,lachapelle2022disentanglement,lachapelle2022partial}.
Other works use more explicit supervision in the form of annotations of the ground truth causal variables or a known causal graph~\citep{shen2022weakly,yang2021causalvae,Liang2023cca}.

\looseness-1 
A different line of work instead approaches causal representation learning from the perspective of causal discovery in the presence of latent variables~\citep{spirtes2001causation}. 
As discussed in~\cref{chap:IMA}, doing so from \textit{purely observational i.i.d.\ data} requires additional constraints on the generative process, such as restrictions on the graph structure or particular parametric and distributional assumptions, and typically leverages the identifiability of linear ICA~\citep{comon1994independent,lewicki2000learning,eriksson2004identifiability}. 
For \textit{linear, non-Gaussian} models, \citet{silva2006learning} show that the causal DAG can be recovered up to Markov equivalence if all observed variables are ``pure'' in that they have a unique latent causal parent. 
\citet{cai2019triad} and \citet{xie2020generalized,xie2022identification} extend this result to identify the full graph given two pure observed children per latent, and \citet{adams2021identification} provide sufficient and necessary graphical conditions for full identification.
For \textit{discrete} causal variables, \citet{kivva2021learning} introduce a similar ``no twins'' condition to reduce the task of learning the number and cardinality of latents and the Markov-equivalence class of the graph to mixture identifiability.

\looseness-1 Most closely related to the work we present here are recent identifiability studies which also leverage multiple environments arising from single node interventions~\citep{squires2023linear,ahuja2022interventional,varici2023score,buchholz2023learning}, thus mirroring in different interventional setups the result of~\citet{brehmer2022weakly} based on counterfactual, multi-view data, which is harder to obtain.
\citet{squires2023linear} provide results 
for linear causal models and linear mixing;
\citet{ahuja2022interventional} consider nonlinear causal models and polynomial mixings, subject to additional constraints on the latent support~\citep{wang2021desiderata}; and~\citet{varici2023score} employ a score-based approach for nonlinear causal models and linear mixing.
\citet{buchholz2023learning} extend the results of~\citet{squires2023linear} to general nonlinear mixings and linear Gaussian causal models.
\Citet{liu2022identifying} leverage recent advances in nonlinear ICA~\citep{hyvarinen2019nonlinear,khemakhem2020variational}
to identify a linear Gaussian causal model with context-dependent weights and nonlinear mixing from sufficiently diverse environments. 

\looseness-1 Another line of work has investigated the relationship between causal models at different levels of coarse-graining or abstraction~\citep{anand2022effect,eberhardt2016green,chalupka2015visual,chalupka2016multi,chalupka2017causal,rubenstein2017causal,beckers2019abstracting,beckers2020approximate,zennaro2022abstraction,zennaro2023jointly}.

A structured overview of and comparison with some existing identifiability results for CRL is provided in~\cref{tab:related_work}.

\begin{table}[tbp]
\newcommand{\obscol}{green!6}
\newcommand{\ivcol}{yellow!6}
\newcommand{\cfcol}{red!6}
\caption[Comparison of Existing Identifiability Results for CRL]{\looseness-1  \textbf{Comparison of Existing Identifiability Results for CRL.} 
All of the listed works assume invertibility (or injectivity) of the mixing function, as well as causal sufficiency (Markovianity) for the causal latent variables.
Most or all of the listed results require additional technical assumptions, and may provide additional results, which we omit for sake of readability; see the references for more details.}
\vspace{1em}
\label{tab:related_work}
\centering
\renewcommand{\arraystretch}{1.5}%
\resizebox{\textwidth}{!}{
\begin{tabularx}{21.5cm}{m{3.4cm} m{2.5cm} m{3.85cm} m{4.5cm} m{5.15cm}}
     \toprule
     \textbf{Work} & \textbf{Layer} & \textbf{Causal Model} & \textbf{Mixing Function} & \textbf{Main Identifiability Result} 
     \\     \midrule 
     \rowcolor{\obscol}
     \citet{cai2019triad}, \citet{xie2020generalized,xie2022identification} & observational & linear, non-Gaussian & linear with non-Gaussian noise s.t.\ each $V_i$ has 2 pure (obs.\ or unobs.) children & number of latents + $G$
     \\\midrule
     \rowcolor{\obscol}
     \citet{kivva2021learning} & observational & discrete, nonparametric & indentifiable mixture model s.t.\ obs.\ children of $V_i$ $\not\subseteq$ obs.\ children of $V_j$ & number, cardinality \& dist.\ of discrete latents + $G$ up to Markov equivalence 
     \\ \midrule
     \rowcolor{\obscol}
     \citet[][Thm.~4]{ahuja2022interventional} & observational & nonlinear w.\ independent support & finite-degree polynomial &  $\Vb$ up to permutation, shift, \& linear scaling
     \\ \midrule
     \rowcolor{\ivcol}
     \citet[][Thms.~1 \& 2]{squires2023linear} & interventional & linear & linear & $G$ and $\Vb$ up to partial-order preserving permutations from obs.\ dist.\ \& all single-node \textit{perfect} interventions
     \\\midrule
      \rowcolor{\ivcol}
     \citet[][Thm.~1]{squires2023linear} & interventional & linear & linear &  $G$ up to transitive closure from obs.\ dist.\ \& all single-node \textit{imperfect} interventions
     \\\midrule
      \rowcolor{\ivcol}
     \citet[][Thm.~16]{varici2023score} & interventional & nonparametric & linear & $G$ and $\Vb$ up to partial-order preserving permutations from obs.\ dist.\ \& all single-node \textit{perfect} interventions  
     \\\midrule
       \rowcolor{\ivcol}
    \citet[][Thm.~2]{ahuja2022interventional} & interventional & nonparametric & finite-degree polynomial &  $\Vb$ up to permutation, shift, and linear scaling from all single-node \textit{perfect} \textit{hard} interventions
      \\\midrule
      \rowcolor{\ivcol}
      \citet{buchholz2023learning} & interventional & linear Gaussian & nonparametric & $G$ and $\Vb$ up to permutation from obs.\ dist.\ \& all single-node \textit{perfect} interventions
      \\\midrule
     \rowcolor{\ivcol}
   \textbf{This Chapter} (\cref{thm:bivariate}) & interventional & nonparametric & nonparametric &  for $n=2$: $G$ and $\Vb$ up to $\sim_\textsc{crl}$ from all single-node \textit{perfect} interventions, subject to genericity~\eqref{eq:genericity_condition}
       \\\midrule
     \rowcolor{\ivcol}
     \textbf{This Chapter} (\cref{thm:general}) & interventional & nonparametric & nonparametric &  $G$ and $\Vb$ up to $\sim_\textsc{crl}$ from two distinct, paired single-node \textit{perfect} interventions  per node
      \\ \midrule
     \rowcolor{\cfcol}
     {\cref{chap:SSL_content_style} \newline \citep{von2021self}} & counterfactual & nonparametric & nonparametric & block of non-descendants $\Vb_{\nd(\Ical)}$ up to invertible function from fat-hand \textit{imperfect} interventions on $\Vb_\Ical$
     \\ \midrule
      \rowcolor{\cfcol}
     \citet{brehmer2022weakly} & counterfactual & nonparametric & nonparametric & $G$ and $\Vb$ up to $\sim_\textsc{crl}$ from all single-node \textit{perfect} interventions
      \\     \bottomrule
\end{tabularx}
}
\end{table}

\section{Nonparametric Causal Representation Learning}
\label{sec:problem_setting}
\looseness-1 In this section, we describe the considered problem setting  and state our main assumptions. First, we specify the assumed data generating process~(\cref{subsec:data_generating_process}) and learning task in the form of a target identifiability class~(\cref{subsec:learning_target}).
We then demonstrate the hardness of our task from i.i.d.\ data or imperfect interventions
and use this to motivate a multi-environment approach with perfect interventions~(\cref{subsec:multi_environment_data}).

\subsection{Data Generating Process}
\label{subsec:data_generating_process}
\looseness-1 The assumed data generating process consists of an acyclic latent  SCM~(\cref{def:SCM}) 
and a mixing function, as illustrated in~\cref{fig:full_crl_setup_left}.
\looseness-1 We place the following additional assumption on the distribution $P_\Vb$ induced by the SCM.
\begin{assumption}[Faithfulness]
\label{ass:faithfulness}
    The \textit{only} (conditional) independence relations satisfied by $P_\Vb$ are those implied by $\{V_i \independent \Vb_{\nd(i)}~|~\Vb_{\pa(i)}\}_{i\in[n]}$, where $\Vb_{\nd(i)}\subseteq\Vb\setminus \{V_i\}$  denotes the non-descendants of $V_i$ in $G$.
\end{assumption}
\Cref{ass:faithfulness} ensures a one-to-one correspondence between (conditional) independence in $P_\Vb$ and graphical separation in $G$ and
is a standard assumption in causal discovery~\citep{spirtes2001causation}. Faithfulness rules out  cancellations of influences along different paths, which  occurs with probability zero for random path-coefficients~\citep{uhler2013geometry}. It can thus also be viewed as a minimality or genericity assumption.

In contrast to classical causal inference, we assume that both the exogenous variables $\Ub$ and the endogenous causal variables $\Vb$ are unobserved. Instead, we will only have access to $d$-dimensional nonlinear mixtures $\Xb$ of $\Vb$. 
We therefore make the following additional assumption.%

\begin{assumption}[Known\texorpdfstring{~$n$}{n}]
\label{ass:known_n}
    The number $n$ of latent causal variables is known.
\end{assumption}
Next, we specify the relationship between the unobserved causal variables $\Vb$ and the observed~$\Xb$.

\begin{definition}[Mixing function]
\label{def:mixing}
  The observations $\Xb$ are deterministically
  generated from $\Vb$ by applying 
a mixing function $\fb:\RR^n \to \mathbb{R}^d$ to $\Vb$, that is $\Xb:=\fb(\Vb)$.
\end{definition}%

\looseness-1 For representation learning scenarios, we are particularly interested in the case $n\ll d$.
To allow for recovery of $\Vb$ from $\Xb$,
we assume that $\fb$ is invertible, which is a standard assumption for identifiability.
\begin{assumption}[Diffeomorphic mixing]
\label{ass:diffeomorphism}
    $\fb$ is a diffeomorphism
    onto its image~$\mathrm{Im}(\fb)=\Xcal\subseteq\RR^{d}$.
\end{assumption}

\subsection{Learning Target: The CRL Identifiability Class}
\label{subsec:learning_target}
\looseness-1 Our goal is to infer the underlying latent causal variables $\Vb=\fb^{-1}(\Xb)$ and their causal relations.
We therefore consider the true unmixing function $\fb^{-1}$ \textit{and} the causal graph $G$ our joint learning target:
$\fb^{-1}$ informs us how to map observations $\Xb$ to causal variables $\Vb$, and $G$ tells us how to factorise the implied joint $p(\vb)$ into the causal mechanisms~$p_i(v_i~|~\vb_{\pa(i)})$ from~\eqref{eq:causal_Markov_factorisation}.
Given only observations of $\Xb$, this is a challenging task since neither $\Vb$ nor $G$ are directly observed or known a priori.
%
%
%
%
%
%

%
%
%
%
%
%

%

%
%
%
%

%
%
%
%
%
%

%
%
%
%

When is a candidate solution $(\hb,G')$ that satisfies a given learning objective (such as maximizing the likelihood, possibly subject to additional constraints)
guaranteed to match the ground truth~$(\fb^{-1},G)$?
This is the subject of identifiability studies
and the main focus of our work.

\looseness-1 For the assumed data generating process~(\cref{subsec:data_generating_process}), the order of the causal variables is arbitrary, since $\Vb$ is unobserved. 
We can therefore assume without loss of generality (w.l.o.g.) that 
the $V_i$'s are
partially 
ordered w.r.t.\ $G$, that is, $V_i\to V_j \implies i<j$. If they were not in such order to begin with, we could apply an appropriate permutation $\tilde\pi$ to $\Vb$ and incorporate the inverse permutation into the unknown mixing $\fb$ without affecting $\Xb=\fb(\Vb)=(\fb\circ \tilde\pi^{-1})(\tilde\pi(\Vb))$~\citep{squires2023linear}. 
Learning $G$ thus reduces to inferring whether the edges $\{V_1, \dots, V_{i-1}\}\to V_i$ exist for $i=2,\dots,n$. The only remaining permutation ambiguity  arises from permutations $\pi$ that preserve the partial order:
for example, if $G$ is given by $V_1\to V_3 \leftarrow V_2$,  the order of $V_1$ and $V_2$ cannot be uniquely determined without further assumptions.
Moreover, the scaling of the causal variables is also arbitrary: any invertible element-wise transformation can be undone as part of $\fb$.
We therefore define the desired identifiability class through the following equivalence relation.\footnote{\looseness-1$\sim_\textsc{crl}$ satisfies symmetry and transitivity because permutations and element-wise functions commute.}

\begin{restatable}[$\sim_\textsc{crl}$-identifiability]{definition}{defcrl}
\label{def:identifiability}
Let $\Hcal$ be a space of unmixing functions  $\hb:\Xcal\to\RR^n$ and let $\Gcal$ be the space of DAGs over $n$ vertices.
Let $\sim_\textsc{crl}$ be the equivalence relation on $\Hcal\times\Gcal$ defined as
\begin{equation}
\label{eq:sim_CRL}
    (\hb_1,G_1)\sim_\textsc{crl}(\hb_2,G_2) \quad  \iff \quad 
    (\hb_2, G_2) = (\Pb_{\pi^{-1}} \circ \phi \circ \hb_1, \pi(G_1))
\end{equation}
\looseness-1 for an element-wise diffeomorphism $\phi(\vb)=(\phi_1(v_1), \dots, \phi_n(v_n))$ of $\RR^n$ and  a permutation~$\pi$ of~$[n]$ such that $\pi:G_1\mapsto G_2$ is a graph isomorphism and $\Pb_\pi$ the corresponding permutation matrix.
\end{restatable}
\begin{remark}
\label{remark:ICA_identifiability}
    When $G$ has no edges, any permutation is admissible and $\sim_\textsc{crl}$ reduces to the standard notion of identifiability up to permutation and element-wise reparametrisation of nonlinear ICA~\citep{hyvarinen2023nonlinear}.
\end{remark}
\looseness-1 The ground truth $(\fb^{-1}, G)$ is identified up to $\sim_\textsc{crl}$ by a given learning objective if any optimal solution $(\hb,G')$ satisfies $(\hb,G')\sim_\textsc{crl} (\fb^{-1}, G)$. We seek to discover suitable conditions that ensure this. 

\subsection{Multi-Environment Setup
}
\label{subsec:multi_environment_data}
\looseness-1 Given only a single dataset of i.i.d.\ observations from $P_\Xb$, there is no hope for $\sim_\textsc{crl}$-identifiability.
Even for observed $\Vb$ (i.e., with $n=d$ and known $\fb=\mathrm{id}$), $G$ can only be identified up to Markov equivalence~\citep{spirtes2001causation}.
With unknown mixing $\fb$, the degree of observational non-identifiability gets even worse: for example, by using the reduced form~\eqref{eq:reduced_form} of the SCM we can express $\Xb$ in terms of the latent exogenous variables $\Ub$ via $\Xb=\fb(\Vb)=\fb\circ \fb_\textsc{rf}(\Ub)$. 
This gives rise to a ``spurious ICA solution'' $(\fb_\textsc{rf}^{-1}\circ \fb^{-1}, G_\textsc{ica})$ where $G_\textsc{ica}$ denotes the empty graph with independent components. 
Due to the non-identifiability of nonlinear ICA~(\cref{sec:background_nonlinear_ICA_nonidentifiable}), however, we cannot even learn the composition $\fb\circ \fb_\textsc{rf}$, let alone separate it into its constituents $\fb$ and $\fb_\textsc{rf}$ to isolate the intermediate causal variables $\Vb$.

\looseness-1 Motivated by these challenges to identifiability from i.i.d.\ data, we instead consider learning from \emph{multiple environments} $e\in \Ecal$. That is, we assume access to heterogenous data from multiple distinct distributions $P^e_\Xb$.
Environments can arise, for example, from different experimental settings or correspond to broader contexts such as climate or time.
Previous work has shown that this
setting can, in principle, provide useful causal learning signals~\citep{Tian2001,kocaoglu2019characterization,kocaoglu2017experimental,scholkopf2012causal,rothenhausler2021anchor,brouillard2020differentiable,eaton2007exact,mooij2020joint,peters2016causal,heinze2018invariant,rojas2018invariant,arjovsky2019invariant,krueger21rex,eastwood2022probable,perry2022causal,jaber2020causal,huang2020causal,squires2023linear,ahuja2022interventional,varici2023score,liu2022identifying}.
However, multi-environment data is not necessarily useful if the corresponding distributions $P^e_\Xb$ are allowed to differ in arbitrary ways.
What makes this setting interesting is the assumption that certain parts of the causal generative process are shared across environments.

\looseness-1 Here, we assume that all environments share the same invariant mixing function and underlying SCM, and that any distribution shifts arise from interventions on some of the causal mechanisms.
We summarise this as follows.%
\begin{assumption}[Shared mixing and mechanisms]
\label{ass:shared_mechs}
Each environment $e$ shares the same mixing~$\fb$, 
\begin{equation*}
    P^e_\Xb=\fb_*(P^e_\Vb)
\end{equation*}
and each $P^e_\Vb$ results from the same SCM through an intervention on a subset of mechanisms $\Ical^e\subseteq[n]$:
\begin{equation*}
\label{eq:shared_mechanisms}
    p^e(\vb)=p^e(v_1, ..., v_n) = 
    \prod_{i\in \Ical^e} p^e_i\left(v_i~|~\vb_{\pa(i)}\right)
    \,
    \prod_{j\in[n]\setminus \Ical^e}p_j\left(v_j~|~\vb_{\pa(j)}\right)\,.
\end{equation*}
Importantly, the intervention targets $\Ical^e$ are not assumed to be known.
\end{assumption}

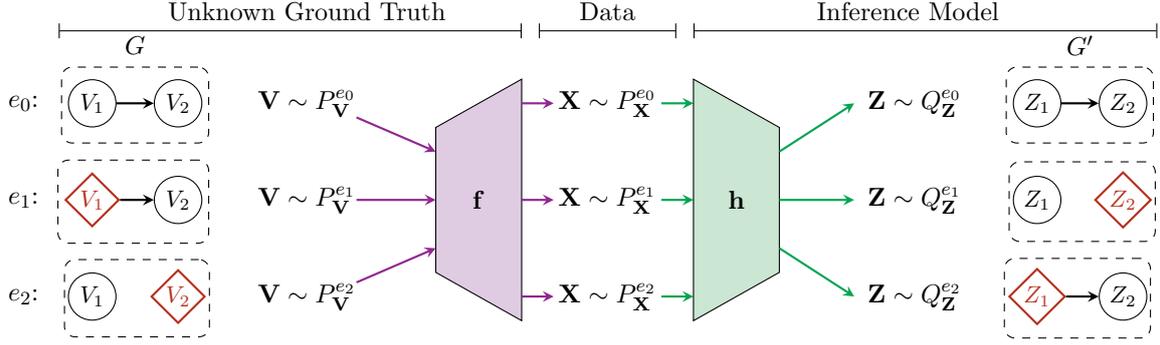
\begin{figure}[t]
    \newcommand{\xshift}{3.3em}
    \newcommand{\yshift}{1.87em}
    \newcommand{\zshift}{0.75em}
    \newcommand{\decodercolor}{Plum}
    \newcommand{\ivcolor}{BrickRed}
    \newcommand{\encodercolor}{Green}
    \centering
    \begin{tikzpicture}[scale=0.89, every node/.style={transform shape}]
        \centering
        \node (G) [const, xshift=.5*\xshift,yshift=1.2*\yshift]{$G$};
        \node (G') [const, xshift=11.5*\xshift,yshift=1.2*\yshift]{$G'$};
        \coordinate (GT_left) at (-0.39*\xshift,1.5*\yshift);
        \coordinate (GT_right) at (5*\xshift,1.5*\yshift);
        \node (GT) [const, xshift=2.5*\xshift,yshift=1.9*\yshift] {Unknown Ground Truth
        };
        \draw[|-|] (GT_left) -- (GT_right);
        
        \coordinate (Data_left) at (5.2*\xshift,1.5*\yshift);
        \coordinate (Data_right) at (6.8*\xshift,1.5*\yshift);
        \node (Data) [const, xshift=6*\xshift,yshift=1.9*\yshift] {Data};
        \draw[|-|] (Data_left) -- (Data_right);
        
        \coordinate (Model_left) at (7*\xshift,1.5*\yshift);
        \coordinate (Model_right) at (12.4*\xshift,1.5*\yshift);
        \node (Model) [const, xshift=9.5*\xshift,yshift=1.9*\yshift] {Inference Model
        };
        \draw[|-|] (Model_left) -- (Model_right);

        \node (V1) [latent] {$V_1$};
        \node (V2) [latent, xshift=\xshift] {$V_2$};
        \draw [-stealth, thick] (V1) -- (V2);
        \plate[inner sep=0.2em, yshift=0.2em, dashed] {plate1}{(V1) (V2) }{};

        \node (Y1) [det,yshift=-2*\yshift,draw=\ivcolor,thick] {\textcolor{\ivcolor}{$V_1$}};
        \node (Y2) [latent, xshift=\xshift, yshift=-2*\yshift] {$V_2$};
        \draw [-stealth, thick] (Y1) -- (Y2);
        \plate[inner sep=0.2em, yshift=0.2em, dashed] {plate2}{(Y1) (Y2)}{};

        \node (W1) [latent, yshift=-4*\yshift] {$V_1$};
        \node (W2) [det, xshift=\xshift, yshift=-4*\yshift, draw=\ivcolor,thick] {\textcolor{\ivcolor}{$V_2$}};
        \plate[inner sep=0.1em, yshift=0.2em, dashed] {plate3}{(W1) (W2)}{};

        \node (e1) [const, xshift=-.8*\xshift]{$e_0$:};
        \node (e2) [const, xshift=-.8*\xshift, yshift=-2*\yshift]{$e_1$:};
        \node (e3) [const, xshift=-.8*\xshift, yshift=-4*\yshift]{$e_2$:};
        \node (p1) [const, xshift=2.5*\xshift]{$\Vb\sim P^{e_0}_\Vb$};
        \node (p2) [const, xshift=2.5*\xshift, yshift=-2*\yshift]{$\Vb\sim P^{e_1}_\Vb$};
        \node (p3) [const, xshift=2.5*\xshift, yshift=-4*\yshift]{$\Vb\sim P^{e_2}_\Vb$};

        \coordinate (A) at (4*\xshift,-0.5*\yshift);
        \coordinate (B) at (5*\xshift,0.5*\yshift);
        \coordinate (C) at (5*\xshift,-4.5*\yshift);
        \coordinate (D) at (4*\xshift,-3.5*\yshift);
        \draw[fill=\decodercolor!20] (A) -- (B) -- (C) -- (D) -- cycle;
        \coordinate (input_1) at (4*\xshift,-1*\yshift);
        \coordinate (input_i) at (4*\xshift,-2*\yshift);
        \coordinate (input_n) at (4*\xshift,-3*\yshift);
        \draw [-stealth, thick, color=\decodercolor] (p1.south east) -- (input_1);
        \draw [-stealth, thick, color=\decodercolor] (p2.east) -- (input_i);
        \draw [-stealth, thick, color=\decodercolor] (p3.north east) -- (input_n);

        \node[const,xshift=4.5*\xshift, yshift=-2*\yshift] {$\fb$};
        \coordinate (output_1) at (5.*\xshift,-0*\yshift);
        \coordinate (output_i) at (5.*\xshift,-2*\yshift);
        \coordinate (output_n) at (5.*\xshift,-4*\yshift);
        \node (D_1) [const, xshift=6*\xshift, yshift=-0*\yshift] {\,$\Xb\sim P^{e_0}_\Xb$\,};
        \node (D_i) [const, xshift=6*\xshift, yshift=-2*\yshift] {\,$\Xb\sim P^{e_1}_\Xb$\,};
        \node (D_n) [const, xshift=6*\xshift, yshift=-4*\yshift] {\,$\Xb\sim P^{e_2}_\Xb$\,};
        \draw [-stealth, thick, color=\decodercolor] (output_1) -- (D_1);
        \draw [-stealth, thick, color=\decodercolor] (output_i) -- (D_i);
        \draw [-stealth, thick, color=\decodercolor] (output_n) -- (D_n);

        \coordinate (encinput_1) at (7.*\xshift,-0*\yshift);
        \coordinate (encinput_i) at (7.*\xshift,-2*\yshift);
        \coordinate (encinput_n) at (7.*\xshift,-4*\yshift);
        \draw [-stealth, thick, color=\encodercolor] (D_1) -- (encinput_1);
        \draw [-stealth, thick, color=\encodercolor] (D_i) -- (encinput_i);
        \draw [-stealth, thick, color=\encodercolor] (D_n) -- (encinput_n);
        
        \coordinate (A') at (8*\xshift,-0.5*\yshift);
        \coordinate (B') at (7*\xshift,0.5*\yshift);
        \coordinate (C') at (7*\xshift,-4.5*\yshift);
        \coordinate (D') at (8*\xshift,-3.5*\yshift);
        \draw[fill=\encodercolor!20] (A') -- (B') -- (C') -- (D') -- cycle;
        \node[const,xshift=7.5*\xshift, yshift=-2*\yshift] {$\hb$};
        \coordinate (encoutput_1) at (8*\xshift,-1*\yshift);
        \coordinate (encoutput_i) at (8*\xshift,-2*\yshift);
        \coordinate (encoutput_n) at (8*\xshift,-3*\yshift);
        \node (q1) [const, xshift=9.5*\xshift]{\, $\Zb\sim Q^{e_0}_\Zb$};
        \node (q2) [const, xshift=9.5*\xshift, yshift=-2*\yshift]{\, $\Zb\sim Q^{e_1}_\Zb$};
        \node (q3) [const, xshift=9.5*\xshift, yshift=-4*\yshift]{\, $\Zb\sim Q^{e_2}_\Zb$};
        \draw [-stealth, thick, color=\encodercolor] (encoutput_1) -- (q1.west);
        \draw [-stealth, thick, color=\encodercolor] (encoutput_i) -- (q2);
        \draw [-stealth, thick, color=\encodercolor] (encoutput_n) -- (q3.west);

        \node (Z1) [latent, xshift=11*\xshift] {$Z_1$};
        \node (Z2) [latent, xshift=12*\xshift] {$Z_2$};

        \draw [-stealth, thick] (Z1) -- (Z2);
        \plate[inner sep=0.2em, yshift=0.2em, dashed] {plate1}{(Z1) (Z2) }{};

        \node (S1) [latent, xshift=11*\xshift,yshift=-2*\yshift] {$Z_1$};
        \node (S2) [det, xshift=12*\xshift, yshift=-2*\yshift, draw=\ivcolor,thick] {\textcolor{\ivcolor}{$Z_2$}};
        \plate[inner sep=0.1em, yshift=0.2em, dashed] {plate3}{(S1) (S2)}{};

        \node (T1) [det,yshift=-4*\yshift,xshift=11*\xshift,draw=\ivcolor,thick] {\textcolor{\ivcolor}{$Z_1$}};
        \node (T2) [latent, xshift=12*\xshift,yshift=-4*\yshift] {$Z_2$};
        \draw [-stealth, thick] (T1) -- (T2);
        \plate[inner sep=0.1em, yshift=0.2em, dashed] {plate2}{(T1) (T2)}{};

    \end{tikzpicture}
    \caption[Multi-Environment Setup]{\looseness-1 %
    \textbf{Multi-Environment Setup with Single-Node Perfect Interventions and Shared Mixing Function.}
    Illustration of the considered  multi-environment setup for $n=2$ causal variables $\Vb=\{V_1,V_2\}$ with graph~$G$ given by $V_1 \to V_2$, shared mixing function $\fb$, %
    and environments $\Ecal=\{e_0,e_1,e_2\}$, corresponding to the observational setting $(e_0)$ and perfect stochastic interventions on $V_1$ (in $e_1$) and $V_2$ (in $e_2$).
    The learnt unmixing function, or encoder, is denoted by $\hb$, and the  inferred latent representation by $\Zb=\hb(\Xb)$.
    The corresponding inferred latent distributions $Q^{e_i}_\Zb=\hb_*(P^{e_i}_\Xb)$ are Markovian w.r.t.\ the candidate graph $G'$ (here, equal to $G$).
    Since the intervention targets are not known, they may in principle differ in $Q^e_\Zb$ as shown here. However, as we show in~\cref{thm:bivariate}, such misalignment is only possible if a certain  genericity condition~\eqref{eq:genericity_condition} is violated.    }
    \label{fig:multi_env}
\end{figure}

\looseness-1 
It has been shown that imperfect interventions are generally insufficient for full identifiability~\citep{brehmer2022weakly}, even in the linear case~\citep{squires2023linear}.
This is intuitive: if arbitrary imperfect interventions were allowed, including ones which preserve $f_i(\Vb_{\pa(i)}, \cdot)$ and only replace $U_i$ with some new $\tilde U_i$, then 
the spurious ICA solution $(f_\textsc{rf}^{-1}\circ f^{-1}, G_\textsc{ica})$ should be indistinguishable from the ground truth.
In line with prior work~\citep{brehmer2022weakly,squires2023linear,ahuja2022interventional,varici2023score}, we therefore assume perfect interventions.

\begin{assumption}[Perfect interventions]
\label{ass:perfect_interventions}
For all $e\in \Ecal$ and $i\in\Ical^e$, we have
    \[p^{e}(v_i~|~\vb_{\pa(i)})=p^{e}(v_i)\,.\]
\end{assumption}

\looseness-1 We will assume single-node (``atomic'') interventions for our main results, as also required for existing results~\citep{brehmer2022weakly,squires2023linear,ahuja2022interventional,varici2023score}.

\looseness-1 As motivated in~\cref{subsec:learning_target}, given data from $\{P^e_\Xb\}_{e\in\Ecal}$ we consider candidate solutions of the form $(\hb,G')$ where $\hb$ is an unmixing function, or encoder, which maps observations $\Xb$ to the inferred latents $\Zb=\hb(\Xb)$, and $G'$ is a causal graph capturing the relations among the $Z_i$.
The corresponding distributions of the inferred latents are thus given by the push-forward
\begin{align*}
\label{eq:QeZ}
    Q_\Zb^e=\hb_*(P^e_\Xb)=(\hb\circ \fb)_*P^e_\Vb
\end{align*}
\looseness-1 The considered multi-environment setup with unknown single-node perfect interventions is illustrated in~\cref{fig:multi_env}.

\section{Identifiability Theory}
\label{sec:theory}
\looseness-1
\label{sec:minimality_of_equivalence_class}
We start by showing that identifiability up to $\sim_\textsc{crl}$ is, in fact, the best we can hope for when learning from interventional data, without any more direct forms of supervision. 
All proofs are provided in~\cref{app:proofs_CRL}.

\begin{restatable}[Minimality of $\sim_\textsc{crl}$]{proposition}{minimality}
\label{prop:minimality_of_CRL}
Let $(\hb,G')\sim_\textsc{crl}(\fb^{-1},G)$ 
with $\pi$ denoting the graph isomorphism mapping $G$ to $G'$ (i.e., a permutation that preserves the partial topological order of~$G$). 
Let $\Zb=\hb(\Xb)$ be the inferred representation with distribution $Q_\Zb=\hb_*(P_\Xb)$ Markov w.r.t.~$G'$ and associated density $q$.
Let $\Ical^e\subseteq[n]$ denote a set of intervention targets, and consider an intervention that changes $p_i(v_i~|~\vb_{\pa(i)})$ to some intervened mechanism $\tilde p_i(v_i~|~\vb_{\pa(i)})$ for all $i\in \Ical^e$, giving rise to the interventional distributions $P^e_\Vb$ and $P^e_\Xb=\fb_*(P^e_\Vb)$.
Then there exist appropriately chosen $\tilde q_{\pi(i)}(z_{\pi(i)}~|~\zb_{\pa(\pi(i),G')})$ for $i\in\Ical^e$ such that the resulting interventional distribution $Q^e_\Zb$ gives rise to the same observed distributions, that is, $P^e_\Xb=\hb^{-1}_*(Q^e_\Zb)$.
\end{restatable}

\label{sec:identifiability_results}
We now present our identifiability results. 
We first study the most fundamental bivariate case with two latent causal variables $V_1$ and $V_2$. This can loosely be seen as the CRL analogue of the widely studied cause-effect problem ($X{\to}Y$ or $Y{\to}X$?) in classical structure learning~\citep{mooij2016distinguishing,peters2017elements}.

\begin{restatable}[Bivariate identifiability up to $\sim_\textsc{crl}$ from one perfect stochastic intervention per node]{theorem}{bivariate}
\label{thm:bivariate}    
\looseness-1 
Suppose that we have access to multiple environments $\{P^e_\Xb\}_{e\in\Ecal}$ generated as described in~\cref{sec:problem_setting} under~\cref{ass:faithfulness,ass:diffeomorphism,ass:shared_mechs,ass:perfect_interventions} with $n=2$.
Let $(\hb,G')$ be any candidate solution such that the inferred latent distributions $Q^e_\Zb=\hb_*(P^e_\Xb)$ of $\Zb=\hb(\Xb)$ and the inferred mixing function~$\hb^{-1}$ satisfy the above assumptions w.r.t.\ the candidate causal graph $G'$.
Assume additionally that
\begin{enumerate}[leftmargin=2.7em]
    \item[(A1)] all densities $p^e$ and $q^e$ are continuously differentiable and fully supported on $\RR^n$;
    \item[(A2)] \looseness-1 we have access to a \emph{known observational environment} $e_0$ and one \emph{single node perfect intervention for each node}, with \emph{unknown targets}: there exist exactly $n+1$ environments $\Ecal=\{e_i\}_{i=0}^n$ such that $\Ical^{e_0}=\varnothing$ and for each $i\in[n]$ we have $\Ical^{e_{i}}=\{\pi(i)\}$
    for an unknown permutation $\pi$ of $[n]$;
    \item[(A3)] 
    for all $i\in[n]$, the intervened mechanisms $\tilde p_{i}(v_i)$ differ from the corresponding base mechanisms $p_i(v_i~|~\vb_{\pa(i)})$ everywhere, 
    in the sense that 
    \begin{equation}
        \label{eq:different_mechanisms}
        \forall \vb: \qquad \frac{\partial}{\partial v_i}\frac{\tilde p_i(v_i)}{p_i(v_i~|~\vb_{\pa(i)})}\neq 0\,;
    \end{equation}
    \item[(A4)] (\textbf{``genericity''}) the base and intervened mechanisms are not fine-tuned to each other, in the sense that there exists a continuous function $\varphi:\RR^+\to\RR$ for which
    \begin{equation}
    \label{eq:genericity_condition}
        \EE_{\vb\sim P^{e_0}_\Vb}\left[\varphi\left(\frac{\tilde p_2(v_2)}{p_2(v_2~|~v_1)}\right)\right]
        \neq 
        \EE_{\vb\sim P^{e_1}_\Vb}\left[\varphi\left(\frac{\tilde p_2(v_2)}{p_2(v_2~|~v_1)}\right)\right]
    \end{equation}
\end{enumerate}
Then the ground truth is identified in the sense of~\cref{def:identifiability}, that is, $(f^{-1},G)\sim_\textsc{crl}(h,G')$.
\end{restatable}
\begin{proof}[Proof sketch (full proof in~\cref{app:proof_bivariate}).]
Consider $V_1\to V_2$ (the proof for $V_1\not\to V_2$ is similar).
Let $\psi=\fb^{-1}\circ \hb^{-1}:\RR^n\to\RR^n$ such that $\Vb=\psi(\Zb)$. By~\cref{ass:diffeomorphism}, $\fb$, $\hb$, and thus also $\hb\circ \fb$  are diffeomorphisms. Hence, $\psi$ is well-defined and also diffeomorphic. 
By the change of variable formula, for all $\zb$
\begin{equation}
\label{eq:change_of_variable}
\textstyle
q^e(\zb)=p^e(\psi(\zb))\abs{\det \Jb_\psi(\zb)}
\end{equation}
where $(\Jb_\psi(\zb))_{ij}=\frac{\partial \psi_i}{\partial z_j}(\zb)$ denotes the Jacobian of $\psi$.
\looseness-1 We consider two separate cases, depending on whether the 
intervention targets in $q^{e_i}$ for $e_i\in\{e_1,e_2\}$ match those in $p^{e_i}$ (Case 1) or not (Case 2).

\textit{Case 1: Aligned Intervention Targets.}
\looseness-1
According to~\cref{ass:shared_mechs} and (A2), \eqref{eq:change_of_variable} applied to the known observational environment $e_0$ and the interventional environments $e_1,e_2$ leads to the following system of equations:
\begin{align}
\label{eq:qe0}
q_1(z_1)q_2(z_2~|~z_{\pa(2;G')})&=
p_1\left(\psi_1(\zb)\right)p_2\left(\psi_2(\zb)~|~\psi_1(\zb)\right) \abs{\det\Jb_{\psi}(\zb)}
\\
\label{eq:qe1}
\tilde q_1(z_1)q_2(z_2~|~z_{\pa(2;G')})&=
\tilde p_1\left(\psi_1(\zb)\right) p_2\left(\psi_2(\zb)~|~\psi_1(\zb)\right) \abs{\det\Jb_{\psi}(\zb)}
\\
\label{eq:qe2}
q_1(z_1)\tilde q_2(z_2)&= 
p_1\left(\psi_1(\zb)\right)\tilde p_2\left(\psi_2(\zb)\right) \abs{\det\Jb_{\psi}(\zb)}
\end{align}
where $z_{\pa(2;G')}$ denotes the parents of $z_2$ in $G'$.
Taking quotients of~\eqref{eq:qe1} and~\eqref{eq:qe0} yields
\begin{equation}
\label{eq:triangularJ}
    \frac{\tilde q_1}{q_1}(z_1)=\frac{\tilde p_{1}}{p_{1}}(\psi_1(\zb))
    \quad
    \overset{\frac{\partial}{\partial z_2}}{\implies}
    \quad 
    0=\left(\frac{\tilde p_{1}}{p_{1}}\right)'\left(\psi_1(\zb)\right)\frac{\partial \psi_1}{\partial z_2}(\zb)
    \quad
    \overset{(A3)}{\implies}
    \quad
    \frac{\partial \psi_1}{\partial z_2}(\zb)=0 \,.
\end{equation}
Thus $V_1=\psi_1(Z_1)$ and $q_1(z_1)=p_1(\psi_1(z_1))\abs{\frac{\partial \psi_1}{\partial z_1}(z_1)}$.
Substitution into~\eqref{eq:qe2} yields
\begin{align}
\label{eq:measure_preserving}
\tilde q_2(z_2)&= \tilde p_2\left(\psi_2(z_1,z_2)\right) \abs{\frac{\partial \psi_2}{\partial z_2}(z_1,z_2)}
\end{align}
\looseness-1 where we have used that, according to~\eqref{eq:triangularJ}, $\Jb_\psi$ is triangular.
According to~\eqref{eq:measure_preserving}, for all $z_1$, the mapping $z_2\mapsto \psi_2(z_1,z_2)$ is measure preserving for $\tilde q_2$ and $\tilde{p}_2$. By~\cref{lemma:brehmer}%
~\cite[\S~A.2, Lemma 2]{brehmer2022weakly},
it follows that $\psi_2$ must be constant in $z_1$.\footnote{This step is where the assumption of perfect interventions~(\cref{ass:perfect_interventions}) is leveraged: the conclusion would not hold for arbitrary imperfect interventions for which~\eqref{eq:measure_preserving} would involve $\tilde q_2(z_2~|~z_1)$ and $p_2\left(\psi_2(z_1,z_2)~|~\psi_1(z_1)\right)$.}
Hence, $\psi$ is an element-wise function.
Finally, $G=G'$ follows from faithfulness~(\cref{ass:faithfulness}), for otherwise $(V_1,V_2)=(\psi_1(Z_1), \psi_2(Z_2))$ would be independent.

\textit{Case 2: Misaligned Intervention Targets.}
If $G'\neq G$, a similar argument to Case 1 (with the roles of $z_1$ and $z_2$ interchanged) also yields a contradiction to faithfulness. This leaves $G=G'$. Writing down the system of equations similar to~\eqref{eq:qe0}--\eqref{eq:qe2}, and then taking ratios of $e_1$ and $e_2$ with $e_0$ yields
\begin{align}
    \label{eq:misaligned_ratios}
    \frac{\tilde q_1}{q_{1}}(z_1)
    =\frac{\tilde p_2\left(\psi_2(\zb)\right)}{p_{2}\left(\psi_2(\zb)~|~\psi_1(\zb)\right)}
    \qquad \mbox{and} \qquad 
    \frac{\tilde q_2(z_2)}{q_{2}(z_2~|~z_1)}
    =\frac{\tilde p_1}{p_1}\left(\psi_1(\zb)\right)\,.
\end{align}
These conditions highlight the misalignment in intervention targets (see~\cref{fig:multi_env}). Unlike in Case 1, they do not directly imply that some elements of $\Jb_\psi$ need to be zero, that is $\Zb$ can be arbitrarily mixed w.r.t.~$\Vb$.
However, \eqref{eq:misaligned_ratios} imposes constraints on the form of $\psi$ that, by exploiting the invariance of $q_1$ across $e_0$ and $e_1$ while $p_1$ changes to $\tilde p_1$, can ultimately be shown to only be satisfied if the two expectations in~\eqref{eq:genericity_condition} are equal for all continuous $\varphi$. However, such fine-tuning is ruled out by (A4).
\end{proof}

\begin{remark}
\label{remark:intervention_targets}
\looseness-1
The main difficulty of the proof is that~\eqref{eq:misaligned_ratios} may, in principle, hold when $(p,\tilde p, q, \tilde q)$ and~$\psi$ are completely unconstrained. 
This does not arise in prior work if the intervention targets are known~\citep{lippe2022citris,lippe2022icitris,Liang2023cca} (Case 1), or the densities or mixing are parametrically constrained~\citep{squires2023linear,varici2023score,ahuja2022interventional}.
\end{remark}

\paragraph{On the Genericity Condition (A4).}
\looseness-1 
The condition in~\eqref{eq:genericity_condition} contrasts  expectations of the same quantity w.r.t.\ 
the observational distribution $P_\Vb^{e_0}$ and the interventional distribution $P_\Vb^{e_1}$ corresponding to an intervention on $V_1$ that turns $p_1$ into $\tilde p_1$.
The shared argument, on the other hand, is a function of the ratio between the intervened mechanism $\tilde p_2$ and its base mechanism $p_2$. 
While the two expectations are always equal for linear $\varphi$, other choices imply non-trivial constraints. For instance, $\varphi(y)=y^2$ yields
\begin{align*}
    \label{eq:genericity_example}
    \int \left(\tilde{p}_1(v_1)-p_1(v_1)\right)
    \int \frac{\tilde{p}_2(v_2)^2}{p_2(v_2\mid v_1)}
    \d v_2\d v_1 \neq 0\,.
\end{align*}
Since $p_1 \neq \tilde p_1$ by assumption (A3), we argue that (A4) should generally hold for randomly chosen $(p_1, p_2, \tilde p_1, \tilde p_2)$ and can only be violated if they are fine-tuned to each other. It can thus be viewed as encoding some notion of genericity---in line with the principle of independent causal mechanisms~\citep{scholkopf2012causal,peters2017elements,janzing2012information,besserve2018group,gresele2021independent}, but also involving the intervened mechanisms.
Interestingly, related genericity conditions also arise in the study of nonlinear cause-effect inference from observational data, where identifiability is often obtained up to a set of pathological (``fine-tuned'') spurious solutions satisfying a partial differential equation involving the original mechanisms~\citep{hoyer2008nonlinear,zhang2009identifiability,immer2022identifiability,strobl2022identifying}.
Further, we note that $\varphi$ in~\eqref{eq:genericity_condition} can also be thought of as a \textit{witness function of genericity}, similar to witness functions in kernel-based two sample and independence testing~\citep{gretton2012kernel}. 

\looseness-1 Next, we provide our identifiability result for an arbitrary number of causal variables.
\begin{restatable}[Identifiability up to $\sim_\textsc{crl}$ from two paired perfect stochastic interventions per node]{theorem}{general}
\label{thm:general}    
\looseness-1 
Suppose that we have access to multiple environments $\{P^e_\Xb\}_{e\in\Ecal}$ generated as described in~\cref{sec:problem_setting} under~\cref{ass:faithfulness,ass:known_n,ass:diffeomorphism,ass:shared_mechs,ass:perfect_interventions}.
Let $(\hb,G')$ be any candidate solution such that the inferred latent distributions $Q^e_\Zb=\hb_*(P^e_\Xb)$ of $\Zb=\hb(\Xb)$ and the inferred mixing function $\hb^{-1}$ satisfy the above assumptions w.r.t.\ the candidate causal graph $G'$.
Assume additionally that 
\begin{enumerate}[leftmargin=2.7em]
    \item[(A1)] all densities $p^e$ and $q^e$ are continuously differentiable and fully supported on $\RR^n$;
    \item[(A2')] 
    \looseness-1 we have access to at least one
    \emph{pair} of single-node perfect interventions per  node, with unknown targets: 
    there exist $m\geq n$ known pairs of environments $\Ecal=\{(e_j,e_j')\}_{j=1}^m$ such that for each $i\in[n]$ there exists some \emph{unknown} $j\in[m]$ for which $\Ical^{e_{j}}=\Ical^{e_{j}'}=\{i\}$; 
    \item[(A3')] for all $i\in[n]$, the intervened mechanisms $\tilde p_{i}(v_i)$ and $\dbtilde p_{i}(v_i)$ differ everywhere, 
    in the sense that 
    \begin{equation}
        \label{eq:distinct_interventions}
        \forall v_i: \qquad \bigg(\frac{\dbtilde p_i}{\tilde p_i}\bigg)'(v_i)\neq 0\,;
    \end{equation}
\end{enumerate}
Then the ground truth is identified in the sense of~\cref{def:identifiability}, that is, $(\fb^{-1},G)\sim_\textsc{crl}(h,G')$.
\end{restatable}

\begin{proof}[Proof sketch (full proof in~\cref{app:proof_general}).]
\looseness-1 
By considering ratios between $e_j$ and $e_j'$, taking partial derivatives w.r.t.\ $z_l$, and using assumptions (A3'), we can identify a subset $\Ecal_n\subseteq \Ecal$ of exactly $n$ pairs of environments corresponding to distinct intervention targets in $p$ (for otherwise $\psi$ cannot be invertible). 
For $(e_i,e_i')\in\Ecal_n$, w.l.o.g.\ fix the intervention targets in $p$ to $\Ical^{e_i}=\Ical^{e_i'}=\{i\}$ and let $\pi$ be a permutation of $[n]$ such that $\pi(i)$ denotes the inferred intervention target in $q$ that by (A2') is shared across $(e_i,e_i')$.
By the same argument as before, we must have $V_i=\psi_i(Z_{\pi(i)})$, that is, $\psi$ is a permutation composed with an element-wise function.
It remains to show that $\pi$ is a graph isomorphism, that is, $V_i\to V_j$ in~$G$ $\iff$ $Z_{\pi(i)}\to Z_{\pi(j)}$ in $G'$. We prove $\implies$; the other direction is analogous. Suppose for a contradiction that there exist $(i,j)$ such that $V_i\to V_j$ in~$G$, but $Z_{\pi(i)}\not\to Z_{\pi(j)}$ in $G'$. 
Consider $e_i$ in which there are perfect interventions on $Z_{\pi(i)}$ and $V_i$.
For $Z_{\pi(k)}\in\Zb_{\pa(\pi(j);G')}$, let $\tilde V_k=\psi_{k}(Z_{\pi(k)})$ and denote $\tilde\Vb=\cup_k \tilde V_k\subset \Vb\setminus \{V_i,V_j\}$.
Since $Z_{\pi(i)}$ and $Z_{\pi(j)}$ are d-separated by $\Zb_{\pa(\pi(j);G')}$ in the post-intervention graph $G'_{\overline Z_{\pi(i)}}$ with arrows pointing into $Z_{\pi(i)}$ removed~\citep{Pearl2009}, it follows by Markovianity of $q$ w.r.t.\ $G'$ that $Z_{\pi(i)}\independent Z_{\pi(j)}~|~\Zb_{\pa(\pi(j);G')}$ in~$Q_\Zb^{e_i}$.
By applying the corresponding diffeomorphic functions $\psi_i$, it follows from~\cref{lemma:preserve_cond_ind} in~\cref{app:lemmas} that $V_i\independent V_j~|~\tilde \Vb$ in $P^{e_i}_\Vb$. This violates faithfulness~(\cref{ass:faithfulness}) of $P_\Vb$ to $G$ since $V_i$ and $V_j$ are d-connected in $G_{\overline V_i}$.
\end{proof}%
\looseness-1  (A2') states that we know that a \textit{pair of datasets} corresponds to two distinct {interventions} on the same underlying variable, even though we may not know the exact target of the intervention.
This situation could arise, for example, if both datasets are collected under the same experimental setup but with varying experimental parameters. 
We stress that this is different from \textit{counterfactual} data consisting of \textit{pairs of views} $(\xb,\tilde\xb)$ sharing the values of some variables, as considered in the previous chapter.
One of the main challenges for our analysis (compared to a counterfactual multi-view setting) thus stems from the lack of correspondences across observations from different datasets.
We also stress that a purely observational environment is not needed in this case, cf.\ (A2) in~\cref{thm:bivariate}.

\looseness-1 (A3') states that the intervention mechanisms are distinct in that their ratio is strictly monotonic, similar to~\eqref{eq:different_mechanisms} in (A3).
This is a slightly stronger version of the assumption of \textit{interventional discrepancy} proposed by~\citet{Liang2023cca},\footnote{Cf.\ the \textit{interventional regularity} assumption of~\citet[][Asm.~2]{varici2023score} which instead considers partial derivatives w.r.t.\ the parents and is related to c-faithfulness~\citep[][Defn.~7]{jaber2020causal}.} which has been shown to be necessary for identifiability even if the graph $G$ is known.
For Gaussian $\tilde p_i, \dbtilde p_i$, (A3') is satisfied, for example, by a shift in mean.
In the proofs, this is used to show that $\psi$ must be an element-wise function, see~\eqref{eq:triangularJ}.
Intuitively, if ${\tilde p_i=\dbtilde p_i}$ in some open set for more than one $i$, then the underlying variables can be nonlinearly mixed by a measure-preserving automorphism within this set without affecting the observed distributions~\citep{Liang2023cca}.

\section{Interpreting Causal Representations}
\label{sec:interpreting}
Suppose that we succeed in identifying $\Vb$ and $G$ up to~$\sim_\textsc{crl}$ (\cref{def:identifiability}). 
How can we use or interpret such a causal representation?
Since the scale of the variables is arbitrary~(\cref{subsec:learning_target}),  we clearly cannot predict the exact outcomes of interventions. 
We therefore seek causal quantities that are preserved by the irresolvable ambiguities of $\sim_\textsc{crl}$. 
A prime candidate for this are interventional causal notions defined in terms of information theoretic quantities~\citep{cover2012elements} and in particular the KL divergence $D_\textsc{kl}$. 
\begin{definition}[Causal influence; based on Defn.~2 of~\citet{janzing2013quantifying}]
\label{def:influence}
Let $P_\Vb$ be Markovian w.r.t.\ a DAG $G$ with vertices $\Vb$. For any $V_i\to V_j$ in $G$, the \emph{causal influence of $V_i$ on $V_j$} is given by
\begin{equation*}
    \mathfrak{C}^{P_\Vb}_{i\to j}:=D_\textsc{kl}\big(P_\Vb~\big\|~ P^{i\to j}_\Vb\big), \quad \mbox{where} \quad p^{i\to j}_j\big(v_j~|~\vb_{\pa(j)\setminus \{i\}}\big)
    =
    \medint\int p_j\big(v_j~|~\vb_{\pa(j)}\big) p_i(v_i) \d v_i
\end{equation*}
\looseness-1
and $P_\Vb^{i\to j}$ is the interventional distribution arising from replacing the $j$\textsuperscript{th} mechanism 
by $p^{i\to j}_j$.
\end{definition}

Intuitively, this intervention captures the process of removing the arrow $V_i\to V_j$ in $G$ and ``feeding'' the conditional $p(v_j~|~\vb_{\pa(j)})$ with an independent copy of $V_i$, distributed according to its marginal, see~\citet{janzing2013quantifying} for details.
\looseness-1 The following result, proven in~\cref{app:proof_influence}, states that the causal influences are invariant to reparametrisation and equivariant to permutations, the two irresolvable ambiguities of the $\sim_\textsc{crl}$ equivalence class.

\begin{restatable}[Preservation of causal influences within $\sim_\textsc{crl}$]{theorem}{influence}
\label{prop:influence}
Let $P_\Vb$ be Markovian w.r.t.~$G$, let $\pi$ be a graph isomorphism of $G$, and let $\phi$ be an element-wise diffeomorphism. Let $\Zb=\Pb_{\pi^{-1}}\circ \phi(\Vb)$ and denote its induced distribution by $Q_\Zb$.
Then for any $V_i\to V_j$ in $G$ we have:
\begin{equation}
\mathfrak{C}^{P_\Vb}_{i\to j}=\mathfrak{C}^{Q_\Zb}_{\pi(i)\to \pi(j)}\,.
\end{equation}
\end{restatable}

\Cref{prop:influence} implies that the strength of causal relations among variables in the inferred graph carry meaning. They can thus be used to uncover changes to the latent causal mechanisms underlying different experimental datasets, for example, to gain scientific insights when combined with domain knowledge. We will return to this point in~\cref{chap:conclusion}.

\section{Learning Objectives}
\label{sec:objectives}
While our main focus is on studying identifiability, our theoretical insights also suggest approaches to learning causal representations from finite interventional datasets sampled from $\{P^e_\Xb\}_{e\in\Ecal}$.
The main idea is to fit the data in a way that preserves the sparsity of interventions~\citep{scholkopf2021toward,perry2022causal} by employing the same (un)mixing function and sharing mechanisms across environments~(\cref{ass:shared_mechs}).
We sketch two approaches, which, according to our theory,  should both asymptotically identify the ground truth up to $\sim_\textsc{crl}$ if the set of available environments $\Ecal$ is sufficiently diverse (and the other assumptions hold).

\begin{enumerate}
    \item \textbf{Autoencoder Framework.}
Jointly learn an encoder~$\hb$, a graph~$G'$, and intervention targets~$\Ical^e$ such that the encoded latents $\Zb=\hb(\Xb)$ can be used to reconstruct the observed $\Xb$ across all environments, while ensuring all but the intervened mechanisms are shared.
Using $E$ as an environment indicator, the latter corresponds to the constraint $Z_i \independent E~|~\Zb_{\pa(i;G')}$ for $i\not\in\Ical^e$, implementable, for example, through a suitable conditional independence test~\citep{fukumizu2007kernel,zhang2011kernel,park2021conditional}.
\item \textbf{Generative Modelling Approach. }
Fit a base generative model $(G',p,\fb)$ and intervention models $(\Ical^e,\{p^e_i\}_{i\in\Ical^e})_{e\in \Ecal}$  by maximizing the likelihood of the multi-environment data. For example, given a candidate graph $G'$ and candidate intervention targets $\{\Ical^e\}_{e\in \Ecal}$, learn the base and intervened mechanisms and mixing; %
then pick the graph and intervention targets that achieve the best fit. 
\end{enumerate}

\section{Experiments}
\label{sec:experiments}
\paragraph{Setup.}
We experimentally pursue the second, generative approach.
Specifically, we model the mixing function generating $\Xb$ from $
\Zb$ as a \textit{normalizing flow}~\citep{papamakarios2021normalizing,durkan2019neural} with different environment-specific base distributions, determined by the underlying causal graph, intervention targets, and (learnt) base and intervened mechanisms.
Here, we focus on the bivariate case with two ground-truth causal latent variables $V_1\to V_2$. According to~\cref{thm:bivariate}, this setting should be identifiable from three environments: an observational one and one perfect intervention on each of $V_1$ and $V_2$.
Our goal is to verify this empirically in light of finite data and optimization issues.
We fix the observation dimension to $d=2$ to facilitate exact likelihood training of the normalizing flows, and fit a separate generative model for each choice of graph and intervention targets. This can be viewed as a nested approach in which the inner loop corresponds to the method for Causal Component Analysis (CauCA) of~\citet{Liang2023cca} (see~\cref{sec:CauCA}), and the outer loop to a search over $G'$ and $\{\Ical^e\}_{e\in\Ecal}$. Code to reproduce our experiments is available at: \href{https://github.com/akekic/causal-component-analysis}{https://github.com/akekic/causal-component-analysis}.
The base and intervened mechanisms are linear Gaussian and the mixing function a three-layer MLP, see~Appendix C.1 of~\citet{von2023nonparametric} for further implementation details.

\begin{figure}[t]
\centering
      \includegraphics[width=\textwidth]{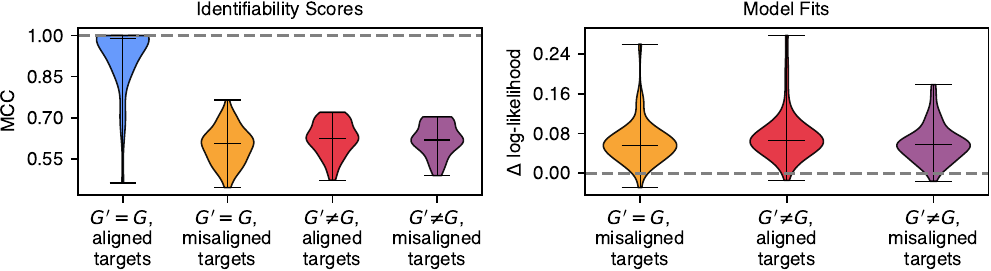}
\caption[Empirical Comparison of Correctly and Incorrectly Specified Models]{\looseness-1 
\textbf{Empirical Comparison of Correctly and Incorrectly Specified Normalizing Flow-Based Models.} 
For $n=2$ latent causal variables with graph $G$ given by $V_1\to V_2$, we compare a generative model based on the correct causal graph $G'=G$ and intervention targets (blue) to other generative models assuming the wrong graph $G'\ne G$ or misaligned intervention targets (yellow, red, purple).
We show mean correlation coefficients (MCCs) between the learned and ground truth latents \textit{(Left)} and the difference in validation model log-likelihood between the well-specified and misspecified models \textit{(Right)}.
Each violin plot is based on 50 different ground truth data generating processes;
the horizontal lines indicate the minimum, median and maximum values.
\label{fig:experiments_main}}
\end{figure}

\paragraph{Results.}
The results are summarized in~\cref{fig:experiments_main}.
Our main findings are two-fold:
First, we observe that, in the majority of cases, the well-specified model attains the highest held-out log-likelihood, as shown in~\cref{fig:experiments_main} \textit{(Right)}. 
This suggests that the likelihood of otherwise comparable generative models can act as a reliable criterion to select the correct causal graph and intervention targets.
Second, we find that the ground truth latent causal variables are approximately identified up to element-wise rescaling (MCC values close to one) by the correctly specified model and not by any other model, as shown in~\cref{fig:experiments_main} \textit{(Left)}.
This indicates that recovering the correct graph and targets is not only sufficient but also necessary for reliable identification of the causal representation.
Taken together, these findings are consistent with our notion of $\sim_\textsc{crl}$-identifiability~(\cref{def:identifiability}) and~\cref{thm:bivariate}.

\clearpage
\section{Discussion}
\label{sec:discussion}
\looseness-1 

\paragraph{Weaker Notions of Identifiability.}
\looseness-1 
Here, we have focused on the strongest notion of identifiability that is achievable in a nonparametric setting~(\cref{def:identifiability}). 
However, subject to the available data and assumptions, identifiability
up to~$\sim_\textsc{crl}$ will not always be possible. In this case, weaker notions of identifiability are of interest.
For example, we may not be able to uniquely recover variables that are not targeted by interventions~\citep{lippe2022citris,ahuja2022interventional,von2021self}, or only recover groups of variables up to (non-)linear mixing~\citep{von2021self,varici2023score,ahuja2022interventional} and the graph up to transitive closure~\citep{squires2023linear} if interventions are imperfect, or soft.
A precise characterization of weaker notions of \textit{nonparametric} identifiability from different types of \textit{interventions} (cf.~\citet{lachapelle2022partial} for a temporal, semi-parametric setting) is an interesting direction for future work.

\paragraph{Known vs.\ Unknown Intervention Targets.}
\looseness-1
When intervention targets can be considered known appears to be a more nuanced concept in CRL than in a fully observed setting, see also \citet[\S E]{Liang2023cca} for an extended discussion.
Recall that we assume w.l.o.g.\ that $V_1\preceq...\preceq V_n$ and only consider graphs respecting this ordering~(\cref{subsec:learning_target}), see also~\citet[][Remark 1]{squires2023linear}.
The intervention targets are then unknown w.r.t.\ the pre-imposed causal ordering.
This is a key aspect that makes our setting more realistic, but also
substantially complicates the analysis (see~\eqref{eq:misaligned_ratios} and~\cref{remark:intervention_targets}).
Similar to~\citet{varici2023score}, for~\cref{thm:bivariate} we require a set of \textit{exactly} $n$ environments (one intervention on each node).\footnote{By dropping the assumption of a fixed ordering and considering all DAGs, in this case one could also call $V_i$ the variable intervened upon in environment $i$, and then consider the targets ``known'' in this sense.}
However, we relax this requirement to mere coverage (``at least one'') in~\cref{thm:general} as shown for linear causal models by~\citet{squires2023linear,buchholz2023learning}.

\paragraph{Identifiability From One Intervention per Node for Any Number of Nodes.}
We conjecture that~\cref{thm:bivariate} can be generalized to $n>2$, subject to a suitably adjusted set of genericity conditions involving several intervened and base mechanisms, akin to~\eqref{eq:genericity_condition} in the bivariate case.
The main challenge to such a generalisation appears to be combinatorial, as there are $n!-1$ ways of misaligning intervention targets across $p$ and $q$.
In~\cref{thm:general}, we sidestep this issue by learning from \textit{pairs} of environments.
Thus, while two single-node perfect interventions are  sufficient, we do not believe this to be necessary. 
\paragraph{On the Assumption of Known $n$ and Its Relation to Markovianity.}
\looseness-1 
\Cref{ass:known_n} relates to the notions of  \textit{causal sufficiency} or \textit{Markovianity} in classical causal inference, which correspond to the assumption of independent $U_i$ in~\cref{def:SCM}, implying the causal Markov factorization~\eqref{eq:causal_Markov_factorisation}. 
With \textit{unobserved} $\Vb$ and \textit{unknown} $n$, the notion of ``unobserved confounders'' gets blurred,
since one can, in principle, always construct a causally sufficient system by increasing~$n$ and adding any causes of two or more endogenous variables to~$\Vb$.
\Cref{ass:known_n} then states that the minimum number of variables required to do so is known.\footnote{Techniques for estimating the intrinsic dimensionality $n$ of the observation manifold $\Xcal\subseteq \RR^d$ or methods rooted in Bayesian nonparametrics could provide a means of relaxing this assumption.}
However, this can lead to very large systems, 
which may in turn challenge the assumption of an invertible mixing~(\cref{ass:diffeomorphism}). 
Extensions of our analysis to unknown $n$, non-Markovian, or non-invertible models constitute an interesting direction for future investigations.

\paragraph{Practicality.}
The method explored in~\cref{sec:experiments} requires searching over graphs and intervention targets, which  gets intractable even for moderate $n$.
Simultaneously learning an (un)mixing function, causal graph,  intervention targets, and mechanisms is challenging. Further work is needed to make methods for nonparametric CRL from multi-environment data more practical, e.g., by exploring the proposed autoencoder framework with a continuous parametrisation of graph~\citep{zheng2018dags,zheng2020learning} and targets~\citep{jang2017categorical}.

\section{Extensions and Connections with Other Work}
\label{sec:extensions_multi_env}
We now briefly sketch connections with some of our other works in a multi-environment setting. 
In~\cref{sec:multi_env_causal_discovery}, we consider classical structure learning (i.e., the causal variables are directly observed)  by leveraging sparse mechanism shifts across environments arising from unknown soft interventions, possibly on multiple  targets.
Conversely, in~\cref{sec:CauCA} we consider a special case of the multi-environment setting studied in this chapter, for which both the causal graph and the intervention targets are known, allowing for stronger identifiability results. 
Finally, in~\cref{sec:supervised_CRL_domain_generalisation} we consider supervised CRL in the form of domain generalisation.

\subsection{Multi-Environment Causal Discovery}
\label{sec:multi_env_causal_discovery}
This subsection is based on the following publication, with all figures therein adopted without further modification. 
We briefly summarise the main points that are relevant to the context of this chapter, and refer to the full paper for further details.
\begin{selfcitebox}
\href{https://arxiv.org/abs/2206.02013}{\ul{Causal discovery in heterogeneous environments under the sparse mechanism shift hypothesis}}
\\
Ronan Perry, \textbf{Julius von K\"ugelgen}$^{\dagger}$, Bernhard Sch\"olkopf$^{\dagger}$ ($^{\dagger}$shared last author)
\\
\textit{Advances in Neural Information Processing Systems (NeurIPS)}, 2022
\end{selfcitebox}

For the main results presented in this chapter, we assumed that the interventions giving rise to different environments are \textit{perfect}~(\cref{ass:perfect_interventions}) and occur on single nodes at a time.
Here, we show that, in the fully observed case (i.e., when the mixing function is the identity), these restrictions are not needed.
To this end, we study a classical structure learning task and show that imperfect, structure-preserving interventions (i.e., $p^e(V_i~|~\PA_i)=\tilde p(V_i~|~\PA_i)$), possibly on multiple nodes simultaneously, provide enough information to infer the causal graph if sufficiently many such environments are available.
Our main idea is to leverage the \textit{sparse mechanism shift hypothesis}~\citep{scholkopf2021toward}, which posits that mechanism changes tend to be sparse w.r.t.\ the true graph.
In the context of~\cref{ass:shared_mechs}, we formalize this as follows.

\begin{figure}[tbp]
    \centering
    \includegraphics[width=\textwidth]{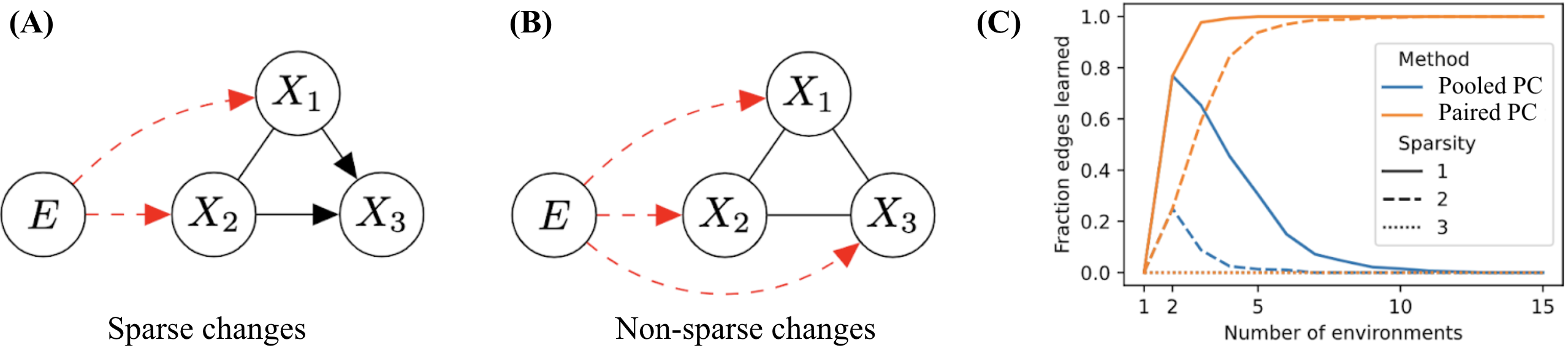}
    \caption[Sparse Shifts Yield Structure Identifiability]{\textbf{Sparse Shifts Yield Structure Identifiability}. In a causal model with a fully connected DAG over $\Xb:=\{X_1,X_2,X_3\}$, no edge directions can be learned from i.i.d.\ observational data.
    Given multiple datasets on $\Xb$ across different environments with possible distribution shifts, the PC algorithm~\citep{spirtes2001causation} can be applied to the pooled data augmented by the environment index variable $E$~\citep{huang2020causal}. \textbf{(A)}~A mechanism shift $E \to X_i$ can allow orientation of some edges. \textbf{(B)}~Dense shifts prohibit any orientations~\citep{huang2020causal}.
    \looseness-1 \textbf{(C)}~Combining results learned across \textit{pairs} of pooled environments leads to identifiability under \textit{sparse} shifts. Pooling all environments leads to dense shifts, even if pairs of environments differ sparsely; under \textit{dense} shifts, only the equivalence class is learned.}
    \label{fig:oracle_sparse_pc}
\end{figure}

\begin{assumption}[Sparse Mechanism Shifts]
\label{assmpt:sparse_change}
\looseness-1 Changes in mechanisms between observed environments are sparse, in the sense that for all $e\in\Ecal$:
\begin{equation}
    \textstyle
    0 < |\Ical^e| < n\,.
\end{equation}%
\end{assumption}%
The value of this assumption when met is illustrated in~\cref{fig:oracle_sparse_pc}.
A key ingredient of our method is a \textit{pair-wise} comparison of environments, which preserves sparsity, whereas pooling environments with an auxiliary environment indicator variable ($E$ in~\cref{fig:oracle_sparse_pc}) does not.
Specifically, we propose a score-based multi-environment causal discovery algorithm based on the \textit{Mechanism Shift Score} (MSS). 
Given a candidate DAG $G$ over $n$ observed variables $\Xb = \{X_1, ..., X_n\}$ and data from multiple environments $\Ecal$, the MSS counts the number of conditionals in the causal Markov factorisation~\eqref{eq:causal_Markov_factorisation} implied by $G$ that change across any pair of environments $e,e'\in \Ecal$,
\begin{align*}\label{eq:min_shift_estimand}
    \text{MSS}(G; \Ecal) &=
    \sum_{i=1}^n
    \sum_{e<e'} \II\left[P^e(X_i~|~\PA_i^G) \neq P^{e'}(X_i~|~\PA_i^G)\right]\,.
\end{align*}
We then show that the true DAG $G^*$ minimizes (not necessarily uniquely) the number of changing conditionals among all DAGs, meaning that changes in any other factorization than the true causal one are at least as dense. 
This can be viewed as a generalisation of the Principle of Minimal Changes~\citep{ghassami2018multi} beyond linear models. 
For a given set of environments~$\Ecal$, the set of all minimisers of MSS corresponds to the $\Psi$-Markov equivalence class w.r.t.\ the true but unknown intervention targets $\{\Ical^e\}_{e\in\Ecal}$~\citep{jaber2020causal}.
Under a probabilistic notion of sparsity (with Nature randomly perturbing mechanisms to produce new environments), we bound the probability that this class reduces to the unique true graph, and show that it converges to one in the limit of infinite environments.
\begin{figure}[tbp]
    \centering
    \vspace{-1em}
    \includegraphics[width=0.5\textwidth]{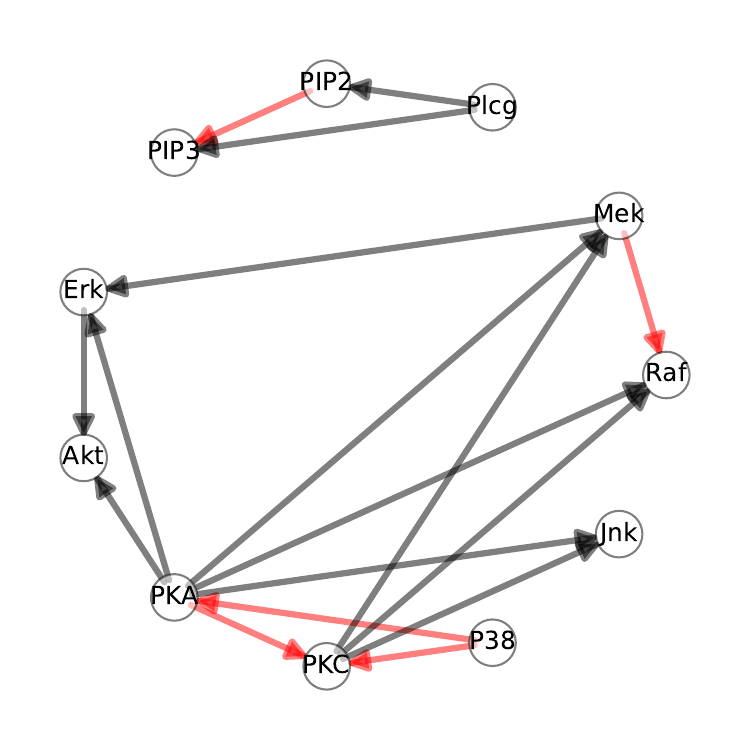}
    \caption[DAG Minimizing the Mechanism Shift Score on the Sachs Protein Dataset]{\textbf{DAG Minimizing the Mechanism Shift Score on a Cytometry Dataset.}  The learned edges mostly match the expert graph~\citep{sachs2005causal}. Non-matching edges (red) have been posited to be involved in cycles or of ambiguous orientation in the literature.}
    \label{fig:cytometry}
\end{figure}

MSS is amenable to various estimators: empirically, we
find the combination of soft, p-value based scores and KCI~\citep{zhang2011kernel} to test for $X_i \independent E~|~\Pa_i$ to perform well across a range of simulated settings. 
We also apply this approach to a well-studied cytometry dataset~\citep{sachs2005causal} consisting of $9$ experimental environments of $11$ cellular proteins.
The DAG which uniquely minimizes the MSS is visualized in~\cref{fig:cytometry}, and mostly matches the expert graph. 

While imperfect interventions are problematic and generally insufficient for CRL (see~\cref{subsec:multi_environment_data} and the discussion before~\cref{ass:perfect_interventions}), our results demonstrate that they suffice for structure learning when the intervened upon variables are directly observed. 

\subsection{Causal Component Analysis (CauCA)}
\label{sec:CauCA}
This subsection is based on the following publication, with all figures therein adopted without further modification. 
We briefly summarise the main points that are relevant to the context of this chapter, and refer to the full paper for further details.
\begin{selfcitebox}
\href{https://arxiv.org/abs/2305.17225}{\ul{Causal component analysis}}
\\
Liang Wendong, Armin Keki\'c, \textbf{Julius von K\"ugelgen}, Simon Buchholz, Michel Besserve, Luigi Gresele$^\dagger$, Bernhard Sch\"olkopf$^\dagger$ ($^\dagger$shared last author)
\\
\textit{Advances in Neural Information Processing Systems (NeurIPS)}, 2023
\end{selfcitebox}

\begin{figure}
    \centering
    \includegraphics[width=\textwidth]{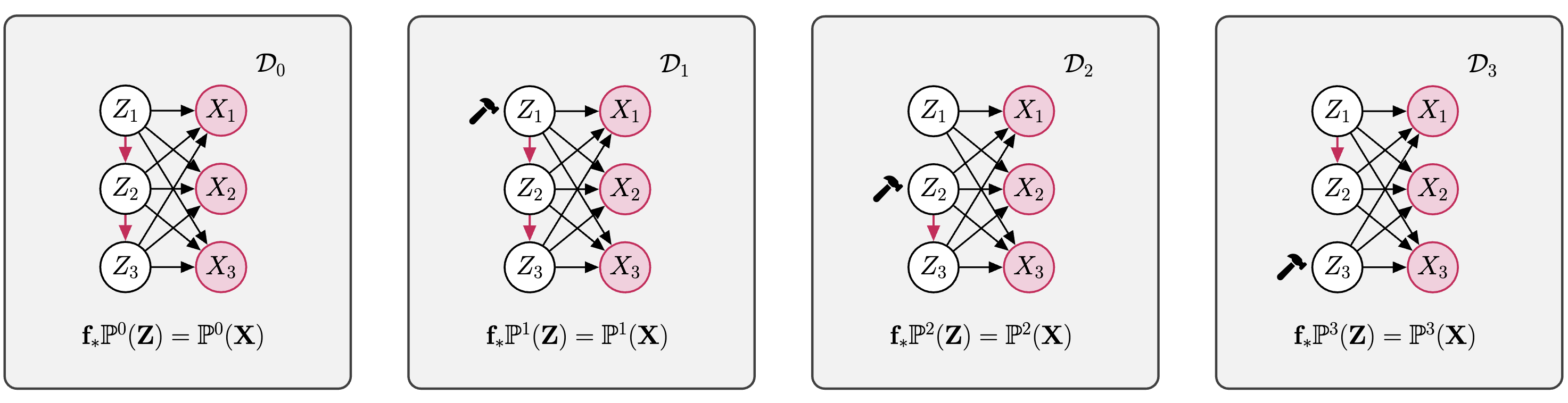}
    \caption[Causal Component Analysis Problem Setting]{\textbf{Causal Component Analysis (CauCA).} 
    In CauCA, observed variables~$\Xb$ are generated through a nonlinear mapping~$\fb$, applied to unobserved latent variables~$\Zb$ which are causally related. 
    The causal structure~$G$ of the latent variables is assumed to be known, while the     causal mechanisms~$P_i(Z_i~|~\Zb_{\text{pa}(i)})$ and the nonlinear mixing function are unknown, and need to be estimated.
    (Known or observed quantities
    are highlighted in red.) 
    CauCA assumes access to multiple datasets~$\Dcal_e$ that result from stochastic interventions on the latent variables.
    }
    \label{fig:CauCA}
\end{figure}

Whereas CRL aims to infer causally dependent latents and the {unknown} graph encoding their relations, ICA assumes that this graph is {known} to be {trivial}. 
We propose to study {Causal} Component Analysis (CauCA) as an intermediate representation learning problem, in which \textit{the causal graph is known, but non-trivial}, see~\cref{fig:CauCA} for an illustration. 
CauCA can be viewed as a causal generalisation of ICA, or as a special case of CRL that focuses only on learning the (un)mixing function and causal mechanisms.
Any negative results regarding identifiability of CauCA therefore also apply for CRL, while possibility results translate directly to ICA and may serve as a stepping stone for extensions to CRL.

\begin{table}[t]
\centering
\caption[Identifiability Results for CauCA and ICA]{
\textbf{Identifiability Results for CauCA and ICA.} 
We illustrate our identifiability guarantees for representations learnt by maximizing the likelihoods $P^e(\Xb)$ across different interventional regimes $e\in\Ecal$ with $n=3$ latents.
CauCA refers to the example DAG from~\cref{fig:CauCA} ($Z_1 \xrightarrow{} Z_2 \xrightarrow{} Z_3$), and ICA to the special case with independent $Z_1,Z_2,Z_3$.
$h_i(\cdot)$ are arbitrary nonlinear functions s.t.\ $\hb(\cdot)=[h_1(\cdot),h_2(\cdot),h_3(\cdot)]$ is invertible, and
$\pi$ is an arbitrary permutation.
}
\resizebox{\textwidth}{!}{
\def\arraystretch{1.15}%
\begin{tabular}{c m{7.25cm} m{6.5cm}}
\toprule%
\textbf{Setting} & \textbf{Interventional Environments} & \textbf{Learned Representation} $\hat{\mathbf{z}}=\hat{\mathbf{f}}^{-1}(\mathbf{x})$ \\[.25em]
\midrule
CauCA & $1$ intervention per node & $\left[h_1(z_1), h_2(z_1,z_2), h_3(z_1,z_2,z_3)\right]$ \\[.5em] 
\hline
\addlinespace
CauCA & $1$ \textit{perfect} intervention per node & $\left[h_1(z_1), h_2(z_2), h_3(z_3)\right]$ \\[.5em] 
\hline
\addlinespace
CauCA & $1$ intervention per node for $z_1$ and $z_2$, plus $|\overline{\pa}(3)|(|\overline{\pa}(3)|{+}1)=2{\times}3$ imperfect interventions on $z_3$ with ``variability'' assumption & $\left[h_1(z_1), h_2(z_2), h_3(z_2,z_3)\right]$ \\[1.5em] 
\hline
\addlinespace
CauCA & $1$ perfect intervention on $z_1$ and $2{+}1{=}3$ perfect fat-hand interventions on $(z_2, z_3)$ & 
$\left[h_1(z_1), h_2(z_2, z_3), h_3(z_2,z_3)\right]$ \\[1em] 
\midrule
\addlinespace
ICA & $1$ intervention per node on any two nodes & $\left[h_1(z_1), h_2(z_2), h_3(z_3)\right]$ 
\\[.5em]
\hline\addlinespace
ICA & $1$ intervention per node on any two nodes with unknown targets & 
$ \pi \left[h_1(z_1),h_2(z_2), h_3(z_3)\right]$ 
\\[1.em]
\hline\addlinespace
ICA & $1$ intervention on $z_1$ and $2$ fat-hand interventions on $(z_2, z_3)$  with  ``block-variability'' assumption & 
$\left[h_1(z_1), h_2(z_2, z_3), h_3(z_2,z_3)\right]$ 
\\[1.em]
\hline\addlinespace
ICA & $1$ intervention on $z_1$ and $4$ fat-hand interventions on $(z_2, z_3)$ with ``block-variability'' assumption & 
$\Bigl[h_1(z_1), \pi[h_2(z_2), h_3(z_3)]\Bigr]$ 
\\[1.em]
\hline\addlinespace
ICA & $6$ fat-hand interventions on $(z_1, z_2, z_3)$ with variability assumption~\citep[][Thm.1]{hyvarinen2019nonlinear} & 
$ \pi \left[h_1(z_1),h_2(z_2), h_3(z_3)\right]$\\
\bottomrule
\end{tabular}
}
\label{table:CauCA_eg}
\end{table}

We study CauCA in a multi-environment setting in which the intervention targets $\{\Ical^e\}_{e\in\Ecal}$ are \textit{known}. However, we do not restrict ourselves to perfect interventions, but also consider imperfect and multi-node (``fat-hand'') interventions. 
We introduce the \textit{interventional discrepancy} assumption discussed in~\cref{sec:theory} (which underlies assumptions (A3) and (A3') of~\cref{thm:bivariate,thm:general}), and
present several identifiability results for different types and combinations of interventions---see~\cref{table:CauCA_eg} for a summary of the main results in the context of~\cref{fig:CauCA} (CauCA) and for the special case of ICA with independent latents.
Similar to~\cref{thm:bivariate} in the full CRL setting, one perfect intervention per node suffices for identifiability of CauCA, though without the need for a genericity condition as the known graph and intervention targets provide additional information.
We also find that partial disentanglement of the causal latents is possible even from imperfect or fat hand interventions. 

Moreover, this causal perspective also leads to new identifiability results for the special case of multi-environment nonlinear ICA.
In particular, we find that linking different datasets to targeted interventions on \textit{subsets} of the latent variables enables identifiability from strictly fewer environments than previous results~\citep{hyvarinen2019nonlinear} that place assumptions on changes at the level of the entire joint distribution of all latents as discussed in~\cref{sec:background_nonlinear_ICA_auxiliary}.

In summary, we find our study of the intermediate CauCA problem to yield new insights on possibilities for ICA, as well as on hopes and fundamental limitations for CRL.

\subsection{Domain Generalisation and Supervised CRL}
\label{sec:supervised_CRL_domain_generalisation}
This subsection is based on the following publications, with all figures therein adopted without further modification.
We briefly summarise the main points that are relevant to the context of this chapter, and refer to the full papers for further details.
\begin{selfcitebox}
\href{https://arxiv.org/abs/2207.09944}{\ul{Probable domain generalisation via quantile risk minimisation}}
\\
Cian Eastwood$^*$, Alexander Robey$^*$, Shashank Singh, \textbf{Julius von K\"ugelgen}, \\Hamed Hassani, George J.\ Pappas, Bernhard Sch\"olkopf ($^*$equal contribution)
\\
\textit{Advances in Neural Information Processing Systems (NeurIPS)}, 2022
\end{selfcitebox}

\begin{selfcitebox}
\href{https://arxiv.org/abs/2307.09933}{\ul{Spuriosity didn't kill the classifier: Using invariant predictions to harness spurious features}}
\\
Cian Eastwood$^*$, Shashank Singh$^*$, Andrei Liviu Nicolicioiu, Marin Vlastelica, \\ \textbf{Julius von K\"ugelgen}, Bernhard Sch\"olkopf ($^*$equal contribution)
\\
\textit{Advances in Neural Information Processing Systems (NeurIPS)}, 2023
\end{selfcitebox}

\begin{figure}[t]
    \centering
    \begin{subfigure}[b]{0.24\linewidth}
        \centering
        \includegraphics[width=\linewidth]{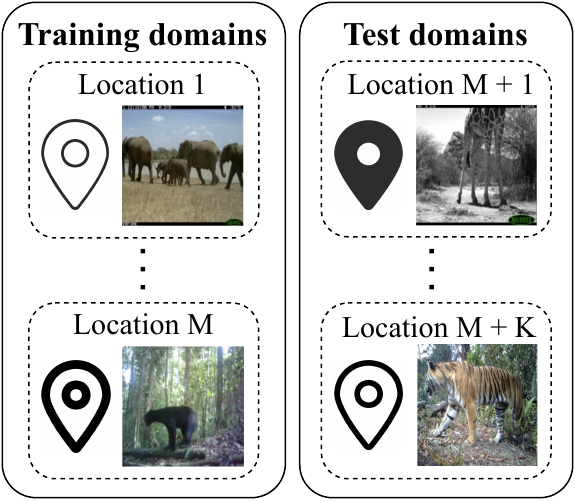}
        \vspace{0.1mm}
        \caption{}
        \label{fig:fig1:train-test}
    \end{subfigure}
    \hfill
    \begin{subfigure}[b]{0.365\linewidth}
        \centering
        \includegraphics[width=0.95\textwidth]{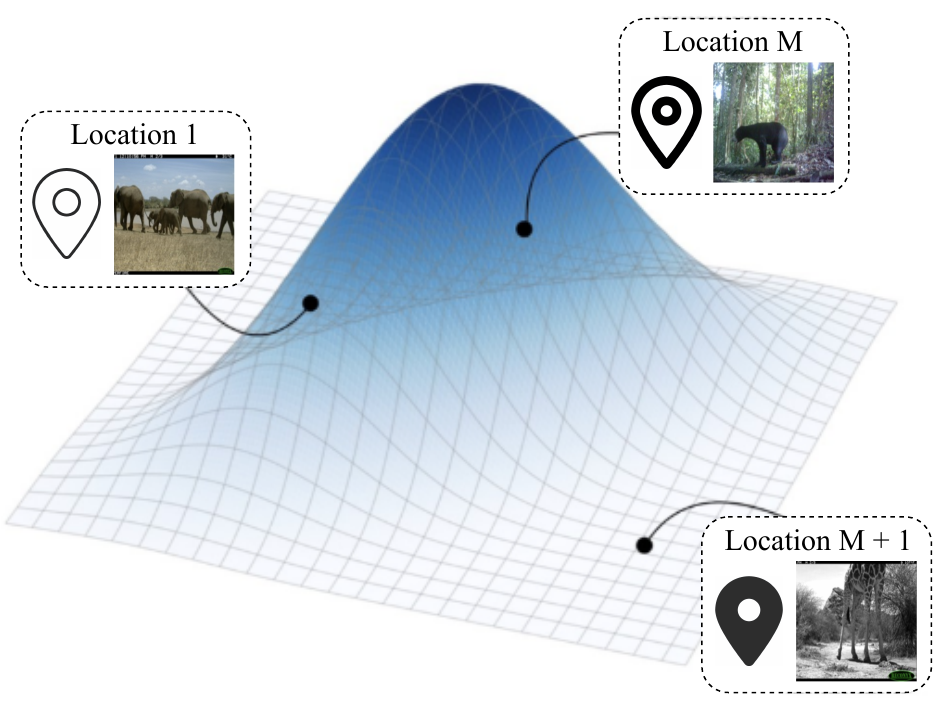}
        \caption{}
        \label{fig:fig1:q-dist}
    \end{subfigure}
    \hfill
    \begin{subfigure}[b]{0.36\linewidth}
        \centering
        \includegraphics[width=\linewidth]{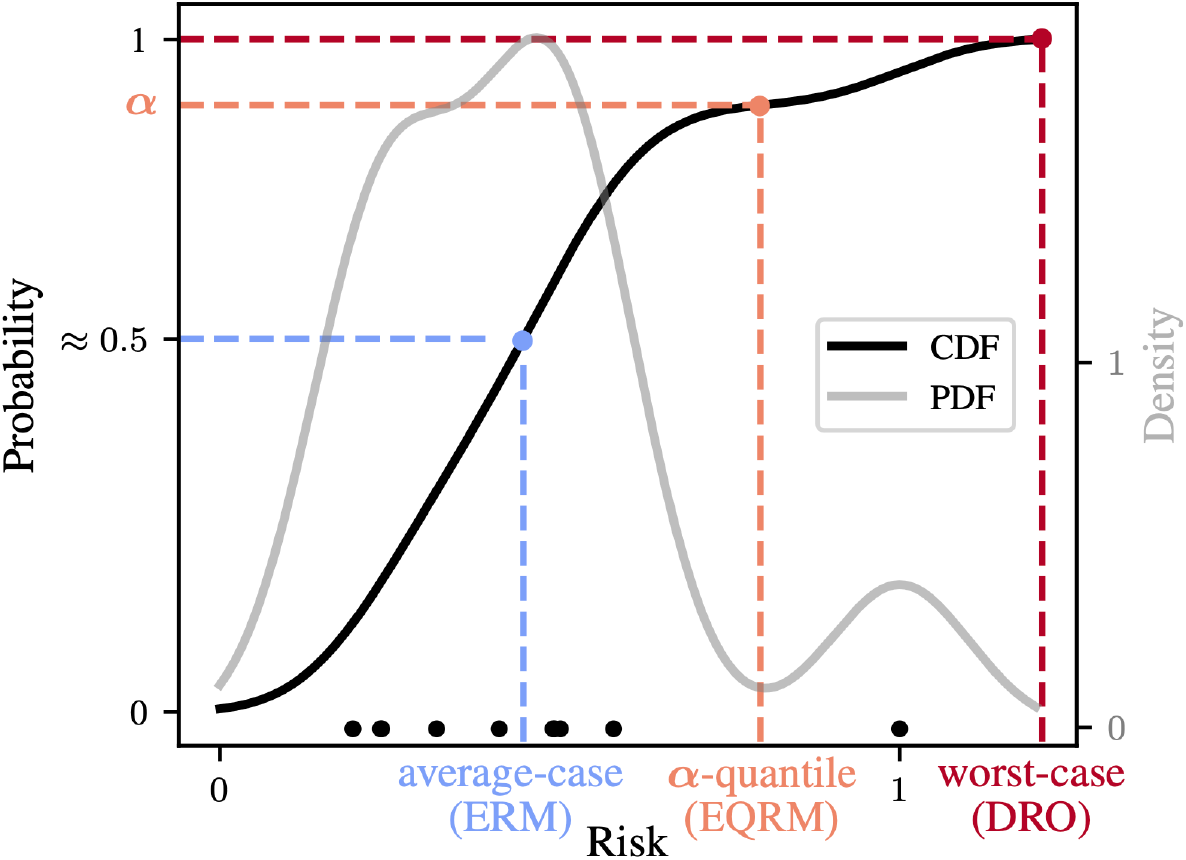}
        \caption{}
        \label{fig:fig1:risk}
    \end{subfigure}
    \caption[Probable Domain Generalisation and Quantile Risk Minimisation (QRM)]{\textbf{Probable Domain Generalisation and Quantile Risk Minimisation (QRM).} (a) In domain generalisation, training and test data are drawn from multiple related distributions or domains. For example, in the \texttt{iWildCam} dataset~\citep{beery2021iwildcam}, which contains camera-trap images of animal species, the domains correspond to the different camera-traps which captured the images. (b) We relate training and test domains as draws from the same underlying (and often unknown) meta-distribution over domains $Q$. (c) We consider a predictor's estimated risk distribution over training domains, naturally-induced by $Q$. \looseness-1 By minimizing the $\alpha$-quantile of this distribution, we learn predictors that perform well with high probability ($\approx \alpha$) rather than on average or in the worst case.
    }
    \label{fig:QRM}
\end{figure}

Throughout this thesis, we have mostly studied \textit{unsupervised} learning problems, in which we only have unlabelled observations $\Xb$---possibly across multiple views or environments---and the goal is to learn about the underlying structure and uncover certain properties of the generative process. 
We now finally turn to a \textit{supervised} setting, in which we additionally observe a label or target variable $Y$.
Specifically, in the spirit of the multi-environment setting studied in this chapter, we consider the problem of \textit{domain generalisation} (DG)~\citep{blanchard2011generalizing,muandet2013domain,gulrajani2020search}.
Given multiple labelled training domains (or environments, we use the two terms interchangeably), the aim of DG is to learn a predictor for $Y$ that performs well on new, unseen test domains, see~\cref{fig:fig1:train-test} for an illustration.
Depending on the interpretation of what it means to ``perform well'' on new domains, different approaches to DG have been explored.
\paragraph{Average-Case Generalisation: ERM.} 
Perhaps the most natural approach is to
minimise the \textit{average} risk across all domains~\citep{blanchard2011generalizing,muandet2013domain}, 
\begin{equation}
\label{eq:domain-gen-average-case}
    \min_{f\in\Fcal} \, \EE_{e \sim Q} \, \left[\Rcal^e(f)\right]\,,
\end{equation}
where $Q$ denotes a meta-distribution over domains $e$~(see~\cref{fig:fig1:q-dist}), $\Fcal$ is the function class of considered predictors $f:\Xcal\to\Ycal$, and 
\begin{equation}
\label{eq:domain_specific_risk}
    \Rcal^e(f) = \mathbb{E}_{(\Xb, Y)\sim P^e} \left[l(f(\Xb), Y)\right]
\end{equation}
denotes the risk of $f$ in domain $e$ with joint distribution $P^e(\Xb,Y)$ and loss function $l$.
In practice, the expectations in~\eqref{eq:domain-gen-average-case} and~\eqref{eq:domain_specific_risk} can be approximated using the available pooled training data, amounting to the standard empirical risk minimisation (ERM) framework.
ERM yields predictors that perform well \textit{on average} w.r.t.\ domains sampled from~$Q$. %
However, such predictors tend to rely on spurious features~(see~\cref{sec:weaknesses_of_current_ML}), lack robustness, and thus might perform poorly on any new, unseen domain~\citep{nagarajan2021understanding,rojas2018invariant}.
This is particularly undesirable in safety-critical applications, as motivated in~\cref{ex:medical_AI}. 

\paragraph{Worst-Case Approach: DRO and IRM.} 
It has therefore been argued that DG should instead be treated as a \textit{worst-case} problem,
\begin{equation}
\label{eq:domain-gen}
    \min_{f\in\Fcal} \, \max_{e \in \text{supp}(Q)} \, \Rcal^e(f).
\end{equation}
Given only finitely many training domains, solving \eqref{eq:domain-gen} is impossible without additional restrictions on how domains may differ~\citep{ben2010theory}. This motivates approaches rooted in distributionally robust optimisation (DRO)~\citep{sagawa2019distributionally} or causality~\citep{christiansen2021causal,rothenhausler2021anchor,krueger21rex}.
In DRO, the set of considered domains is typically specified directly, e.g., by bounding a distance to the observed (empirical) distributions. 
Causal approaches instead specify domain shifts indirectly through interventions in an underlying causal model, which can be seen as a special case of DRO~\citep{meinshausen2018causality}.
The method of invariant causal prediction~\citep[ICP;][]{peters2016causal,heinze2018invariant} is based on the insight that only predictors that rely exclusively on causal parents $\PA_Y$ of the target are invariant to arbitrary interventions (except those directly affecting $Y$). 
While originally proposed for settings in which the causal variables, including $\PA_Y$, are directly observed, ICP has also inspired methods for supervised CRL~\citep{lu2021invariant}.  A notable approach in the context of DG from high-dimensional observations is \textit{invariant risk minimisation}~\citep[IRM;][]{arjovsky2019invariant}, which aims to learn an $f$ s.t.\ the optimal linear predictor $\wb^\top f(\Xb)$ is the same for all training domains. 
However, striving for perfect invariance---thereby optimizing for worst-case performance---can lead to discarding a lot of information and thus results in overly conservative solutions, seeing as worst-case, truly adversarial shifts tend to be unlikely in practice~\citep{tsipras2019robustness, raghunathan2019adversarial}.

\paragraph{\textit{Probable} Domain Generalisation via Quantile Risk Minimisation (QRM).}
To address the shortcomings of the average and worst-case approaches pursued by ERM and DRO/IRM, respectively, we propose to instead approach DG as a probabilistic problem and seek predictors that perform well \emph{with high probability}.
The key idea is that distribution shifts seen during training should inform us of the types of shifts that are likely to occur at test time.
Specifically, we consider the \textit{risk profile} of a given predictor, defined as the distribution of risks across domains induced by the meta-distribution $Q$.
To learn a predictor that achieves low risk with high probability, we propose to \textit{minimise the $\alpha$ quantile} of this risk profile by solving
\begin{equation}
\label{eq:prob_gen}
	\min_{f\in\Fcal,\, t \in \RR}  \, t  \qquad \text{subject to} \qquad  \Pr_{e\sim Q} \left\{\Rcal^e(f) \leq t \right\} \geq \alpha \,.
\end{equation}
We refer to this approach as \textit{quantile risk minimisation} (QRM), and say that a solution to~\eqref{eq:prob_gen} ``generalizes with risk at most~$t$ 
with probability at least~$\alpha$ over domains sampled from $Q$''.
 The conservativeness hyper-parameter $\alpha$  thus determines the probability of generalisation, with $\alpha=1$ corresponding to the worst case problem, whereas $\alpha=0.5$ optimises for the median risk (thus recovering the average-case behaviour of ERM for symmetric risk profiles), see~\cref{fig:fig1:risk} for an illustration. 
 To solve QRM in practice, we propose the \textit{Empirical QRM}~(EQRM) algorithm which relies on an empirical or kernel density estimate of the risk CDF based on the training domains to approximately solve~\eqref{eq:prob_gen}. 
 Importantly, this approach avoids having to reason explicitly about $Q$: instead, we only care about its push-forward through the risk functional. 
 We establish a generalisation bound for EQRM, which states that, given sufficiently many domains and samples, the empirical $\alpha$-quantile risk is a good estimate of the population $\alpha$-quantile risk.
Moreover, we show that as $\alpha \to 1$, EQRM learns a predictor with minimal, invariant risk over domains which leads to recovering the causal predictor $f(\PA_Y)$ under weaker assumptions than prior work~\citep{peters2016causal,krueger21rex}.
Empirically, we find EQRM to mostly outperform state-of-the-art baselines on typical DG benchmarks.

\begin{figure}[tb]
    \centering
    \begin{subfigure}[b]{0.275\linewidth}
        \centering
        \includegraphics[width=0.9\linewidth]{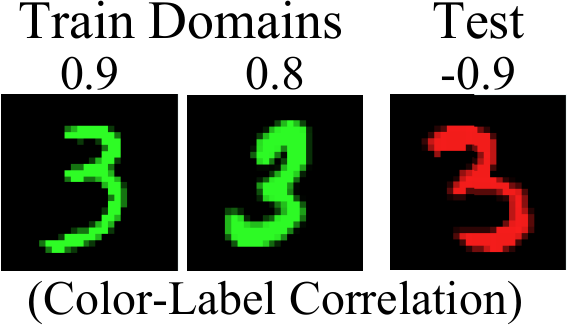}\vspace{5mm}
        \caption{}
        \label{fig:motiv-example:dataset}
    \end{subfigure}\hfill
    \begin{subfigure}[b]{0.4\linewidth}
        \centering
        \includegraphics[width=0.95\linewidth]{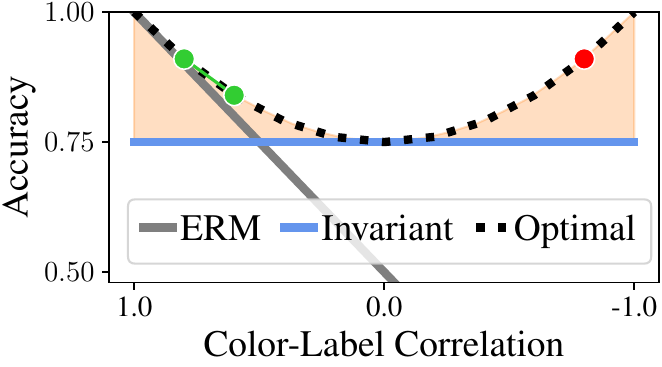}
        \caption{}
        \label{fig:motiv-example:ideal-results}
    \end{subfigure}\hfill
    \begin{subfigure}[b]{0.275\linewidth}
        \centering
        \includegraphics[width=0.6\linewidth]{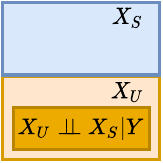}\vspace{5mm}
        \caption{}
        \label{fig:motiv-example:mutual_info}
    \end{subfigure}%
    \caption[Invariant (Stable) and Spurious (Unstable) Features]{\textbf{Invariant (Stable) and Spurious (Unstable) Features.} (a) Representative images from three different domains in the \text{CMNIST} dataset.
    (b) Idealised accuracies (y-axis) for different methods over test domains of varying colour-label correlation (x-axis). The green and red dots indicate the domains from (a). The dotted line corresponds to an oracle that has access to test-domain labels to use both the stable shape feature \textit{and} the unstable colour feature optimally in the test domain, thereby boosting performance (orange region) over an invariant model (blue line).
    (c) Generally, invariant models use only the \textit{stable} component $\Xb_S$ of $\Xb$, discarding the spurious or \textit{unstable} component $\Xb_U$. We show that predictions based on~$\Xb_S$ can be used as pseudo-labels to safely harness a sub-component of $\Xb_U$ (dark-orange region) and thus partly close the gap to the oracle behaviour from (b) even \textit{without test-domain labels}.}
    \label{fig:spuriosity_didnt_kill}
\end{figure}

\paragraph{Stable Feature Boosting: Harnessing Spurious Features via Invariant Predictions.}
The features $\Xb=(\Xb_S,\Xb_U)$ available in DG tasks consist of invariant or \textit{stable} features~$\Xb_S$ (such as the direct causes $\PA_Y$ of the target) and spurious or \textit{unstable} features~$\Xb_U$, whose relationship with the target may change in new domains.
As an illustrative example, consider the \textit{CMNIST} dataset~\citep{arjovsky2019invariant} containing images of coloured handwritten characters as shown in~\cref{fig:motiv-example:dataset}. The task is to classify the digit shape ($0$--$4$ vs.\ $5$--$9$, with $25$\% label noise) and domains correspond to different correlations between the label and digit colour (red or green).
As illustrated in~\cref{fig:motiv-example:ideal-results}, average-case methods like ERM use any features that aid average performance across the training domains, including the unstable colour feature, 
leading to a dramatic drop in performance in test domains where the colour-label correlation is reversed.
Worst-case methods like IRM instead seek to discard unstable features like colour and use only stable features ($\Xb_S$) such as shape, leading to an invariant model that achieves $75$\% across all domains. 
QRM (not shown) would interpolate between the two approaches, with the amount of colour used controlled by~$\alpha$.
However, all these approaches are suboptimal for test domains where the colour-label correlation is strong, but flipped w.r.t.\ the training domains.
Despite their instability, spurious features often provide \textit{complementary} information that could boost performance substantially (dotted line in~\cref{fig:motiv-example:ideal-results}) if a predictor could \textit{re-learn how to use these features in the right way}, rather than discard them. 
The main problem is that we do not have test-domain labels that could guide a correct use of these features. 

When and how can informative but unstable features be safely harnessed \textit{without labels} $Y$?
Our main idea is to use invariant predictions (obtained, e.g., by IRM) as \textit{pseudo-labels} $\hat{Y}$ to train domain-specific unstable predictors that can boost the stable ones.
However, it is a priori unclear whether this can be expected to work or how exactly the unstable models should be trained and combined with the stable one.  
We show that \textit{complementary} unstable features $\Xb_U$ which are conditionally independent of $\Xb_S$ given $Y$ (see~\cref{fig:motiv-example:mutual_info}) can indeed be leveraged with pseudo-labels as long as $\Xb_S$ is informative of $Y$: in this case, $P(Y~|~\Xb_U$) can  be recovered from $P(\hat Y~|~\Xb_U)$ through a bias correction, and $P(Y~|~\Xb_S,\Xb_U)$ can be obtained by combining the stable and unstable predictions in logit-space.
Based on these insights, we propose the Stable Feature Boosting (SFB) algorithm to extract and separate stable and conditionally-independent unstable features from $\Xb$, and then use stable predictions to adapt the unstable classifier for each test domain.
Empirically, we find SFB to be highly effective on real and synthetic data, where it approaches oracle performance as was the initial goal.

Overall, taking a causal perspective on multi-domain data appears to also be useful for supervised settings in which a target or label $Y$ is observed alongside $\Xb$. While prior work has sought to learn stable models through invariant causal prediction, we have proposed: QRM for interpolating between stable and unstable-but-informative models 
(when shifts are not expected to be adversarial, or perfect robustness is not required), and SFB for leveraging stable models to make safe use of unstable-but-informative features.

\section{Summary}
The world is full of domain shifts and different environments. 
Often, we cannot pin down what exactly differs between two domains, but  it may reasonably be modelled as a change in some causal mechanisms.
Our main results of this chapter demonstrate that such interventional, multi-environment data can, under the right conditions, be sufficient for identifying causal representations---even in the nonparametric case in which the spaces of mixing functions and mechanisms are infinite dimensional.
This highlights non-identically distributed, heterogeneous data as a promising learning signal for nonparametric CRL, which is particularly relevant since counterfactual, multi-view data as considered in~\cref{chap:SSL_content_style} is typically hard or impossible to obtain, and imposing constraints on the model class as explored in~\cref{chap:IMA} is often undesirable.
In this sense, our findings could also be considered a step towards justifying the use of expressive ML methods for learning interpretable causal representations from high-dimensional experimental data in situations where parametric assumptions are undesirable, for example, for complex systems in physics, biology, or medicine.
At the same time, the requirement for single-node, perfect interventions is highly idealised and not realistic for most settings of interest. 
While we have shown that these restrictions can be relaxed if the causal variables are directly observed or the graph is known, doing so in a full CRL setting in which both are unknown appears to be fundamentally hard, and further work is needed to try to exploit additional learning signals such as meta-data or task information.

%% file: Chapter9/chapter9.tex
 \graphicspath{{Chapter9/Figs/}}
\chapter{Conclusion and Future Directions}
\label{chap:conclusion}
In this thesis, we have investigated the question of when causal representations can provably be learnt.
Such identifiable CRL is an extremely challenging task, even when some parts of the problem are fixed (as in~\cref{chap:IMA}, where the graph is mostly assumed to be trivial) or only parts of the generative process are of interest (as in~\cref{chap:SSL_content_style}, which focuses mostly on recovering non-descendants of style variables). 
At the same time, we have found that \textit{there are several different routes to identifiability in CRL}---provided that the right assumptions are met or the right type of data is available. 
The settings studied in the main chapters of this thesis cover all three rungs of the ladder of causation~(see~\cref{tab:structure}): 
in~\cref{chap:IMA}, we have considered purely observational i.i.d.\ data, whereas in~\cref{chap:SSL_content_style,chap:CRL} we have linked violations of the i.i.d.\ assumption to multiple counterfactual views or interventional environments, respectively.

Throughout, a common theme has been to avoid overly restrictive assumptions like linearity of the mixing or causal relations, which is an important simplification leveraged in some existing identifiability studies~\citep{comon1994independent,squires2023linear,varici2023score,buchholz2023learning}.
Our progress in~\cref{chap:IMA} relies on constraining the mixing function while still allowing for nonlinearity, whereas both the mixing and causal mechanisms are unrestricted in~\cref{chap:SSL_content_style,chap:CRL}. 
This complicates the analysis, but may also make the considered problem settings and resulting insights more relevant to modern ML practice, where the encountered data is typically complex and the employed estimation methods highly expressive.

We also note that, while some parts of~\cref{chap:IMA,chap:SSL_content_style} could, in principle, also be presented without \textit{explicitly} mentioning causality, the key insights presented therein were arrived at by thinking about the respective data generating processes in causal terms (independent influences inspired by the ICM principle in~\cref{chap:IMA} and multiple augmented views as counterfactuals in~\cref{chap:SSL_content_style}). 
In this sense, CRL may generally offer a useful perspective for formalising representation learning problems and coming up with suitable assumptions to address them, even if the resulting statistical footprints are ultimately the main objects of interest.

Beyond our own related works on unsupervised~(\cref{sec:extensions_ima}), multi-view~(\cref{sec:multiview_extensions}), and multi-environment CRL~(\cref{sec:extensions_multi_env}), several other works have extended or built upon the main contributions presented in this thesis in meaningful ways.
Some examples we are aware of include the following:
\begin{itemize}
\item
In the context of the IMA constraint developed in~\cref{chap:IMA}, \citet{buchholz2022function} have established full identifiability for the subclass of conformal maps, and local identifiability for the IMA function class, thus generalizing our results from~\cref{sec:ima_theory} that only rule out certain spurious solutions. 
Other works have extended IMA to higher-dimensional observations~\citep{ghosh2023independent}, empirically probed its robustness to violations of the IMA principle~\citep{sliwa2022probing}, and explored a related objective for uncovering the structure of data manifolds~\citep{cunningham2022principal}. 
\item 
In the context of the counterfactual multi-view setting studied in~\cref{chap:SSL_content_style}, \citet{brehmer2022weakly} have proven a comprehensive identifiability result showing that all causal latents (not just content) and their graph can be recovered if pairs of pre- and post-intervention views arise from a mixture of all single-node perfect interventions (rather than from imperfect style interventions), similar to the interventional CRL studies discussed in~\cref{chap:CRL}.
Our content identifiability results have been extended to a multi-modal setting with view-specific mixing functions and private latent components~\citep{daunhawer2023identifiability}, including identifiability of independent style latents~\citep{lyu2021understanding}, and the notion of identifying blocks rather than individual latents has also been widely used.
\item 
In the context of the nonparametric multi-environment CRL setting studied in~\cref{chap:CRL}, \citet{varici2023general} have proposed a score-based estimation procedure and extended our identifiability result from two perfect interventions per node~(\cref{thm:general}) by relaxing the requirements of paired environments and faithfulness, provided that we instead have access to an observational environment and exactly $2n$ interventional ones,  structured into two sets of size $n$ with one intervention per node in each set.
\end{itemize}

\looseness-1 We hope that the insights presented in this thesis can help further advance the field of \textit{identifiable CRL} towards more realistic settings.
The multi-environment setting studied in~\cref{chap:CRL} is perhaps particularly well-suited for this goal. As we have shown, access to multiple interventional domains can help overcome some of the fundamental challenges faced when learning from i.i.d.\ data~(\cref{chap:IMA}), while at the same time being less restrictive than the counterfactual data that is typically required in multi-view settings~(\cref{chap:SSL_content_style}).
To this end, recent work has made some progress in relaxing the arguably unrealistic requirement of access to all single-node, perfect interventions, instead learning from imperfect single~\citep{zhang2023identifiability} or multi-node~\citep{ahuja2023multi} interventions
by imposing restrictions on the latent support~\citep{wang2021desiderata,roth2023disentanglement} and considering fixed-degree polynomials as mixings~\citep{ahuja2022interventional}.
The IMA function class may constitute another useful case of constrained but nonlinear functions. 
Future work should also focus on developing more \textit{practical CRL algorithms}, for example, by drawing on developments in other areas of ML. 
\textit{Self-supervised} approaches have shown great promise for learning representations of images or text, and could thus become an important ingredient for new CRL methods as discussed in~\cref{chap:SSL_content_style}. 
In terms of learning \textit{discrete} causal concepts, recent work has established connections between CRL and classical literature on identifiable \textit{mixture models}~\citep{kivva2021learning}.
The field would likely also benefit from a tighter integration with \textit{Bayesian or probabilistic ML}, for example, by learning a distribution over candidate causal graphs~\citep{deleu2022bayesian,lorch2021dibs} that takes into account different sources of uncertainty, including both a lack of data and potential non-identifiability issues.
Such posterior beliefs may in turn be leveraged for \textit{active learning} rooted in optimal Bayesian experimental design~\citep{toth2022active}. 
A related promising direction is the combination of CRL with \textit{agent-based frameworks} such as reinforcement learning~\citep{lippe2023biscuit,cohen2022towards}: while actions have a natural interpretation as interventions, the exact
mapping of which and how available actions influence different subsets of the latent causal variables is typically unknown and also needs to be included in the model~\citep{lachapelle2022disentanglement}.

Finally, we believe that further work---both conceptual and technical---is needed to make CRL more useful for practical applications, such as  learning causal world models or gaining new scientific insights.
Regarding the latter, using a learnt causal representation seems to at least involve: linking the extracted high-level latent causal variables (numbers on a computer) to patterns in the low-level observations (e.g., voxels in an MRI scan), and then matching those to meaningful real-world objects or concepts (e.g., a brain activity pattern). 
The former could be achieved through an integration of CRL methods with explainability tools, and the latter through working together with domain experts. 
The inferred qualitative and quantitative high-level causal influences could then yield new insights about the underlying mechanisms.

%% file: Appendix1/appendix1.tex
\chapter{Proofs} 
This appendix contains the proofs of theoretical results from the main chapters.

\section{Proofs for~\texorpdfstring{\cref{chap:IMA}}{}}
\label{app:proofs_IMA}

\subsection{Proof of\texorpdfstring{~\Cref{prop:local_IMA_contrast_properties}}{}}
Before giving the proof, it is useful to rewrite the local IMA constrast~\eqref{eq:adm_single_point} as follows:
\begin{align}
    c_\IMA(\fb,\sb) =& \sum_{i = 1}^n \log \norm{\frac{\partial \fb}{\partial s_i}(\sb)} - \log \left|\Jb_\fb(\sb)\right| \nonumber\\
    =& \frac{1}{2} \left( \log \left|\diag \left(\Jb^\top_\fb(\sb) \Jb_\fb(\sb)\right) \right| - \log \left|\Jb^\top_\fb(\sb) \Jb_\fb(\sb) \right| \right) \nonumber \\
     =& \frac{1}{2} D_{KL}^{\text{left}}\left(\Jb^\top_\fb(\sb) \Jb_\fb(\sb)\right)\,,
    \label{eq:dkl_diag}
\end{align}
where the quantity in~\eqref{eq:dkl_diag} is called the left KL measure of diagonality of the matrix $\Jb^\top_\fb(\sb) \Jb_\fb(\sb)$~\citep{alyani2017diagonality} (see~\cref{remark:left_KL}):
\begin{align*}
    D_{KL}^{\text{left}}(\Ab) &= - \log | (\diag(\Ab))^{-\frac{1}{2}} \Ab (\diag(\Ab))^{-\frac{1}{2}} |\\
                              &= \log |\diag(\Ab)| - \log |\Ab|\,.
\end{align*}
From~\eqref{eq:dkl_diag}, it can be seen that $c_\IMA(\fb,\sb)$ is a function of $\Jb_\fb(\sb)$ only through $\Jb^\top_\fb(\sb) \Jb_\fb(\sb)$. 

\impropties*
\begin{proof} 
For ease of exposition, we denote the value of the Jacobian of $\fb$ evaluated at the point $\sb$ by $\Jb_\fb(\sb) = \Wb$.
The two properties can then be proved as follows:
\begin{enumerate}[(i)] 
        \item This is a consequence of Hadamard's inequality, applied to the expression on the RHS of~\eqref{eq:adm_single_point}, which states that, for a matrix $\Wb$ with columns $\wb_i$, $\sum_{i = 1}^n \log \norm{\wb_i} \geq \log |\Wb|$; equality in Hadamard's inequality is achieved iff. the vectors $\wb_i$ are orthogonal.
        \item We split the proof in three parts.
        \begin{enumerate}[a.]
            \item \textit{Invariance to left multiplication by an orthogonal matrix:}\\
            Let $\Wbt = \Ob \Wb$, with $\Ob$ an orthogonal matrix, i.e., $\Ob\Ob^\top=\Ib$.
            Then the property follows from  writing $c_\IMA(\fb,\sb)$ as in~\eqref{eq:dkl_diag}:
            \begin{align*}
                \frac{1}{2}\leftkl(\Wbt^\top \Wbt) =\frac{1}{2} \leftkl(\Wb^\top \Ob^\top  \Ob \Wb) 
                =\frac{1}{2} \leftkl(\Wb^\top \Ib \Wb) 
                = \frac{1}{2}\leftkl(\Wb^\top \Wb) %
            \end{align*}
            \item
            \textit{Invariance to right multiplication by a permutation matrix:}\\
            Let $\Wbt = \Wb \Pb$, with $\Pb$ a permutation matrix. Then $\Wbt$ is just $\Wb$ with permuted columns. Clearly, the sum of the log-column-norms does not change by changing the order of the summands.
            Further, $\log|\tilde{\Wb}| = \log|\Wb| + \log|\Pb| =\log|\Wb|$, because the absolute value of the determinant of a permutation matrix is one. 
            \item \textit{Invariance to right multiplication by a diagonal matrix}:\\ Let $\Wbt = \Wb \Db$, with $\Db$ a diagonal matrix. Consider the two terms on the RHS of~\eqref{eq:adm_single_point}.
            For the first term, we know that the columns of~$\tilde{\Wb}$ are scaled versions of the columns of $\Wb$, that is $\tilde{\wb}_i = d_i \wb_i$, where $d_i$ denotes the $i^\text{th}$ diagonal element of~$\Db$.
            Then $\norm{\tilde{\wb}_i} = |d_i|\norm{\wb_i}$. 
            For the second term, we use the decomposition of the determinant: \[\log|\tilde{\Wb}| = \log|\Wb| + \log|\Db| = \log|\Wb| + \sum_{i=1}^n\log|d_i|.\]
            Taken together, we obtain
            \begin{align*}
            \sum_{i = 1}^n \log \norm{\wbt_i} - \log |\Wbt| 
            &= \sum_{i = 1}^n \log \left(|d_i| \norm{\wb_i}\right) -\left( \log |\Wb| + \sum_{i=1}^n\log|d_i|\right)\\
            &= \sum_{i = 1}^n \log \norm{\wb_i} + \sum_{i = 1}^n \log |d_i|  - \log |\Wb| - \sum_{i=1}^n\log|d_i|\\
            &= \sum_{i = 1}^n \log \norm{\wb_i}   - \log |\Wb|%
            \end{align*}%
        \end{enumerate}%
    \end{enumerate}%
\end{proof}

\subsection{Proof of\texorpdfstring{~\Cref{prop:global_IMA_contrast_properties}}{}}
\admproperties*
\begin{proof}
 The properties can be proved as follows:
    \begin{enumerate}[(i)]
        \item From property \textit{(i)} of~\cref{prop:local_IMA_contrast_properties}, we know that $c_\IMA(\fb, \sbb) \geq 0$. Hence, $C_\IMA(\fb, p(\sbb)) \geq 0$ follows as a direct consequence of integrating the non-negative quantity $c_\IMA(\fb, \sbb)$.
        
        Equality is attained iff.\ $c_\IMA(\fb, \sbb)=0$ almost surely w.r.t.\ $p_\sb$, which according to property \textit{(i)} of~\cref{prop:local_IMA_contrast_properties} occurs 
        iff.\  the columns of $\Jb_{\fb}(\sb)$ are orthogonal almost surely w.r.t.\ $p_\sb$.
        
        It remains to show that this is the case iff.\ $\Jb_{\fb}(\sb)$ can be written as $\Ob(\sb) \Db(\sb)$, with $\Ob(\sb)$ and $\Db(\sb)$  orthogonal and diagonal matrices, respectively.
        (To avoid confusion, note that \textit{orthogonal columns} need not have unit norm, whereas an \textit{orthogonal matrix} $\Ob$ satisfies $\Ob\Ob^\top=\Ib$.)
        
        The \textit{if} is clear since right multiplication by a diagonal matrix merely re-scales the columns, and hence does not affect their orthogonality. 
        
        For the \textit{only if}, let $\Jb_\fb(\sb)$ be any matrix with orthogonal columns $\jb_i(\sb)$, $\jb_i(\sb)^\top \jb_j(\sb)=0, \forall i\neq j$, and denote the column norms by $d_i(\sb)=||\jb_i(\sb)||$. Further denote the normalised columns of $\Jb_\fb(\sb)$ by $\ob_i(\sb)=\jb_i(\sb) / d_i(\sb)$ and let $\Ob(\sb)$ and $\Db(\sb)$ be the orthogonal and diagonal matrices with columns $\ob_i(\sb)$ and diagonal elements $d_i(\sb)$, respectively. Then $\Jb_\fb(\sb)=\Ob(\sb)\Db(\sb)$.
        \item Let $\fbt=\fb\circ\hb^{-1}\circ \Pb^{-1}$ and $\sbt = \Pb\hb(\sb)$,  where $\Pb\in\RR^{n\times n}$ is a permutation matrix and $\hb(\sb)=(h_1(s_1), ..., h_n(s_n))$ is an invertible element-wise function.
        Then
        \begin{equation}
        \label{eq:equality_of_global_contrasts}
            C_\IMA(\fbt,p_\sbt)
            =\int c_\IMA(\fbt,\sbt)p_\sbt(\sbt) d\sbt
            =\int c_\IMA(\fbt,\sbt)p_\sb(\sb) d\sb
        \end{equation}
        where, for the second equality, we have used the fact that 
        \[p_\sbt(\sbt)d\sbt = p_\sb(\sb)d\sb\,.\]
        since $\Pb \circ \hb$ is an invertible tranformation (see, e.g., ~\citet{rezendeshort}).
        It thus suffices to show that
        \begin{equation}
        \label{eq:equality_of_local_contrasts}            c_\IMA(\fbt,\sbt)=c_\IMA(\fb,\sb).
        \end{equation}
        at any point $\sbt=\Pb\hb(\sb)$.
        To show this, we write
        \begin{align}
            \Jb_\fbt(\sbt)
            &=\Jb_{\fb\circ\hb^{-1}\circ\Pb^{-1}}(\Pb\hb(\sb)) \nonumber \\
            &=\Jb_{\fb\circ\hb^{-1}}\left(\Pb^{-1}\Pb\hb(\sb)\right)\, \Jb_{\Pb^{-1}}\left(\Pb\hb(\sb)\right) \nonumber\\
            &=\Jb_{\fb\circ\hb^{-1}}(\hb(\sb))\, \Jb_{\Pb^{-1}}(\Pb\hb(\sb)) \nonumber\\
            &=\Jb_{\fb}(\hb^{-1}\circ\hb(\sb))\, \Jb_{\hb^{-1}}(\hb(\sb)) \, \Jb_{\Pb^{-1}}(\Pb\hb(\sb)) \nonumber\\
            &=\Jb_{\fb}(\sb) \, \Db(\sb) \Pb^{-1}
            \label{eq:Jacobian_relation}
        \end{align}
        where we have repeatedly used the chain rule for Jacobians, as well as that $\Pb^{-1}\Pb=\Ib$; that permutation is a linear operation, so $\Jb_\Pb(\sb)=\Pb$ for any $\sb$; and that $\hb$ (and thus $\hb^{-1}$) is an element-wise transformation, so the Jacobian $\Jb_{\hb^{-1}}$ is a diagonal matrix $\Db(\sb)$.
        
        The equality in~\eqref{eq:equality_of_local_contrasts} then follows from~\eqref{eq:Jacobian_relation} by applying property \textit{(ii)} of~\cref{prop:local_IMA_contrast_properties}, according to which $c_\IMA$ is invariant to right multiplication of the Jacobian $\Jb_\fb(\sb)$ by diagonal and permutation matrices.
        
        Substituting~\eqref{eq:equality_of_local_contrasts} into the RHS of~\eqref{eq:equality_of_global_contrasts}, we finally obtain
        \begin{equation*}
            C_\IMA(\fbt,p_\sbt)=C_\IMA(\fb,p_\sb).
        \end{equation*}%
        \end{enumerate}%
\end{proof}

\subsection{Proof of\texorpdfstring{~\Cref{thm:adm_darmois}}{}}
Before proving the main theorem, we first introduce some additional details on the Jacobian of the Darmois construction~\citep{hyvarinen1999nonlinear} which will be important for the proof.

\paragraph{Jacobian of the Darmois construction for $n=2$.}
\label{app:Jacobian_Darmois}
Consider the Darmois construction for $n=2$, 
\begin{align*}
     y_1&=g^{\text{D}}_1(x_1):=F_{X_1}(x_1)=\PP_{X_1}(X_1\leq x_1)\\
     y_2&=g^{\text{D}}_2(y_1, x_2):=F_{X_2|Y_1=y_1}(x_2)=\PP_{X_2|Y_1=y_1}(X_2\leq x_2|Y_1=y_1)
\end{align*}

Its Jacobian takes the form
\begin{equation}
\label{eq:general_Jacobian_Darmois_2d}
    \Jb_{\gb^\text{D}} (\xb) =
    \begin{pmatrix}
    p(x_1) & 0\\
    c_{21}(\xb) & p(x_2|x_1)
    \end{pmatrix}\,,
\end{equation}
where 
\[
c_{21}(\xb)=\frac{\partial }{\partial x_1} \int_{-\infty}^{x_2} p(x'_2|x_1)dx'_2\,.
\]

\paragraph{Jacobian of the Darmois construction: general case.}
In the general case, the Jacobian of the Darmois construction will be 
\begin{equation}
\Jb_{\gb^\text{D}}(\xb) = 
\begin{pmatrix} 
p(x_1) & \cdots & 0 \\
 & \ddots & \vdots \\ 
\mathbf{C}(\xb) &  & p(x_n |x_1, \ldots, x_{n-1} )
\end{pmatrix} 
\label{eq:full_jacobian_darmois}
\end{equation}
where the components $c_{ji}(\xb_{1:j})$ of $\mathbf{C}(\xb)$ for all $i<j$ %
are defined  by
\begin{align*}
c_{ji}(\xb_{1:j})&=\frac{\partial }{\partial x_i} \int_{-\infty}^{x_j} p(x'_j| \xb_{1:j-1} ) dx'_j \,.
\end{align*}

It is additionally useful to introduce the following lemmas.

\begin{restatable}{lemma}{trijaccima}
\label{lemma:trijaccima}
A function $\fb$ with triangular Jacobian has $C_\IMA(\fb, p_\sb) = 0$ iff. its Jacobian is diagonal almost everywhere. Otherwise,  $C_\IMA(\fb, p_\sb) > 0$.%
\end{restatable}%
\begin{proof}
    Let $\fb$ have lower triangular Jacobian at $\sb$, and denote $\Jb_\fb(\sb) = \Wb$. Then we have 
    \[
    c_\IMA(\fb,\sb) = \sum_{i = 1}^n \log \left( \sqrt{\sum_{j=i}^n w^2_{ji} } \right) - \sum_{i = 1}^n \log \left| w_{ii} \right|\,,
    \]
    where $w_{ji} = [\Wb]_{ji} $. Since the logarithm is a strictly monotonically increasing function and since \[\sqrt{\sum_{j=1}^n w^2_{ji} } \geq | w_{ii} |\,,\] with equality iff.\ $w_{ji}=0, \forall j \neq i$ (i.e., iff.\ $\Wb$ is a diagonal matrix), we must have $c_\IMA(\fb,\sb)=0$ iff.\ $\Wb$ is diagonal. 
    
    $C_\IMA(\fb,p_\sb)$ is therefore equal to zero iff. $\fb$ has diagonal Jacobian almost everywhere, and it is strictly larger than zero otherwise. 
\end{proof}

\begin{restatable}{lemma}{elemwise}
\label{lemma:elemwise}
A smooth function $\fb:\RR^n \rightarrow \RR^n$ whose Jacobian is diagonal everywhere is an element-wise function, $\fb(\sb) = (f_1(s_1), ..., f_n(s_n))$.
\end{restatable}%
\begin{proof}
Let $\fb$ be a smooth function with diagonal Jacobian everywhere.

Consider the function $f_i(\sb)$ for any $i\in\{1, ..., n\}$. Suppose \textit{for a contradiction} that $f_i$ depends on $s_j$ for some $j\neq i$. Then there must be at least one point $\sb^*$ such that $\nicefrac{\partial f_i}{\partial s_j} (\sb^*)\neq 0$.
However, this contradicts the assumption that $\Jb_\fb$ is diagonal everywhere (since $\nicefrac{\partial f_i}{\partial s_j}$ is an off-diagonal element for $i\neq j$).
Hence, $f_i$ can only depend on $s_i$ for all $i$, i.e., $\fb$ is an element wise function.
\end{proof}%

We can now restate and prove~\Cref{thm:adm_darmois}.

\admdarmois*

\begin{proof}
First, the Jacobian $\Jb_{\gb^\text{D}}(\xb)$ of the Darmois construction $\gb^\text{D}$ is lower triangular $\forall\xb$, see~\eqref{eq:full_jacobian_darmois}.

Because CDFs are monotonic functions (strictly monotonically increasing given our assumptions on $\fb$ and $p_\sb$), $\gb^\text{D}$ is invertible.

We can thus apply the inverse function theorem (with $\fb^\text{D}=(\gb^\text{D})^{-1}$) to write
\[
\Jb_{\fb^\text{D}}(\yb)=\left(\Jb_{\gb^\text{D}}(\xb)\right)^{-1}
\]
Since the inverse of a lower triangular matrix is lower triangular, we conclude that $\Jb_{\fb^\text{D}}(\yb)$ is lower triangular for all $\yb=\gb^\text{D}(\xb)$.

Now, according to~\Cref{lemma:trijaccima}, we have
$C_\IMA(\fb^\text{D}, p_{\ub}) > 0$, unless $\Jb_{\fb^\text{D}}$ is diagonal almost everywhere.

Suppose \textit{for a contradiction} that $\Jb_{\fb^\text{D}}$ is diagonal almost everywhere.

Since $\fb$ and $p_\sb$ are smooth by assumption, so is the push-forward $p_\xb=\fb_*p_\sb$, and thus also $\gb^\text{D}$ (CDF of a smooth density) and its inverse $\fb^\text{D}$. 
Hence, the partial derivatives $\nicefrac{\partial f^\text{D}_i}{\partial y_j}$, i.e., the elements of $\Jb_{\fb^\text{D}}$ are continuous.

Consider an off-diagonal element $\nicefrac{\partial f^\text{D}_i}{\partial y_j}$ for $i\neq j$. Since these are zero almost everywhere, and because continuous functions which are zero almost everywhere must be zero everywhere, we conclude that $\nicefrac{\partial f^\text{D}_i}{\partial y_j}=0$ everywhere for $i\neq j$, i.e., the Jacobian $\Jb_{\fb^\text{D}}$ is \textit{diagonal everywhere}. 

Hence, we conclude from~\cref{lemma:elemwise} that $\fb^\text{D}$ must be an element-wise function, $\fb^\text{D}(\yb)=(f^\text{D}_1(y_1), ..., f^\text{D}_1(y_n))$.

Since $\yb$ has independent components by construction, it follows that $x_i=f^\text{D}_i(y_i)$ and $x_j=f^\text{D}_j(y_j)$ are independent for any $i\neq j$. 

However, this constitutes a contradiction to the assumption that $x_i \not\independent x_j$ for some $x_j$. 

We conclude that $\Jb_{\fb^\text{D}}$ cannot be diagonal almost everywhere, and hence, by~\cref{lemma:trijaccima}, we must have $C_\IMA(\fb^\text{D},p_\ub)>0$.
\end{proof}

\subsection{Proof of\texorpdfstring{~\Cref{cor:IMA_identifiability_of_conformal_maps}}{}}
\confmapsadm*
\begin{proof}
The proof follows from property \textit{(i)} of~\Cref{prop:global_IMA_contrast_properties}: by definition, the Jacobian of conformal maps at any point $\sb$ can be written as $\Ob(\sb) \lambda(\sb)$, with $\lambda: \RR^n \rightarrow \RR$, which is a special case of $\Ob(\sb) \Db(\sb)$, with $\Db(\sb) = \lambda(\sb) \Ib $.
\end{proof}

\subsection{Proof of\texorpdfstring{~\Cref{cor:IMA_identifiability_of_linear_ICA}}{}}
\admidentlinear*
\begin{proof}
Since, by assumption, the mixing matrix is non-trivial (i.e., not the product of a diagonal and permutation matrix), and  at most one of the $s_i$ is Gaussian, according to the identifiability of linear ICA~\citep[][Thm.~11]{comon1994independent} there must be at least one pair $x_i, x_j$, with $i \neq j$, such that $x_i \nindep x_j$.

We can then use the same argument as in the proof of~\cref{thm:adm_darmois}
to show that the Darmois construction has nonzero $C_\IMA$, whereas the linear orthogonal transformation $\Ab$ has orthogonal Jacobian, and thus $C_\IMA=0$ by property \textit{(i)} of~\cref{prop:global_IMA_contrast_properties}.
\end{proof}

\subsection{Proof of\texorpdfstring{~\Cref{thm:IMA_identifiability_measure_preserving_automorphism}}{}}
\thmMPA*
\begin{proof}
Recall the definition 
\[
 \ab^{\Rb}(p_\sb) =\Fb_\sb^{-1} \circ \bm\Phi \circ \Rb \circ \bm\Phi^{-1} \circ \Fb_\sb.
\]
For notational convenience, we denote $\sib = \bm\Phi^{-1} \circ \Fb_\sb$
and write
\[
 \ab^{\Rb}(p_\sb) =\sib^{-1} \circ \Rb \circ \sib.
\]
Note that, since both $\Fb_\sb$ and $\bm\Phi$ are element-wise transformations, so is $\sib$. 

First, by using property \textit{(ii)} of~\cref{prop:global_IMA_contrast_properties} (invariance of $C_\IMA$ to element-wise transformation), we obtain
\[
    C_\IMA(\fb\circ  \ab^{\Rb}(p_\sb), p_{\sb})
    =C_\IMA(\fb\circ \sib^{-1} \circ \Rb \circ \sib, p_{\sb})
    = C_\IMA(\fb\circ \sib^{-1} \circ \Rb,p_\zb)\,,
\]
with $\zb=\sib(\sb)$ such that $p_\zb$ is an isotropic Gaussian distribution.

Suppose \textit{for a contradiction} that $C_\IMA(\fb\circ \sib^{-1} \circ \Rb,p_\zb)=0$.

According to property \textit{(i)} of~\cref{prop:global_IMA_contrast_properties},
this entails that the matrix
\begin{equation}\label{eq:diagMP}
\Jb_{\fb\circ \sib^{-1} \circ \Rb}(\zb)^\top\Jb_{\fb\circ \sib^{-1} \circ \Rb}(\zb)=
\Rb^\top 
\, \Jb_{\sib^{-1}}(\zb)^\top 
\, \Jb_\fb(\sib^{-1}(\zb))^\top  
\, \Jb_\fb(\sib^{-1}(\zb)) 
\, \Jb_{\sib^{-1}}(\zb)
\,  \Rb\,
\end{equation}
is diagonal almost surely w.r.t.\ $p_\zb$.
Moreover, smoothness of $p_{\sb}$ and $\fb$ implies the matrix expression of~\eqref{eq:diagMP} is a continuous function of $\zb$. Thus \eqref{eq:diagMP} actually needs to be diagonal for all $\zb\in \RR^n$, i.e., \textit{everywhere} (c.f., the argument used in the proof of~\cref{thm:adm_darmois}).

Since $(\fb,p_\sb)\in\Mcal_\IMA$ by assumption, by property \textit{(i)} of~\cref{prop:global_IMA_contrast_properties}, the inner term on the RHS of~\eqref{eq:diagMP},
\[
\Jb_\fb( \sib^{-1}(\zb))^\top 
\,  
\Jb_\fb( \sib^{-1}(\zb)),
\]
is diagonal. 
Moreover, since $\sib$ is an element-wise transformation, 
$\Jb_{\sib^{-1}}(\zb)^\top$ and $\Jb_{\sib^{-1}}(\zb)$ 
are also diagonal. 
Taken together, this implies that
\begin{equation}
\label{eq:diagonal_inner_term}
\Jb_{\sib^{-1}}(\zb)  
\, 
\Jb_\fb( \sib^{-1}(\zb))^\top 
\,  
\Jb_\fb( \sib^{-1}(\zb))
\, 
\Jb_{\sib^{-1}}(\zb)    
\end{equation}
is diagonal (i.e., ~\eqref{eq:diagMP} is of the form $\Rb^\top \Db(\zb)\Rb$ for some diagonal matrix $\Db(\zb)$).

Without loss of generality, we assume the first component $s_1$ of $\sb$ is non-Gaussian and satisfies the assumptions stated relative to $\Rb$ (axis not invariant nor sent to another canonical axis).

Now, since both the Gaussian CDF $\bm \Phi$ and the CDF $\Fb_\sb$ are smooth (the latter by the assumption that $p_\sb$ is a smooth density), $\sib$ is a smooth function and thus has continuous partial derivatives.

By continuity of the partial derivative, the first diagonal element $\frac{\partial \sigma_1^{-1}}{\partial z_1}$ of $\Jb_{\sib^{-1}}$ must be strictly monotonic in a neighborhood of some $z_1^0$ (otherwise $\sigma_1$ would be an affine transformation, which would contradict non-Gaussianity of $s_1$).

On the other hand, our assumptions relative to $\Rb$ entail that there are at least two non-vanishing coefficients in the first row of $\Rb$ (i.e., first column of $\Rb^\top$).\footnote{In short, if this were not the case, this column would have a single non-vanishing coefficient, which would need to be one due to the unit norm of the rows of this orthogonal matrix. 
Such structure of the matrix $\Rb$ would entail that  the associated canonical basis vector $\eb_1$ is transformed by $\Rb^{-1}=\Rb^\top$ into a canonical basis vector $\eb_j$ which contradicts the assumptions.} 
Let us call $i\neq j$ such pair of coordinates, i.e., $r_{1j}\neq 0$ and $r_{1i}\neq 0$. 

Now consider the off-diagonal term $(i,j)$ of~\eqref{eq:diagMP}, which we assumed (for a contradiction) must be zero almost surely w.r.t.\ $p_\zb$.
Since the term in~\eqref{eq:diagonal_inner_term} is diagonal, this off-diagonal term is given by:
\[
\sum_{k=1}^n 
\left(\frac{d\sigma_k^{-1}}{dz_k}(z_k)\right)^2 \norm{\frac{\partial \fb}{ds_k}\circ \sib^{-1}(\zb)}^2 r_{ki} r_{kj}
=
\sum_{k=1}^n  \left(\frac{d\sigma_k^{-1}}{dz_k}(z_k)\right)^2 \lambda(\sib^{-1}(\zb))^2 r_{ki} r_{kj}=0\,.
\]
where for the first equality we have used the fact that $\fb$ is a conformal map with conformal factor~$\lambda(\sb)$ (by assumption), and 
where the second equality must hold
almost surely w.r.t.\ $p_\zb$.

Since $\fb$ is invertible, it has non vanishing Jacobian determinant. Hence, the conformal factor $\lambda$ must be a strictly positive function, so
\[
\lambda(\sib^{-1}(\zb))^2>0, \, \forall \zb.
\]
Thus, for almost all $\zb$, we must have:
\begin{equation}
\label{eq:contradiction_expression}
\sum_{k=1}^n  \left(\frac{d\sigma_k^{-1}}{dz_k}(z_k)\right)^2  r_{ki} r_{kj}=0\,.
\end{equation}
Now consider the first term $\left(\frac{d\sigma_1^{-1}}{dz_1}(z_1)\right)^2  r_{1i} r_{1j}$  in the sum.  

Recall that $r_{1i}r_{1j}\neq 0$,
and that $\frac{d\sigma_1^{-1}}{dz_1}(z_1)$
is strictly monotonic  on a neighborhood of $z_1^0$.

As  a consequence, $\left(\frac{d\sigma_1^{-1}}{dz_1}(z_1)\right)^2  r_{1i} r_{1j}$ is also strictly monotonic with respect to $z_1$  on a neighborhood of $z_1^0$ (where the  other variables $(z_{2},...,z_{n})$ are left  constant), while the other
terms in the sum in~\eqref{eq:contradiction_expression} are left constant because $\sib$ is an element-wise transformation.

This leads to a contradiction as~\eqref{eq:contradiction_expression} (which should be satisfied for all $\zb$) cannot stay constantly zero as $z_1$ varies within the neighbourhood of $z_1^0$. 

Hence our assumption that $C_\IMA(\fb\circ  \ab^{\Rb}(p_\sb), p_{\sb})=0$ cannot hold.

We conclude that $C_\IMA(\fb\circ  \ab^{\Rb}(p_\sb), p_{\sb})>0$.
\end{proof}

\section{Proofs for~\texorpdfstring{\cref{chap:SSL_content_style}}{}}
\label{app:proofs_SSL}
Here, we present the full detailed proofs of the three theorems which were briefly sketched in~\cref{chap:SSL_content_style}. We remark that these proofs build on each other, in the sense that the (main) step 2 of the proof of~\cref{thm:main} is also used  in the proofs of~\cref{thm:CL,thm:CL_MaxEnt}.

\subsection{Proof of\texorpdfstring{~\Cref{thm:main}}{thmmain}}
\label{app:proof_generative}

\generative*

\begin{proof}
The proof consists of two main steps.

In the first step, we use assumption \textit{(i)} and the matching likelihoods to show that the representation $\zbh=\gb(\xb)$ extracted by $\gb=\hat{\fb}^{-1}$ is related to the true latent $\zb$ by a smooth invertible mapping $\hb$, and that $\zbh$ must satisfy invariance across $(\xb,\xbt)$ in the first $n_c$ (content) components almost surely (a.s.) with respect to (w.r.t.) the true generative process.

In the second step, we then use assumptions \textit{(ii)} and \textit{(iii)}
to  prove (by contradiction) that $\cbh:=\zbh_{1:n_c}=\hb(\zb)_{1:n_c}$ can, in fact, only depend on the true content $\cb$ and not on the true style $\sb$, for otherwise the invariance established in the first step would have be violated with probability greater than zero.

To provide some further intuition for the second step, the assumed generative process implies that $(\cb,\sb,\sbt)|A$ is constrained to take values (a.s.) in a subspace $\Rcal$ of $\Ccal\times\Scal\times\Scal$ of dimension $n_c+n_s+|A|$ (as opposed to dimension $n_c+2n_s$ for $\Ccal\times\Scal\times\Scal$). In this context, assumption \textit{(iii)} implies that  $(\cb,\sb,\sbt)|A$ has a density with respect to a measure on this subspace equivalent to the Lebesgue measure on $\RR^{n_c+n_s+|A|}$. This equivalence implies, in particular, that this ``subspace measure'' is strictly positive: it takes strictly positive values on open sets of $\Rcal$ seen as a topological subspace of $\Ccal\times\Scal\times\Scal$. These open sets are defined by the induced topology: they are the intersection of the open sets of $\Ccal\times\Scal\times\Scal$ with $\Rcal$. An open set $B$ of $V$ on which $p(\cb,\sb,\sbt|A) >0$ then satisfies $P(B|A)>0$. We look for such an open set to prove our result.

\paragraph{Step 1.}

From the assumed data generating process described in~\cref{sec:problem_formulation}---in particular, from the form of the model conditional $\hat{p}_{\zbt|\zb}$ described in~\cref{ass:content_invariance,ass:style_changes}---it follows that
\begin{align}
    \label{eq:invariance_constraint}
    \gb(\xb)_{1:n_c} &= \gb(\xbt)_{1:n_c}
\end{align}
a.s., i.e., with probability one, w.r.t.\ the model distribution $\hat{p}_{\xb,\xbt}$.

Due to the assumption of matching likelihoods, the invariance in~\eqref{eq:invariance_constraint} must also hold (a.s.) w.r.t.\ the true data distribution $p_{\xb,\xbt}$.

Next, since $\fb,\hat{\fb}:\Zcal\rightarrow\Xcal$ are smooth and invertible functions by assumption \textit{(i)}, there exists
a  smooth and invertible function $\hb=\gb\circ \fb:\Zcal\rightarrow\Zcal$ such that
\begin{equation}
\label{eq:relation_between_f_and_g}
\gb=\hb\circ\fbinv.
\end{equation}

Substituting~\eqref{eq:relation_between_f_and_g} into~\eqref{eq:invariance_constraint}, we obtain (a.s.\ w.r.t.\ $p$):
\begin{equation}
\label{eq:cbh}
\cbh
:=\zbh_{1:n_c}=\gb(\xb)_{1:n_c}
=\hb(\fbinv(\xb))_{1:n_c}
=\hb(\fbinv(\xbt))_{1:n_c}
\end{equation}

Substituting $\zb=\fbinv(\xb)$ and $\zbt=\fbinv(\xbt)$ into~\eqref{eq:cbh}, we obtain (a.s.\ w.r.t.\ $p$)
\begin{equation}
\label{eq:contradicted_expression}
    \cbh
    =\hb(\zb)_{1:n_c}
    =\hb(\zbt)_{1:n_c}.
\end{equation}

It remains to show that $\hb(\cdot)_{1:n_c}$ can only be a function of $\cb$, i.e., does not depend on any other (style) dimension of $\zb=(\cb,\sb)$.

\paragraph{Step 2.}
Suppose \textit{for a contradiction} that $\hb_c(\cb,\sb):=\hb(\cb,\sb)_{1:n_c}=\hb(\zb)_{1:n_c}$ depends on some component of the style variable $\sb$:
\begin{equation}
    \label{eq:contradiction_assumption}
    \exists l\in\{1, ..., n_s\}, (\cb^*,\sb^*)\in \Ccal\times \Scal,
    \quad \quad
    \text{s.t.}
    \quad \quad
    \frac{\partial \hb_c}{\partial s_{l}}(\cb^*,\sb^*)\neq 0,
\end{equation}
that is, we assume that the partial derivative of $\hb_c$ w.r.t.\ some style variable $s_l$ is non-zero at some point $\zb^*=(\cb^*,\sb^*)\in\Zcal=\Ccal\times\Scal$.

Since $\hb$ is smooth, so is $\hb_c$. Therefore, $\hb_c$ has continuous (first) partial derivatives.

By continuity of the partial derivative, $\frac{\partial \hb_c}{\partial s_{l}}$ must be non-zero in a neighbourhood of $(\cb^*,\sb^*)$, i.e.,
\begin{equation}
\label{eq:monotonicity}
\exists \eta>0 
\quad \quad \text{s.t.} \quad \quad 
s_{l} \mapsto \hb_c\big(\cb^*,(\sb^*_{-l},s_l)\big)
\quad  \text{is strictly monotonic on} \quad
(s^*_l-\eta,s^*_l+\eta),
\end{equation}
where $\sb_{-l}\in\Scal_{-l}$ denotes the vector of remaining style variables except $s_l$. 

Next, define the auxiliary function $\psi:\Ccal\times \Scal\times \Scal\rightarrow \RR_{\geq 0}$ as follows:
\begin{equation}
\label{eq:psi_def}
    \psi(\cb,\sb,\sbt):=|\hb_c(\cb,\sb)-\hb_c(\cb,\sbt)|\geq0\,.
\end{equation}

To obtain a contradiction to the invariance condition~\eqref{eq:contradicted_expression} from Step 1 under  assumption~\eqref{eq:contradiction_assumption},
it remains to show that $\psi$ from~\eqref{eq:psi_def} is \textit{strictly positive} with probability greater than zero (w.r.t.\ $p$).

First, the strict monotonicity from~\eqref{eq:monotonicity}
implies that 
\begin{equation}
\label{eq:positivity}
\psi\big(\cb^*,(\sb_{-l}^*,s_l),(\sb_{-l}^*,\tilde{s}_l)\big)
>0\,,\quad \forall (s_l,\tilde{s}_l)\in (s^*_l,s^*_l+\eta)\times(s^*_l-\eta,s^*_l) \,.
\end{equation}
Note that in order to obtain the strict inequality in~\eqref{eq:positivity}, it is important that $s_l$ and $\tilde{s}_l$ take values in \textit{disjoint} open subsets of the interval $(s^*_l-\eta,s^*_l+\eta)$ from~\eqref{eq:monotonicity}.

Since $\psi$ is a composition of continuous functions (absolute value of the difference of two continuous functions), $\psi$ is continuous.

Consider the open set $\RR_{>0}$, and recall that, under a continuous function, pre-images (or inverse images) of open sets are always \textit{open}.

Applied to the continuous function $\psi$, this pre-image corresponds to an \textit{open} set 
\begin{equation}
    \Ucal\subseteq \Ccal\times \Scal\times \Scal
\end{equation}
in the domain of $\psi$ on which $\psi$ is strictly positive. 

Moreover, due to~\eqref{eq:positivity}:
\begin{equation}
\label{eq:U_nonempty}
    \{\cb^*\}\times\left(\{\sb^*_{-l}\}\times(s^*_l,s^*_l+\eta)\right) \times \left(\{\sb^*_{-l}\}\times(s^*_l-\eta,s^*_l)\right)\subset \Ucal,
\end{equation}
so $\Ucal$ is \textit{non-empty}.

Next, by assumption \textit{(iii)}, there exists at least one subset $A\subseteq\{1, ..., n_s\}$ of changing style variables such that $l\in A$ and $p_A(A)>0$;
pick one such subset and call it $A$.

Then, also by assumption \textit{(iii)}, for any $\sb_A\in\Scal_A$, there is an open subset $\Ocal(\sb_A)\subseteq \Scal_A$ containing $\sb_A$, such that $p_{\sbt_\A|\sb_\A}(\cdot|\sb_A)>0$ within $\Ocal(\sb_A)$. 

Define the following space
\begin{equation}
    \Rcal_A:=\{(\sb_A,\sbt_A):\sb_A\in\Scal_A,\sbt_A\in\Ocal(\sb_A)\}
\end{equation}
and, recalling that $\Ac=\{1, ..., n_s\} \setminus A$ denotes the complement of $A$, define
\begin{equation}
    \Rcal := \Ccal\times \Scal_{\Ac}\times \Rcal_A
\end{equation}
which is a topological subspace of
$\Ccal\times\Scal\times\Scal$.

By assumptions \textit{(ii)} and \textit{(iii)},
$p_\zb$ is smooth and fully supported, and $p_{\sbt_A|\sb_A}(\cdot|\sb_A)$ is smooth and fully supported on $\Ocal(\sb_A)$ for any $\sb_A\in\Scal_A$.
Therefore,
the measure
$\mu_{(\cb,\sb_{\Ac},\sb_{A},\sbt_{A})|A}$ has fully supported, strictly-positive density on $\Rcal$
w.r.t.\ a strictly positive measure on $\Rcal$.
In other words, $p_\zb \times p_{\sbt_A|\sb_A}$ is fully supported (i.e., strictly positive) on $\Rcal$.

Now consider the intersection $\Ucal\cap \Rcal$ of 
the open set $\Ucal$ 
with the topological subspace $\Rcal$.

Since $\Ucal$ is open, by the definition of topological subspaces, the intersection $\Ucal\cap \Rcal\subseteq \Rcal$ is \textit{open} in $\Rcal$, (and thus has the same dimension as $\Rcal$ if non-empty). 

Moreover, since $\Ocal(\sb_A^*)$ is open containing $\sb_A^*$, there exists $\eta'>0$ such that $\{\sb^*_{-l}\}\times(s^*_l-\eta',s^*_l)\subset \Ocal(\sb_A^*)$. Thus, for $\eta''=\min(\eta,\eta')>0$,
\begin{equation}
     \{\cb^*\}\times\{\sb^*_{\Ac}\}
     \times\left(\{\sb^*_{A\setminus\{l\}}\}\times (s^*_l,s^*_l+\eta)\right)
     \times \left(\{\sb^*_{A\setminus\{l\}}\}\times(s^*_l-\eta'',s^*_l)\right)\subset \Rcal.
\end{equation}
In particular, this implies that 
\begin{equation}
\label{eq:R_nonempty} 
    \{\cb^*\}\times\left(\{\sb^*_{-l}\}\times(s^*_l,s^*_l+\eta)\right) \times \left(\{\sb^*_{-l}\}\times(s^*_l-\eta'',s^*_l)\right)\subset \Rcal,
\end{equation}
Now, since $\eta''\leq\eta$, the LHS of~\eqref{eq:R_nonempty} is also in $\Ucal$ according to~\eqref{eq:U_nonempty}, so the intersection $\Ucal\cap\Rcal$ is \textit{non-empty}.

In summary, the intersection $\Ucal\cap\Rcal\subseteq \Rcal$:
\begin{itemize}
    \item is non-empty (since both $\Ucal$ and $\Rcal$ contain the LHS of~\eqref{eq:U_nonempty});
    \item is an open subset of the topological subspace $\Rcal$ of $\Ccal\times\Scal\times\Scal$ (since it is the intersection of an open set, $\Ucal$, with $\Rcal$);
    \item satisfies $\psi>0$ (since this holds for all of $\Ucal$);
    \item is fully supported w.r.t. the generative process (since this holds for all of $\Rcal$).
\end{itemize}

As a consequence,
\begin{equation}
    \PP\left(\psi(\cb,\sb,\sbt)>0
    |A\right)\geq\PP(\Ucal\cap \Rcal)>0,
\end{equation}
where $\PP$ denotes probability w.r.t.\ the true generative process $p$.

Since $p_A(A)>0$, this is a \textbf{contradiction} to the invariance~\eqref{eq:contradicted_expression} from Step 1.

Hence, assumption~\eqref{eq:contradiction_assumption} cannot hold, i.e., $\hb_c(\cb,\sb)$ does not depend on any style variable $s_l$. It is thus only a function of $\cb$, i.e., $\cbh=\hb_c(\cb)$.

Finally, smoothness and invertibility of $\hb_c:\Ccal\rightarrow\Ccal$ follow from smoothness and invertibility of $\hb$, as established in Step 1.

This concludes the proof that $\cbh$ is related to the true content $\cb$ via a smooth invertible mapping.
\end{proof}

\subsection{Proof of\texorpdfstring{~\Cref{thm:CL}}{thmcl}}
\label{app:proof_discriminative}
\discriminative*

\begin{proof}
As in the proof of~\cref{thm:main}, the proof again consists of two main steps.

In the first step, we show that the representation $\zbh=\gb(\xb)$ extracted by any $\gb$ that minimises $\Lcal_\mathrm{Align}$ is related to the true latent $\zb$ through a smooth invertible mapping $\hb$, and that $\zbh$ must satisfy invariance across $(\xb,\xbt)$ in the first $n_c$ (content) components almost surely (a.s.) with respect to (w.r.t.) the true generative process.

In the second step, we use the same argument by contradiction as in Step 2 of the proof of~\cref{thm:main}, to show that $\cbh=\hb(\zb)_{1:n_c}$ can only depend on the true content $\cb$ and not on style $\sb$.

\paragraph{Step 1.}
From the form of the objective~\eqref{eq:CL_MSE_objective}, it is clear that $\Lcal_\mathrm{Align}\geq 0$ with equality if and only if $\gb(\xbt)_{1:n_c}=\gb(\xb)_{1:n_c}$ for all $(\xb, \xbt)$ s.t. $p_{\xb, \xbt}(\xb, \xbt)>0$.

Moreover, it follows from the assumed generative process that the global minimum of zero is attained by the true unmixing $\fbinv$ since
\begin{equation}
\fbinv(\xb)_{1:n_c}=\cb=\cbt=\fbinv(\xbt)_{1:n_c}
\end{equation}
holds a.s.\ (i.e., with probability one) w.r.t.\ the true generative process $p$. 

Hence, there exists at least one smooth invertible function ($\fbinv$) which attains the global minimum.

Let $\gb$ be \textit{any} function attaining the global minimum of $\Lcal_\mathrm{Align}$ of zero.

As argued above, this implies that (a.s.\ w.r.t.\ $p$):
\begin{equation}
    \gb(\xbt)_{1:n_c}=\gb(\xb)_{1:n_c}.
\end{equation}

Writing $\gb=\hb \circ \fbinv$, where $\hb$ is the smooth, invertible function $\hb=\gb \circ \fb$
we obtain (a.s.\ w.r.t.\ $p$):
\begin{equation}
    \cbh=\hb(\zbt)_{1:n_c}=\hb(\zb)_{1:n_c}.
\end{equation}

Note that this is the same invariance condition as~\eqref{eq:contradicted_expression} derived in Step 1 of the proof of~\cref{thm:main}.

\paragraph{Step 2.}
It remains to show that $\hb(\zb)_{1:n_c}$ can only depend on the true content $\cb$ and not on any of the style variables $\sb$.
To show this, we use the same Step 2 as in the proof of~\cref{thm:main}.
\end{proof}

\subsection{Proof of\texorpdfstring{~\Cref{thm:CL_MaxEnt}}{thmclmaxent}}
\label{app:proof_CL_MaxEnt}
\discriminativeMaxEnt*

\begin{proof}
The proof consists of three main steps.

In the first step, we show that the representation $\cbh=\gb(\xb)$ extracted by any smooth function $\gb$ that minimises~\eqref{eq:CL_MSE_MaxEnt_objective} 
is related to the true latent $\zb$ through a smooth mapping $\hb$; that $\cbh$ must satisfy invariance across $(\xb,\xbt)$ almost surely (a.s.) with respect to (w.r.t.) the true generative process $p$; and that $\cbh$ must follow a uniform distribution on $(0,1)^{n_c}$.

In the second step, we use the same argument by contradiction as in Step 2 of the proof of~\cref{thm:main}, to show that $\cbh=\hb(\zb)$ can only depend on the true content $\cb$ and not on style $\sb$.

Finally, in the third step, we show that $\hb$ must be a bijection, i.e., invertible, using a result from~\cite{zimmermann2021contrastive}.

\paragraph{Step 1.}

The global minimum of $\Lcal_{\mathrm{AlignMaxEnt}}$ is reached when the first term (alignment) is minimised (i.e., equal to zero) and the second term (entropy) is maximised.

Without additional moment constraints, the \textit{unique} maximum entropy distribution on $(0,1)^{n_c}$ is the uniform distribution~\cite{jaynes1982rationale,cover2012elements}.

First, we show that there exists a smooth function $\gb^*:\Xcal\rightarrow(0,1)^{n_c}$ which attains the global minimum of $\Lcal_{\mathrm{AlignMaxEnt}}$.

To see this, consider the function $\fbinv_{1:n_c}:\Xcal\rightarrow\Ccal$, i.e., the inverse of the true mixing~$\fb$, restricted to its first $n_c$ dimensions.
This exists and is smooth since $\fb$ is smooth and invertible by assumption \textit{(i)}.
Further, we have $\fbinv(\xb)_{1:n_c}=\cb$ by definition.

We now build a function $\db:\Ccal\rightarrow(0,1)^{n_c}$ which maps $\cb$ to a uniform random variable on $(0,1)^{n_c}$ using a recursive construction known as the \textit{Darmois construction}~\cite{darmois1951construction,hyvarinen1999nonlinear}. 

Specifically, we define
\begin{equation}
    d_i(\cb) := F_i(c_i|\cb_{1:i-1})=\PP(C_i\leq c_i|\cb_{1:i-1}), 
    \quad \quad \quad i=1, ..., n_c,
\end{equation}
where $F_i$ denotes the conditional cumulative distribution function (CDF) of $c_i$ given $\cb_{1:i-1}$.

By construction, $\db(\cb)$ is uniformly distributed on $(0,1)^{n_c}$~\cite{darmois1951construction,hyvarinen1999nonlinear}. 

Further, $\db$ is smooth by the assumption that $p_\zb$ (and thus $p_\cb$) is a smooth density.

Finally, we define
\begin{equation}
\label{eq:global_min_max_ent}
    \gb^*:=\db\circ\fbinv_{1:n_c}:\Xcal\rightarrow(0,1)^{n_c},
\end{equation}
which is a smooth function since it is a composition of two smooth functions. 

\begin{claim}
\label{claim:global_min_max_ent}
$\gb^*$ as defined in~\eqref{eq:global_min_max_ent} attains the global minimum of $\Lcal_{\mathrm{AlignMaxEnt}}$.
\end{claim}

\textbf{Proof of Claim~\ref{claim:global_min_max_ent}.}
Using $\fbinv(\xb)_{1:n_c}=\cb$ and $\fbinv(\xbt)_{1:n_c}=\cbt$, we have
\begin{align}
    \Lcal_{\mathrm{AlignMaxEnt}}(\gb^*)
    &=
    \EE_{(\xb,\xbt)\sim p_{(\xb, \xbt)}}
\left[
\bignorm{
\gb^*(\xb)-\gb^*(\xbt)
}_2^2
\right] - H\left(\gb^*(\xb)\right)
\\
&=
\EE_{(\xb,\xbt)\sim p_{(\xb, \xbt)}}
\left[
\bignorm{
\db(\cb)-\db(\cbt)
}^2_2
\right] - H\left(\db(\cb)\right)
\\
&= 0 
\end{align}
where in the last step we have used the fact that $\cb=\cbt$ almost surely w.r.t.\ to the ground truth generative process $p$ described in~\cref{sec:problem_formulation}, so the first term is zero; and the fact that $\db(\cb)$ is uniformly distributed on $(0,1)^{n_c}$ and the uniform distribution on the unit hypercube has zero entropy, so the second term is also zero.

Next, let $\gb:\Xcal\rightarrow(0,1)^{n_c}$ be \textit{any} smooth function which attains the global minimum of~\eqref{eq:CL_MSE_MaxEnt_objective}, i.e.,
\begin{equation}
\label{eq:g_align_maxent}
\Lcal_{\mathrm{AlignMaxEnt}}(\gb)=\EE_{(\xb,\xbt)\sim p_{(\xb, \xbt)}}
\left[
\bignorm{
\gb(\xb)-\gb(\xbt)
}^2_2
\right] - H\left(\gb(\xb)\right)
=0.
\end{equation}

Define
    $\hb:=\gb\circ\fb:\Zcal\rightarrow (0,1)^{n_c}$
which is smooth because both $\gb$ and $\fb$ are smooth.

Writing $\xb=\fb(\zb)$, ~\eqref{eq:g_align_maxent} then implies in terms of $\hb$:
\begin{align}
\label{eq:MaxEntinvariance}
\EE_{(\xb,\xbt)\sim p_{(\xb, \xbt)}}
\left[
\bignorm{
\hb(\zb)-\hb(\zbt)
}^2_2
\right]
&= 0\, ,
\\
\label{eq:MaxEntuniformity}
H\left(\hb(\zb)\right)
&=
0\, .
\end{align}

Equation \eqref{eq:MaxEntinvariance} implies that the same invariance condition~\eqref{eq:contradicted_expression} used in the proofs of~\cref{thm:main,thm:CL} must hold (a.s. w.r.t. $p$), and~\eqref{eq:MaxEntuniformity} implies that $\cbh=\hb(\zb)$ must be uniformly distributed on $(0,1)^{n_c}$. 

\paragraph{Step 2.}
Next, we show that $\hb(\zb)=\hb(\cb,\sb)$ can only depend on the true content $\cb$ and not on any of the style variables $\sb$. For this we use the same Step 2 as in the proofs of~\cref{thm:main,thm:CL}.

\paragraph{Step 3.}
Finally, we show that the mapping $\cbh=\hb(\cb)$ is invertible.

To this end, we make use of the following result from~\cite{zimmermann2021contrastive}.

\begin{proposition}[Proposition 5 of~\cite{zimmermann2021contrastive}]
\label{prop:bijectivity}
Let $\Mcal, \Ncal$ be simply connected and oriented $\Ccal^1$ manifolds without boundaries and $h : \Mcal \rightarrow \Ncal$ be a differentiable map.
Further, let the random variable $\zb \in \Mcal$ be distributed according to $\zb \sim p(\zb)$ for a regular density function $p$, i.e., $0 < p < \infty$.
If the pushforward $p_{\#h}(\zb)$ of $p$ through $h$ is also a regular density, i.e., $0 < p_{\#h} < \infty$,
then $h$ is a bijection.
\end{proposition}

We apply this result to the simply connected and oriented $\Ccal^1$ manifolds without boundaries $\Mcal=\Ccal$ and  $\Ncal=(0,1)^{n_c}$, and the smooth (hence, differentiable) map $\hb:\Ccal\rightarrow (0,1)^{n_c}$ which maps the random variable $\cb$ to a uniform random variable $\cbh$ (as established in Step 1). 

Since both $p_\cb$ (by assumption) and the uniform distribution (the pushforward of $p_\cb$ through $\hb$) are regular densities in the sense of~\cref{prop:bijectivity}, we conclude that $\hb$ is a bijection, i.e., invertible.

We have shown that for any smooth $\gb:\Xcal\rightarrow(0,1)^{n_c}$ which minimises $\Lcal_{\mathrm{AlignMaxEnt}}$, we have that $\cbh=\gb(\xb)=\hb(\cb)$ for a smooth and invertible $\hb:\Ccal\rightarrow(0,1)^{n_c}$, i.e., $\cb$ is block-identified by $\gb$.
\end{proof}

\section{Proofs from~\texorpdfstring{\cref{chap:CRL}}{}}
\label{app:proofs_CRL}

\subsection{Proof of~\texorpdfstring{\cref{prop:minimality_of_CRL}}{}}
\minimality*
\begin{proof}
Since $(h,G')\sim_\textsc{crl}(f^{-1},G)$, by~\cref{def:identifiability} we have
\begin{align}
\label{eq:Z_in_CRL}
    \Zb=\Pb_{\pi^{-1}}\circ \phi (\Vb)
\end{align}
\looseness-1 for some element-wise diffeomorphism $\phi$ with inverse $\psi=\phi^{-1}$. Then~\eqref{eq:Z_in_CRL} implies that for all $i\in[n]$
\begin{align}
\label{eq:element_wise_change_of_var}
    V_i=\psi_i(Z_{\pi(i)})
\end{align}

According to~\eqref{eq:element_wise_change_of_var}, each conditional in the Markov factorization of $Q_\Zb$ is given in terms of $p$ by
\begin{align}
    \label{eq:transformed_mechanisms}
    q_{\pi(i)}\left(z_{\pi(i)}~|~\zb_{\pa(\pi(i);G')}\right)
    =p_i\left(\psi_i\left(z_{\pi(i)}\right)~|~\psi_{\pa(i)}\left(\zb_{\pa(\pi(i);G')}\right)\right)
    \abs{\frac{\d \psi_i}{\d z_{\pi(i)}}\left(z_{\pi(i)}\right)}
\end{align}
where we have used the change of variables in~\eqref{eq:element_wise_change_of_var} and the fact that $\pi(\pa(i))=\pa(\pi(i);G')$ since $\pi:G\mapsto G'$ is a graph isomorphism.

\looseness-1 Consider an intervention that changes $p_i(v_i~|~\vb_{\pa(i)})$ to some intervened mechanism $\tilde p_i(v_i~|~\vb_{\pa(i)})$ for all $i\in \Ical^e$.
Denote the corresponding intervened joint distribution by $P^e_\Vb$ with joint density $p^e$ given by
\begin{equation}
\label{eq:soft_intervention}
    p^e(\vb)= 
    \prod_{i\in \Ical^e} \tilde p_i\left(v_i~|~\vb_{\pa(i)}\right)
    \,
    \prod_{j\in[n]\setminus \Ical^e}p_j\left(v_j~|~\vb_{\pa(j)}\right)\,.
\end{equation}

\looseness-1 Denote by $Q^e_\Zb=(\Pb_{\pi^{-1}}\circ \phi)_*(P^e_\Vb)$ the corresponding distribution over $\Zb$ with joint density given by $q^e$
\begin{align}
    q^e(\zb)&=p^e(\psi\circ\Pb_{\pi}(\zb))\abs{\det\Jb_{\psi\circ\Pb_{\pi}}(\zb)}
    \\
    &=
     \prod_{i\in \Ical^e} \tilde p_i\left(\psi_i\left(z_{\pi(i)}\right)~|~\psi_{\pa(i)}\left(\zb_{\pa(\pi(i);G')}\right)\right)
    \abs{\frac{\d \psi_i}{\d z_{\pi(i)}}\left(z_{\pi(i)}\right)}
    \,
    \prod_{j\in[n]\setminus \Ical^e}q_{\pi(j)}\left(z_{\pi(j)}~|~\zb_{\pa(\pi(j);G')}\right)\,,
\end{align}
where we have used~\eqref{eq:transformed_mechanisms}, \eqref{eq:soft_intervention}, and the fact that $\Jb_\psi$ is diagonal.

By defining 
\begin{align}
    \label{eq:induced_intervened_mechanism_q}
    \tilde q_{\pi(i)}\left(z_{\pi(i)}~|~\zb_{\pa(\pi(i);G')}\right)
    :=
    \tilde p_i\left(\psi_i\left(z_{\pi(i)}\right)~|~\psi_{\pa(i)}\left(\zb_{\pa(\pi(i);G')}\right)\right)
    \abs{\frac{\d \psi_i}{\d z_{\pi(i)}}\left(z_{\pi(i)}\right)}
\end{align}
we finally arrive at 
\begin{align}
    q^e(\zb)=
     \prod_{i\in \Ical^e} \tilde q_{\pi(i)}\left(z_{\pi(i)}~|~\zb_{\pa(\pi(i);G')}\right)
    \,
    \prod_{j\in[n]\setminus \Ical^e}q_{\pi(j)}\left(z_{\pi(j)}~|~\zb_{\pa(\pi(j);G')}\right)\,.
\end{align}

This shows that any intervention on $\{V_i\}_{i\in\Ical^e}\subseteq \Vb$ which replaces
\begin{equation}
    \left\{p_i(v_i~|~\vb_{\pa(i)}) \right\}_{i\in\Ical^e} \mapsto \left\{\tilde p_i(v_i~|~\vb_{\pa(i)})\right\}_{i\in\Ical^e}\,,
\end{equation}
can equivalently be captured by an intervention on $\{Z_{\pi(i)}\}_{i\in\Ical^e}\subseteq\Zb$ which replaces
\begin{equation}
    \left\{q_{\pi(i)}\left(z_{\pi(i)}~|~\zb_{\pa(\pi(i);G')}\right)\right\}_{i\in\Ical^e} \mapsto \left\{\tilde q_{\pi(i)}\left(z_{\pi(i)}~|~\zb_{\pa(\pi(i);G')}\right)\right\}_{i\in\Ical^e}\,.
\end{equation}
with $\tilde q_i$ defined according to~    \eqref{eq:induced_intervened_mechanism_q}.
\end{proof}

\subsection{Auxiliary Lemmata}
\label{app:lemmas}
\begin{lemma}[Lemma~2 of~\citet{brehmer2022weakly}]
\label{lemma:brehmer}
\looseness-1 Let $A = C = \RR$ and $B = \RR^n$. Let $f: A\times B \to C$ be differentiable. Define two differentiable
measures $p_A$ on $A$ and $p_C$ on $C$. Let $\forall b \in B$, $f(\cdot, b): A \to C$ be measure-preserving, in the sense that the pushforward of $p_A$ is always $p_C$. Then $f(\cdot, b)$ is constant in $b$ on $B$.
\begin{proof}
    See Appendix A.2 of~\citet{brehmer2022weakly}.
\end{proof}
\end{lemma}

\begin{lemma}[Preservation of conditional independence under invertible transformation.]
\label{lemma:preserve_cond_ind}
Let $X$ and $Y$ be continuous real-valued random variables, and let $\Zb$ be a continuous random vector taking values in $\RR^n$. Suppose that $(X,Y,\Zb)$ have a joint density w.r.t.\ the Lebesgue measure.
Let $f:\RR\to\RR$, $g:\RR\to\RR$, and $h:\RR^n\to\RR^n$ be diffeomorphisms. Then $X\independent Y ~|~ \Zb \implies f(X)\independent g(Y)~|~h(\Zb)$.
\begin{proof}
    Denote by $p(x,y,\zb)$ the density of $(X,Y,\Zb)$. Then $X\independent Y ~|~ \Zb$ implies that for all $(x,y,\zb)$, $p$ can be factorized as follows:
    \begin{align}
    \label{eq:cond_ind_factorisation}
        p(x,y,\zb)=p_\zb(\zb)p_x(x~|~\zb)p_y(y~|~\zb)\,. 
    \end{align}
Let $(A, B, \Cb)=(f(X),g(Y),h(\Zb))$, and write $\tilde f=f^{-1}$, $\tilde g=g^{-1}$, and $\tilde h=h^{-1}$.

Then $(A,B,\Cb)$ also has a density $q(a,b,\cbb)$, which for all $(a,b,\cbb)$ is given by the change of variable formula:
\begin{align}
\label{eq:block_diag}
    q(a,b,\cbb)&=p\left(\tilde f(a), \tilde g(b), \tilde h(\cbb)\right)\abs{\frac{\d \tilde f}{\d a}(a)\frac{\d \tilde g}{\d b}(b) \det \Jb_{\tilde h}(\cbb)} 
    \\
    &= p_\zb\left(\tilde h(\cbb)\right)\abs{\det \Jb_{\tilde h}(\cbb)} 
    \, p_x\left(\tilde f(a)~|~\tilde h(\cbb)\right)\abs{\frac{\d \tilde f}{\d a}(a)}
    \, p_y\left(\tilde g(b)~|~\tilde h(\cbb)\right)\abs{\frac{\d \tilde g}{\d b}(b)}
    \label{eq:p_factor}
\end{align}
where in~\eqref{eq:block_diag} we have used the fact that $(X, Y, \Zb)\mapsto(f(X),g(Y),h(\Zb))$ has block-diagonal Jacobian, and in~\eqref{eq:p_factor} that $p$ factorises as in~\eqref{eq:cond_ind_factorisation}. Next, define
\begin{align}
    q_{\cbb}(\cbb)&:=p_\zb\left(\tilde h(\cbb)\right)\abs{\det \Jb_{\tilde h}(\cbb)}\,,
    \\
    q_{a}(a~|~\cbb)&:=p_x\left(\tilde f(a)~|~\tilde h(\cbb)\right)\abs{\frac{\d \tilde f}{\d a}(a)}\,,
    \\
    q_{b}(b~|~\cbb)&:= p_y\left(\tilde g(b)~|~\tilde h(\cbb)\right)\abs{\frac{\d \tilde g}{\d b}(b)}\,.
\end{align}
Since $p_\zb$, $p_x$, and $p_y$ are valid densities (non-negative and integrating to one), so are $q_\cbb$, $q_a$, and~$q_b$.
Substitution into~\eqref{eq:p_factor} yields for all $(a,b,\cbb)$,
\begin{align}
    q(a,b,\cbb)=q_{\cbb}(\cbb)q_{a}(a~|~\cbb) q_{b}(b~|~\cbb)\,,
\end{align}
which implies that $A\independent B~|~\Cb$, concluding the proof.
\end{proof}
\end{lemma}

\subsection{Proof of\texorpdfstring{~\cref{thm:bivariate}}{thmbivariate}}
\label{app:proof_bivariate}
\bivariate*

\begin{proof}
From the assumption of a shared mixing $f$ and shared encoder $h$ across all environments, we have that
\begin{equation}
    \Zb=h(\Xb)=h(f(\Vb))=h\circ f(\Vb)\,.
\end{equation}
Let $\psi=f^{-1}\circ h^{-1}:\RR^n\to\RR^n$ such that $$\Vb=\psi(\Zb)\,.$$ 
By~\cref{ass:diffeomorphism}, $f$, $h$, and thus also $h\circ f$  are diffeomorphisms. Hence, $\psi$ is well-defined and also diffeomorphic.

By the change of variable formula,  for all $e$ and all $\zb$ the density $q^e(\zb)$ is given by
\begin{align}
\label{eq:app_change_of_variable}
q^e(\zb)=p^e(\psi(\zb))\abs{\det \Jb_\psi(\zb)}
\end{align}
where $(\Jb_\psi(\zb))_{ij}=\frac{\partial \psi_i}{\partial z_j}(\zb)$ denotes the Jacobian of $\psi$.

We now consider two separate cases, depending on whether the 
intervention targets in $q^{e_i}$ for $e_i\in\{e_1,e_2\}$ match those in $p^{e_i}$ (Case 1) or not (Case 2).

\paragraph{Case 1: Aligned Intervention Targets.}
\looseness-1
According to~\cref{ass:shared_mechs} and (A2), \eqref{eq:app_change_of_variable} applied to the known observational environment $e_0$ and the interventional environments $e_1,e_2$ leads to the system of equations:
\begin{align}
\label{eq:app_qe0}
q_1(z_1)q_2(z_2~|~z_{\pa(2;G')})&=
p_1\left(\psi_1(\zb)\right)p_2\left(\psi_2(\zb)~|~\psi_{\pa(2)}(\zb)\right) \abs{\det\Jb_{\psi}(\zb)}
\\
\label{eq:app_qe1}
\tilde q_1(z_1)q_2(z_2~|~z_{\pa(2;G')})&=
\tilde p_1\left(\psi_1(\zb)\right) p_2\left(\psi_2(\zb)~|~\psi_{\pa(2)}(\zb)\right) \abs{\det\Jb_{\psi}(\zb)}
\\
\label{eq:app_qe2}
q_1(z_1)\tilde q_2(z_2)&= 
p_1\left(\psi_1(\zb)\right)\tilde p_2\left(\psi_2(\zb)\right) \abs{\det\Jb_{\psi}(\zb)}
\end{align}
where $z_{\pa(2;G')}$ denotes the parents of $z_2$ in $G'$, and $\pa(2)$ denotes the parents of $V_2$ in $G$.

Note that neither side of the previous equations can be zero due to the full support assumption\footnote{This can also be relaxed to fully supported on a Cartesian product of intervals.} (A1) and $\psi$ being diffeomorphic implying the determinant is non-zero. 

Taking quotients of~\eqref{eq:app_qe1} and~\eqref{eq:app_qe0}, yields
\begin{equation}
    \frac{\tilde q_1}{q_1}(z_1)=\frac{\tilde p_{1}}{p_{1}}(\psi_1(\zb))\,.
\end{equation}
Next, we take the partial derivative w.r.t.\ $z_2$ on both sides and use the chain rule to obtain:
\begin{equation}
\label{eq:app_zero_product}
    0=\left(\frac{\tilde p_{1}}{p_{1}}\right)'\left(\psi_1(\zb)\right)\frac{\partial \psi_1}{\partial z_2}(\zb)\,.
\end{equation}
Now, by assumption (A3), the first term on the RHS of~\eqref{eq:app_zero_product} is non-zero everywhere. Hence,
\begin{equation}
\label{eq:app_triangularJ}
    \forall \zb: \qquad 
    \frac{\partial \psi_1}{\partial z_2}(\zb)=0 \,.
\end{equation}
It follows that $\psi_1$ is, in fact, a scalar function, and  
\begin{equation}
\label{eq:app_reparam_V1}
    V_1=\psi_1(Z_1)\,. 
\end{equation}
Since $\psi$ is a diffeomorphism, $\psi_1$ must also be diffeomorphic. Hence, by the change of variable formula applied to~\eqref{eq:app_reparam_V1}, the marginal density $q_1(z_1)$ is given by
\begin{equation}
\label{eq:app_marginal_density}
    q_1(z_1)=p_1(\psi_1(z_1))\abs{\frac{\partial \psi_1}{\partial z_1}(z_1)}\,.
\end{equation}

Further, since $\Jb_\psi$ is triangular due to~\eqref{eq:app_triangularJ}, its determinant is given by
\begin{equation}
\label{eq:app_det_triangluar}
    \abs{\det\Jb_\psi(\zb)}=\abs{\frac{\partial \psi_1}{\partial z_1}(z_1)\, \frac{\partial \psi_2}{\partial z_2}(z_1,z_2)}\,.
\end{equation}

Substituting~\eqref{eq:app_marginal_density} and ~\eqref{eq:app_det_triangluar} into~\eqref{eq:app_qe2} yields (after cancellation of equal terms):
\begin{align}
\label{eq:app_measure_preserving}
\tilde q_2(z_2)&= \tilde p_2\left(\psi_2(z_1,z_2)\right) \abs{\frac{\partial \psi_2}{\partial z_2}(z_1,z_2)}\,.
\end{align}
The expression in~\eqref{eq:app_measure_preserving} implies that, for all $z_1$, the mapping $\psi_2(z_1,\cdot):\RR\to\RR$ is measure preserving for the differentiable $\tilde q_2$ and $\tilde{p}_2$. By~\cref{lemma:brehmer} (Lemma~2 of~\citet[][\S~A.2]{brehmer2022weakly}),
it then follows that $\psi_2$ must, in fact, be constant in $z_1$, that is
\begin{equation}
    \label{eq:app_psi_diagonal}
    \forall \zb: \qquad 
    \frac{\partial \psi_2}{\partial z_1}(\zb)=0 \,.
\end{equation}

Note that this last step is where the assumption of perfect interventions~(\cref{ass:perfect_interventions}) is leveraged: the conclusion would not hold for arbitrary imperfect interventions for which~\eqref{eq:measure_preserving} would involve $\tilde q_2(z_2~|~z_1)$ and $p_2\left(\psi_2(z_1,z_2)~|~\psi_{1}(z_1)\right)$.

Hence, we have shown that $\psi$ is an element-wise function:
\begin{equation}
\label{eq:app_element_wise_psi}
    \Vb=(V_1,V_2)=\psi(\Zb)=(\psi_1(Z_1), \psi_2(Z_2))\,.
\end{equation}

Finally, since $\psi$ is a diffeomorphism, \eqref{eq:app_element_wise_psi} implies that
\begin{equation}
    V_1 \independent V_2 \iff Z_1\independent Z_2\,.
\end{equation}
It then follows from the faithfulness assumption~(\cref{ass:faithfulness}) that we also must have $G=G'$.

This concludes the proof of Case 1, as we have shown that $ (h,G')\sim_\textsc{crl}(f^{-1},G)$ with $G'=\pi(G)=G$ ($\pi$ being the identity permutation) and $h\circ f=\psi^{-1}=:\phi$ an element-wise diffeomorphism. 

\paragraph{Case 2: Misaligned Intervention Targets.}
Writing down the system of equations similar to~\eqref{eq:app_qe0}--\eqref{eq:app_qe2}, but for the case with misaligned intervention targets across $p$ and $q$ yields:
\begin{align}
\label{eq:app_2qe0}
q_1(z_1)q_2(z_2~|~z_{\pa(2;G')})&=
p_1\left(\psi_1(\zb)\right)p_2\left(\psi_2(\zb)~|~\psi_{\pa(2)}(\zb)\right) \abs{\det\Jb_{\psi}(\zb)}
\\
\label{eq:app_2qe1}
\tilde q_1(z_1)q_2(z_2~|~z_{\pa(2;G')})&=
 p_1\left(\psi_1(\zb)\right) \tilde p_2\left(\psi_2(\zb)\right)  \abs{\det\Jb_{\psi}(\zb)}
\\
\label{eq:app_2qe2}
q_1(z_1)\tilde q_2(z_2)&= 
\tilde p_1\left(\psi_1(\zb)\right)p_2\left(\psi_2(\zb)~|~\psi_{\pa(2)}(\zb)\right) \abs{\det\Jb_{\psi}(\zb)} \,.
\end{align}

Taking ratios of $e_1$ and $e_2$ with $e_0$ yields
\begin{align}
    \label{eq:app_misaligned_ratio_1}
    \frac{\tilde q_1}{q_{1}}(z_1)
    &=\frac{\tilde p_2\left(\psi_2(\zb)\right)}{p_{2}\left(\psi_2(\zb)~|~\psi_{\pa(2)}(\zb)\right)}
\\
    \label{eq:app_misaligned_ratio_2}
\frac{\tilde q_2(z_2)}{q_{2}(z_2~|~z_{\pa(2;G')})}
    &=\frac{\tilde p_1}{p_1}\left(\psi_1(\zb)\right)\,.
\end{align}

We separate the remainder of the proof of Case 2 into different subcases depending on $G$ and $G'$: as we will see, we can use a similar reasoning as in Case 1, except for the case where both $G$ and $G'$ are missing no edge.

\textit{Case 2a: $V_1\not\to V_2$ in $G$, that is $\pa(2)=\varnothing$.} Then we can proceed similarly to Case 1. First, we take the partial derivative of~\eqref{eq:app_misaligned_ratio_1} w.r.t.\ $z_2$  to arrive at:
\begin{equation}
\label{eq:app_2zero_product}
    0=\left(\frac{\tilde p_{2}}{p_{2}}\right)'\left(\psi_2(\zb)\right)\frac{\partial \psi_2}{\partial z_2}(\zb)\,.
\end{equation}
Using (A3), this implies that $\psi_2$ does not depend on $Z_2$, that is, $V_2=\psi_2(Z_1)$.

Next, we again write $q(z_1)$ in terms of $p_2(\psi_2(z_1))$ using the univariate change of variable formula, substitute into~\eqref{eq:app_2qe2}, cancel the corresponding terms, and arrive at:
\begin{equation}
\tilde q_2(z_2)= 
\tilde p_1\left(\psi_1(z_1,z_2)\right) \abs{\frac{\partial \psi_1}{\partial z_2}(z_1,z_2)}
\end{equation}
\cref{lemma:brehmer} applied to $\psi_1(z_1,\cdot)$ which preserves $\tilde q_2$ and $\tilde p_1$ for all $z_1$ shows that $\psi_1$ is constant in $Z_1$, that is
\begin{equation}
\label{eq:app_permutation}
    \Vb=(V_1,V_2)=\psi(\Zb)=(\psi_1(Z_2),\psi_2(Z_1))\,.
\end{equation}
Since $V_1\independent V_2$ by the assumption of Case 2a, it follows from the invertible element-wise reparametrisation above that $Z_1 \independent Z_2$ and hence, by faithfulness, $Z_1\not\to Z_2$ in $G'$.

Finally, note that there is no partial order on the empty graph and so $G'=\pi(G)=G$ and $\Zb=\Pb_{\pi^{-1}}\cdot \psi^{-1}(\Vb)$ where $\pi$ is the nontrivial permutation of $\{1,2\}$.

\textit{Case 2b: $V_1\to V_2$ in $G$, that is $\pa(2)=\{1\}$. }
If $G'\neq G$, that is $Z_1\not\to Z_2$ in $G'$, then the same argument as in Case 2a, this time starting by taking the partial derivative of~\eqref{eq:app_misaligned_ratio_2} w.r.t.\ $z_1$, can be used to reach the same conclusion in~\eqref{eq:app_permutation}. However, this contradicts faithfulness since $V_1\not\independent V_2$ in $G$. 

Hence, we must have $G'=G$, and the following two equations must hold for all $\zb$:
\begin{align}
    \frac{\tilde q_1(z_1)}{q_{1}(z_1)}
    &=\frac{\tilde p_2\left(\psi_2(\zb)\right)}{p_{2}\left(\psi_2(\zb)~|~\psi_1(\zb)\right)}
    \label{eq:case2be2}
    \\
    \frac{\tilde q_2(z_2)}{q_{2}(z_2~|~z_1)}
    &=\frac{\tilde p_1\left(\psi_1(\zb)\right)}{p_1\left(\psi_1(\zb)\right)}
    \label{eq:case2be1}
\end{align}

The remainder of the proof consists of exploring the implications of~\eqref{eq:case2be1} and ~\eqref{eq:case2be2}, ultimately resulting in a violation of the genericity condition (A4).

To ease notation, define the following auxiliary functions:
\begin{align}
    a(z_1)&:=\frac{\tilde q_1(z_1)}{q_{1}(z_1)}\,,
    \\
    b(\vb)&:=\frac{\tilde p_2(v_2)}{p_2(v_2~|~v_1)}\,,
    \\
    c(\zb)&:=\frac{\tilde q_2(z_2)}{q_{2|1}(z_2~|~z_1)}\,,
    \\
    d(v_1)&:=\frac{\tilde p_1\left(v_1\right)}{p_1\left(v_1\right)}\,.
\end{align}

With this, \eqref{eq:case2be2} and \eqref{eq:case2be1}   take the following form:
\begin{align}
    \label{eq:KR}
    a(z_1)&=b(\psi(\zb))\,.
    \\
    \label{eq:QS}
    c(\zb)&=d(\psi_1(\zb))\,,
\end{align}

Next, define the following maps:
\begin{align}
    \label{eq:kappa}
    \kappa&: \zb \mapsto \begin{bmatrix}
        a(z_1) 
        \\
        c(\zb)
    \end{bmatrix}
    \\
    \label{eq:rho}
    \rho&:\vb \mapsto
    \begin{bmatrix}
    b(\vb) \\
    d(v_1)
    \end{bmatrix}
\end{align}

Then, \eqref{eq:KR} and \eqref{eq:QS} together imply that
\begin{equation}
\label{eq:identity_kappa_rho_psi}
    \kappa=\rho \circ \psi\,.
\end{equation}

Recalling that by (A1) all densities are continuously differentiable, the Jacobians of $\kappa$ and $\rho$ are given by:
\begin{align}
    \Jb_\kappa(\zb)
    &=
    \begin{bmatrix}
        a'(z_1)
        & 
        0\\
        \frac{\partial c}{\partial z_1}(\zb) &  \frac{\partial c}{\partial z_2}(\zb)
    \end{bmatrix}\,,
    \\
    \Jb_\rho(\vb)
    &=
    \begin{bmatrix}
    \frac{\partial b}{\partial v_1}(\vb) &  \frac{\partial b}{\partial v_2}(\vb)\\
    d'(v_1)
        & 
        0
    \end{bmatrix}\,,
\end{align}
and the corresponding determinants are given by
\begin{align}
    \abs{\det \Jb_\kappa(\zb)}&=\abs{a'(z_1)\frac{\partial c}{\partial z_2}(\zb)}\neq 0
    \\
    \abs{\det \Jb_\rho(\vb)}&=\abs{d'(v_1)\frac{\partial b}{\partial v_2}(\vb)} \neq 0
\end{align}
where the inequalities for all $\zb$ follow since, by assumption (A3), the derivatives of ratios of intervened and original mechanisms are non-vanishing everywhere:
\begin{align}
\label{eq:non_vanishing_derivatives}
    a'(z_1)\neq 0 \neq \frac{\partial c}{\partial z_2}(\zb)
    \qquad \mbox{and} \qquad 
    d'(v_1)\neq 0 \neq \frac{\partial b}{\partial v_2}(\vb)\,,
\end{align}
This implies that the following families of maps are continuously differentiable, monotonic, and invertible, 
\begin{align}
    a&: z_1\mapsto a(z_1)\,,\\
    b_{v_1}&: v_2 \mapsto b(v_1,v_2)\,,\\
    c_{z_1}&: z_2 \mapsto c(z_1,z_2)\,,\\
    d&: v_1 \mapsto d(v_1)\,,
\end{align}
with continuously differentiable inverses
\begin{align}
    a^{-1}&: w_1\mapsto a^{-1}(w_1)\,,\\
    b_{v_1}^{-1}&: w_1 \mapsto b^{-1}_{v_1}(w_1)\,,\\
    c_{z_1}^{-1}&: w_2 \mapsto c^{-1}_{z_1}(w_2)\,,\\
    d^{-1}&: w_2 \mapsto d^{-1}(w_2)\,.
\end{align}

This implies that $\rho$ and $\kappa$ are valid diffeomorphisms onto their image and their inverses are given by: 
\begin{align}
\kappa^{-1}&: \wb \mapsto 
    \begin{bmatrix}
        a^{-1}(w_1) 
        \\
        c^{-1}_{a^{-1}(w_1)}(w_2)
    \end{bmatrix}\,,
    \\
    \rho^{-1}&:\wb \mapsto
    \begin{bmatrix}
    d^{-1}(w_2)\\
    b^{-1}_{d^{-1}(w_2)}(w_1)
    \end{bmatrix}\,.
\end{align}

Since $\Vb=\psi(\Zb)$, by~\eqref{eq:identity_kappa_rho_psi} we have 
\begin{equation}
\label{eq:W_def}
    \Wb:=\rho(\Vb)=\rho\circ \psi(\Zb)=\kappa(\Zb)\,.
\end{equation}

Denote the distributions of $\Wb$ by $R_\Wb$ and its density by $r(\wb)$. Since for all $e$, we have 
\begin{equation}
    P^e_\Vb=\psi_*(Q^e_\Zb)
\end{equation}
it follows from~\eqref{eq:W_def} that
\begin{equation}
    R_\Wb^e=\rho_*(P^e_\Vb)=\kappa_*(Q^e_\Zb)\,.
\end{equation}

This provides two different ways of applying the change of variable formula to compute $r(\wb)$.

First, we consider the pushforward of $Q^{e_0}_\Zb$ by $\kappa$:
\begin{align}
    r(\wb)
    &=
    q\left(\kappa^{-1}(\wb)\right) \abs{\det \Jb_{\kappa^{-1}}(\wb)}
    \\
    &=
    q_1\left(a^{-1}(w_1)\right)
    q_2\left(c^{-1}_{a^{-1}(w_1)}\left(w_2\right)~\mid~a^{-1}(w_1)\right)\abs{\frac{\d}{\d w_1}a^{-1}(w_1)\frac{\d}{\d w_2}c^{-1}_{a^{-1}(w_1)}(w_2)}\,
    \label{eq:r_w_q}
\end{align}
By integrating this joint density with respect to $w_2$, we obtain the following expression for the marginal $r_1(w_1)$:
\begin{align}
    r_1(w_1)
    &=
    \abs{\frac{\d}{\d w_1}a^{-1}(w_1)}q_1\left(a^{-1}(w_1)\right)\int 
    q_2\left(c^{-1}_{a^{-1}(w_1)}\left(w_2\right)~\mid~a^{-1}(w_1)\right)\abs{\frac{\d}{\d w_2}c^{-1}_{a^{-1}(w_1)}(w_2)}\d w_2\,.
\end{align}

By the diffeomorphic change of variable $z_2=c^{-1}_{a^{-1}(w_1)}\left(w_2\right)$, \footnote{Note that: $\int q_2(z_2(w_2))\abs{\frac{\d z_2}{\d w_2}}\d w_2=\int q_2(z_2)\d z_2$\,.}
this can be written as
\begin{align}
    r_1(w_1)
    &=
    \abs{\frac{d}{dw_1}a^{-1}(w_1)}q_1\left(a^{-1}(w_1)\right)\int 
    q_2\left(z_2~\mid~a^{-1}(w_1)\right)dz_2\,\\
    &=\abs{\frac{d}{dw_1}a^{-1}(w_1)}q_1\left(a^{-1}(w_1)\right)\label{eq:r1e0q}
\end{align}

Next, we carry out the same calculation for the pushforward of $P^{e_0}_\Vb$ by $\rho$:
\begin{align}
    r(\wb)
    &=
    p\left(\rho^{-1}(\wb)\right) \abs{\det \Jb_{\rho^{-1}}(\wb)}
    \\
    &=p_1\left( d^{-1}(w_2)\right)
    p_2\left(b^{-1}_{d^{-1}(w_2)}(w_1)~\mid~d^{-1}(w_2)\right)
    \abs{\frac{\d }{\d w_2}
    d^{-1}(w_2)\frac{\d }{\d w_1}
    b^{-1}_{d^{-1}(w_2)}(w_1)}\,,
\end{align}
leading to the marginal
\begin{align}
    r_1(w_1)
    &=
    \int p_1\left( d^{-1}(w_2)\right)p_2\left(b^{-1}_{d^{-1}(w_2)}(w_1)~\mid~d^{-1}(w_2)\right)
    \abs{\frac{\d }{\d w_1}
    b^{-1}_{d^{-1}(w_2)}(w_1)}\abs{\frac{\d }{\d w_2}
    d^{-1}(w_2)}\d w_2\\
    &=
    \int p_1(v_1)p_2\left(b^{-1}_{v_1}(w_1)~\mid~v_1\right)
    \abs{\frac{\d }{\d w_1}
    b^{-1}_{v_1}(w_1)}\d v_1\label{eq:r1e0p}\,,
\end{align}
where the second line is obtained by the diffeomorphic change of variable $v_1=d^{-1}(w_2)$. 

Equating the two expressions for $r(w_1)$ in $e_0$ in~\eqref{eq:r1e0p} and~\eqref{eq:r1e0q}, we obtain for all $w_1$:
\begin{equation}
    \abs{\frac{\d }{\d w_1}a^{-1}(w_1)}q_1\left(a^{-1}(w_1)\right)=\int p_1(v_1)p_2\left(b^{-1}_{v_1}(w_1)~\mid~v_1\right)
    \abs{\frac{\d }{\d w_1}
    b^{-1}_{v_1}(w_1)}\d v_1\label{eq:r1e0}\,.
\end{equation}

Applying the same approach to the environment in which $V_1$ is intervened upon changing $p_1$ to $\tilde p_1$ while $Z_2$ is intervened upon leaving $q_1$ invariant,
yields for all $w_1$:
\begin{equation}
    \abs{\frac{\d }{\d w_1}a^{-1}(w_1)}q_1\left(a^{-1}(w_1)\right)=\int \tilde{p}_1(v_1)p_2\left(b^{-1}_{v_1}(w_1)~\mid~v_1\right)
    \abs{\frac{\d }{\d w_1}
    b^{-1}_{v_1}(w_1)}\d v_1\label{eq:r1e1}\,.
\end{equation}

Finally, by equating \eqref{eq:r1e0} and \eqref{eq:r1e1}, we arrive at the following expression which must hold for all $w_1$:
\begin{equation}
    \int p_1(v_1)p_2\left(b^{-1}_{v_1}(w_1)~\mid~v_1\right)
    \abs{\frac{\d }{\d w_1}
    b^{-1}_{v_1}(w_1)}\d v_1=\int \tilde{p}_1(v_1)p_2\left(b^{-1}_{v_1}(w_1)~\mid~v_1\right)
    \abs{\frac{\d }{\d w_1}
    b^{-1}_{v_1}(w_1)}\d v_1\label{eq:r1diff}
\end{equation}
which we can rewrite as
\begin{equation}
    \int \left(\tilde{p}_1(v_1)-p_1(v_1)\right)p_2\left(b^{-1}_{v_1}(w_1)~\mid~v_1\right)
    \abs{\frac{\d }{\d w_1}
    b^{-1}_{v_1}(w_1)}\d v_1=0%
    \,.
\end{equation}

Multiplying by any continuous function $\varphi(w_1)$,
integrating w.r.t.\ $w_1$ and applying the diffeomorphic change of variable $v_2=b^{-1}_{v_1}(w_1)$, this can be expressed as:
\begin{align}
0 &=    \int \varphi(w_1) 
\int \left(\tilde{p}_1(v_1)-p_1(v_1)\right)p_2\left(b^{-1}_{v_1}(w_1)~\mid~v_1\right)
    \abs{\frac{\d }{\d w_1}
    b^{-1}_{v_1}(w_1)}\d v_1
\d w_1\label{eq:r1int}
\\
    &=    \int\int \varphi\left(b_{v_1}(v_2)\right) \left(\tilde{p}_1(v_1)-p_1(v_1)\right)p_2(v_2~\mid~v_1)
    \d v_2 \d v_1\\
 &=    \int\int \varphi\left(\frac{\tilde{p}_2(v_2)}{p_2(v_2\mid v_1)}\right) \left(\tilde{p}_1(v_1)-p_1(v_1)\right)p_2(v_2~\mid~v_1)
    \d v_2 \d v_1
\end{align}
where we have resubstituted the expression for $b_{v_1}(v_2)$ in the last line. 

Equivalently, this can be written as: \textit{for any continuous function} $\varphi$,
\begin{equation}
    \EE_{\vb\sim P^{e_0}_\Vb}\left[\varphi\left(\frac{\tilde p_2(v_2)}{p_2(v_2~|~v_1)}\right)\right]
    =
    \EE_{\vb\sim P^{e_1}_\Vb}\left[\varphi\left(\frac{\tilde p_2(v_2)}{p_2(v_2~|~v_1)}\right)\right] \, .
\end{equation}

However, the genericity condition (A4) precisely rules this out, since the above equality must be violated for at least one $\varphi$, concluding this last case. 
    
To sum up, all cases either lead to a contradiction, or imply the conclusion that $(f^{-1},G)\sim_\textsc{crl}(h,G')$, concluding the proof.
\end{proof}

\subsection{Proof of\texorpdfstring{~\cref{thm:general}}{thmgeneral}}
\label{app:proof_general}
\general*
\begin{proof}
First, we show that we can extract from the $m\geq n$ available pairs of environments a suitable subset $\Ecal_n$ of exactly $n$ pairs, containing one pair of interventional environments for each node.

Let $\Ecal_n\subseteq \Ecal$ be a subset of $n$ pairs of environments which are assumed to correspond to distinct targets in the model $q$, and suppose for a contradiction that this is not actually the case for the ground truth $p$ (i.e., there are duplicate and missing interventions w.r.t.\ $p$).
Then there must be two pairs of environments $(e_a, e_a'),(e_b, e_b')\in \Ecal_n$, both corresponding to interventions on some $V_i$ in $p$, but which are modelled as interventions on distinct nodes $Z_j$ and $Z_k$ with $j\neq k$ in $q$. 
We show that this implies that $V_i$ must simultaneously be a deterministic function of only $Z_j$ and only $Z_k$.
Similar to the proof of~\cref{thm:bivariate},
we obtain the following equations,
\begin{align}
    \frac{\dbtilde q_{j}}{\tilde q_{j}}\left(z_{j}\right)&=\frac{\dbtilde p_i}{\tilde p_i}\left(\psi_i(\zb)\right)\,,\\
    \frac{\dbtilde q_{k}}{\tilde q_{k}}\left(z_{k}\right)&=\frac{\hat{\hat{p_i}}}{\hat p_i}\left(\psi_i(\zb)\right)\,.
\end{align}
By taking partial derivatives w.r.t.\ $z_l$ and applying assumption (A3'), we find that 
\begin{align}
   \frac{\partial \psi_i}{\partial z_l}=0 \qquad \forall l\neq j\,,\\
   \frac{\partial \psi_i}{\partial z_l}=0 \qquad \forall l\neq k\,.
\end{align}
Since $j\neq k$, this implies that $\partial \psi_i / \partial z_l =0$ for all $l$ which contradicts invertibility of $\psi$.
Thus, by contradiction, we find that $\Ecal_n$ must contain exactly one pair of intervention per node also w.r.t.\ $p$. 
For the remainder of the proof, we only consider $\Ecal_n$.

W.l.o.g., for any $(e_i,e_i')\in\Ecal_n$ we now fix the intervention targets in $p$ to $\Ical^{e_i}=\Ical^{e_i'}=\{i\}$ and let $\pi$ be a permutation of $[n]$ such that $\pi(i)$ denotes the inferred intervention target in $q$ that by (A2') is shared across $(e_i,e_i')$.  (We will show later that not all permutations are admissible, but only ones that preserve the partial order of $G$.)

The first part of the proof is similar to Case 1 in the proof of~\cref{thm:bivariate}.
Consider the densities in environments $e_i$ and $e_i'$, which are related through the change of variable formula by:
\begin{align}
    \tilde q_{\pi(i)}\left(z_{\pi(i)}\right) \prod_{j\in [n]\setminus\{\pi(i)\}} q_j\left(z_j~|~\zb_{\pa(j; G')}\right)
    &=
    \tilde p_i\left(\psi_i(\zb)\right) \prod_{j\in [n]\setminus\{i\}} p_j\left(\psi_j(\zb)~|~\psi_{\pa(j)}(\zb)\right)
    \abs{\det \Jb_\psi(\zb)}\,,
    \\
    \dbtilde q_{\pi(i)}\left(z_{\pi(i)}\right) \prod_{j\in [n]\setminus\{\pi(i)\}} q_j\left(z_j~|~\zb_{\pa(j; G')}\right)
    &=
    \dbtilde p_i\left(\psi_i(\zb)\right) \prod_{j\in [n]\setminus\{i\}} p_j\left(\psi_j(\zb)~|~\psi_{\pa(j)}(\zb)\right)
    \abs{\det \Jb_\psi(\zb)}\,,
\end{align}
where $\Zb_{\pa(j;G')}\subseteq \Zb\setminus \{Z_j\}$ denotes the parents of $Z_j$ in $G'$.

Taking the quotient of the two equations yields
\begin{align}
    \frac{\dbtilde q_{\pi(i)}}{\tilde q_{\pi(i)}}\left(z_{\pi(i)}\right)&=\frac{\dbtilde p_i}{\tilde p_i}\left(\psi_i(\zb)\right)\,.
\end{align}

Next, for any $j\neq \pi(i)$, taking partial derivatives w.r.t.\ $z_j$ on both sides yields
\begin{align}
\label{eq:psi_permutation}
    0&=\left(\frac{\dbtilde p_i}{\tilde p_i}\right)'\left(\psi_i(\zb)\right)\frac{\partial \psi_i}{\partial z_j}(\zb)\,.
\end{align}

By assumption (A3'), the first term on the RHS is non-zero everywhere. Hence, \eqref{eq:psi_permutation} implies
\begin{align}
    \forall j\neq \pi(i), \, \forall \zb: \quad \frac{\partial \psi_i}{\partial z_j}(\zb)=0
\end{align}
from which we can conclude that 
\begin{align}
\label{eq:ViZpii}
    V_i=\psi_i\left(Z_{\pi(i)}\right) 
\end{align}
for all $i\in[n]$. 
That is, $\psi$ is the composition of the permutation $\pi$ with an element-wise reparametrisation.

It remains to show that $\pi$ must, in fact, be a graph isomorphism, which is equivalent to the statement 
\begin{equation}
    V_i\to V_j \quad \mbox{in} \quad G \quad \iff \quad Z_{\pi(i)}\to Z_{\pi(j)}\quad \mbox{in} \quad G'. 
\end{equation}
($\implies$) Suppose for a contradiction that there exist $(i,j)$ such that $V_i\to V_j$ in~$G$, but $Z_{\pi(i)}\not\to Z_{\pi(j)}$ in $G'$. 

The main idea is to demonstrate that the lack of such direct arrow implies a certain conditional independence which, by faithfulness, would contradict the unconditional dependence of $V_i$ and $V_j$.

Consider environment $e_i$ in which there are perfect interventions on $Z_{\pi(i)}$ and $V_i$, which has the effect of removing all incoming arrows to $Z_{\pi(i)}$ and $V_i$ in the respective post-intervention graphs $G'_{\overline Z_{\pi(i)}}$ and  $G_{\overline V_i}$.

As a result of this and the lack of direct arrow by assumption, any d-connecting path between $Z_{\pi(i)}$ and $Z_{\pi(j)}$ must enter the latter via $\Zb_{\pa(\pi(j);G')}$~\citep{Pearl2009}.

It then follows from Markovianity of $q$ w.r.t.\ $G'$ that the following holds in~$Q_\Zb^{e_i}$:
\begin{equation}
\label{eq:app_cond_indep_Z}
Z_{\pi(i)}\independent Z_{\pi(j)}~|~\Zb_{\pa(\pi(j);G')}\,.
\end{equation}

We now consider the corresponding implication for $P^{e_i}_\Vb$.
Define
\begin{align}
\tilde\Vb=\left\{V_k=\psi_k\left(Z_{\pi(k)}\right):Z_{\pi(k)}\in\Zb_{\pa(\pi(j);G')}\right\}\subseteq\Vb\setminus\{V_i,V_j\}\,,
\end{align}
and note that by assumption, $Z_{\pi(i)}\not\in \Zb_{\pa(\pi(j);G')}$ and hence $V_i\not\in\tilde\Vb$.

By applying the corresponding diffeomorphic functions $\psi_i$ from~\eqref{eq:ViZpii} to~\eqref{eq:app_cond_indep_Z}, it follows from~\cref{lemma:preserve_cond_ind} that 
\begin{equation}
 V_i\independent V_j~|~\tilde \Vb   
\end{equation}
in $P^{e_i}_\Vb$. 
However, this violates faithfulness~(\cref{ass:faithfulness}) of $P_\Vb$ to $G$ since $V_i$ and $V_j$ are d-connected in $G_{\overline V_i}$.

Thus, by contradiction, we must have $Z_{\pi(i)}\to Z_{\pi(j)}$  in $G'$.

($\Longleftarrow$) Now, suppose for a contradiction that there exist $(i,j)$ such that $Z_{\pi(i)}\to Z_{\pi(j)}$ in $G'$, but $V_i\not \to V_j$ in~$G$.

By the same argument as before, we find that
\begin{equation}
    V_i\independent V_j ~|~ \Vb_{\pa(j)}
\end{equation}
in $P^{e_i}_\Vb$, and thus by~\cref{lemma:preserve_cond_ind}
\begin{equation}
    Z_{\pi(i)}\independent Z_{\pi(j)}~|~\tilde \Zb
\end{equation}
in $Q_\Zb^{e_i}$ where 
$$\tilde \Zb=\left\{Z_{\pi(k)}:V_k\in\Vb_{\pa(j)}\right\}\subseteq\Zb\setminus \{Z_{\pi(i)}, Z_{\pi(j)}\}\,.$$
However, this contradicts faithfulness of $Q_\Zb$ to $G'$. 
Hence, we must have that $V_i \to V_j$ in~$G$.

This shows that $\pi$ must be a graph isomorphism, thus concluding the proof.
\end{proof}

\subsection{Proof of\texorpdfstring{~\cref{prop:influence}}{Proposition 4.2}}
\label{app:proof_influence}
\influence*

\begin{proof}
First, recall that according to~\cref{def:influence}, 
\begin{equation}
\mathfrak{C}^{P_\Vb}_{i\to j}:=D_\textsc{kl}\big(P_\Vb~\big\|~ P^{i\to j}_\Vb\big),
\end{equation}
where $P^{i\to j}_\Vb$ denotes the interventional distribution obtained by replacing $p_j\left(v_j~|~\vb_{\pa(j)}\right)$ with
\begin{equation}
\quad p^{i\to j}_j\big(v_j~|~\vb_{\pa(j)\setminus \{i\}}\big)
=
\int_{\Vcal_i} p_j\big(v_j~|~\vb_{\pa(j)}\big) p_i(v_i) \d v_i\,.
\end{equation}
Writing out the KL divergence and noting that all terms except the interved mechanism $j$ cancel inside the $\log$, we obtain
\begin{equation}
    \mathfrak{C}^{P_\Vb}_{i\to j}
    =
    \int_{\Vcal} \log \left(\frac{p_j\left(v_j~|~\vb_{\pa(j)}\right)}{\int_{\Vcal_i} p_j\big(v_j~|~\vb_{\pa(j)}\big) p_i(v_i) \d v_i}\right)p(\vb)\d\vb \,.
\end{equation}
and similarly 
\begin{equation}
    \mathfrak{C}^{Q_\Zb}_{\pi(i)\to \pi(j)}
    =
    \int_{\Zcal} \log \left(\frac{q_{\pi(j)}\left(z_{\pi(j)}~|~\zb_{\pa(\pi(j);G')}\right)}{\int_{\Zcal_{\pi(i)}} q_{\pi(j)}\left(z_{\pi(j)}~|~\zb_{\pa(\pi(j);G')}\right) q_{\pi(i)}(z_{\pi(i)}) \d z_{\pi(i)}}\right)q(\zb)\d\zb \,.
\end{equation}

Since $\Zb=\Pb_{\pi^{-1}}\circ \phi(\Vb)$, we have $V_{i}=\psi_i(Z_{\pi(i)})$ for all $i\in[n]$ where $\psi=\phi^{-1}$.

Thus, by the change of variable formula, and using the fact that $\pi(\pa(i))=\pa(\pi(i);G')$ since $\pi:G\mapsto G'$ is a graph isomorphism, we have for all $i\in[n]$:
\begin{align}
    \label{eq:app_transformed_mechanisms}
    q_{\pi(i)}\left(z_{\pi(i)}~|~\zb_{\pa(\pi(i);G')}\right)
    =p_i\left(\psi_i\left(z_{\pi(i)}\right)~|~\psi_{\pa(i)}\left(\zb_{\pa(\pi(i);G')}\right)\right)
    \abs{\frac{\d \psi_i}{\d z_{\pi(i)}}\left(z_{\pi(i)}\right)}\,,
\end{align}
as well as for the marginal density
\begin{align}
    \label{eq:app_transformed_mechanisms_marginal}
    q_{\pi(i)}\left(z_{\pi(i)}\right)
    =p_i\left(\psi_i\left(z_{\pi(i)}\right)\right)
    \abs{\frac{\d \psi_i}{\d z_{\pi(i)}}\left(z_{\pi(i)}\right)}\,,
\end{align}
and 
\begin{align}
    q(\zb)=p(\psi\circ\Pb_{\pi}(\zb))\abs{\det \Jb_\psi(\zb)}\,.
\end{align}

Substitution into the expression for $\mathfrak{C}^{Q_\Zb}_{\pi(i)\to \pi(j)}$ yields:
\begin{align}
    \mathfrak{C}^{Q_\Zb}_{\pi(i)\to \pi(j)}
    &=
    \int_{\Zcal} \log \left(\frac{p_{j}\left(\psi_j(z_{\pi(j)})~|~\psi_{\pa(j)}(\zb_{\pa(\pi(j);G')})\right)}
    {\int_{\Zcal_{\pi(i)}} p_{j}\left(\psi_j(z_{\pi(j)})~|~\psi_{\pa(j)}(\zb_{\pa(\pi(j);G')})\right) p_{i}(\psi_i(z_{\pi(i)}))\abs{\frac{\d \psi_i}{\d z_{\pi(i)}}(z_{\pi(i)})} \d z_{\pi(i)}}\right)
    \\
    & \qquad p(\psi\circ\Pb_{\pi}(\zb))\abs{\det \Jb_\psi(\zb)}\d\zb \,.
    \\
    &=\int_{\Vcal} \log \left(\frac{p_j\left(v_j~|~\vb_{\pa(j)}\right)}{\int_{\Vcal_i} p_j\big(v_j~|~\vb_{\pa(j)}\big) p_i(v_i) \d v_i}\right)p(\vb)\d\vb 
    \\
    &=\mathfrak{C}^{P_\Vb}_{i\to j}\,.
\end{align}
where the second to last line follows by integration by substitution, applied to both integrals.
\end{proof}